%% file: main.tex
\documentclass[10pt,twocolumn,letterpaper]{article}

\usepackage[pagenumbers]{cvpr} 

\input{preamble}

\definecolor{cvprblue}{rgb}{0.21,0.49,0.74}
\usepackage[pagebackref,breaklinks,colorlinks,citecolor=cvprblue]{hyperref}

\def\paperID{14402} 
\def\confName{CVPR}
\def\confYear{2024}

\input{math_commands}

\title{\vspace{-.25in}PFStorer: Personalized Face Restoration and Super-Resolution\vspace{-.1in}}

\author{
	Tuomas Varanka$^{*1}$ ~~~ Tapani Toivonen$^2$ ~~~ Soumya Tripathy$^2$ ~~~ Guoying Zhao$^1$ ~~~ Erman Acar$^2$ \\
	$^1$ University of Oulu ~~~~~~~ $^2$ Huawei Finland\\
    {\tt\small tuomas.varanka@oulu.fi}
}

\begin{document}
\input{figures/teaser.tex}
\let\thefootnote\relax\footnotetext{$^*$This research was performed during an internship at Huawei Finland.}
\maketitle

\input{sec/0_abstract}    
\input{sec/1_intro}

\input{sec/2_related}
\input{sec/3_method}

\input{sec/4_experiments}
\input{sec/5_conclusion}
{
    \small
    \bibliographystyle{ieeenat_fullname}
    \bibliography{main}
}

\input{sec/6_appendix}

\end{document}

%% file: preamble.tex
\usepackage[dvipsnames]{xcolor}
\usepackage{wrapfig}


%% file: math_commands.tex
\usepackage{amsmath,amsfonts,bm}
\usepackage{mathtools}

\def\eqref#1{equation~\ref{#1}}

\def\1{\bm{1}}

\def\eps{{\epsilon}}

\DeclareMathAlphabet{\mathsfit}{\encodingdefault}{\sfdefault}{m}{sl}
\SetMathAlphabet{\mathsfit}{bold}{\encodingdefault}{\sfdefault}{bx}{n}



%% file: figures/teaser.tex
\twocolumn[{
\renewcommand\twocolumn[1][]{#1}
\maketitle
\begin{center}
    \centering
    \vspace*{-.8cm}
    \includegraphics[width=\textwidth]{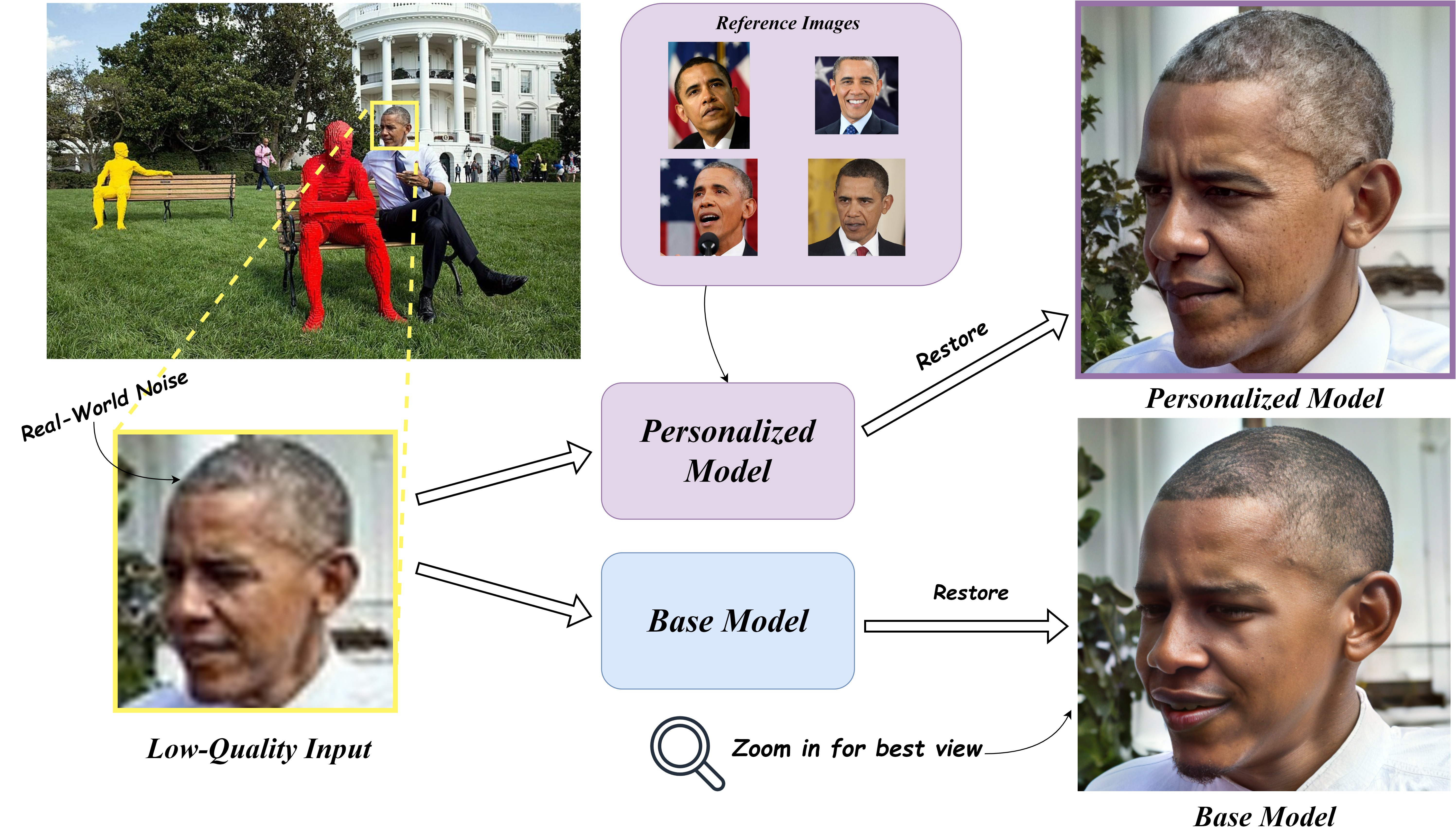}
    \vspace*{-.6cm}
    \captionof{figure}{Imagine wanting to restore a photo of yourself, only for the resulting image to not be you, but someone else! By utilizing a few high-quality reference images, we can faithfully restore images with fine-grained details. Best viewed by zooming in.}
\label{fig:teaser}
\end{center}
}]

%% file: sec/0_abstract.tex
\begin{abstract}
Recent developments in face restoration have achieved remarkable results in producing high-quality and lifelike outputs. The stunning results however often fail to be faithful with respect to the identity of the person as the models lack necessary context. In this paper, we explore the potential of personalized face restoration with diffusion models. In our approach a restoration model is personalized using a few images of the identity, leading to tailored restoration with respect to the identity while retaining fine-grained details. By using independent trainable blocks for personalization, the rich prior of a base restoration model can be exploited to its fullest. To avoid the model relying on parts of identity left in the conditioning low-quality images, a generative regularizer is employed. With a learnable parameter, the model learns to balance between the details generated based on the input image and the degree of personalization. Moreover, we improve the training pipeline of face restoration models to enable an alignment-free approach. We showcase the  robust capabilities of our approach in several real-world scenarios with multiple identities, demonstrating our method's ability to generate fine-grained details with faithful restoration. In the user study we evaluate the perceptual quality and faithfulness of the genereated details, with our method being voted best 61\% of the time compared to the second best with 25\% of the votes.

\end{abstract}

%% file: sec/1_intro.tex
\section{Introduction}
Face restoration aims to recover HQ (high-quality) face images from degraded observations, such as blur, low-resolution, noise and compression artifacts. In real-world scenarios, the task is even more challenging, due to more complex degradations and variations in illumination and pose.

\begin{figure}
	\centering
    \includegraphics[width=0.48\textwidth]{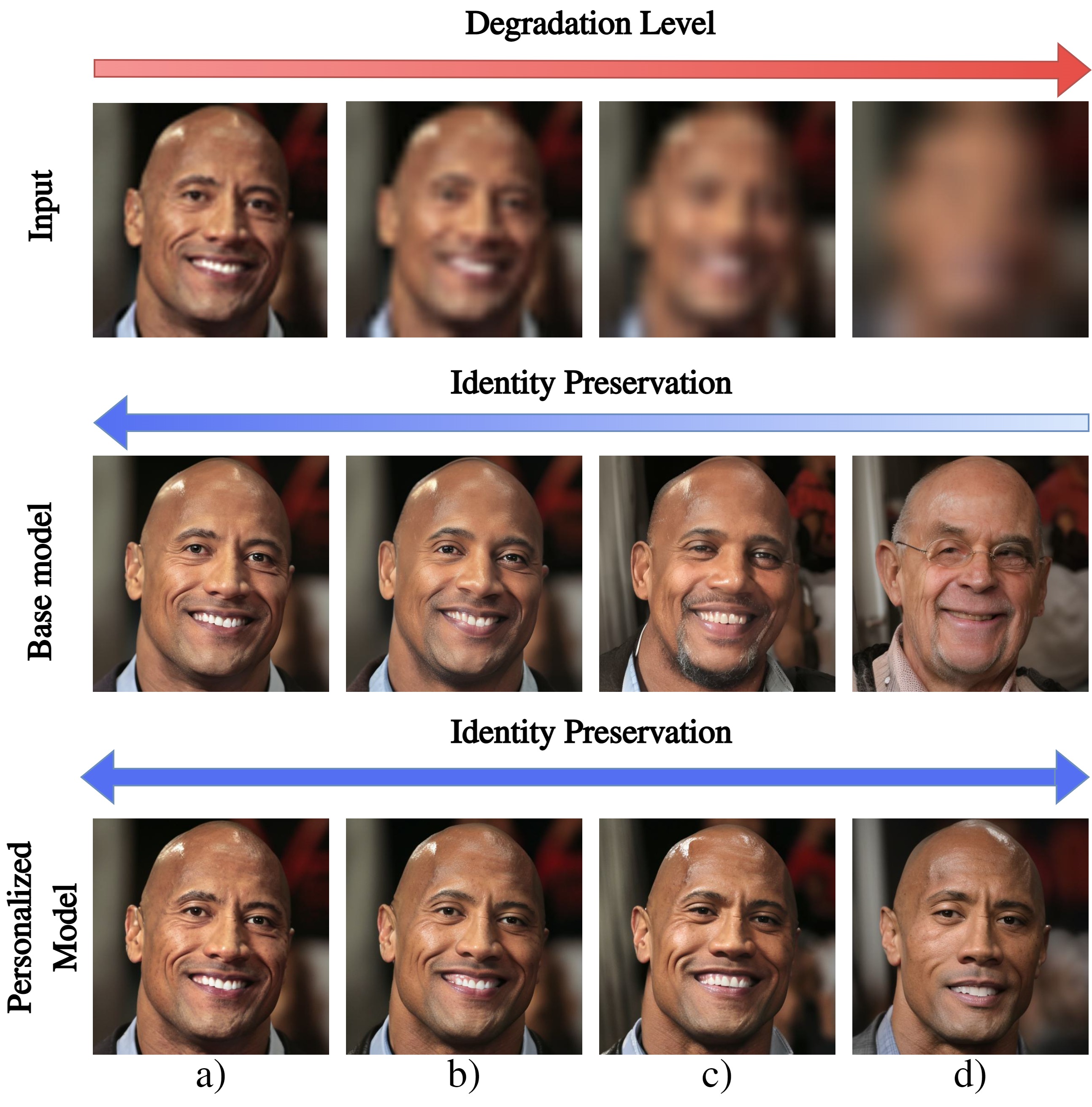}
    \captionof{figure}{Results under increasing levels of degradation. a) With only minor degradation, both base and personalized model are capable of restoration. b) The base model incorrectly restores fine-grained details such as the nose and skin texture. c) More identity details such as eyes and facial hair are lost. d) Base model outputs a completely different identity, while the personalized model retains details of the identity, even if the semantics are not entirely correct due to the extreme low-quality input image. Best viewed by zooming in.}
	\label{fig:degradation_level}
	\vspace{-3mm}
\end{figure}

Restoration of faces is a highly ill-posed problem with multiple solutions to a given LQ (low-quality) input. Compared to natural images humans are very sensitive to subtle differences with facial images. Even small variations in the shape, size or color of eyes, nose, lips, \textit{etc}., can cause a shift in the identity of the person, see top and middle row of \cref{fig:degradation_level}. Furthermore, if we are familiar with identity of a specific person, we are even more prone to spotting subtle differences. We show an example in \cref{fig:teaser}. A restoration model needs to not only output a realistic image, but also one that is faithful to the identity of the person.

Recent research on face restoration has seen great progress towards higher visual quality results. Many of the techniques exploit a generative prior such as GAN \cite{gfp-gan,psfrgan,gwainet}, codebooks \cite{dfdnet,codeformer, vqfr} or diffusion models \cite{difface, dr2}. Generative prior methods have been trained under a generative task prior to being modified to restoration models and as a result they are capable of outputting realistic face images. However, often the outputs can be inauthentic as the models lack crucial context about the identity. To combat this, a reference prior has been used \cite{asffnet, dmdnet}, which uses a HQ reference image of the same identity, leading in theory to high fidelity. However, in practice transferring the identity from a reference image is difficult due to differences in pose, illumination and semantics between the LQ and reference image.

To alleviate the ill-posedness of the restoration problem we fully exploit the reference images by creating a \textit{neural representation} of the identity. We propose PFStorer (Personalized Face Restorer) to restore LQ face images while retaining the identity by personalization. Given a few HQ reference images (\eg 3-5 selfies from a photo gallery) a restoration model is fine-tuned to a personalized restoration model. The reference images can have significantly different illumination, pose, expression and do not have to be aligned with the LQ image. Our goal is to personalize a restoration model such that it is able to restore person specific images while producing realistic images and being faithful to the identity.
 
As opposed to personalized generative models that personalize a generative model, we use a base face restoration model as our foundation. The base model is capable of realistic outputs, however the fidelity may suffer due to the ill-posedness of the task. Our strategy is to perform \textit{personalized restoration} by fine-tuning a base restoration model with a few HQ reference images. However, a naive fine-tuning strategy can destroy the strong existing priors present in the base restoration model. To avoid catastrophic forgetting when performing fine-tuning for personalization, an adapter is used to keep the priors intact. Adapters are trainable blocks that can be used to \textit{adapt} the flow of a model. By freezing the base model and only training the adapter blocks, existing priors can be preserved. To avoid the rapid change of intermediate outputs caused by adapters a learnable parameter is used. The learnable parameter also controls the amount of injection of personalization for different layers of the network, leading to more fine-grained control.

During training, we observe an issue where the model learns to rely too much on the LQ image ignoring the reference images. This is due to the majority of the training samples having low degredation, which can preserve identity information sufficient to restore the face without using the reference images. To alleviate the issue we design a generative regularizer, in which no conditional LQ image is given and the model is forced to generate the identity using only reference images. This approach encourages the model to learn a robust neural representation of the identity, as in generative personalization.

Furthermore, for the base face restoration model we fine-tune a general purpose restoration model on a face dataset to improve its restorative capabilities. During training, instead of resizing all images to a specific size and aligning them, random crops of the face images are used. This has several benefits: 1) The model has access to higher resolution patches. 2) The model is more robust to varying poses. 3) The model is capable of super-resolution through the use of tiling.

We experiment with our technique using both synthetic and real-world data. The user study confirms that our method is able to improve results over previous methods.

%% file: sec/2_related.tex
\section{Related work}

\begin{figure*}
	\centering
    \includegraphics[width=0.99\textwidth]{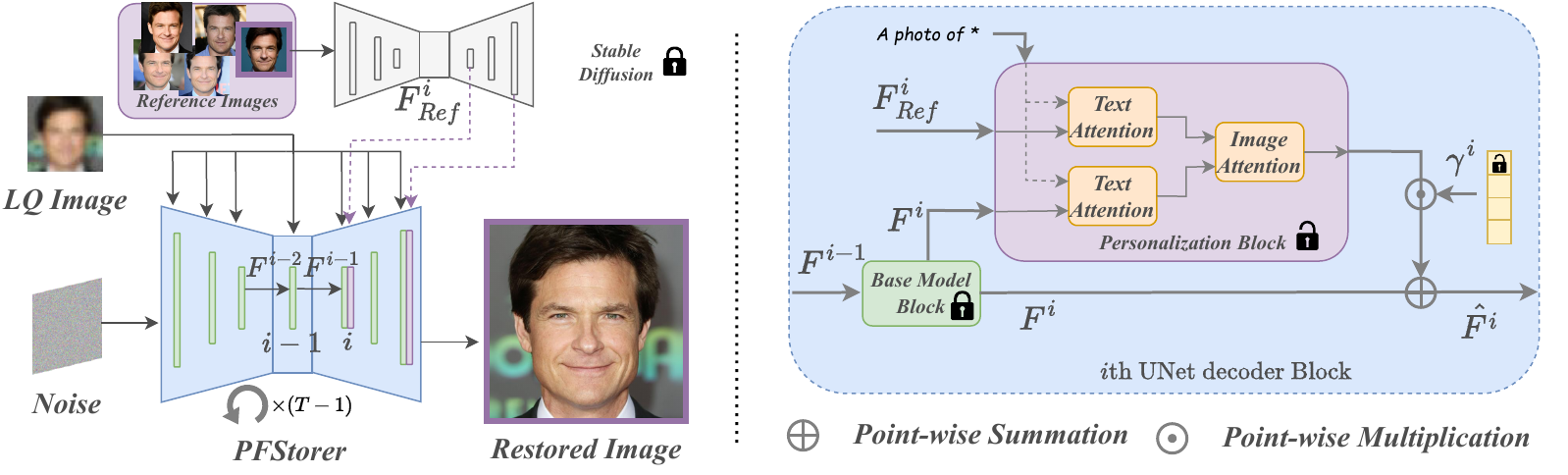}
    \captionof{figure}{(\textbf{Left}) PFStorer restores an image with a diffusion process conditioned on the LQ and the reference image. Base Model blocks are visualized in green and Personlization blocks in purple. StableDiffusion \cite{stablediffusion} is used to extract features $F^i_{Ref}$ from the reference image. During training the reference image is randomly sampled from a set of reference images for each training iteration. During inference, no reference images are required as the identity is learned in the personalization blocks as a neural representation. (\textbf{Right}) $i$th UNet block containing the Base Model Block \cite{stablesr} and Personalization Block \cite{vico}. The Base Model Blocks contain the normal Stable Diffusion blocks with SFT (spatial feature transformation) \cite{sft} blocks from StableSR \cite{stablesr}. After the Base Model block, the intermediate features $F^i$ go to a trainable Personalization Block, which contains cross-attention between the text-embedding and reference image features $F^i_{Ref}$. A learnable adapter vector $\gamma^i$ balances the contribution between the base model and personalization.}
	\label{fig:network_diagram}
	\vspace{-3mm}
\end{figure*}

\paragraph{Face Restoration}
Most recent face restoration approaches include a \textit{generative prior}, where the model has been trained in a generative manner \textit{prior} to training it for restoration. GFP-GAN \cite{gfp-gan} uses the generator of StyleGAN2 \cite{stylegan2} that has been trained on facial images to generate high-quality features. More recently, methods such as VQFR \cite{vqfr} and CodeFormer \cite{codeformer} have been using discrete codebooks with vector quantization and adversarial training \cite{vqgan} to ``store'' high-quality data. Some latest approaches use diffusion models \cite{ddpm} for restoration. DifFace \cite{difface} transforms the LQ input to the manifold of high-quality images with an arbitrary restoration model, which is followed by a forward and backward diffusion, bringing in fine-grained details. DR2 \cite{dr2} uses a similar approach of first performing forward diffusion, after which backward diffusion is performed with guidance from the matching steps of forward diffusion. Zhao \etal \cite{idm} note that around 15\% of the commonly used training data from FFHQ \cite{ffhq} is not necessarily high-quality. To improve the data quality they propose a two-step training process where after the first step, the training data is enhanced, using the model trained in the first step. Concurrent to our work is DiffBFR \cite{diffbfr}, which separates identity and texture restoration using cascaded diffusion models. Another concurrent work \cite{cohen2023posterior}, also emphasizes the ill-posedness of the problem, but goes in the opposite direction to us, encouraging diversity, instead of personalization.
\vspace{-3mm}

\paragraph{Personalization}
With the seminal work of DreamBooth \cite{dreambooth} personalization of generative models with just a few images was made possible. DreamBooth can generate novel scenes of a specific object or a concept. It achieves this by fine-tuning a text-to-image diffusion model while overwriting a rare text embedding. Several works have since followed \cite{custom-diffusion, vico, blip-diffusion, neti, subject-diffusion, hyperdreambooth, fastcomposer, key-locked, photoswap, e4t} that attempt to improve personalization. Custom-Diffusion \cite{custom-diffusion} only fine-tunes the attention layers present in the UNet architecture of StableDiffusion \cite{stablediffusion}, significantly reducing the training time and model size. Perfusion \cite{e4t} goes further and only does rank-1 updates of the attention layers, while locking the keys of cross-attention layers, significantly reducing compute and memory. ViCo \cite{vico} adds image-attention adapters to the cross-attention layers to learn cross-attention between the reference images and predicted image. It also learns a text embedding and applies a regularization using the class-token to avoid overfitting, leading to high-quality results.

On a high-level similar to our approach is RealFill \cite{realfill}, which personalizes a pre-trained in-painting model to perform authentic in- and out-painting. Both MyStyle \cite{mystyle} and IdentityEncoder \cite{identity_encoder} first personalize the model and then transform it to perform tasks such as face in-painting, super-resolution and semantic editing. Compared to their approach, we transform a restoration model to a \textit{personalized restoration model} as opposed to transforming a personalized model to a personalized super-resolution model.
\vspace{-3mm}

\paragraph{Reference-Based Face Restoration}
Reference-based approaches use reference images from the same identity in the restoration process. GFRNet \cite{gfrnet} uses a single reference image and learns a warping between the LQ and reference image. ASFFNet \cite{asffnet} selects the most similar reference image to reduce misalignment and uses adaptive feature fusion for the restoration. DMDNet \cite{dmdnet} constructs a dictionary of deep features from important cropped regions (\eg, eyes, nose, mouth). An alignment module is then used to align the features of the input and reference images, resulting in a fusion of the features to the output image. These methods however struggle when the reference image and LQ input are not aligned or not similar enough. Compared to these approaches we learn a neural representation of the identity, enabling more robust restoration.

%% file: sec/3_method.tex
\section{Method}
\label{sec:method}
We design a face restoration method capable of generating realistic imagery, while still being faithful to the identity of the person in a given image. We begin by analyzing the situation formally (\cref{sec:personal_prior}) and conclude that a personal prior is required for faithful reconstruction in certain situations. Next, we present a method (\cref{sec:personalization}) that preserves existing priors by utilizing adapters for personalized face restoration. To further enhance the results a generative regularizer is proposed to enable robust fine-grained restoration. We name this method \textbf{PFStorer (ours)}. Beyond personalization (\cref{sec:finetune}), we show simple modifications to the training pipeline of general face restoration methods that enable super-resolution and an alignment-free approach. We refer to this improved restoration model without personalization as the \textbf{Base Model}, which is used as a base for personalization. Background for diffusion models and personalization is given in the supplementary material.

\subsection{The Need for a Personal Prior}
\label{sec:personal_prior}
Restoration of low quality images is naturally an ill-posed problem. Assume a degradation function $\mathcal{D}:\mathcal{I} \times \mathbb{R} \rightarrow \mathcal{I}$ that takes in a face image $I \in \mathcal{I}$ and a value of degradation $d \in \mathbb{R}$. A higher degradation value $d$ indicates a higher degraded output image. When $d$ approaches infinity the resulting image will be close to pure noise and restoring the image faithfully is no longer possible, $id(\mathcal{R}(\mathcal{D}(I; d))) \neq id(I)$, where $id$ is a function that returns the identity of a face image and $\mathcal{R}$ is a restoration model. There exists a value $d_f < \infty$ after which faithful restoration is no longer possible. However, with additional personal prior $p_{id}$ the restoration can be made faithfully:
\begin{equation}
	\label{eq:constraint}
	id\left(\mathcal{R}(\mathcal{D}(I; d_f); p_{id})\right) = id(I),
\end{equation}
as $p_{id}$ is unchanged with any value of degradation $d$. In this paper the personal prior $p_{id}$ is learned from a set of reference images using a diffusion model.

\subsection{Personalized Face Restoration}
\label{sec:personalization}

The main idea is to use high-quality images of an individual in aid when restoring LQ images. We start with a restoration model, which is fine-tuned with a personalization technique using the reference images. The personalization is performed for each individual once, after which it can be used for inference as many times as wanted. In essence, the model is trained to add personal details, when the base restoration model is insufficient, due to the ill-posed nature of the problem. The architecture of the model can be seen from \cref{fig:network_diagram}.

During the personalization fine-tuning, the model takes as input a synthesized LQ image $I_{LQ}$ and a reference image $I_{Ref}$ sampled from the set of reference images $\{I_{Ref}^k\}$. A modified diffusion model loss
\begin{equation}
	\mathcal{L}_{Diff} = \mathbb{E}_{z, t, I_{LQ}, I_{Ref}, \eps} \|\eps - \eps_\theta(z_t, c, I_{LQ}, I_{Ref}) \|_2,
\end{equation}
with the addition of the LQ and reference image, is used. Here $\eps_{\theta}$ is the diffusion model, $z_t$ the latent code at time $t$, $c$ the conditioning text embedding and $\eps$ the sampled noise from an Isotropic Gaussian distribution.
\vspace{-3mm}

\begin{figure*}[t]
	\centering
    \includegraphics[width=\textwidth]{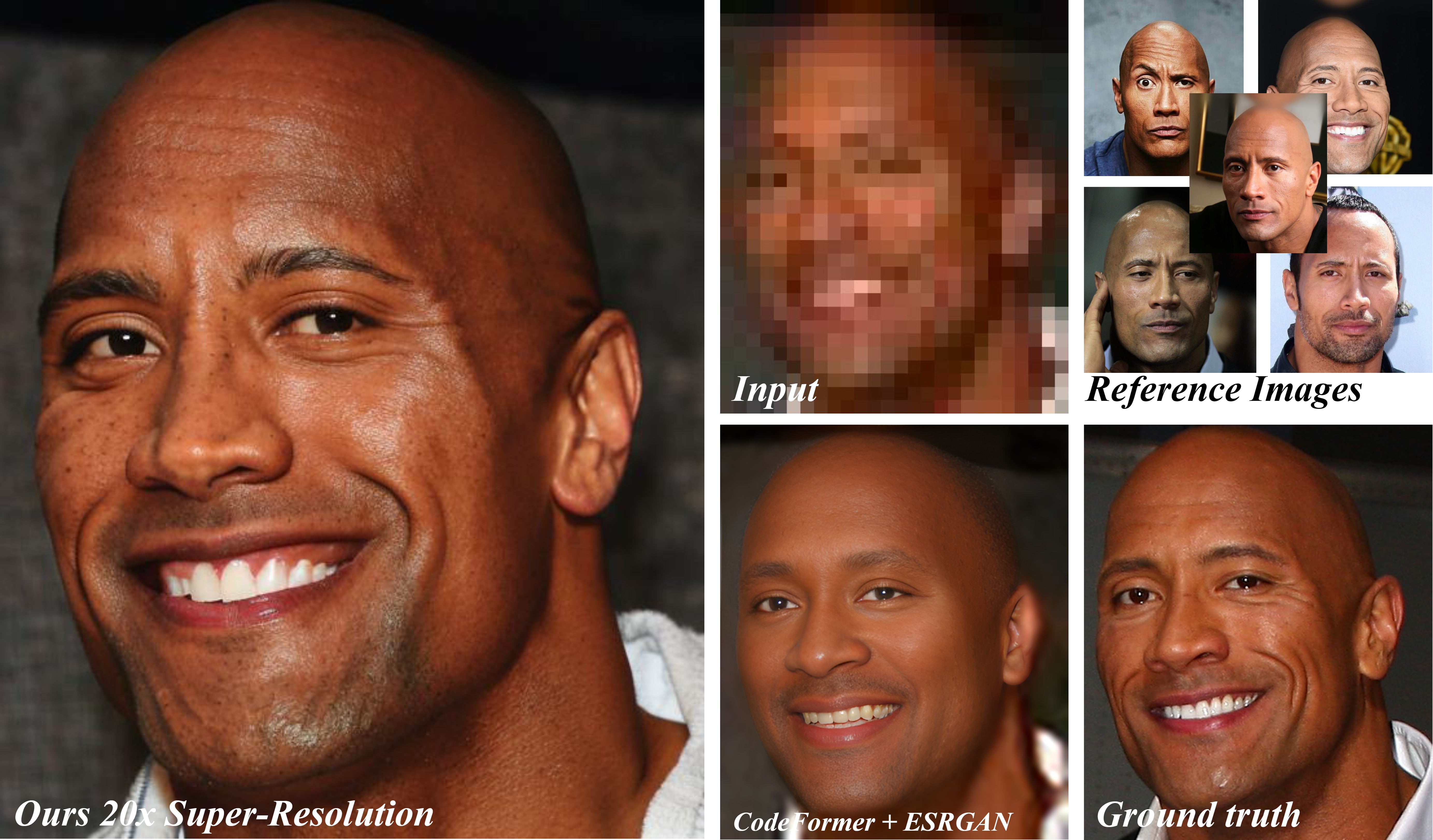}
    \captionof{figure}{20x Super-resolution of a low-quality image. Super-resolution for images larger than $512 \times 512$ using a tiling approach from \cite{stablesr}. Image edited from Vecteezy.com.}
	\label{fig:super-resolution}
	\vspace{-3mm}
\end{figure*}

\paragraph{Personalization}
We initially attempt to fine-tune with prior-preservation regularization \cite{dreambooth}, but find that it fails to properly capture the fine-grained identity details as well as diminishes the results from restoration due to modifying existing priors. This motives the need for preserving the priors completely, leaving the priors untouched. Therefore, we prefer to utilize adapter blocks, which do not modify the existing priors at all, retaining their rich abilities to restore and generate. In order to implement this, we employ text and image cross-attentions between the learnable text-embedding \cite{textual_inversion}, reference image features $F^i_{Ref}$ and intermediate restored image features $F^i$ of the layer $i$ , as used similarly in \cite{vico} and shown on \cref{fig:network_diagram} right. The reference image features $F^i_{Ref}$ are obtained from a frozen StableDiffusion \cite{stablediffusion}, in practice they are fed through part of the Base Model in same batch as the LQ image.  We refer to this as the \textit{Personalization Block} (see \cref{fig:network_diagram} right).
\vspace{-3mm}

\paragraph{Controlled Adaptation}
The simple addition of the personalization block however results in distorted outputs. This is due to the sudden additional data being added to the intermediate features of the Base Model from the personalization block. In order to avoid the personalization block from changing the outputs too much, a learnable vector $\gamma = \textbf{0}$ can be used to initialize the outputs from the adapter, as in \cite{segmentation_adapter}. To further control the effect of personalization we introduce seperate $\gamma$ for each personalization block applied at different resolution of PFStorer. Mathematically, each layer's output can be expressed as:

\begin{equation}
	\hat{F}^i = F^i + \gamma^i \odot \text{Personalization-Block}(F^i, F_{ref}^i),
\end{equation}
where $\text{Personalization-Block}$ is the adapter, consisting of cross-attentions as shown on right of \cref{fig:network_diagram}.
\vspace{-3mm}

\begin{figure*}
	\begin{center}
    	\setlength{\tabcolsep}{1pt}
        \begin{subfigure}{0.98\textwidth}
        \hspace{-0.30cm}
        	\begin{tabular}{*6c}
        		\includegraphics[width=0.167\textwidth]{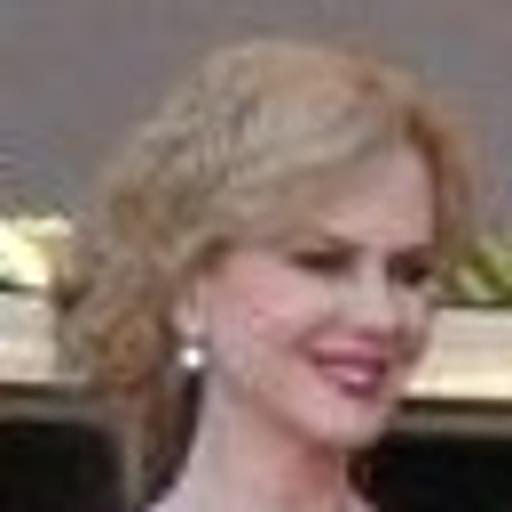} &
        		\includegraphics[width=0.167\textwidth]{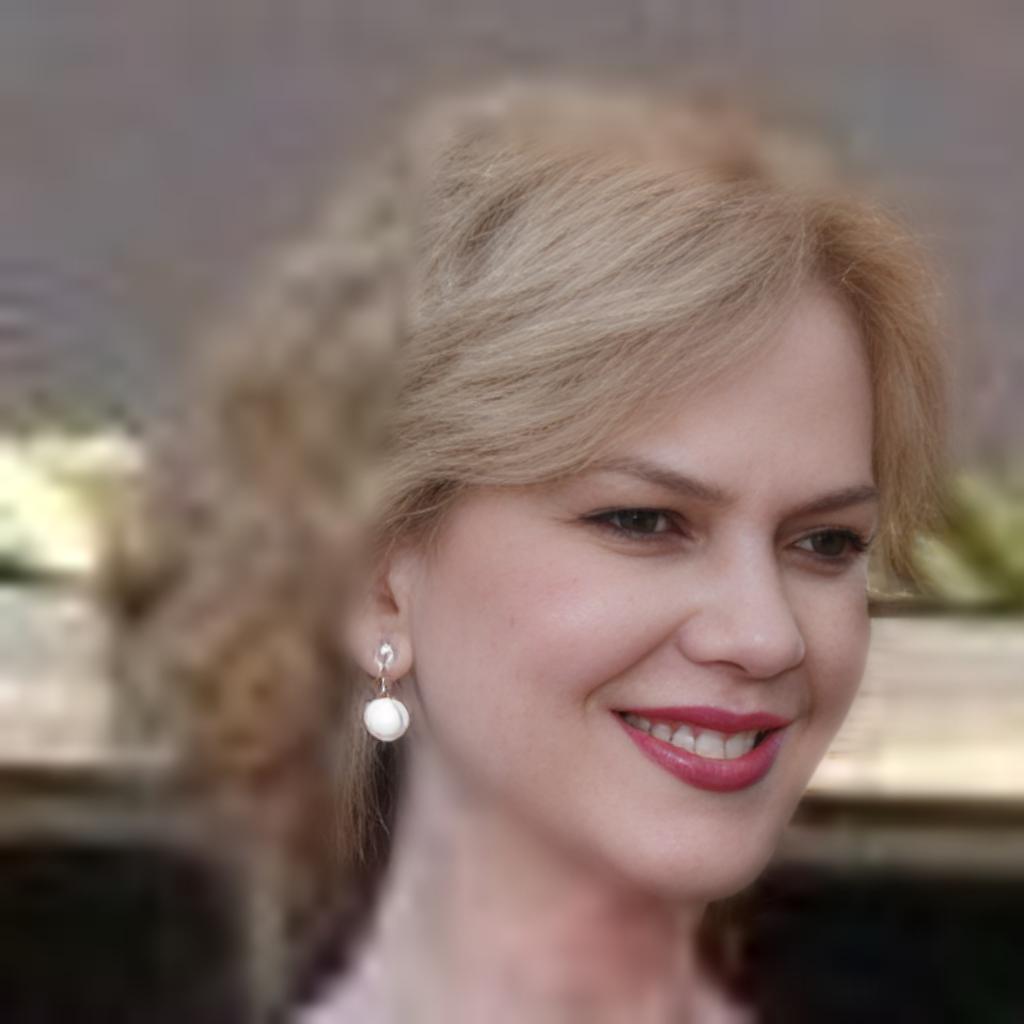} &
        		\includegraphics[width=0.167\textwidth]{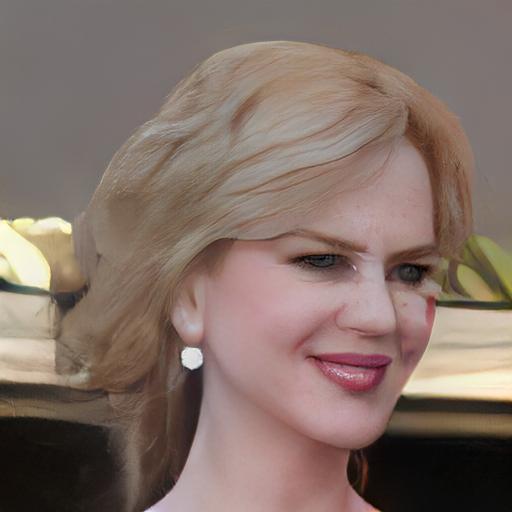} &
        		\includegraphics[width=0.167\textwidth]{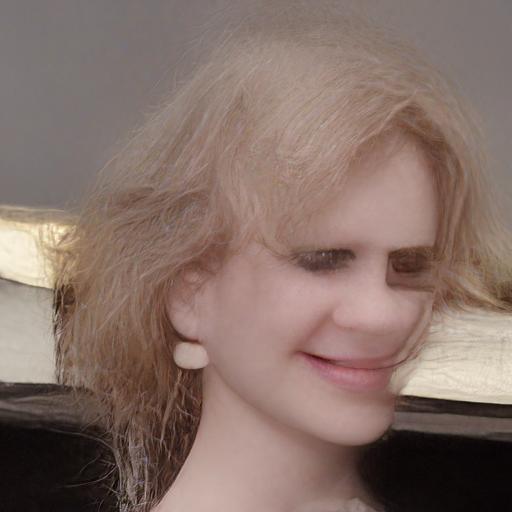} &
        		\includegraphics[width=0.167\textwidth]{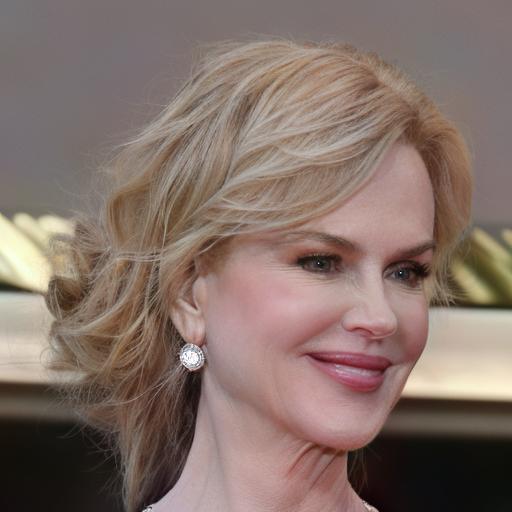} &
        		\includegraphics[width=0.167\textwidth]{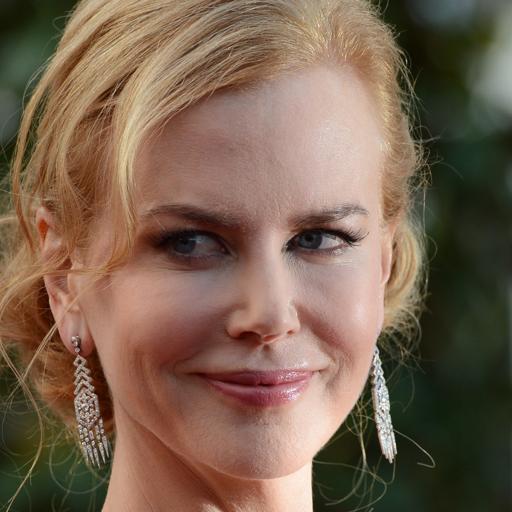}
        		\\
        		\includegraphics[width=0.167\textwidth]{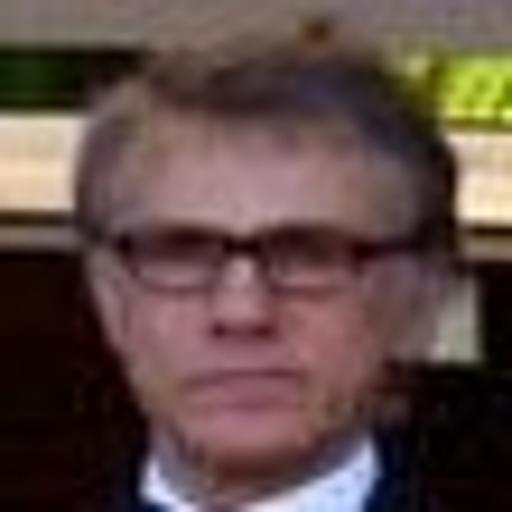} &
        		\includegraphics[width=0.167\textwidth]{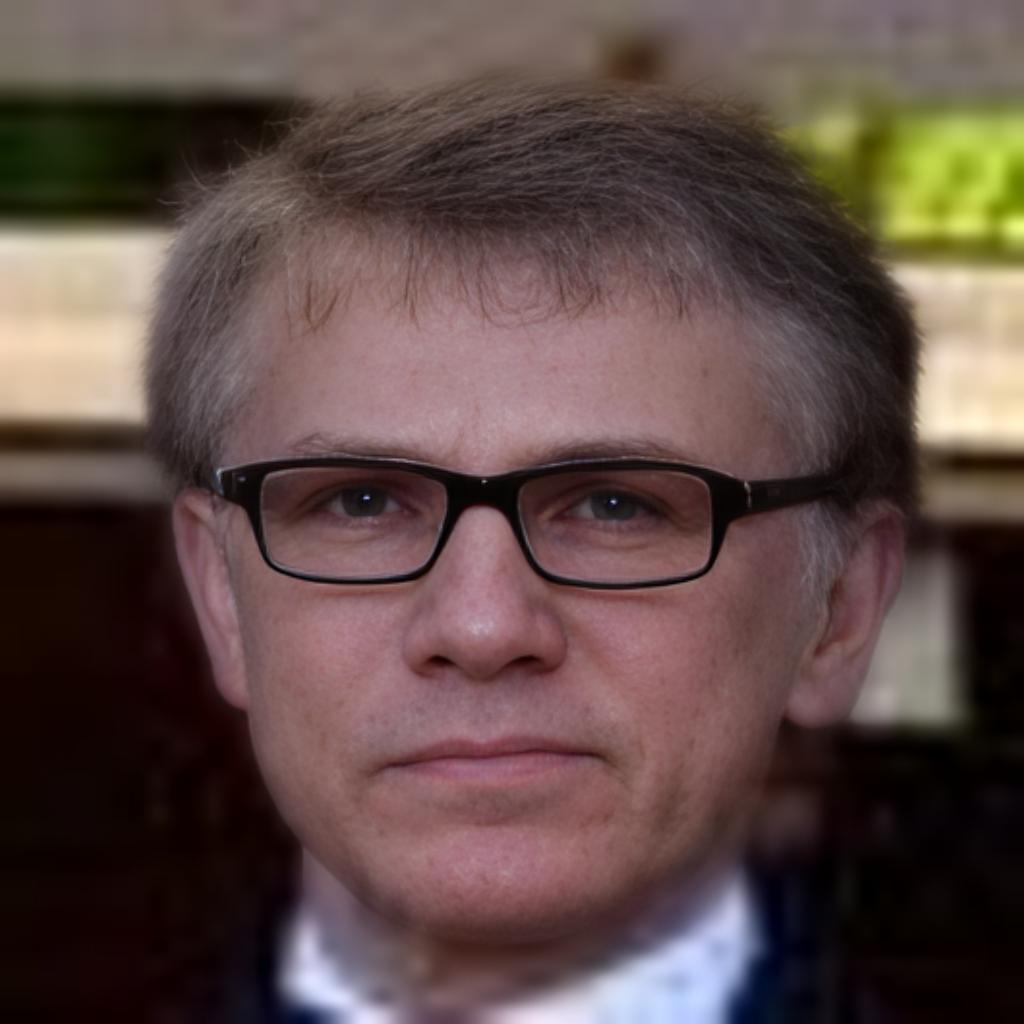} &
        		\includegraphics[width=0.167\textwidth]{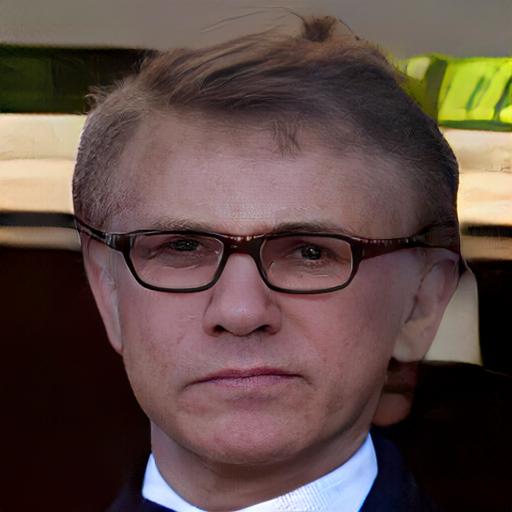} &
        		\includegraphics[width=0.167\textwidth]{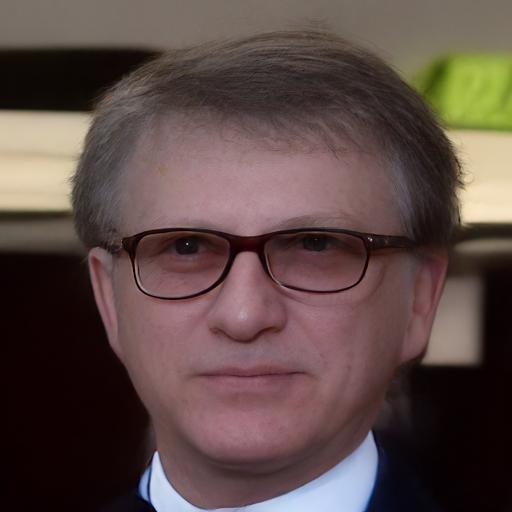} &
        		\includegraphics[width=0.167\textwidth]{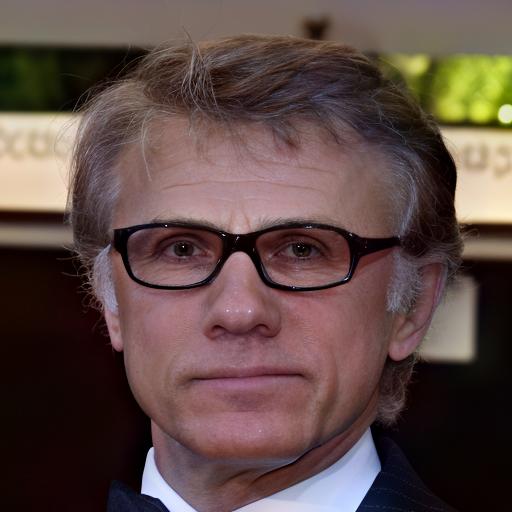} &
        		\includegraphics[width=0.167\textwidth]{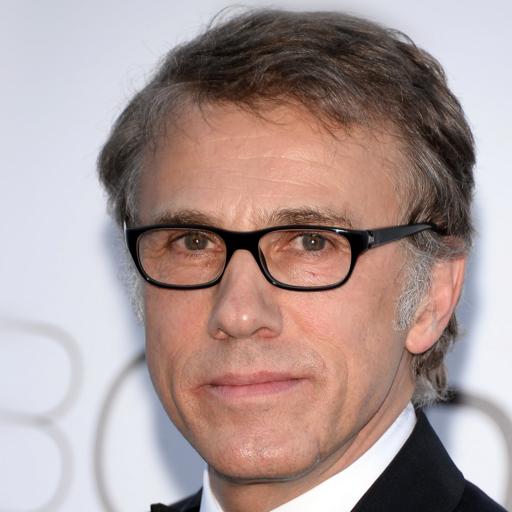}
        		\\
        		\footnotesize{Input} & \footnotesize{CodeFormer} & \footnotesize{DMDNet} & \footnotesize{DR2 + SPAR} & \footnotesize{Ours} & \footnotesize{Pseudo-GT} \\
        		\vspace{-1cm}
        	\end{tabular}
        \end{subfigure}%
	\end{center}
	\caption{
	Qualitative comparison with state-of-the-art restoration models on real-world images. Images from Wikimedia Commons.
	}
	\label{fig:celeb_ref_real}
	\vspace{-3mm}
\end{figure*}

\paragraph{Generative Regularization}
\label{sec:generative_regularization}
Compared to personalized generative models our personalized restoration model has one additional signal, the low-quality image $I_{LQ}$. It guides the general structure of the restoration output and it may contain some information from the identity depending on the severity of the degradation. During training, the additional input can make the task of outputting personalized restored images easier, but it can also introduce shortcuts for the model as the model can rely on information from the additional input. This leaking of identity information from the input can lead to the model not fully learning a representation of the identity during training, hence leading to poor performance on difficult unseen cases, \eg atmospheric turbulence. 

To mitigate the above issue, we propose a generative regularizer that encourages the model to learn a more robust identity representation. A regularizing loss

\begin{equation}
	\mathcal{L}_{Gen} = \mathbb{E}_{z, t, I_{LQ}, I_{Ref}, \eps} \| \eps - \eps_\theta(z_t, c, \varnothing, I_{Ref})\|_2.
\end{equation}
is added to the original training loss, where a null input $\varnothing$ is given as the conditioning LQ image. This forces the model to fully hallucinate the identity without any help from a conditioning image, encouraging a more robust representation of the identity. The final loss is then
\begin{equation}
	\mathcal{L} = \mathcal{L}_{Diff} + \lambda_{Gen} \mathcal{L}_{Gen} + \lambda_{Pers} \mathcal{L}_{Pers}
\end{equation}
where $\lambda_{Gen}$ controls the weight of the generative term and $\lambda_{Pers} \mathcal{L}_{Pers}$ regularizes the cross-attention maps for the learnable text embedding token, which enforces personalization \cite{vico} (see the supplementary for $\mathcal{L}_{Pers}$). The trainable parameters $\theta$ from $\eps_{\theta}$ consist of the personalization blocks and their accompanying vectors $\gamma^i$.

\subsection{Improving Face Restoration Diffusion Models}
\label{sec:finetune}
To integrate personalization into a restoration model, we first need a strong base restoration model. We train our model with the facial dataset FFHQ using the steps described below, which is initialized from the pre-trained StableSR \cite{stablesr}. We refer to the trained model as \textbf{Base Model}, as it has not been personalized to any specific person.
\vspace{-3mm}

\paragraph{Existing Priors}
Many recent face restoration methods have used generative priors \cite{gfp-gan, codeformer}. We go further, and start our training on face images with a restoration model pre-trained on generic natural images, namely StableSR \cite{stablesr}. As the model is not trained from scratch on a new task, the training time is decreased and the model is more robust.
\vspace{-3mm}

\paragraph{Alignment Free Approach}
Cropping and alignment is commonly used in face processing for standardizing input. However, delicate cropping and alignment using facial landmarks is prone to errors when face detection models fail. This is especially true in real world images. To avoid such approach we train our technique with a combination of random crops and resizing, following the training strategy of \cite{stablesr}. The random crops make the model more robust while also providing higher resolution inputs as details are not lost in the resizing operation.
\vspace{-3mm}

\paragraph{Synthetic Noise Generation}
In order to generate LQ images for training, most previous face restoration approaches have used a simple first-order degradation that may not encompass all noises present in real-world images. We use a second order noise model from \cite{stablesr}, ISP model from \cite{scunet} and add motion blur and median blur to better simulate real-world conditions. As noted in \cite{idm}, given a high-quality input, a restoration model should not lose details in the restoration process. We enforce this by directly feeding the high-quality input as is with a probability of $p_{HQ}$, which is set to a low value of 0.03 in all of our experiments.

%% file: sec/4_experiments.tex
\section{Experiments} 
\label{sec:experiments}

\begin{figure}
	\begin{center}
    	\setlength{\tabcolsep}{1pt}
        \begin{subfigure}{0.46\textwidth}
        \hspace{-0.2cm}
        	\begin{tabular}{*6c}
        		\includegraphics[width=0.167\textwidth]{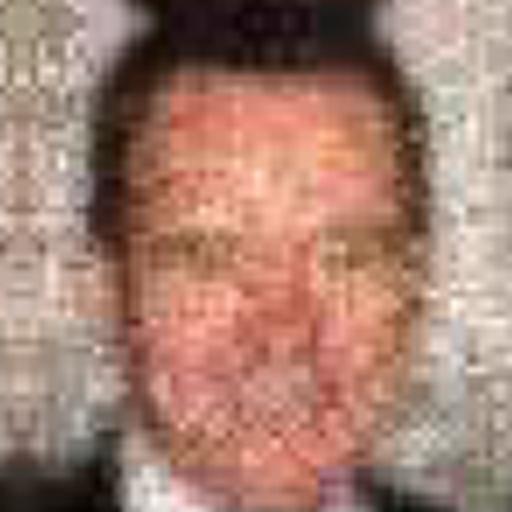} &
        		\includegraphics[width=0.167\textwidth]{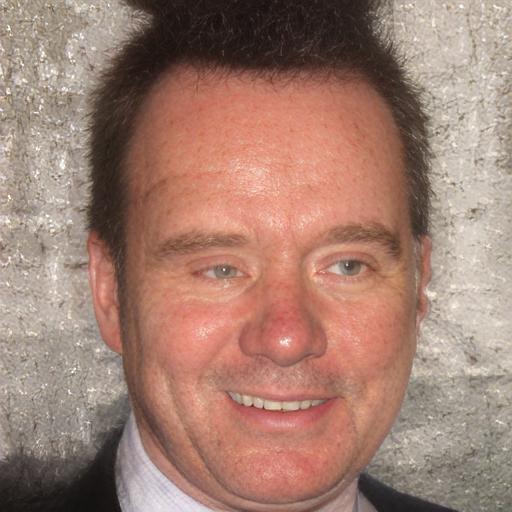} &
        		\includegraphics[width=0.167\textwidth]{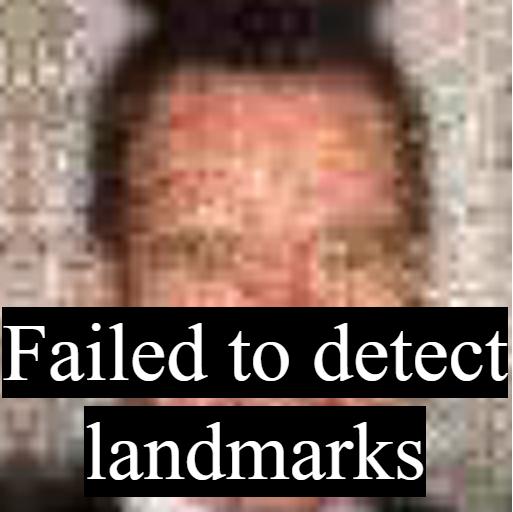} &
        		\includegraphics[width=0.167\textwidth]{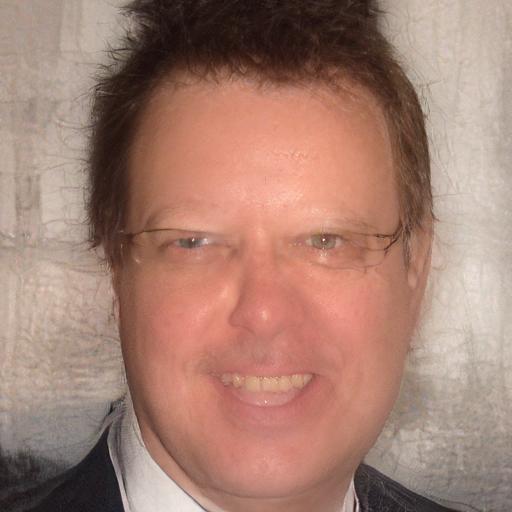} &
        		\includegraphics[width=0.167\textwidth]{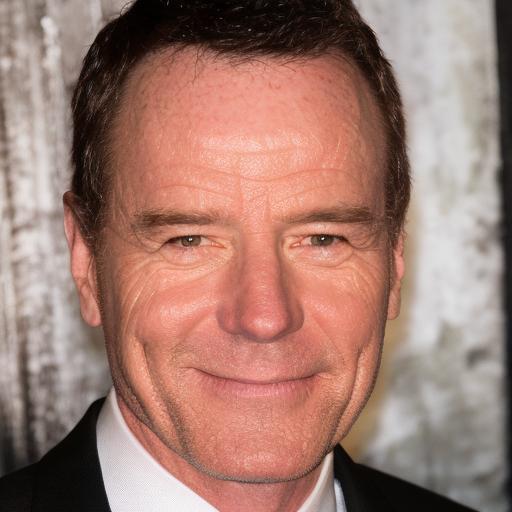} &
        		\includegraphics[width=0.167\textwidth]{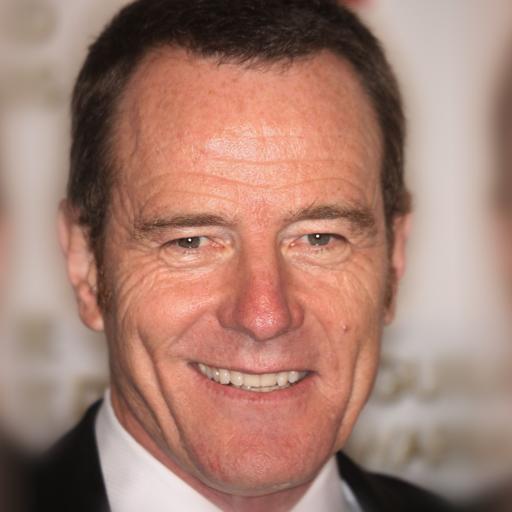}
        		\\
        		\includegraphics[width=0.167\textwidth]{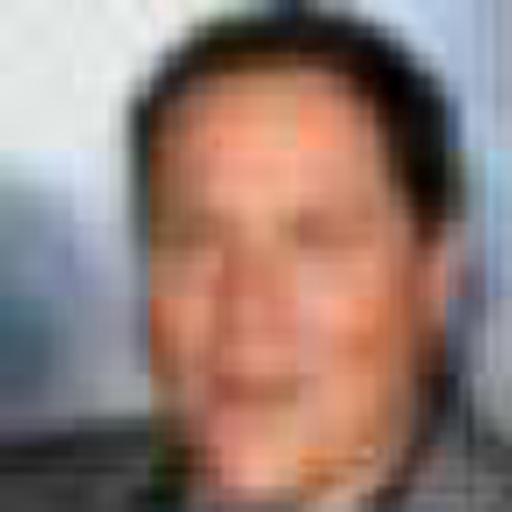} &
        		\includegraphics[width=0.167\textwidth]{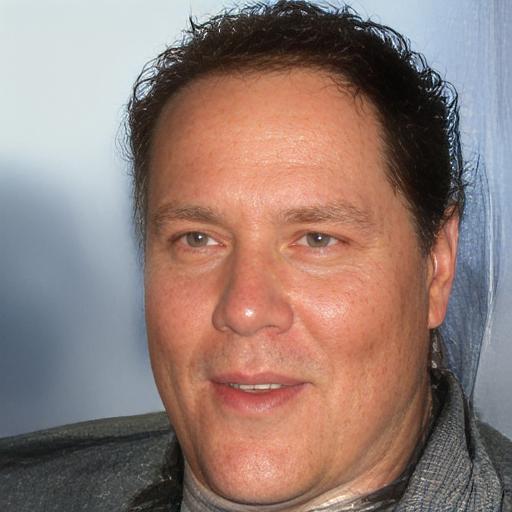} &
        		\includegraphics[width=0.167\textwidth]{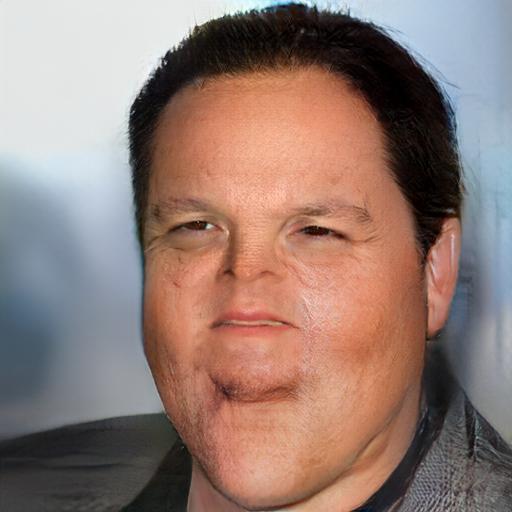} &
        		\includegraphics[width=0.167\textwidth]{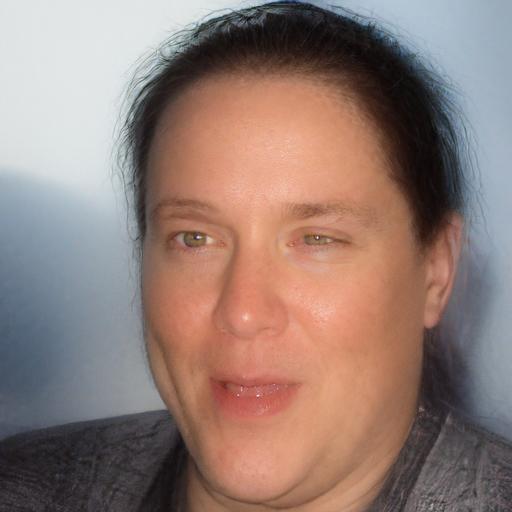} &
        		\includegraphics[width=0.167\textwidth]{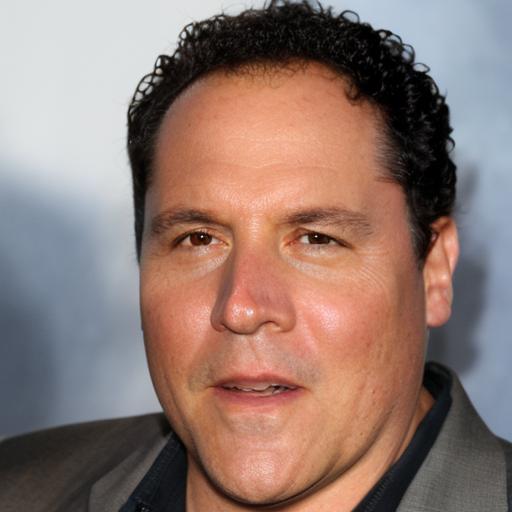} &
        		\includegraphics[width=0.167\textwidth]{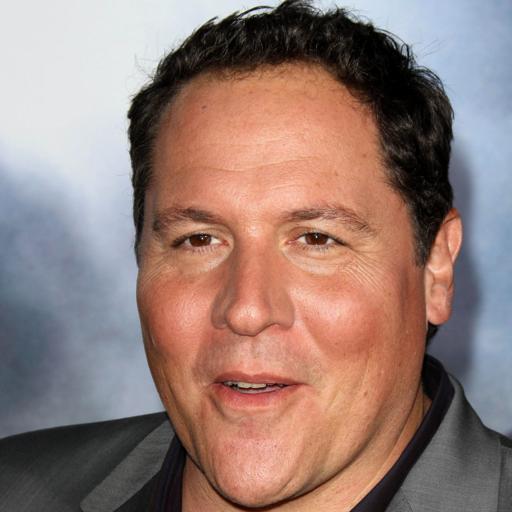}
        		\\
        		\footnotesize{Input} & \footnotesize{CodeFormer} & \footnotesize{DMDNet} & \footnotesize{DR2+SPAR} & \footnotesize{Ours} & \footnotesize{GT} \\
        		\vspace{-1cm}
        	\end{tabular}
        \end{subfigure}%
	\end{center}
	\caption{
	Qualitative comparison with state-of-the-art restoration models on Celeb-Ref dataset \cite{dmdnet} with heavy synthetic degradation. Best viewed zoomed in.
	}
	\label{fig:celeb_ref_heavy}
	\vspace{-3mm}
\end{figure}

\paragraph{Datasets}
For evaluation we use Celeb-Ref \cite{dmdnet} and real-world images collected from the internet. Due to the large computational cost of diffusion models we choose a small subsection of the original Celeb-Ref. For synthetic data evaluation that contains the ground truth, we randomly choose 20 identities with at least 10 images each, for a total of 342 images. For each identity we reserve 5 images for the personalization, leaving a total 242 images for the testing. We further use two variations, \textit{light} and \textit{heavy} degradation sets, see the supplementary for details. For real-world data we again randomly choose 20 identities from Celeb-Ref, reverse search the identities using LAION-5B-KNN \cite{laion-knn} and collect one image for each identity from online. We focus on high-quality images, where the subject is far away and/or out of focus and/or with poor illumination to best simulate real-world applications.
\vspace{-3mm}

\begin{figure*}
	\begin{center}
    	\setlength{\tabcolsep}{1pt}
        \begin{subfigure}{0.98\textwidth}
        \hspace{-0.3cm}
        	\begin{tabular}{*6c}
        		\includegraphics[width=0.167\textwidth]{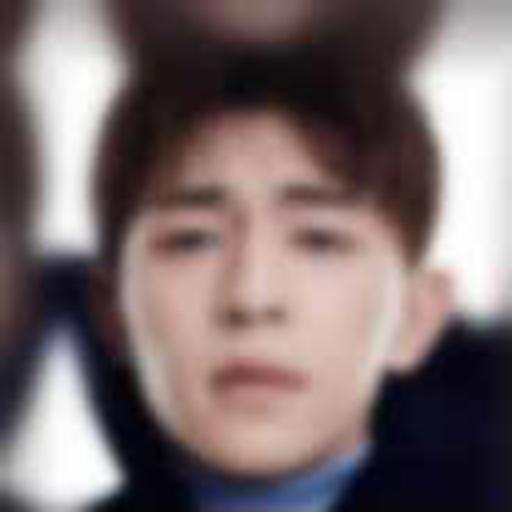} &
        		\includegraphics[width=0.167\textwidth]{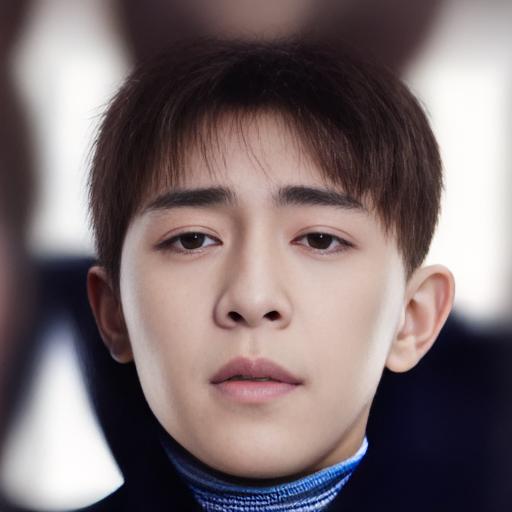} &
        		\includegraphics[width=0.167\textwidth]{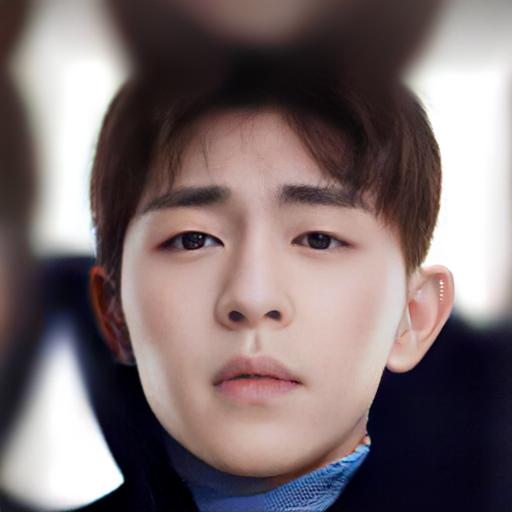} &
        		\includegraphics[width=0.167\textwidth]{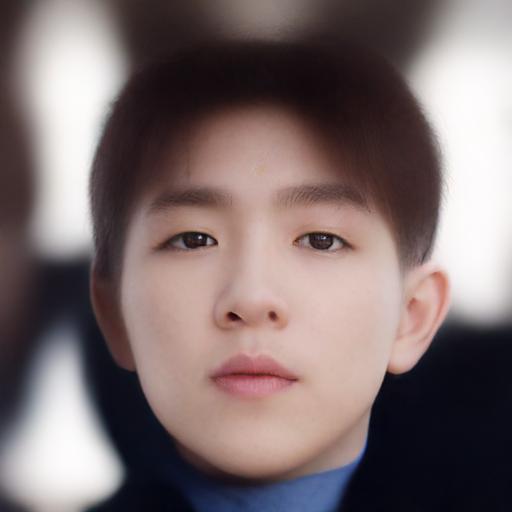} &
        		\includegraphics[width=0.167\textwidth]{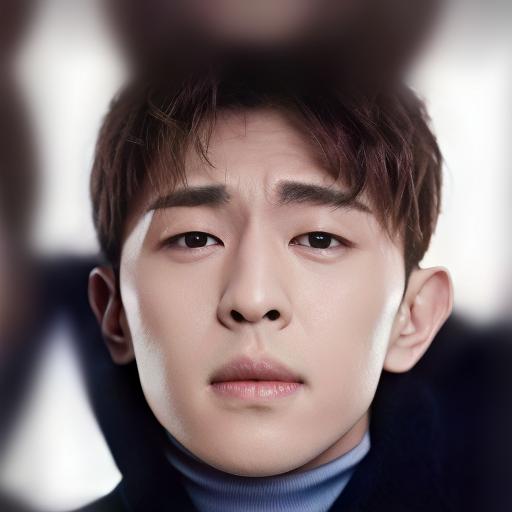} &
        		\includegraphics[width=0.167\textwidth]{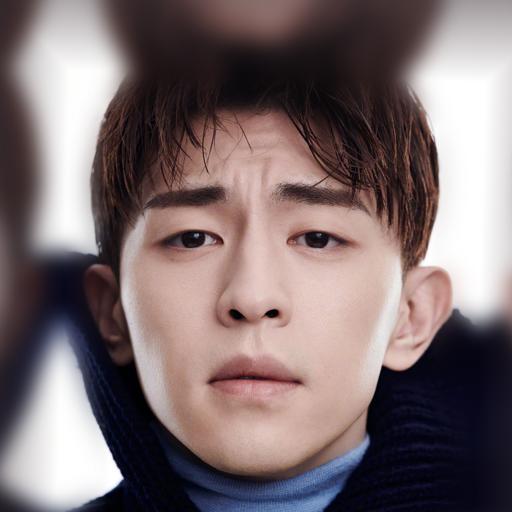}
        		\\
        		\includegraphics[width=0.167\textwidth]{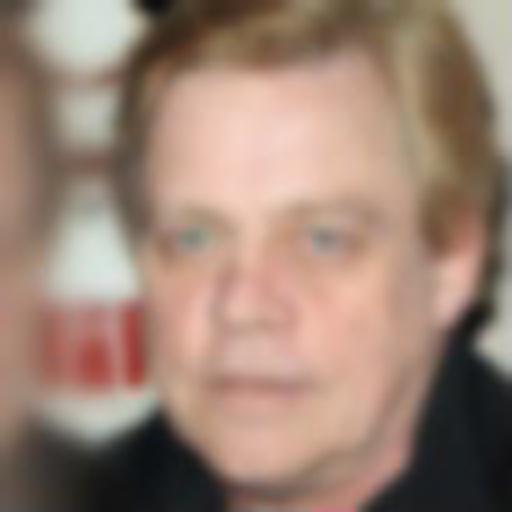} &
        		\includegraphics[width=0.167\textwidth]{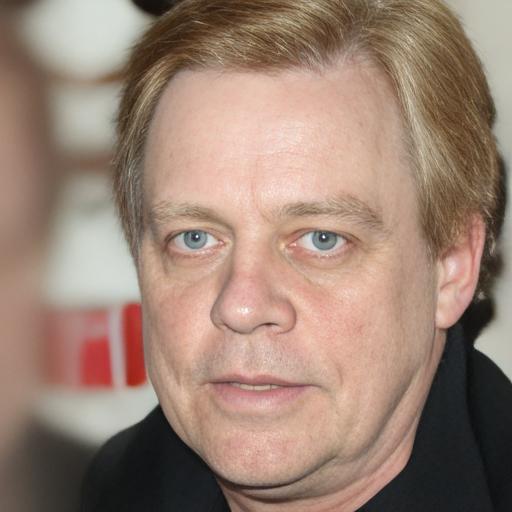} &
        		\includegraphics[width=0.167\textwidth]{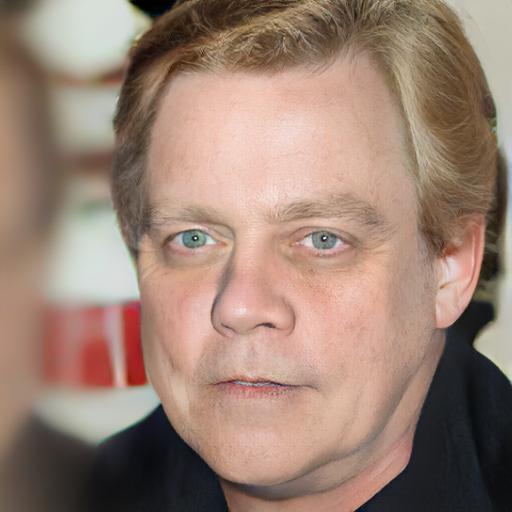} &
        		\includegraphics[width=0.167\textwidth]{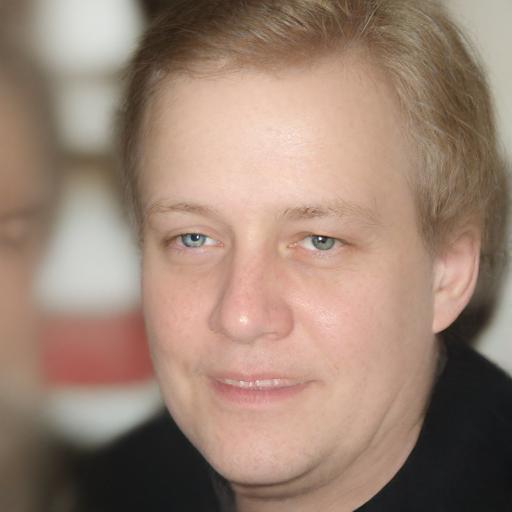} &
        		\includegraphics[width=0.167\textwidth]{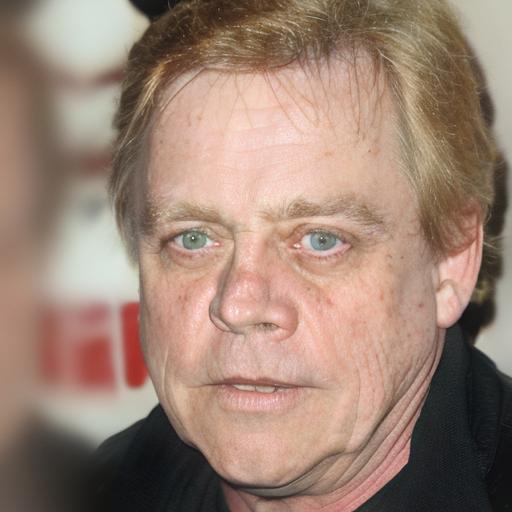} &
        		\includegraphics[width=0.167\textwidth]{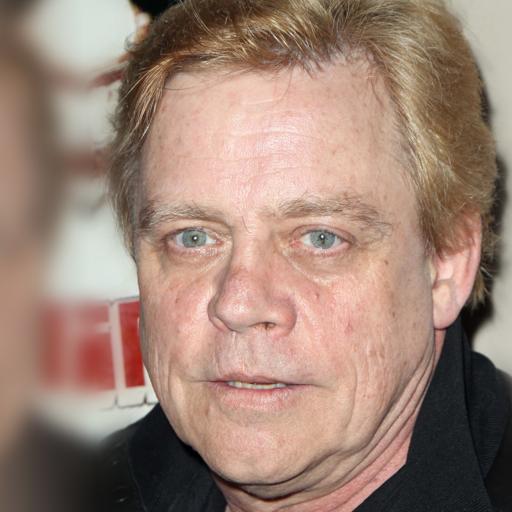}
        		\\
        		\footnotesize{Input} & \footnotesize{CodeFormer} & \footnotesize{DMDNet} & \footnotesize{DR2 + SPAR} & \footnotesize{Ours} & \footnotesize{GT} \\
        		\vspace{-1cm}
        	\end{tabular}
        \end{subfigure}%
	\end{center}
	\caption{
	Qualitative comparison with state-of-the-art restoration models on Celeb-Ref dataset \cite{dmdnet} with light synthetic degradation.
	}
	\label{fig:celeb_ref_light}
	\vspace{-3mm}
\end{figure*}

\paragraph{Baselines}
CodeFormer \cite{codeformer} is state-of-the-art technique for face restoration and it uses a codebook. DR2 \cite{dr2} is based on a diffusion model and is meant for extreme degradations. For DR2, we use the provided SPAR enhancer and empirically find the optimal hyperparameters. DMDNet \cite{dmdnet} is state-of-the-art method for reference-based face restoration, for which we use the same set of 5 reference images as for the proposed method.
\vspace{-3mm}

\paragraph{Evaluation Metrics}
For quantitative evaluation we use PSNR, SSIM, LPIPS \cite{lpips}, MUSIQ (KonIQ) \cite{musiq}, LMSE (Landmark MSE) \cite{farl}, and ID (cosine similarity with ArcFace \cite{arcface}) as metrics.
\vspace{-3mm}

\paragraph{Settings}
For methods that use reference images, 5 images are randomly sampled. For PFStorer, the personalization fine-tuning is done for 500 iterations, which corresponds to 10 minutes on a single A100. For all of our experiments we set the same settings, hyperparameters and a single seed. For detailed experimental settings see the supplementary material.

\subsection{Comparisons}

\begin{figure}
	\begin{center}
    	\setlength{\tabcolsep}{1pt}
        \begin{subfigure}{0.5\textwidth}
        \hspace{-0.2cm}
        	\begin{tabular}{*2c}
        		\includegraphics[width=0.5\textwidth]{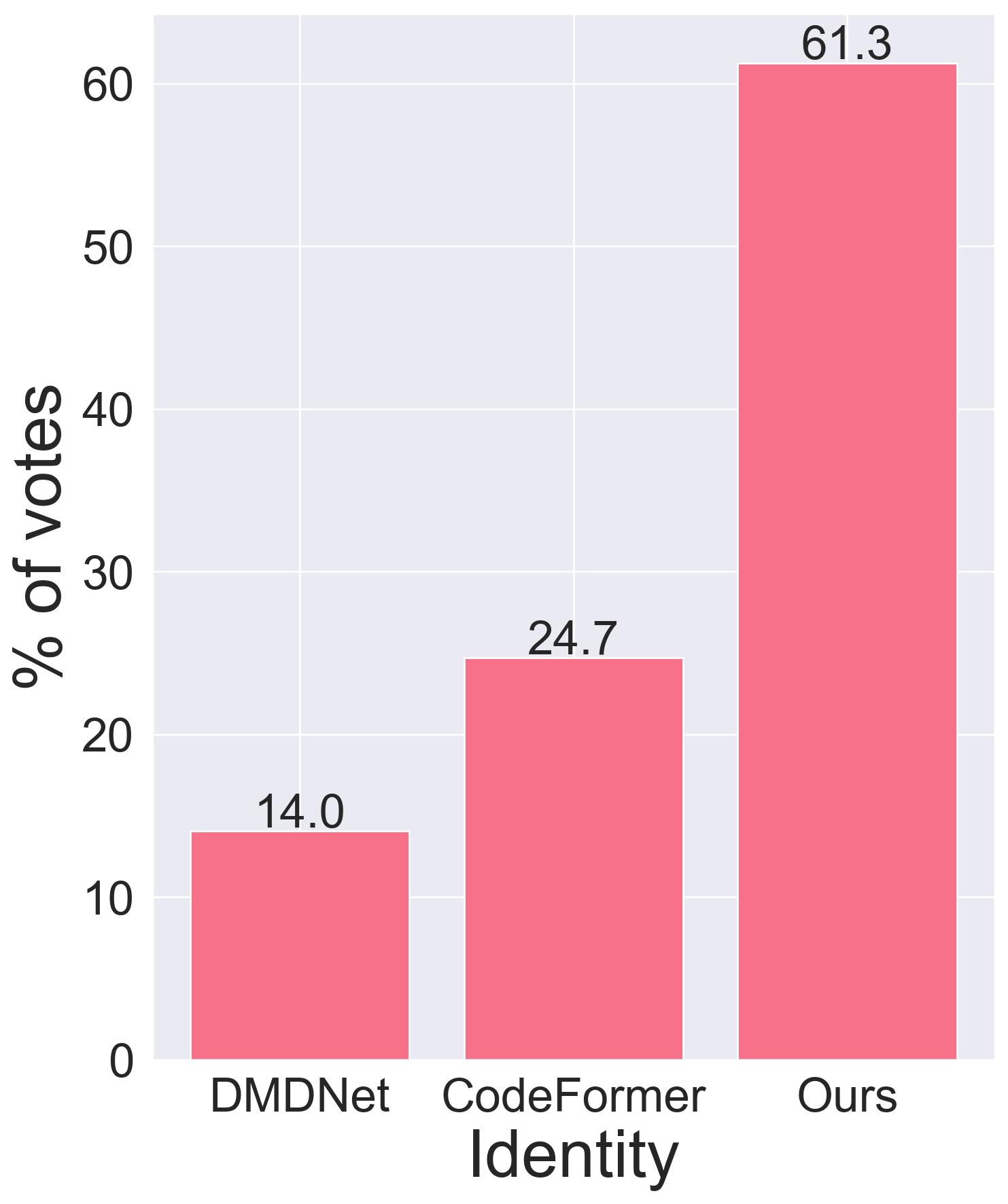} &
        		\includegraphics[width=0.5\textwidth]{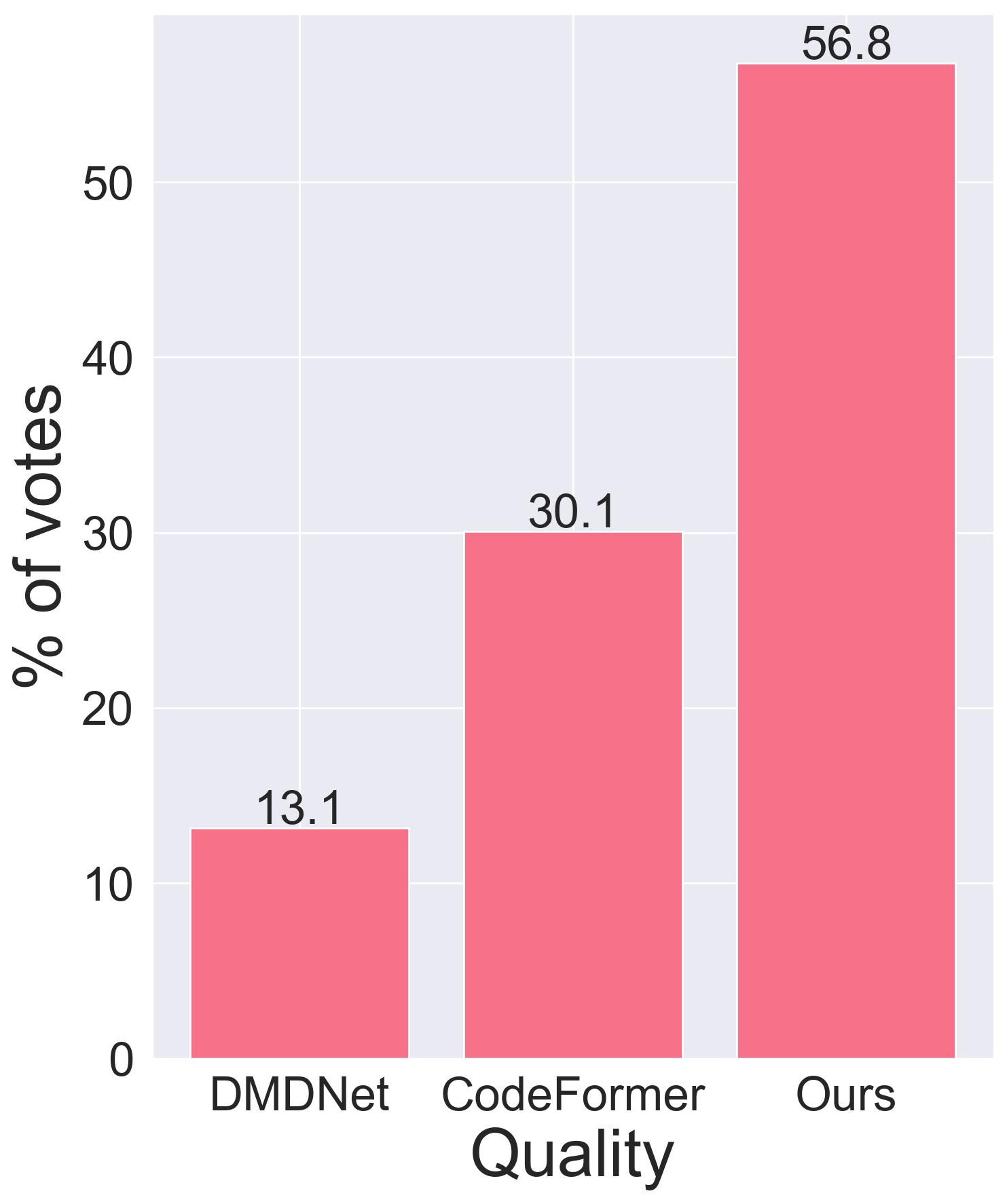}
        		\\
        		\vspace{-1cm}
        	\end{tabular}
        \end{subfigure}%
	\end{center}
	\caption{
	User study results.
	}
	\label{fig:user_study}
	\vspace{-3mm}
\end{figure}

\paragraph{Qualitative}
To evaluate the effectiveness of the proposed method we show visual results in \cref{fig:super-resolution,fig:celeb_ref_real,fig:celeb_ref_heavy,fig:celeb_ref_light}, for real-world, low-quality images collected from real-world, corrupted with heavy and light degradations. For the real-world sample we provide a pseudo-GT that can be used to compare with the identity. It can be observed from \cref{fig:celeb_ref_real} that the baseline methods fail in preserving the identity and producing a high-quality image. Despite the difficult case on first row \cref{fig:celeb_ref_real}, where the head pose is atypical, the proposed method is able to restore the image faithfully, thanks to the learned representation of the identity. \Cref{fig:celeb_ref_heavy} shows examples with heavy synthetic degradation. Even under heavy degradation the proposed method is able to restore the image faithfully, while other methods struggle with retaining the identity and outputting a realistic image. Under light degradation in \cref{fig:celeb_ref_light}, CodeFormer is able to output a high-quality image while mostly retaining the identity. Our method is able to retain even small details such as the wrinkles and skin texture.
\vspace{-3mm}

\paragraph{Quantitative}
Quantitative results on the heavily degraded images can be seen from \cref{tab:quantitative_heavy}. The pixel-wise metrics PSNR and SSIM as well as the perceptual metric LPIPS have relatively similar values across the best performing methods, with slight differences. Notably, the big difference is in the ID metric, where the proposed method obtains a similarity of 57.18\%, almost 20 percentage points higher than the next best performing method. This result showcases the benefit of personalization for retaining identity features. Another major improvement can be seen in the LMSE with almost half the error compared to CodeFormer. This is due to the combination of a strong base model and personalization. See supplementary for the real-world and lightly degraded samples.
\vspace{-3mm}

\begin{table}[h]
\caption{Quantitative results for images with heavy degradation. \color{red} Red \color{black} indicates the best and \color{blue} blue \color{black} indicates the second best. Ref indicates whether the model uses reference images}
\resizebox{0.48\textwidth}{!}{%
\begin{tabular}{c|c|cc|cc|c|c} 
 \hline
 Methods & Ref & PSNR $\uparrow$ & SSIM $\uparrow$ & LPIPS $\downarrow$ & MUSIQ $\uparrow$ & LMSE $\downarrow$ & ID $\uparrow$ \\ 
 \hline
 Input & & 22.56 & \color{red} 0.719 & 0.615 & 58.83 & 80.98 & 21.85 \\
 \hline
 DMDNet \cite{dmdnet} & \checkmark & \color{red} 22.64 & 0.684 & 0.491 & 47.17 & 89.26 & 29.51 \\
 DR2 + SPAR \cite{dr2} & & 22.17 & \color{blue} 0.701 & 0.449 & 47.36 & 40.82 & 30.01 \\
 CodeFormer \cite{codeformer} & & 22.26 & 0.642 & \color{blue} 0.422 & \color{blue} 60.92 & \color{blue} 33.34 & \color{blue} 38.33 \\
 \hline
 PFStorer (Ours) & \checkmark & \color{blue} 22.62 & 0.679 & \color{red} 0.414 & \color{red} 64.04 & \color{red} 18.37 & \color{red} 57.18 \\
 \hline
 GT & & $\infty$ & 1 & 0 & 62.37 & 0 & 100 \\
\end{tabular}}
\label{tab:quantitative_heavy}
\vspace{-3mm}
\end{table}

\paragraph{User Study}
As the quantitative metrics are not fully able to capture the nuances of human preferred perceptual quality, a user study is conducted. We use all three partitions of the data. We randomly pick 100 images. To attain statistical significance we recruit 40 users, following \cite{user_study}. With two questions we have a total of 8000 answers from users. We compare our method to only CodeFormer and DMDNet, as DR2 often produces low-quality images. We ask users to choose between the best image in terms of quality and identity with respect to a reference image.

The results are shown in \cref{fig:user_study}. Our method obtains the highest number of votes in both perceived identity and quality. Our method is especially good in capturing the identity, gaining 36.6 percentage points over the next best method, CodeFormer. This result resonates with both the qualitative results and quantitative metrics.

\begin{figure}
	\begin{center}
    	\setlength{\tabcolsep}{1pt}
        \begin{subfigure}{0.47\textwidth}
        \hspace{-0.2cm}
        	\begin{tabular}{*5c}
        		\includegraphics[width=0.2\textwidth]{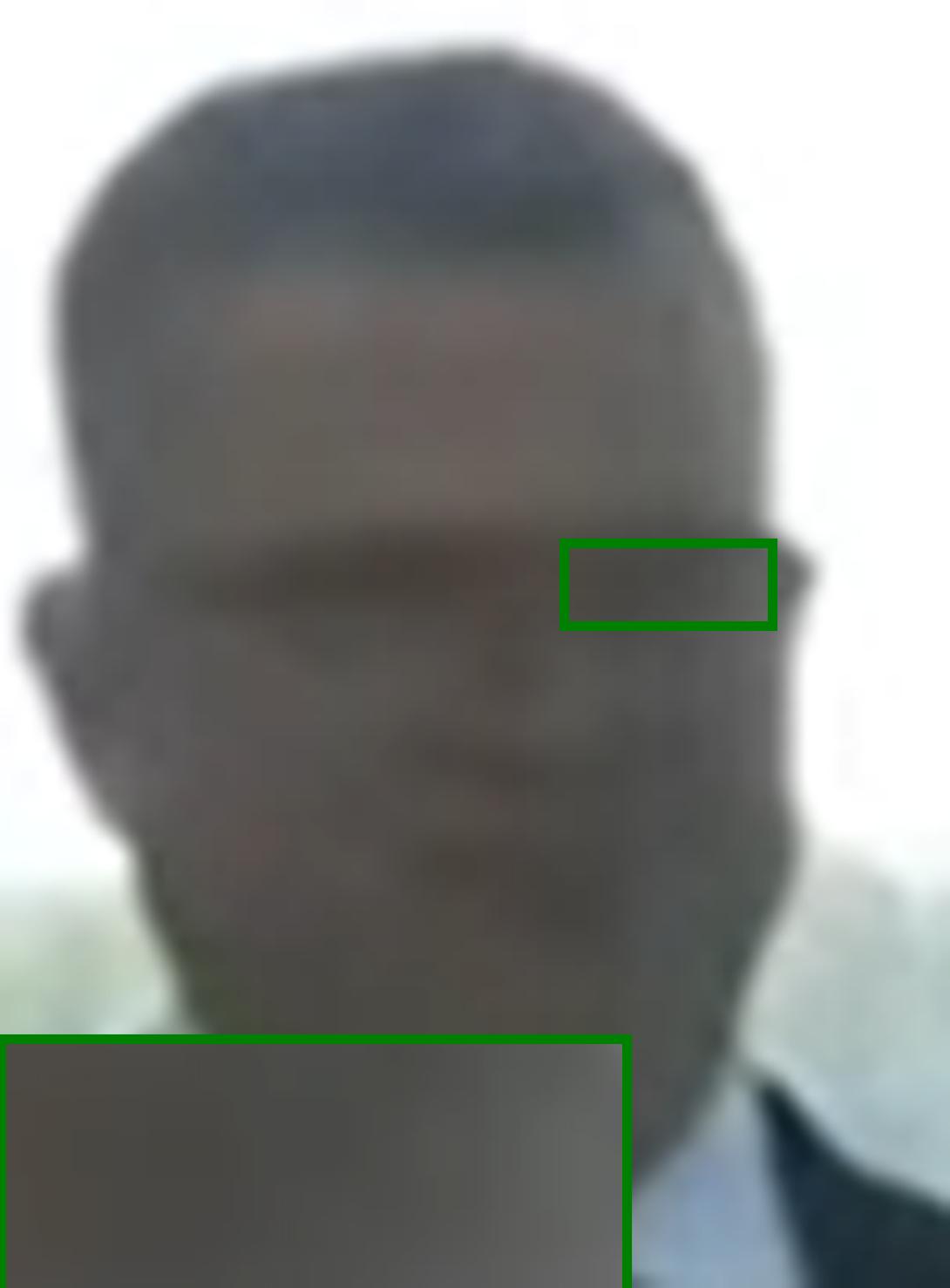} &
        		\includegraphics[width=0.2\textwidth]{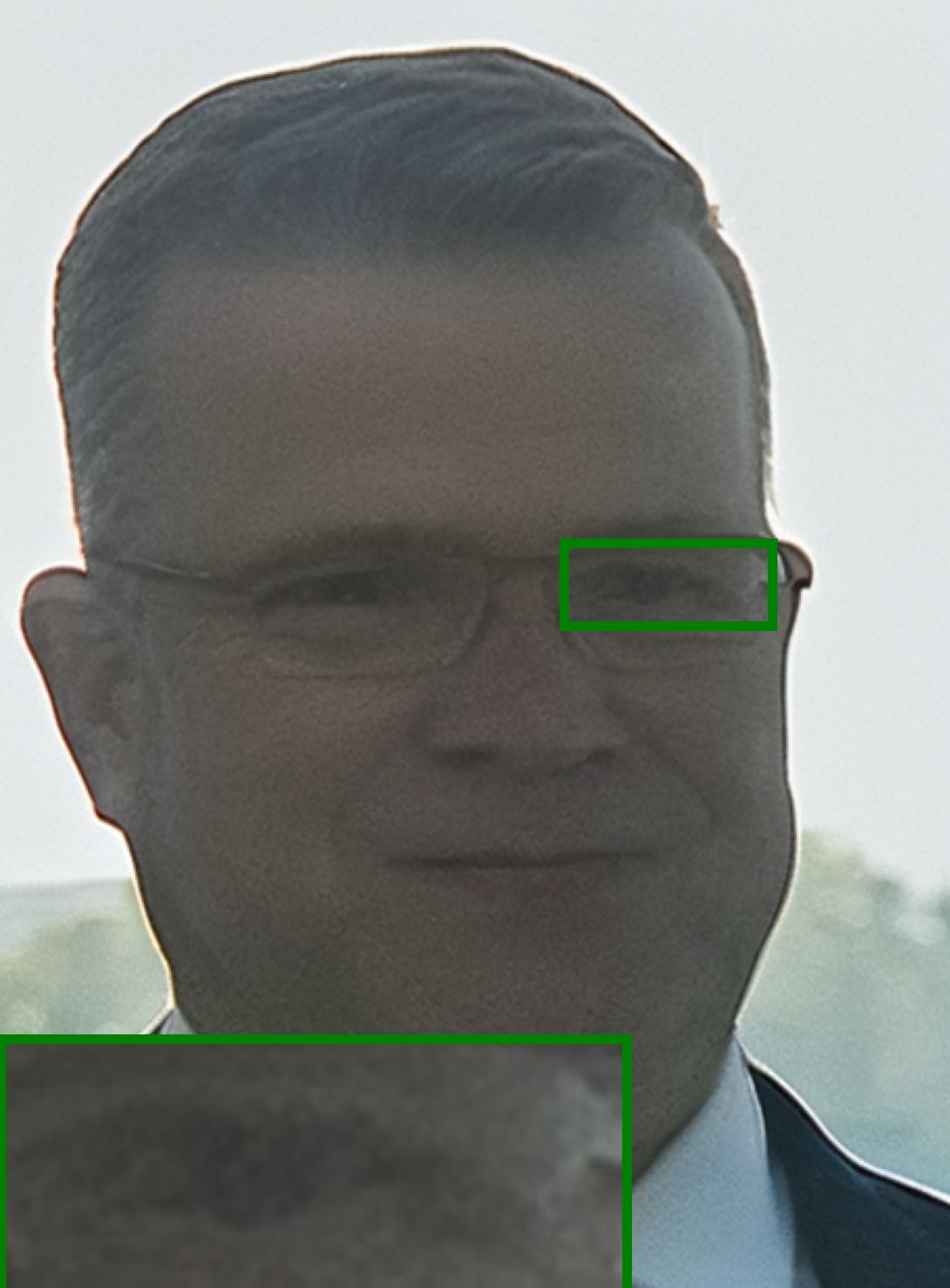} &
        		\includegraphics[width=0.2\textwidth]{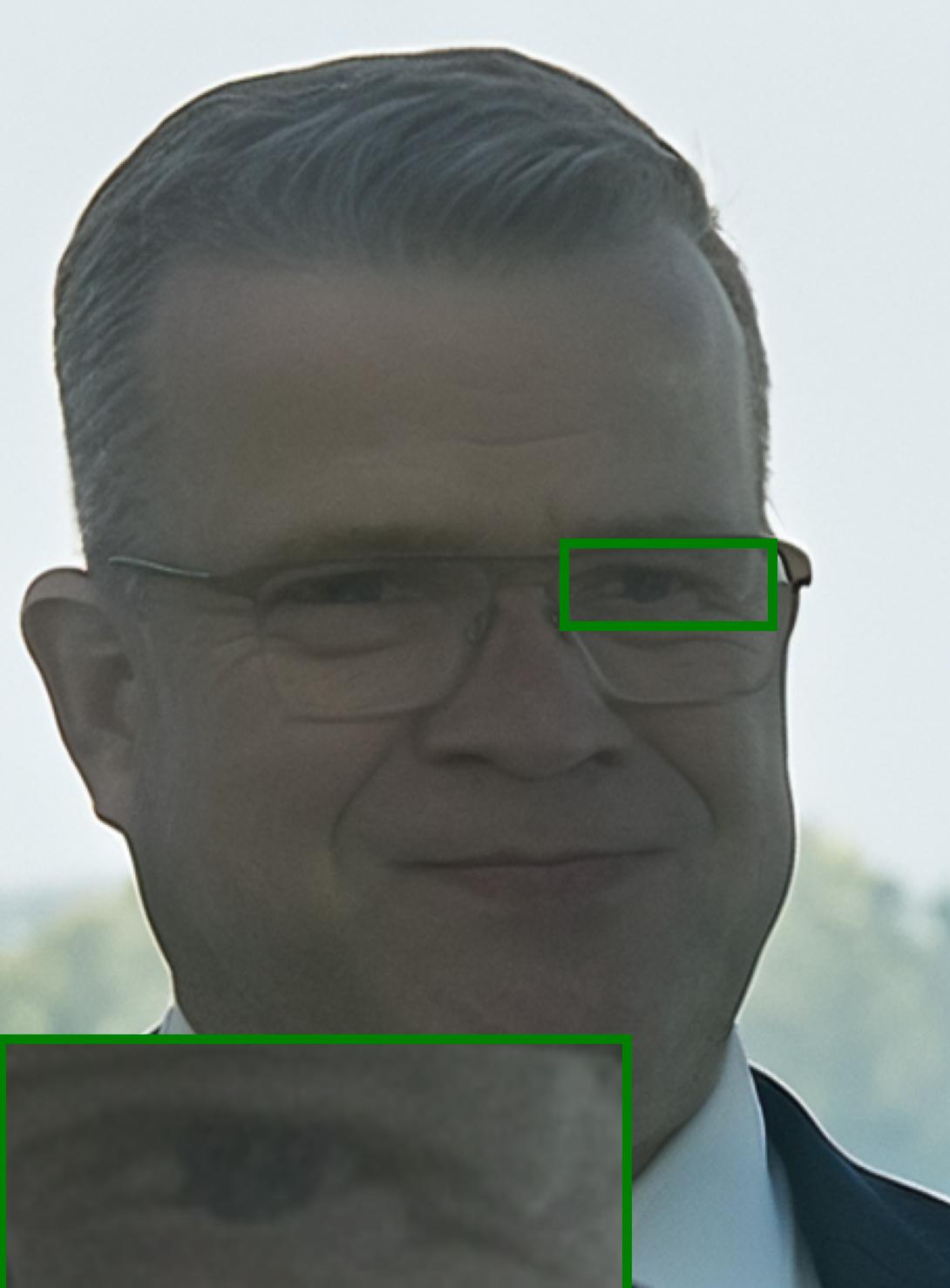} &
        		\includegraphics[width=0.2\textwidth]{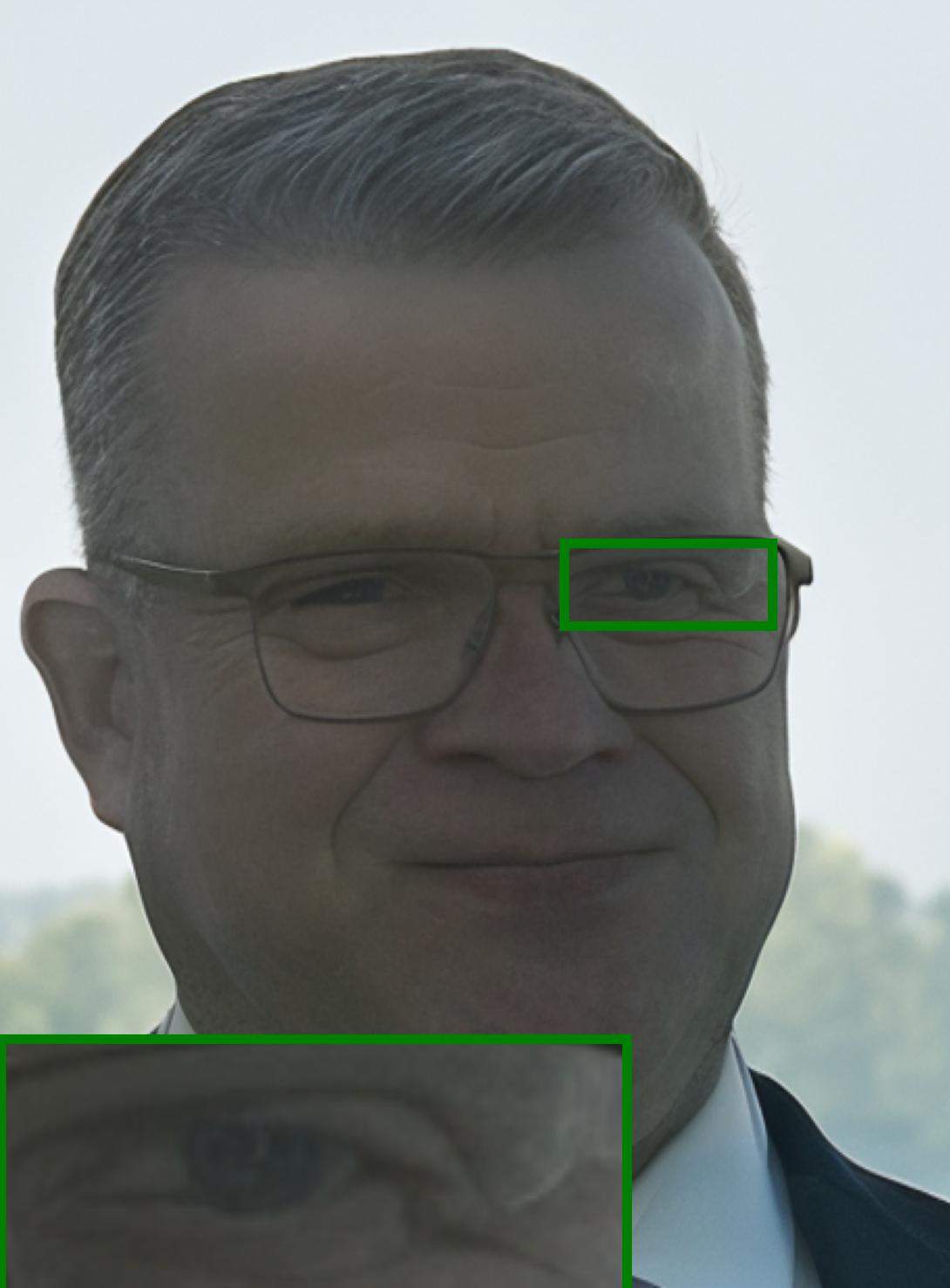} &
        		\includegraphics[width=0.2\textwidth]{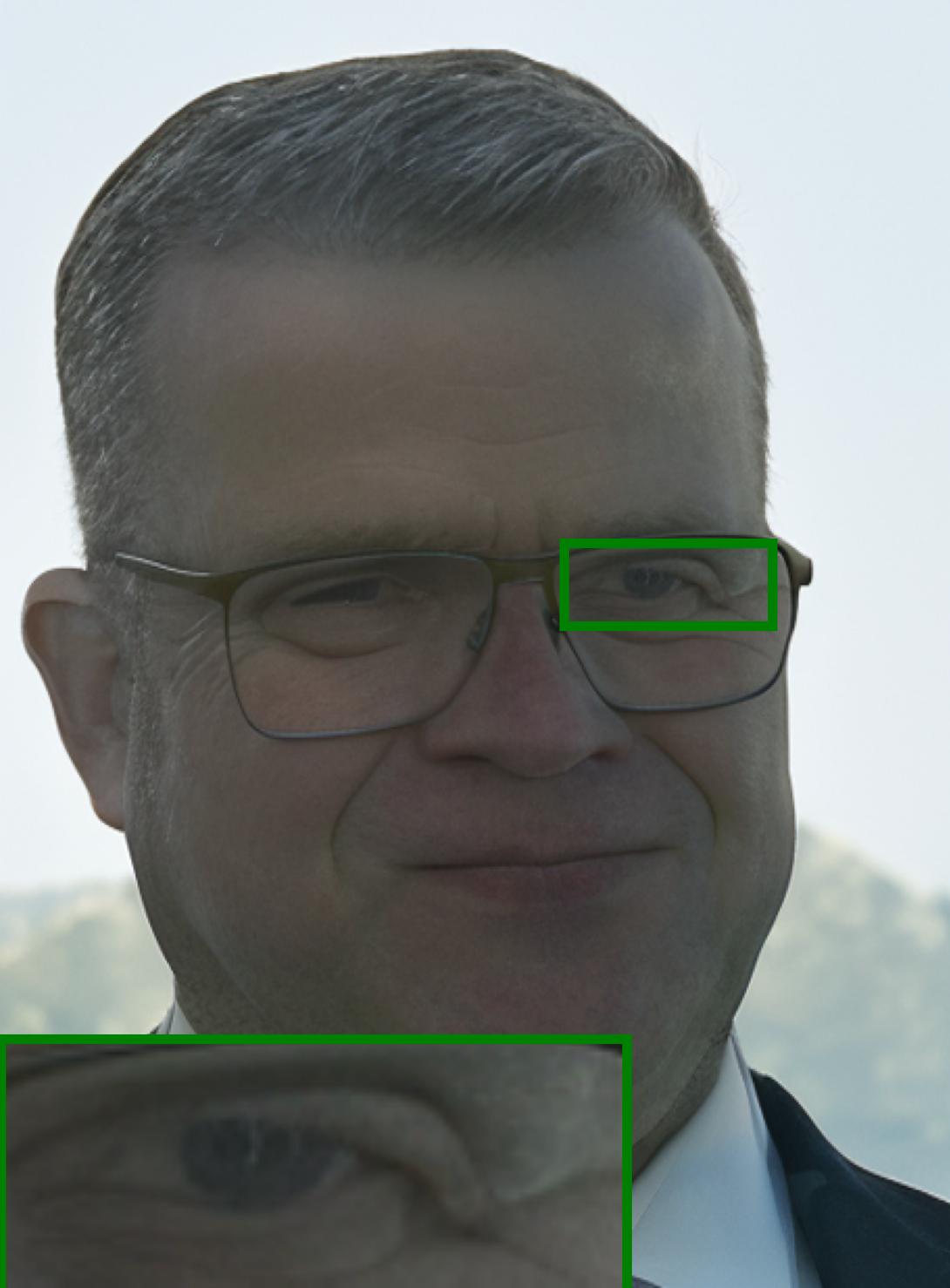}
        		\\
        		\includegraphics[width=0.2\textwidth]{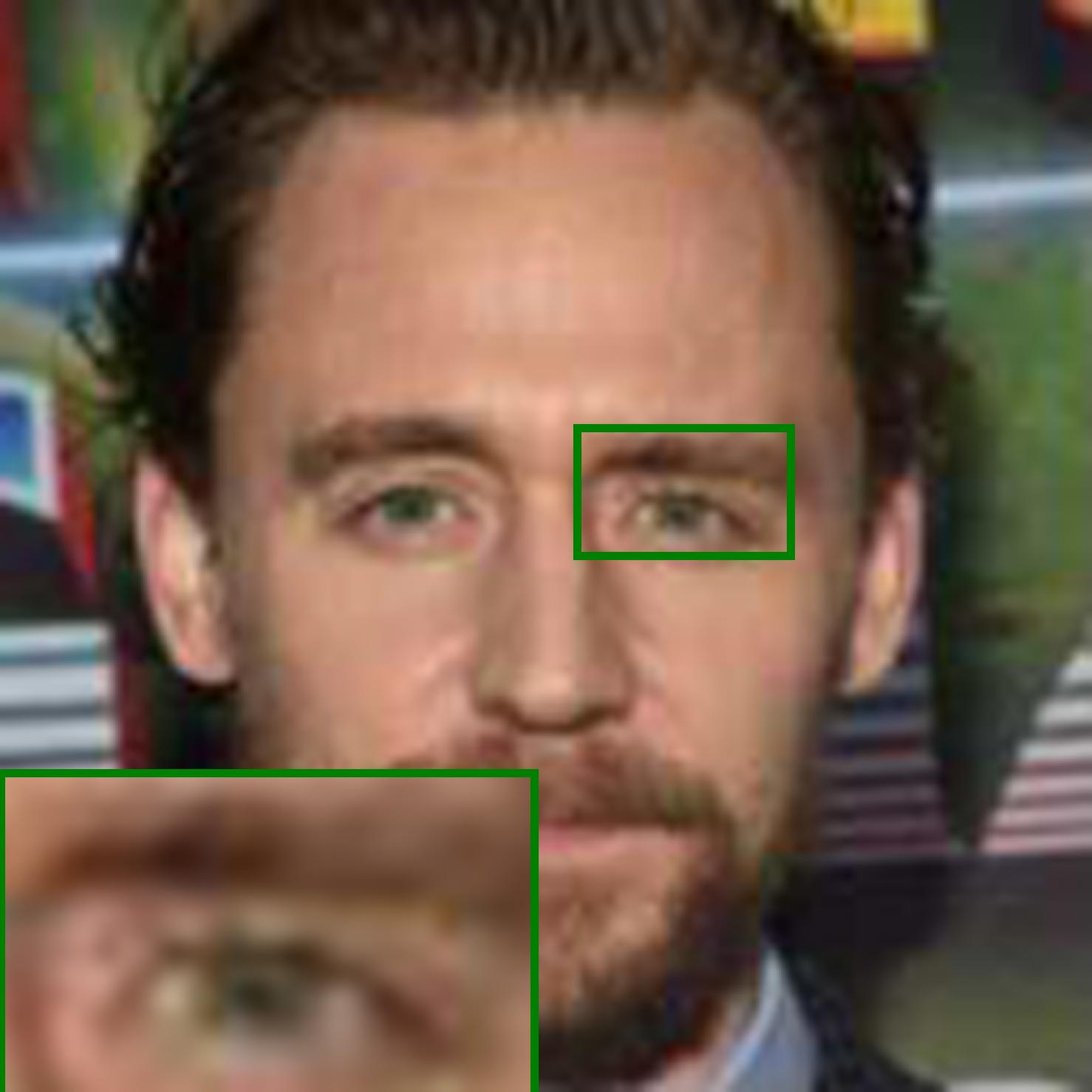} &
        		\includegraphics[width=0.2\textwidth]{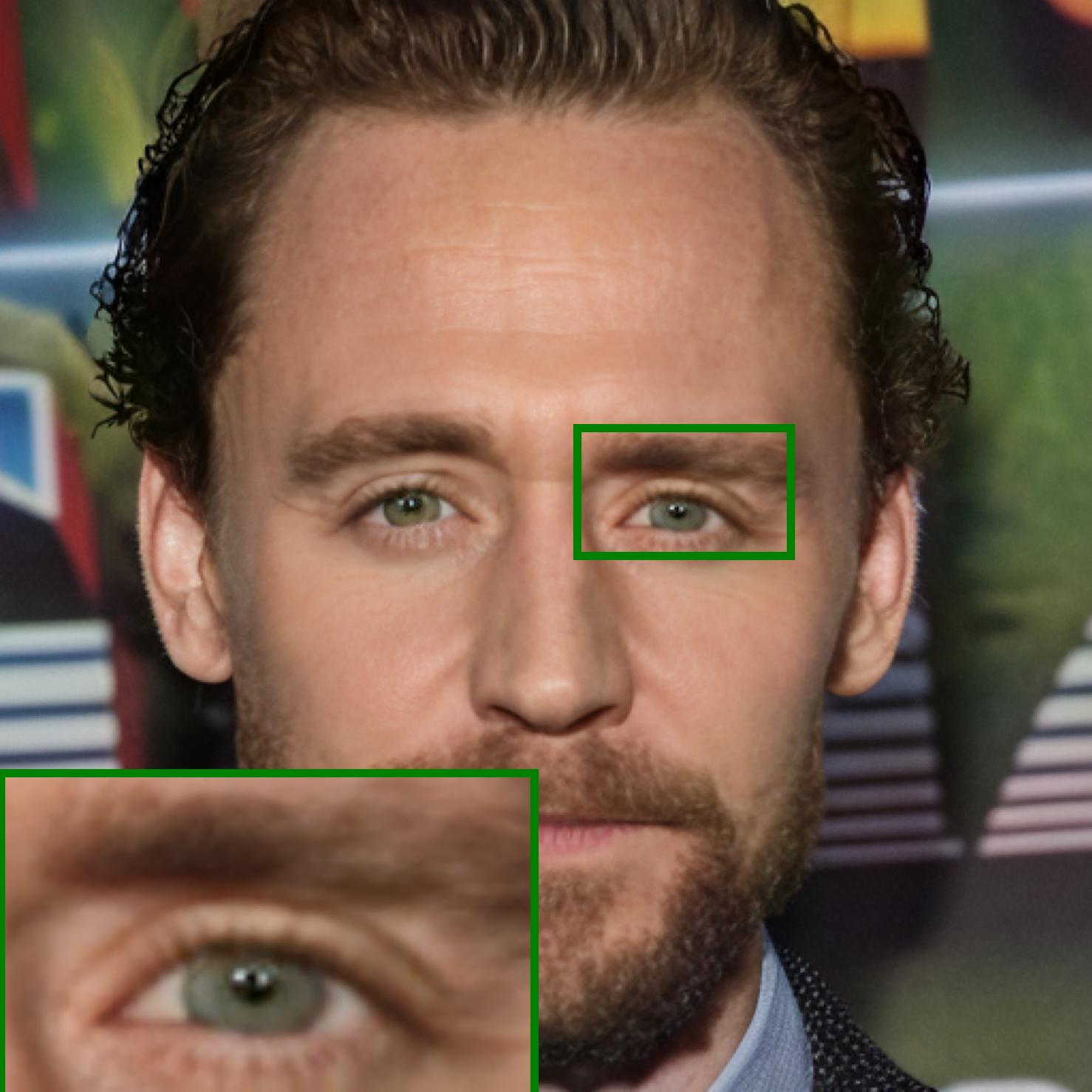} &
        		\includegraphics[width=0.2\textwidth]{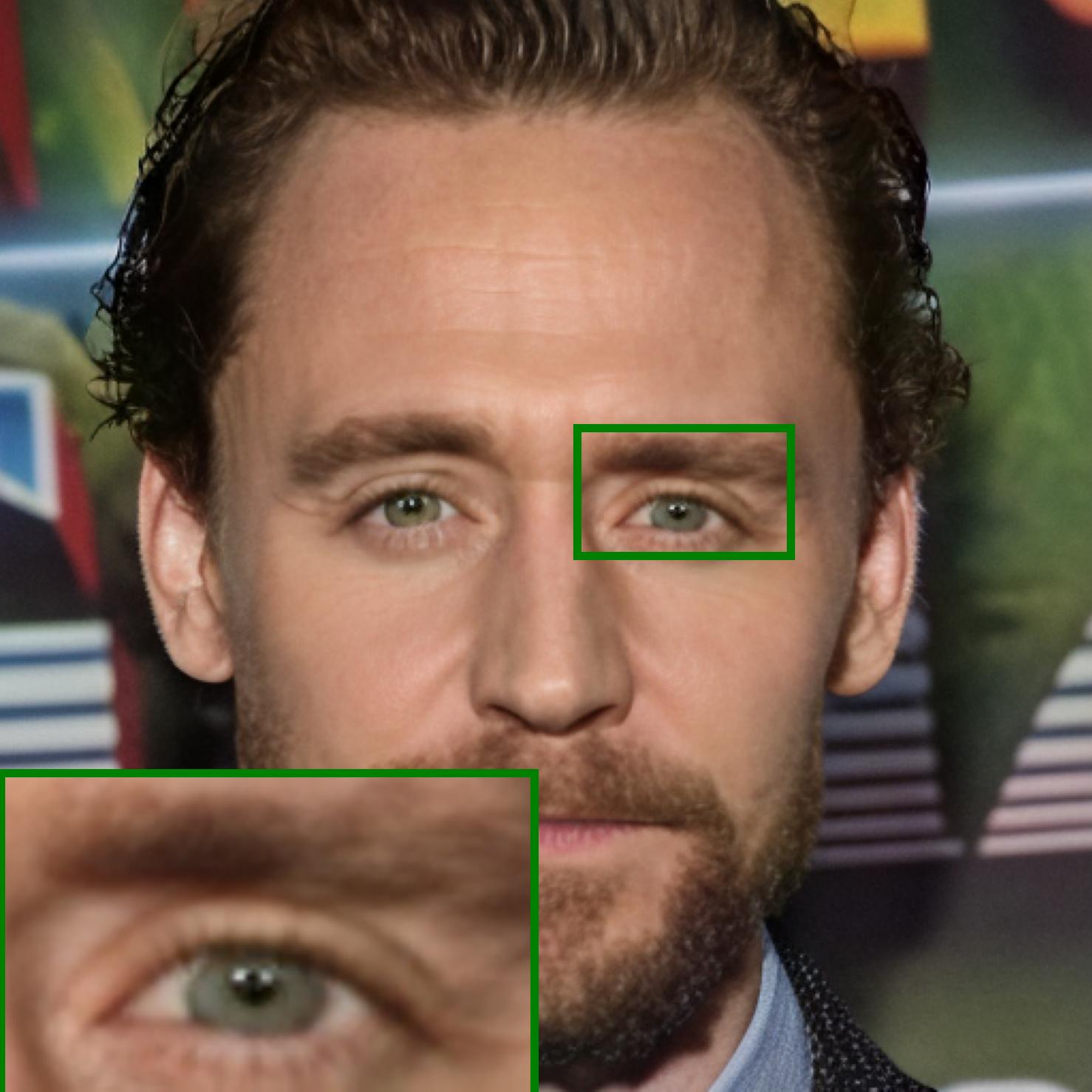} &
        		\includegraphics[width=0.2\textwidth]{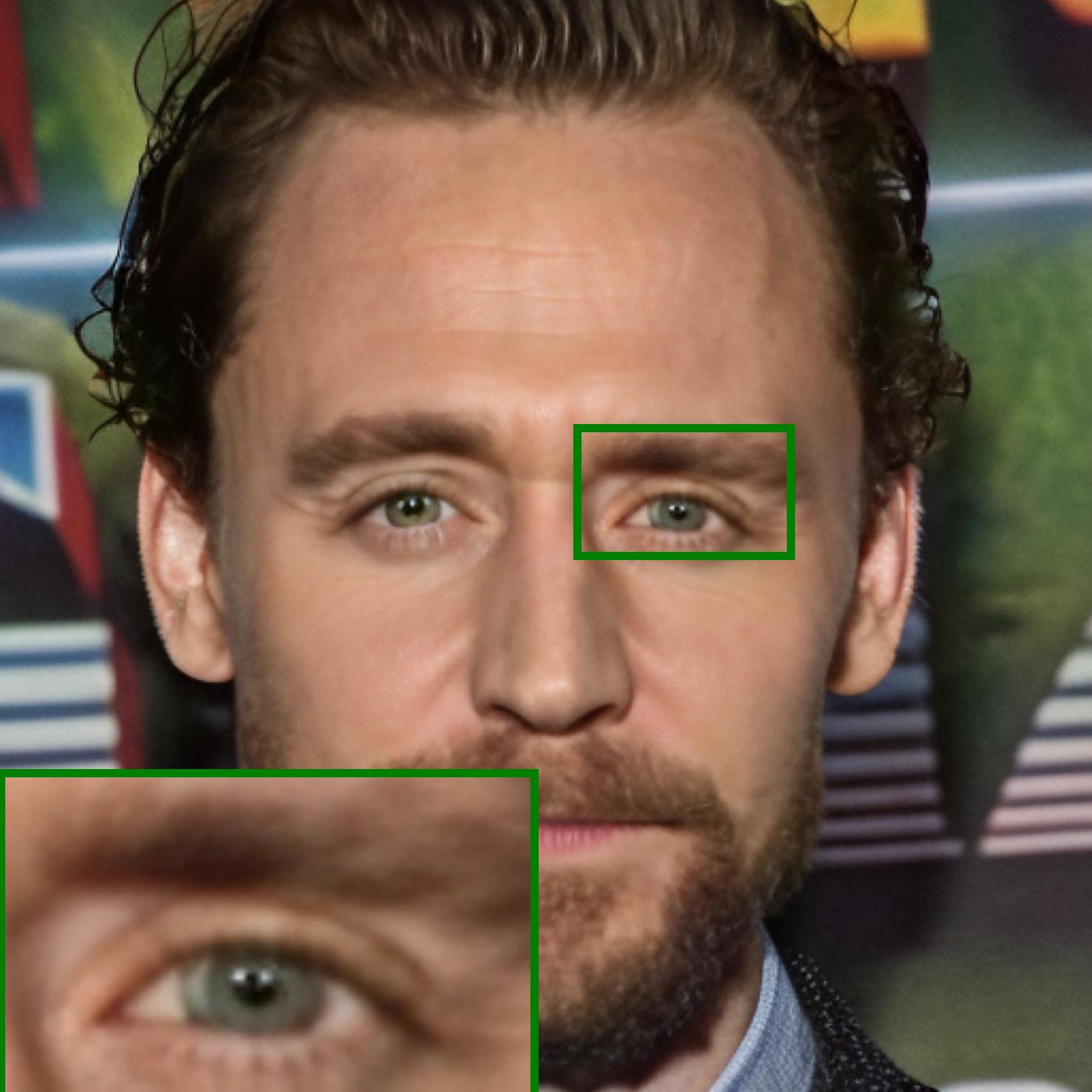} &
        		\includegraphics[width=0.2\textwidth]{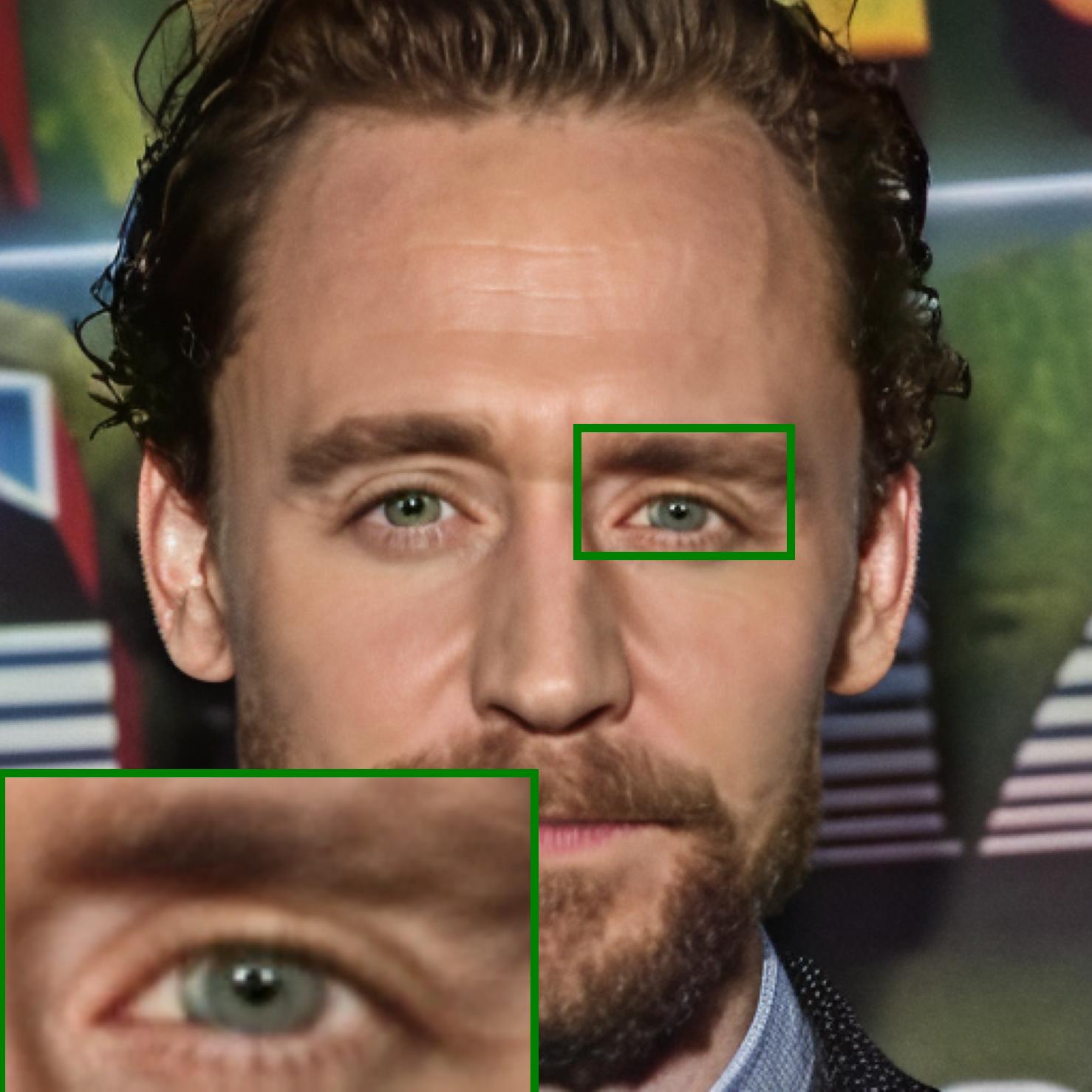}
        		\\
        		\footnotesize{Input} & \footnotesize{$\lambda_{Gen} = 0$} & \footnotesize{$\lambda_{Gen} = 0.1$} & \footnotesize{$\lambda_{Gen} = 0.5$} & \footnotesize{$\lambda_{Gen} = 1$} \\
        		\vspace{-1cm}
        	\end{tabular}
        \end{subfigure}%
	\end{center}
	\caption{
	(Top) In the presence of heavy degradation a larger $\lambda_{Gen}$ is able to improve results. (Bottom) With minor degradation, a larger $\lambda_{Gen}$ can deteriorate results.
	}
	\label{fig:lambda_comparison}
	\vspace{-3mm}
\end{figure}

\subsection{Further analysis}

\paragraph{Personalization}
\Cref{tab:quantitative_ablation} demonstrates the improvements of the proposed method for personalization. Without personalization, the Base Model with the improved training mechanism is able to improve over StableSR \cite{stablesr} in all metrics. However, the results fall behind largely when personalization is added. Base Model + DreamBooth \cite{dreambooth} and Base Model + ViCo \cite{vico} attain similar metrics, however a drop in the PSNR value even below StableSR \cite{stablesr} and the increase in LMSE compared to Base Model, signifies how fine-tuning the whole model can hurt the existing priors. For a fair comparison Base + ViCo also contains generative regularization and other proposed training method proposed and only lacks the learnable $\gamma$ compared to PFStorer. The $\gamma$ provides important balance over the personalized and restored features.
\vspace{-3mm}

\begin{table}[h]
\caption{Quantitative results for different personalization methods on the heavy portion. \color{red} Red \color{black} indicates the best and \color{blue} blue \color{black} indicates the second best}
\resizebox{0.48\textwidth}{!}{%
\begin{tabular}{c|c|cc|cc|c|c} 
 \hline
 Methods & Ref & PSNR $\uparrow$ & SSIM $\uparrow$ & LPIPS $\downarrow$ & MUSIQ $\uparrow$ & LMSE $\downarrow$ & ID $\uparrow$ \\ 
 \hline
 Input & & \color{blue} 22.56 & \color{red} 0.719 & 0.615 & 58.83 & 80.98 & 21.85 \\
 \hline
 StableSR \cite{stablesr} & & 21.68 & 0.601 & 0.605 & 38.55 & 93.82 & 22.87 \\
 Base Model & & 22.15 & 0.661 & 0.449 & 64.33 & 32.83 & 33.90 \\
 Base + DreamBooth \cite{dreambooth} & \checkmark & 21.13 & 0.659 & 0.487 & 62.65 & 37.48 & 52.72 \\
 Base + ViCo \cite{vico} & \checkmark & 22.14 & 0.664 & \color{blue} 0.423 & \color{red} 65.23 & \color{blue} 20.26 & \color{blue} 53.92 \\
 \hline
 PFStorer (Ours) & \checkmark & \color{red} 22.62 & \color{blue} 0.679 & \color{red} 0.414 & \color{blue} 64.04 & \color{red} 18.37 & \color{red} 57.18 \\
 \hline
 GT & & $\infty$ & 1 & 0 & 62.37 & 0 & 100 \\
\end{tabular}}
\label{tab:quantitative_ablation}
\vspace{-3mm}
\end{table}

\paragraph{Alignment-Free Training and Existing Priors}
An immediate benefit to our landmark- and alignment-free approach is that it can be run even when the landmark model fails, as can be seen from the top row of \cref{fig:celeb_ref_heavy}. Furthermore, due to the existing priors of the Base Model, the model is able to restore details from the full head and not only the face, see the result from CodeFormer from  \cref{fig:celeb_ref_real} top.
\vspace{-3mm}

\paragraph{Generative Regularization} \Cref{fig:lambda_comparison} showcases results with different values of the weight $\lambda_{Gen}$ of generative regularization. A larger $\lambda_{Gen}$ encourages more hallucination, which is beneficial for unseen cases, while a smaller $\lambda_{Gen}$ focuses more on the restoration. To balance the effects we use a default $\lambda_{Gen} = 0.1$ for all of our experiments based on empirical observations.

\subsection{Limitations}
\begin{figure}[h]
	\begin{center}
    	\setlength{\tabcolsep}{1pt}
        \begin{subfigure}{0.47\textwidth}
        \hspace{-0.2cm}
        	\begin{tabular}{*4c}
        		\begin{tabular}{*2c}
        		\includegraphics[width=0.117\textwidth]{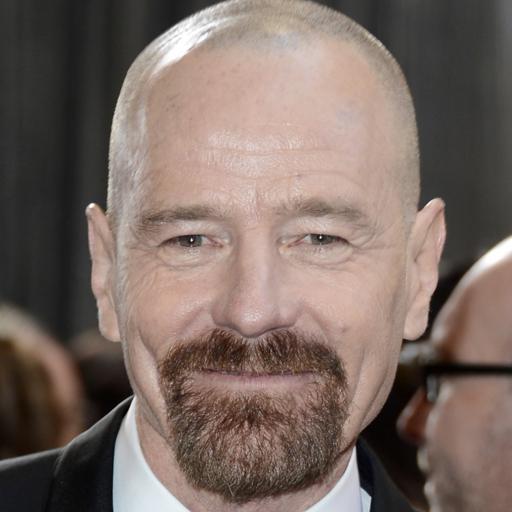} &
        		\includegraphics[width=0.117\textwidth]{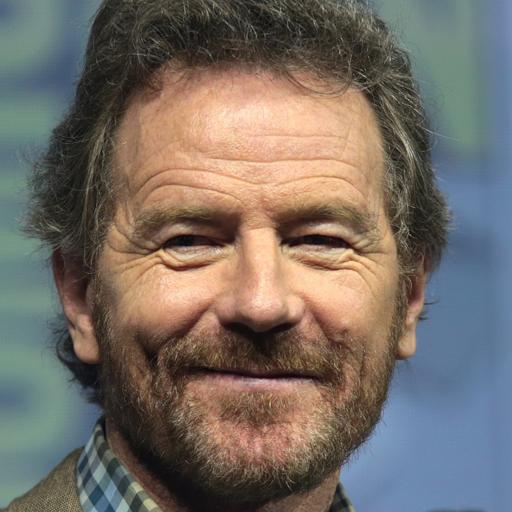}
        		\\
        		\includegraphics[width=0.117\textwidth]{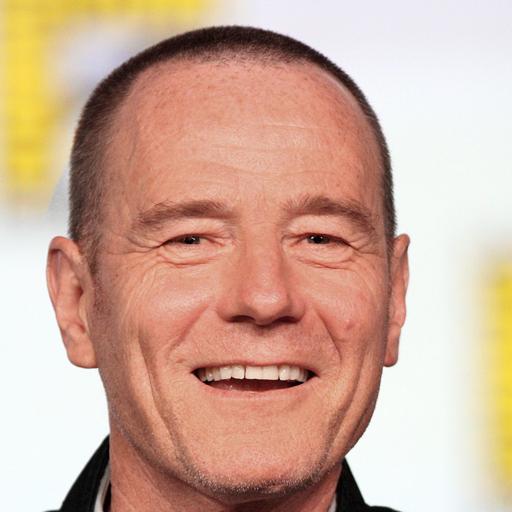} &
        		\includegraphics[width=0.117\textwidth]{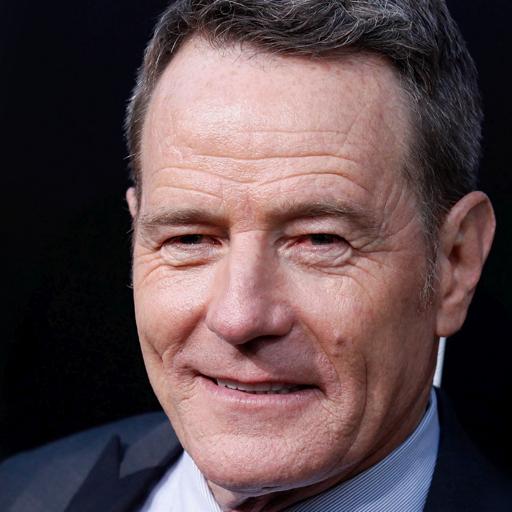}
        		\vspace{1.75cm}
        	\end{tabular}
        	
        			& 
        		\includegraphics[width=0.25\textwidth]{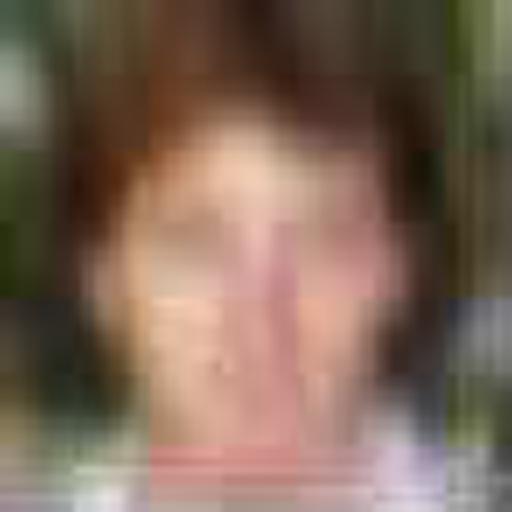} &
        		\includegraphics[width=0.25\textwidth]{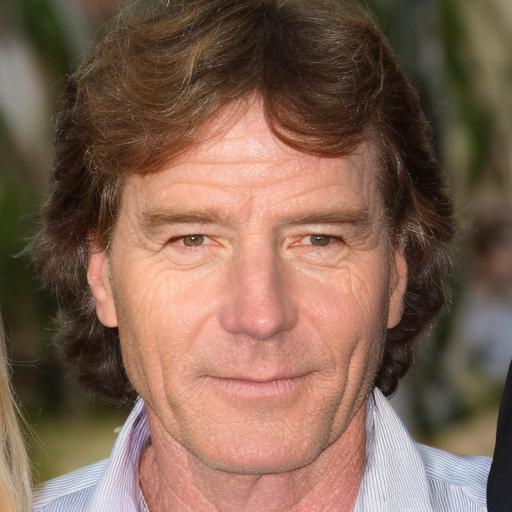}
        			&
        		\includegraphics[width=0.25\textwidth]{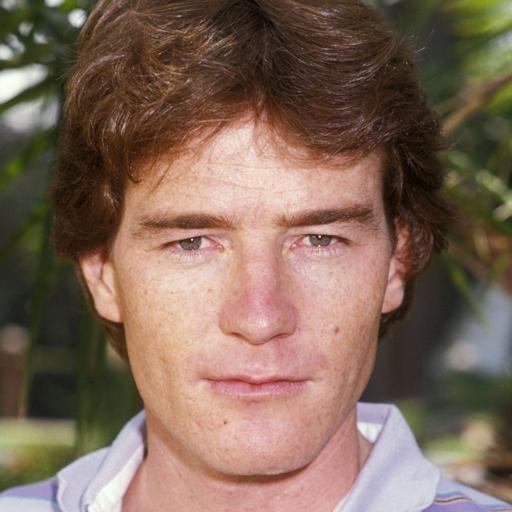}
        		\vspace{-1.8cm}
        		\\
        		\footnotesize{References} & \footnotesize{Input} & \footnotesize{PFStorer} & \footnotesize{GT} \\
        		\vspace{-0.8cm}
        	\end{tabular}
        \end{subfigure}%
	\end{center}
	\caption{
	The output is as accurate as the given reference images are.
	}
	\label{fig:limitations_comparison}
\end{figure}

We show an example of a limitation in \cref{fig:limitations_comparison}. The output is faithful to the given reference images, hence if there are changes in the appearance between references and the input the result may be unwanted. As the model is based on Stable Diffusion it inherits its limitations of slow sampling speed and occasional unwanted artifacts and hallucinations due to the stochasticity. As a possible solution to stochasticity, concurrent work \cite{emu} guides the model towards visually appealing results.

\vspace{-3mm}

%% file: sec/5_conclusion.tex
\vspace{-3mm}
\section{Conclusions}
In this work, we introduce the use of \textit{personalization} for the task of face restoration, where a restoration model is personalized using a few images of a person. We postulate that the problem of face restoration is an ill-posed problem and requires the use of a personal prior for faithful results. We propose the use of a personalization adapter that preserves existing priors of the base restoration model. To enhance the training generative regularization is designed. We showcase our method's abilities through qualitative, quantitative and a user study.
\vspace{-3mm}

\paragraph{Acknowledgements}
This work was supported by the Research Council of Finland Academy Professor project EmotionAI (grants 336116, 345122), ICT 2023 project TrustFace (grant 345948), the University of Oulu \& Research Council of Finland Profi 7 (grant 352788), and by Infotech Oulu.

%% file: sec/6_appendix.tex
\clearpage
\vspace{-50.0em}
\twocolumn[{
\renewcommand\twocolumn[1][]{#1}
\centering
\Large
\textbf{\thetitle}\\
\vspace{1.5em}Supplementary Material \\
\vspace{1.0em}
}]

\begin{figure*}
	\includegraphics[width=1\textwidth]{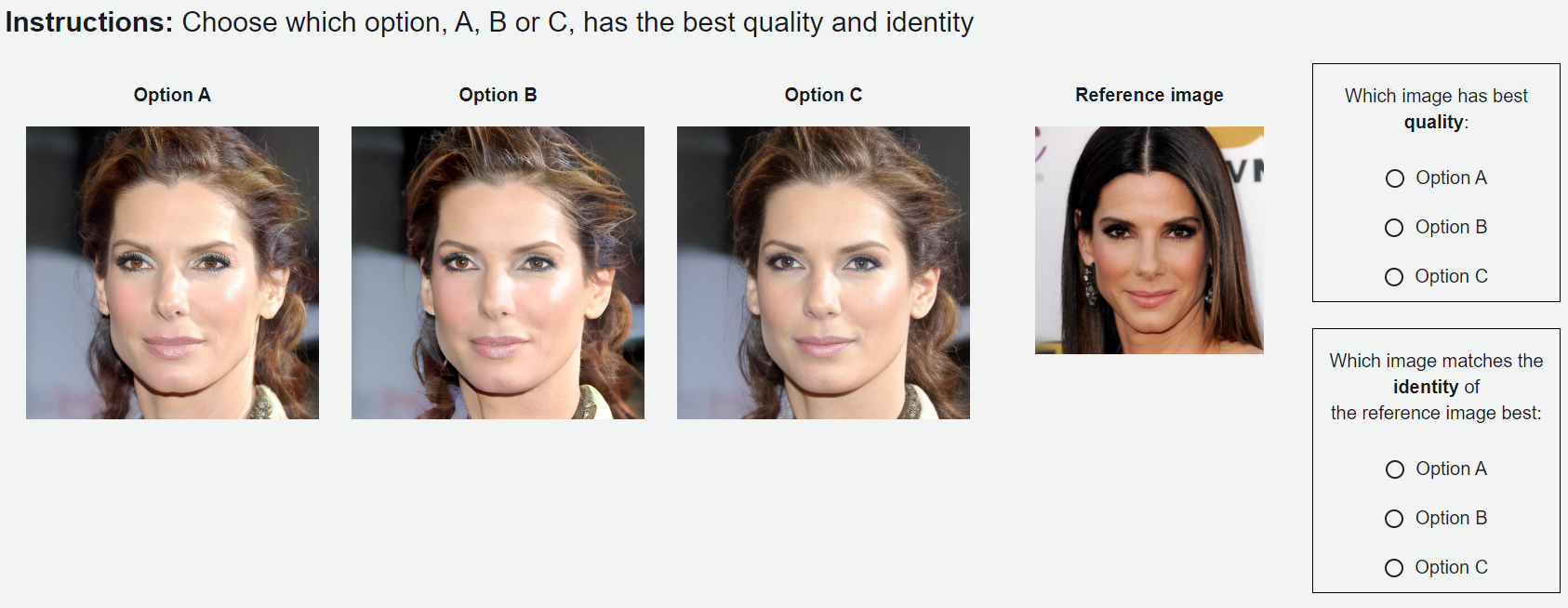}
	\caption{User interface used in the user study.}
	\label{fig:user_study_ui}
\end{figure*}

This supplementary material contains the following sections. First, background is provided for the used models. Next, further details of the user study are provided. Then, full experimental details and additional experiments of the personalized model are shown. Finally, the training details and further experiments of the non-personalized base restoration model are discussed. As the last section, societal impact is discussed.

\section*{Background}
\label{app:background}
In this section we provide sufficient background to keep the paper self-sustained. We first introduce latent diffusion models, namely Stable Diffusion \cite{stablediffusion}, as both the used methods, StableSR \cite{stablesr} and ViCo \cite{vico} are based off of it. Next we provide further details of the base model, StableSR and the personalization technique ViCo.
\paragraph{Latent Diffusion Models}
As oppose to diffusion models \cite{ddpm}, LDMs \cite{stablediffusion} (latent diffusion models) perform the diffusion steps in a latent space. In Stable Diffusion \cite{stablediffusion} an encoder $\mathcal{E}$ is first trained to map input images $x\in \mathbb{R}^{H \times W \times 3}$ to a latent code $z = \mathcal{E}(x) \in \mathbb{R}^{(H/8) \times (W/8) \times 4}$, which can be approximately reconstructed with a decoder $\mathcal{D}$. The diffusion forward and backward steps are then performed within the latent space. To perform conditional generation using text $y$, it is first transformed to an embedding $c(y)$ with a text embedder. The training loss is then given by:
\begin{equation}
	\mathcal{L} = \mathbb{E}_{z\sim\mathcal{E}(x), y, \eps\sim\mathcal{N}(0, 1), t}\left [||\eps - \eps_{\theta}(z_t, c(y), t)||^2_2\right].
\end{equation}
Above, at timestep $t$ the diffusion model $\eps_{theta}$ denoises the added noise from $z_t$ conditioned on the text embedding $c(y)$ and the timestep $t$.
After training the model can be used to generate images with a text prompt $y$ and starting from a $z_t \sim \mathcal{N}(0, 1)$ iteratively until all noise is removed at $t=0$.

\paragraph{StableSR}
To exploit the rich generative prior in Stable Diffusion \cite{stablediffusion} for image restoration, StableSR \cite{stablesr} uses conditioning of low-quality images. The entire Stable Diffusion is kept frozen, while a time-aware encoder and spatial feature transformations are added as adapters to condition the low-quality images. The encoder takes in a low-quality image and the current time-step of the diffusion model and outputs feature maps at different resolutions, corresponding to the ones in Stable Diffusion's UNet. The features are then fed through spatial feature transforms and added to the output of the original UNet layer's intermediate outputs.

The model is trained on high-quality generic natural images of 2k and 8k resolutions, which are randomly cropped to $512 \times 512$. This training enables the use of arbitrary size super-resolution using aggregation sampling. Here, the input is split to overlapping tiles, which are processed by the model independently. To avoid border artifacts a Gaussian kernel is used for the fusion of the tiles.

\paragraph{ViCo}
For efficient and accurate personalization, ViCo \cite{vico} also uses Stable Diffusion as its base. Similarly to StableSR, ViCo keeps the entire Stable Diffusion frozen. Only added adapter blocks and a single text-embedding are trained. Similar to \cite{textual_inversion}, a text-embedding is made learnable that can be associated with the subject. Additionally, image cross-attention adapters are added to four blocks of the UNet. These cross-attention layers take the current time-step's intermediate result and the intermediate features of a reference image, which has also gone through the UNet.  To enhance the result, a mask is used to ignore the background. The mask is obtained from an attention map $A = \mathrm{softmax}\left(\frac{QK^T}{\sqrt{d_k}}\right)$, which is a product of the reference image and the text-embedding. A regularizer
\begin{equation}
	\mathcal{L}_{Pers} = ||A_{\star} / \max(A_{\star}) - A_{\mathrm{EOT}} / \max(A_{\mathrm{EOT}}) ||_2^2,
\end{equation}
where $A_{\star}$ are the similarity logits corresponding to the learnable token of the text-embedding and $A_{\mathrm{EOT}}$ is the end-of-text token, is used to avoid overfitting. The end-of-text token $\mathrm{<|EOT|>}$ captures a global representation, which retains good semantics of the personalizable object through training.

\section*{User Study Details}
\label{app:user_study}
From the light and heavy partitions we randomly select two images for each identity. From the real data, we select all of the images. In total we have $20 \times 2 + 20 \times 2 + 20 \times 1 = 100$ images. With 40 users and two tasks we have a total of 8000 unique answers. As identifying fine-grained details of an identity, especially not a familiar one, can be difficult, we chose four images from each identity (light and heavy). This way the users can get more accustomed to the identities and make more accurate evaluations for the similarity of the identity features.

To ensure that the users are being accurate with their annotations, we use five control tasks. Here a ground-truth image is paired with extremely poor quality images. If the user fails in these tasks, their annotations are likely to be inaccurate and their results can be potentially invalidated. To avoid biases with users always choosing A, B or C, we randomly shuffle the model's positions.

\Cref{fig:user_study_ui} displays the user interface used for the study. Users have to read the full instructions before taking the study. The instructions detail the two different tasks and how they should be evaluated.

Amazon Mechanical Turk is used for the study. We follow principles from \cite{user_study}. We filter users based on the Master certificate to ensure quality annotations. For each task we pay \$ 0.04 as suggested in \cite{user_study}. 

\section*{Personalized Model Additional Experiments}
\label{app:additional_experiments}
In this section we detail the full experimental settings and conduct additional experiments of hyperparameters for the personalized model. In all experiments, excluding the parameter to be studied other hyperparameters are kept constant. With further fine-tuning of hyperparameters, results for specific individuals and inputs can be improved.

\paragraph{Settings}
Followed by community findings that prompts can improve quality of the restored image, we use both a positive and a negative prompt. For the positive prompt we use \textit{"a Photo of * , masterpiece, best quality, realistic, very clear, professional"} and for the negative prompt we use \textit{"3d, cartoon, anime, sketches, worst quality, low quality"}. We note that including semantic changes in the prompt like \textit{"red hair"} does not have an effect. This is due to the restoration blocks fusing the low-quality image with the denoised image directly. This is in line with the goal of the paper, as it is a restoration method, not an editing method.

For the classifier-free guidance value we set 4 as a default for all experiments. Standard DDPM \cite{ddpm} sampling is used with 200 steps as in StableSR \cite{stablesr}. As the personalization fine-tuning approach is about learning a single identity and not multiple parts of an identity, we find that not using the 50\% random crops improves the results slightly. For the baseline method DR2 + SPAR \cite{dr2} we empirically experimented with several hyperparameters values of $N$ and $\tau$ that are crucial for the performance of the method. $N$ is a downsampling factor and $\tau$ is the output step after which generation is started. We set $N = 8$ and $\tau = 40$ as we found it performed the best across different levels of degradations. For LMSE (Landmark MSE) \cite{farl} was used to obtain landmarks. In cases where landmarks could not be found due to the image being severely degraded the MSE was set to 128. Similarly in cases where the MSE was for an image was more than 128 it was capped to 128 to avoid outliers due to numerical errors or other errors.

\paragraph{Degradations}
During testing we synthesized a light and a heavy degradation to better evaluate our algorithm in different situations. During training we use the heavy setting. We use the settings from StableSR as a base and modify them. To better suit for real-world applications we include motion and median blur, as well as adding ISP (Image Signal Processing) noise \cite{scunet}.

To ensure our method works in less severe cases, we also include a light partition during testing. Here, we only include a first-order noise similar to CodeFormer. 

The light portion follows:
\begin{equation}
	I_D = \{[(I \circledast k_\sigma)_{\downarrow_r} + n_\delta]_{\text{JPEG}_q}\}_{\uparrow_r},
	\label{eq:light_degradation}
\end{equation}
where $k_\sigma$ is Gaussian blur kernel, $\downarrow_r$ and $\uparrow_r$ are the downsamping and upsampling operators, $n_\delta$ additive Gaussian noise and $[\cdot]_{\text{JPEG}_q}$ is JPEG compression. We sample uniformly $\sigma$, $r$, $\delta$ and $q$ from $[0.1, 10]$, $[1, 4]$, $[0, 2]$ and $[30, 100]$, respectively. The additive Gaussian noise has a probability of 40\% and downsampling a probablity of 70\%, while filtering and JPEG compression occur always.

The heavy portion first applies ISP model \cite{scunet} with a 50\% probability, followed by motion and median blur with 5\% and 10\% probabilities. Next, we use \eqref{eq:light_degradation} and the same settings except, $r$ and $\delta$ are chosen from $[1, 10]$ and $[0, 15]$, respectively, followed by a sinc filter \cite{stablesr}. Finally, \eqref{eq:light_degradation} is applied a second time with a 90\% probability.

\paragraph{Classifier-Free Guidance Value}
To further emphasize the conditional element, CFG \cite{cfg} can be used to guide the denoising process. As mentioned earlier, we use a negative prompt instead of a null one. The formula is given by
\begin{equation}
	\tilde{X} = X + \lambda_{cfg}(X(p_{pos}, I_{LQ}) - X(p_{neg}, I_{LQ})),
	\label{eq:cfg}
\end{equation}
where $p_{neg}$ and $p_{pos}$ correspond to the positive and negative prompts. We also experimented with null conditioning the low-quality image
\begin{equation}
	\tilde{X} = X + \lambda_{cfg}(X(p_{pos}, I_{LQ}) - X(p_{neg}, \varnothing)),
\end{equation}
but found the results to be of lower-quality, as emphasizing the low-quality image may exaggerate blurry features.

\begin{figure}[h]
	\begin{center}
    	\setlength{\tabcolsep}{1pt}
        \begin{subfigure}{0.48\textwidth}
        \hspace{-0.2cm}
        	\begin{tabular}{*5c}
        		\includegraphics[width=0.19\textwidth]{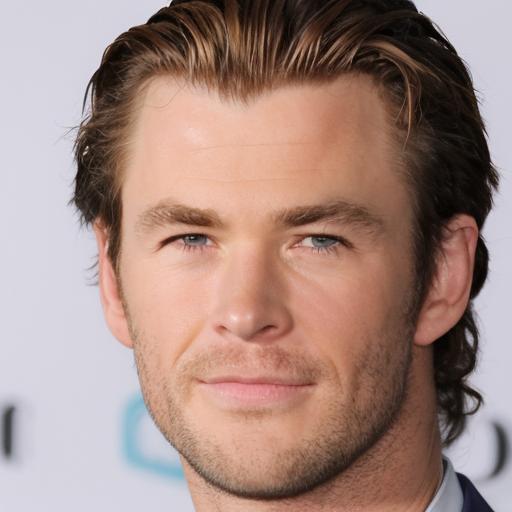} &
        		\includegraphics[width=0.19\textwidth]{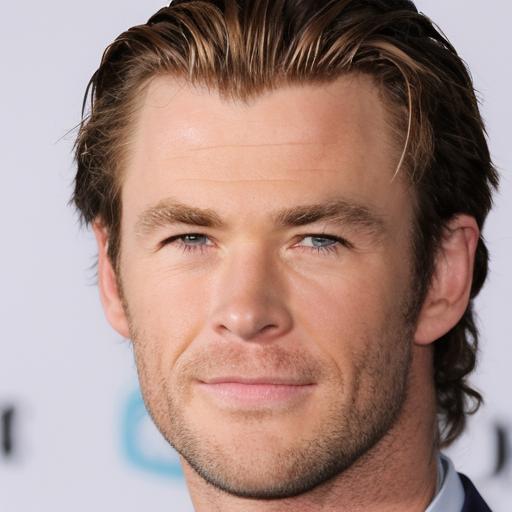} &
        		\includegraphics[width=0.19\textwidth]{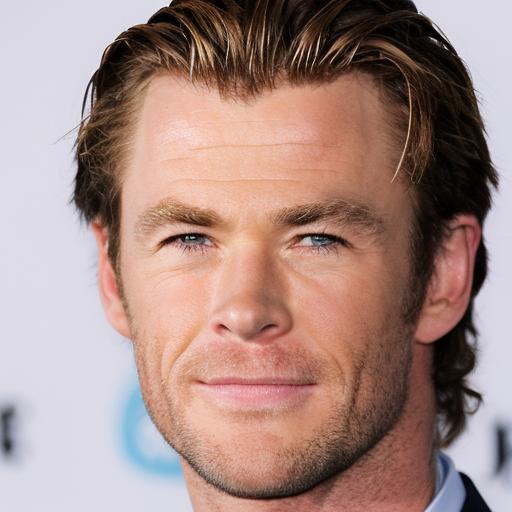} &
        		\includegraphics[width=0.19\textwidth]{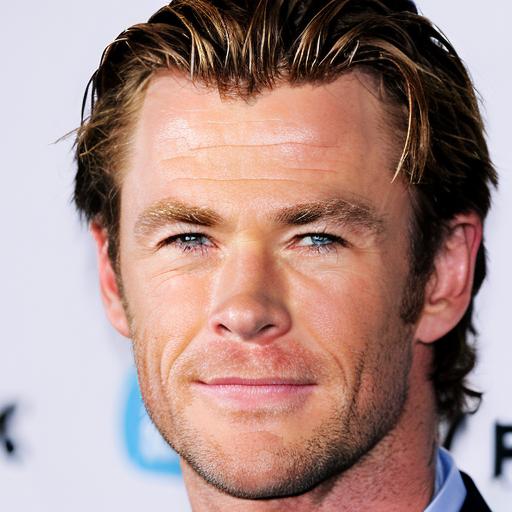} &
        		\includegraphics[width=0.19\textwidth]{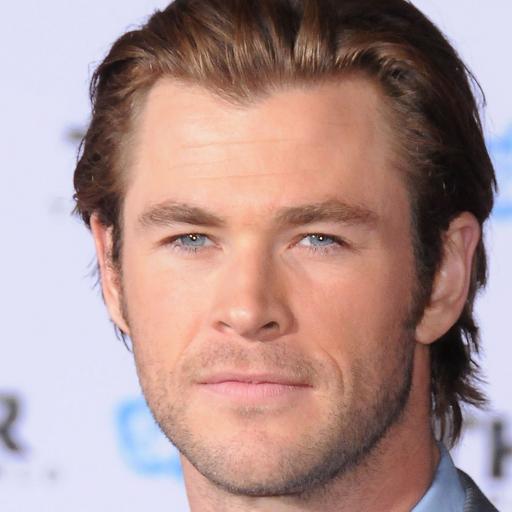}
        		\\
        		\footnotesize{$\lambda_{cfg} = 1$} & \footnotesize{$\lambda_{cfg} = 2$} & \footnotesize{$\lambda_{cfg} = 4$} & \footnotesize{$\lambda_{cfg} = 7.5$} & \footnotesize{GT}
        		\\
        		\includegraphics[width=0.19\textwidth]{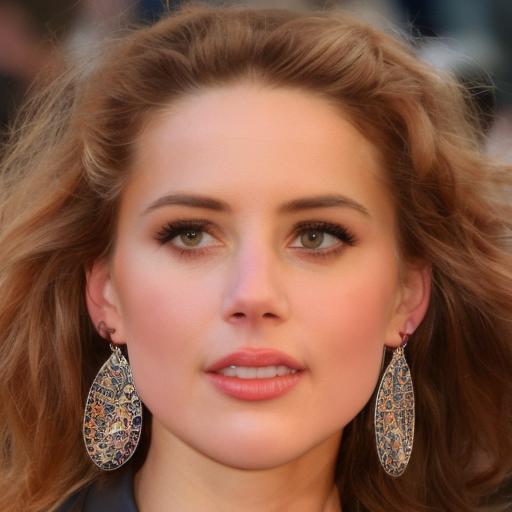} &
        		\includegraphics[width=0.19\textwidth]{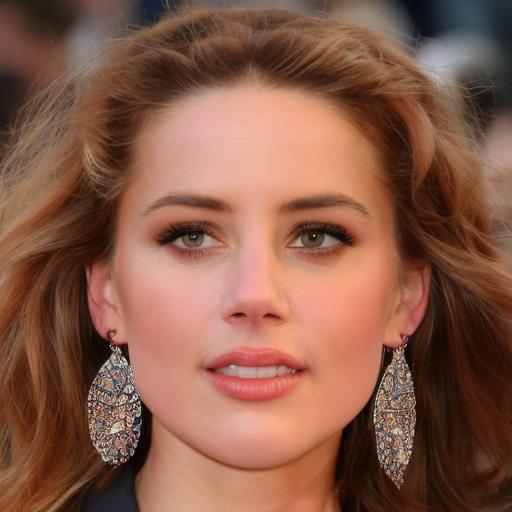} &
        		\includegraphics[width=0.19\textwidth]{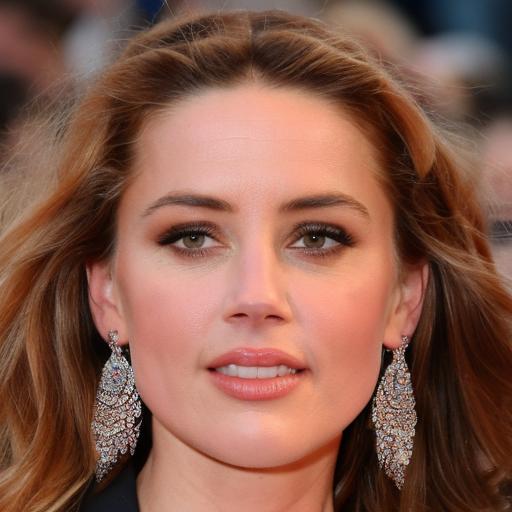} &
        		\includegraphics[width=0.19\textwidth]{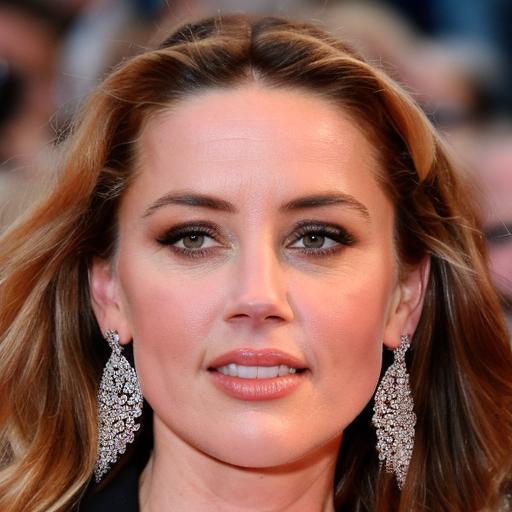} &
        		\includegraphics[width=0.19\textwidth]{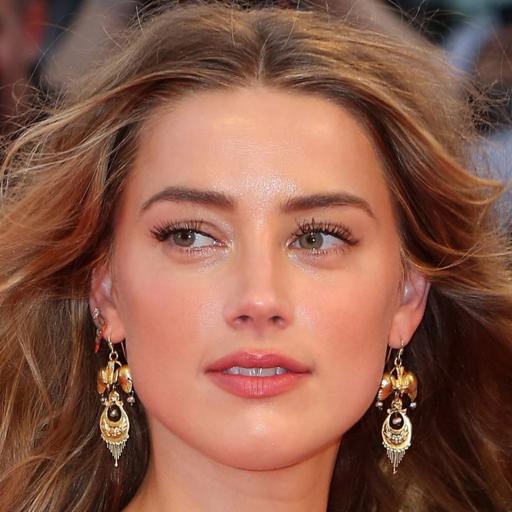}
        		\\
        		\footnotesize{$\lambda_{cfg} = 1$} & \footnotesize{$\lambda_{cfg} = 2$} & \footnotesize{$\lambda_{cfg} = 4$} & \footnotesize{$\lambda_{cfg} = 7.5$} & \footnotesize{GT}
        		\\
        		\vspace{-1cm}
        	\end{tabular}
        \end{subfigure}%
	\end{center}
	\caption{
	Experiments with different classifier-free guidance values.
	}
	\label{fig:cfg}
\end{figure}

We experiment using \cref{eq:cfg} with different CFG values $\lambda_{cfg}$ in \cref{fig:cfg} and note that $\lambda_{cfg} = 1$ corresponds to not using guidance at all. It can be seen that a higher $\lambda_{cfg}$ can oversature, as in the top row. In the lower row, a low $\lambda_{cfg}$ loses identity features, whereas in the top row it is more subtle. From our experiments we observe that different identities behave differently with different $\lambda_{cfg}$. A common value in text-to-image applications is $\lambda_{cfg} = 7.5$, but to avoid saturation we default to $\lambda_{cfg}=4$ in all of our other experiments.

\begin{figure}[h]
	\begin{center}
    	\setlength{\tabcolsep}{1pt}
        \begin{subfigure}{0.48\textwidth}
        \hspace{-0.2cm}
        	\begin{tabular}{*5c}
        		\includegraphics[width=0.19\textwidth]{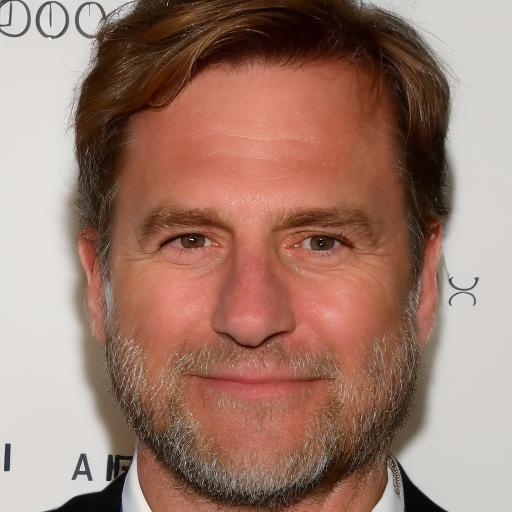} &
        		\includegraphics[width=0.19\textwidth]{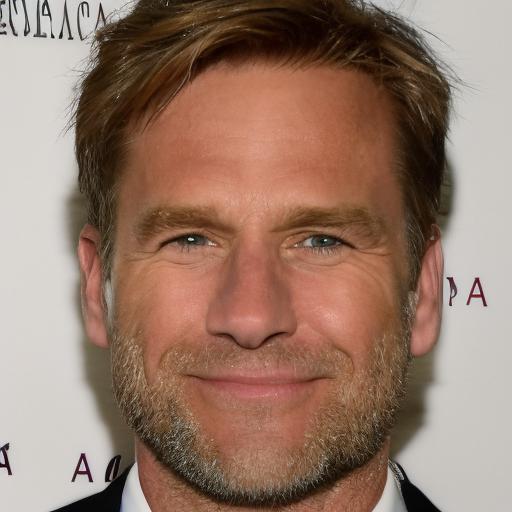} &
        		\includegraphics[width=0.19\textwidth]{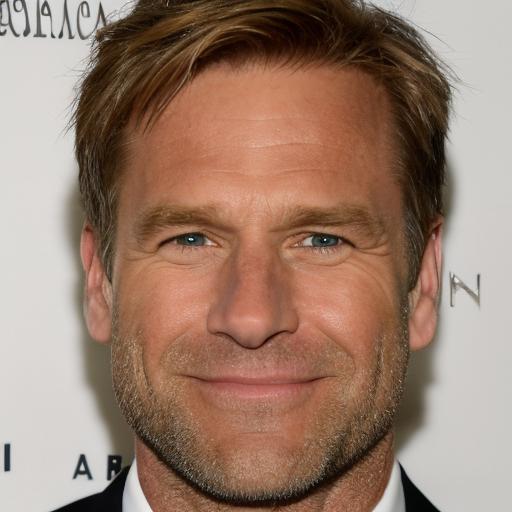} &
        		\includegraphics[width=0.19\textwidth]{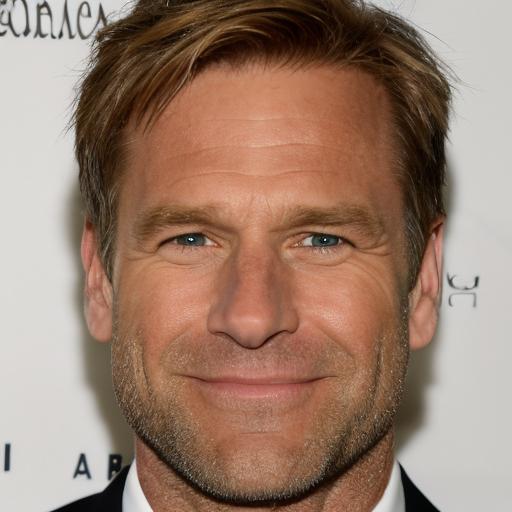} &
        		\includegraphics[width=0.19\textwidth]{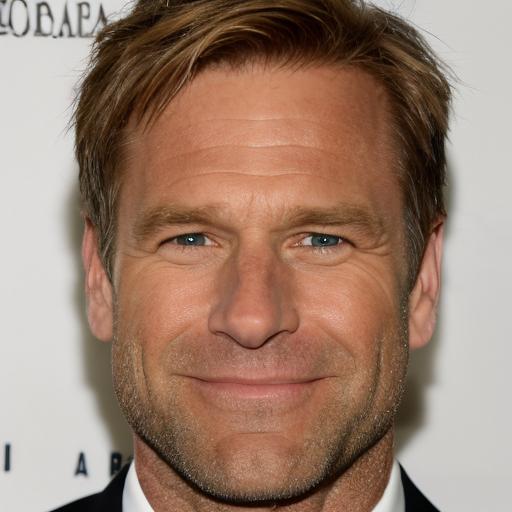}
        		\\
        		\footnotesize{Default} & \footnotesize{$\lambda_{att} = -0.5$} & \footnotesize{$\lambda_{att} = 0$} & \footnotesize{$\lambda_{att} = 0.1$} & \footnotesize{$\lambda_{att} = 0.25$}
        		\\
        		\includegraphics[width=0.19\textwidth]{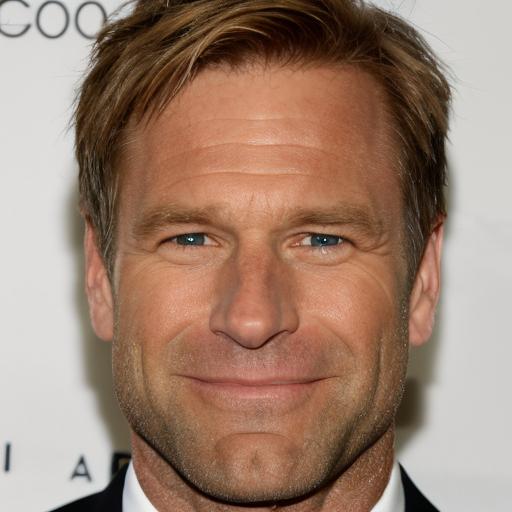} &
        		\includegraphics[width=0.19\textwidth]{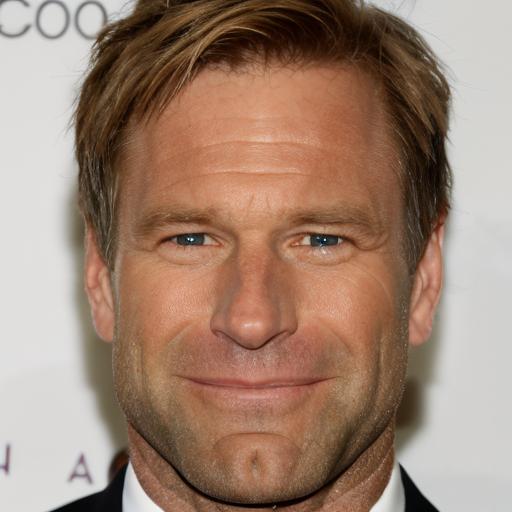} &
        		\includegraphics[width=0.19\textwidth]{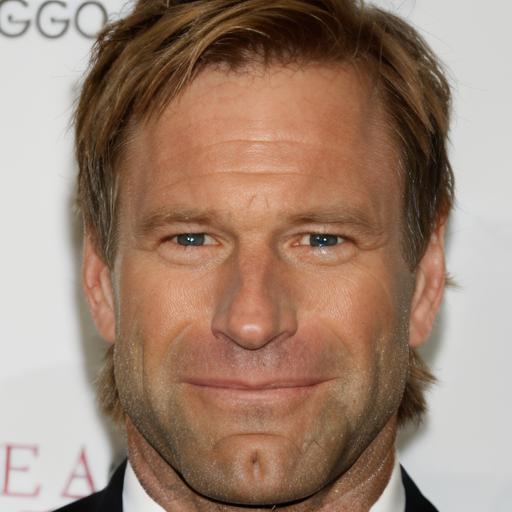} &
        		\includegraphics[width=0.19\textwidth]{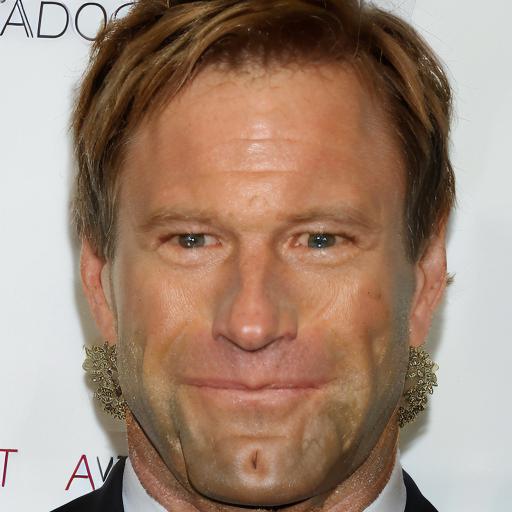} &
        		\includegraphics[width=0.19\textwidth]{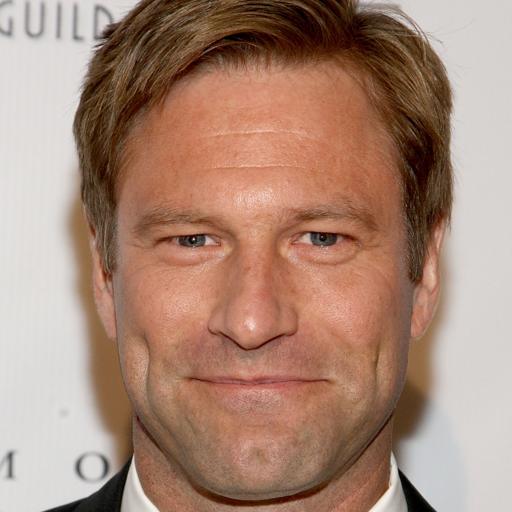}
        		\\
        		\footnotesize{$\lambda_{att} = 0.5$} & \footnotesize{$\lambda_{att} = 0.75$} & \footnotesize{$\lambda_{att} = 1$} & \footnotesize{$\lambda_{att} = 1.5$} & \footnotesize{GT}
        		\\
        		\vspace{-1cm}
        	\end{tabular}
        \end{subfigure}%
	\end{center}
	\caption{
	Controlling identity. The default corresponds to non-personalized output. Samples with $\lambda_{att}$, use the personalized token in the prompt.
	}
	\label{fig:identity}
\end{figure}

\paragraph{Controlling Identity}
The used personalization technique consists of two parts. 1) a learnable text-embedding and 2. image cross-attention. We experiment with different values of $\lambda_{att}$, which controls the weight of the cross-attention layers, in \cref{fig:identity}. The sample with $\lambda_{att} = 0$ corresponds to only using the learnable text-embedding. We can see that it contains some identity at a high-level but is missing details such as wrinkles and dip in the chin. With increasing $\lambda_{att}$ the identity features become more prominent, even to a degree of exaggeration. Similar to CFG values, we have observed that for different individuals and depending on the noise levels of the input, $\lambda_{att}$ acts differently. We chose $\lambda_{att} = 1$ as a default value, although in some cases, like this sample, the optimal results can be something different.

\begin{figure}[h]
	\begin{center}
    	\setlength{\tabcolsep}{1pt}
        \begin{subfigure}{0.48\textwidth}
        \hspace{-0.2cm}
        	\begin{tabular}{*4c}
        		\includegraphics[width=0.24\textwidth]{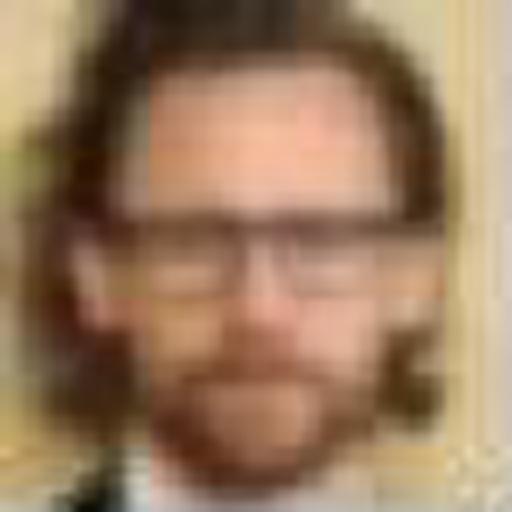} &
        		\includegraphics[width=0.24\textwidth]{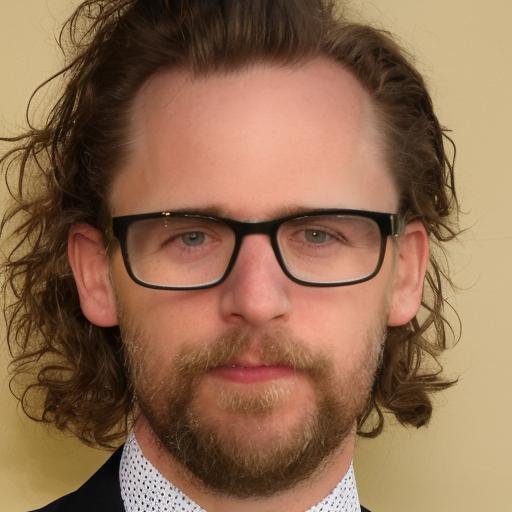} &
        		\includegraphics[width=0.24\textwidth]{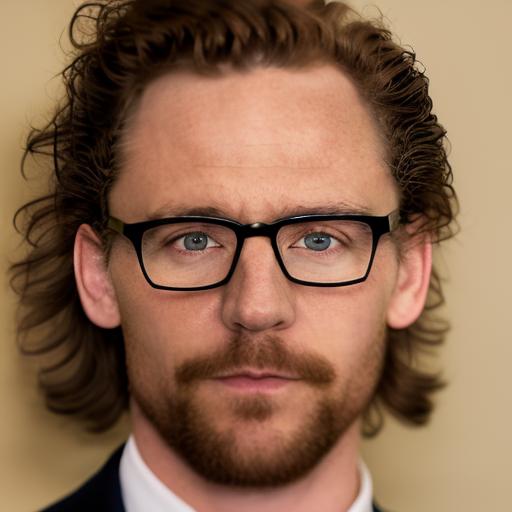} &
        		\includegraphics[width=0.24\textwidth]{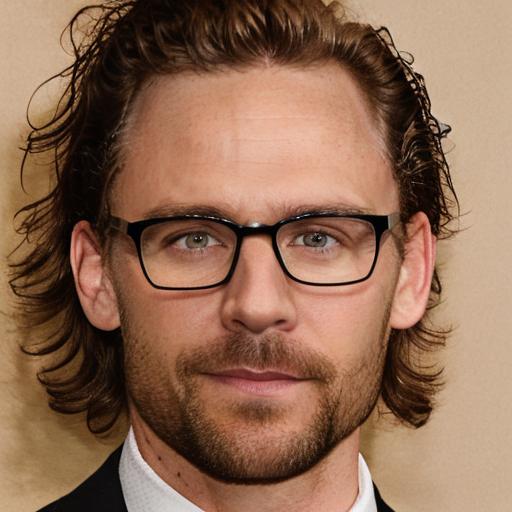}
        		\\
        		\footnotesize{Input} & \footnotesize{$n_{img} = 0$} & \footnotesize{$n_{img} = 1$} & \footnotesize{$n_{img} = 2$}
        		\\
        		\includegraphics[width=0.24\textwidth]{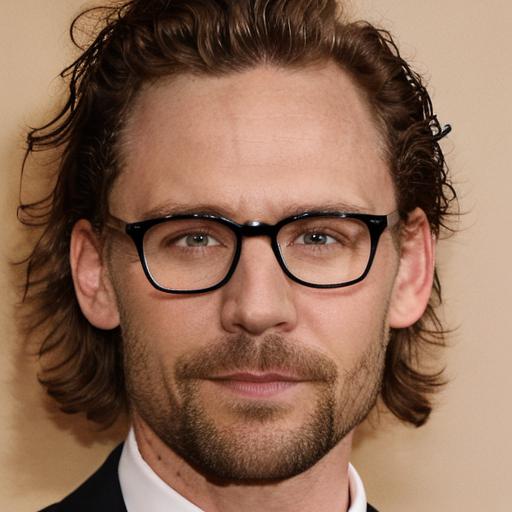} &
        		\includegraphics[width=0.24\textwidth]{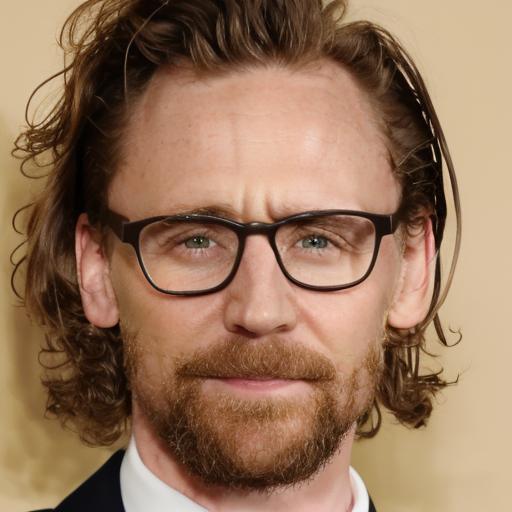} &
        		\includegraphics[width=0.24\textwidth]{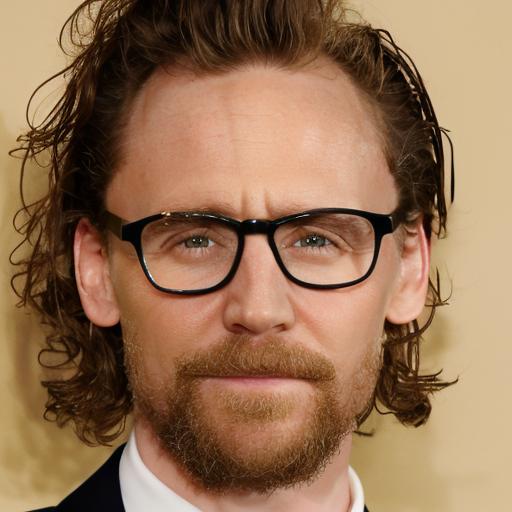}&
        		\includegraphics[width=0.24\textwidth]{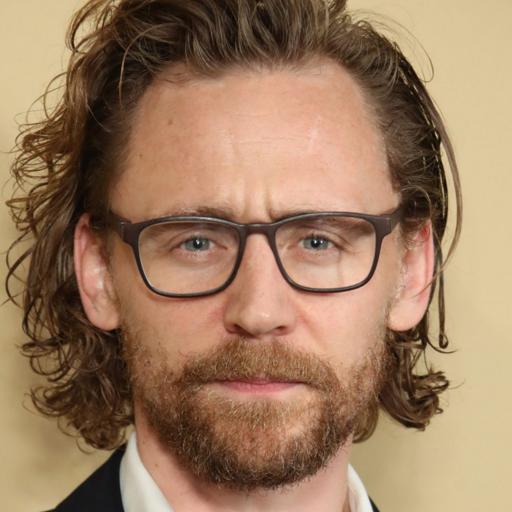}
        		\\
        		\footnotesize{$n_{img} = 3$} & \footnotesize{$n_{img} = 5$} & \footnotesize{$n_{img} = 10$} & \footnotesize{GT} \\
        		\vspace{-1cm}
        	\end{tabular}
        \end{subfigure}%
	\end{center}
	\caption{
	Number of images used for personalization. $n_{img} = 0$ refers to no personalization.
	}
	\label{fig:numpic}
\end{figure}

\paragraph{Number of Reference Images}
In \cref{fig:numpic} we experiment with how many reference images are required to accurately capture the identity. With $n_{img} = 0$, \textit{i.e.} no personalization, high-level features matching the input can be observed. With just one reference image, the eyes, eyebrows and other finer details start to appear. We default to using $n_{img} = 5$ as it often performs sufficiently and the addition of more images has less noticeable effect. For some individuals we found that even three images can be sufficient, but it should be noted that the similarities between input image and the reference images affect the results.

\begin{figure}[h]
	\begin{center}
    	\setlength{\tabcolsep}{1pt}
        \begin{subfigure}{0.48\textwidth}
        \hspace{-0.2cm}
        	\begin{tabular}{*6c}
        		\includegraphics[width=0.16\textwidth]{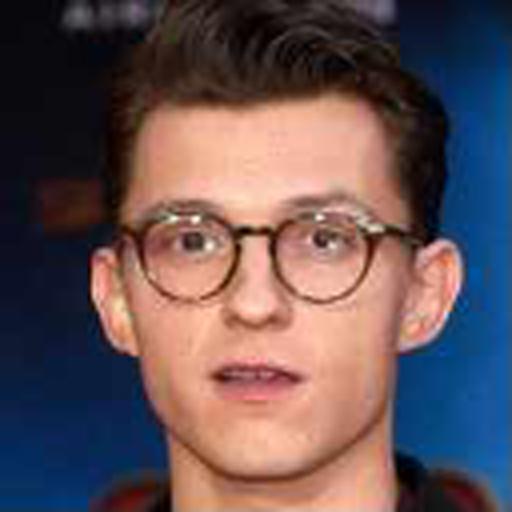} &
        		\includegraphics[width=0.16\textwidth]{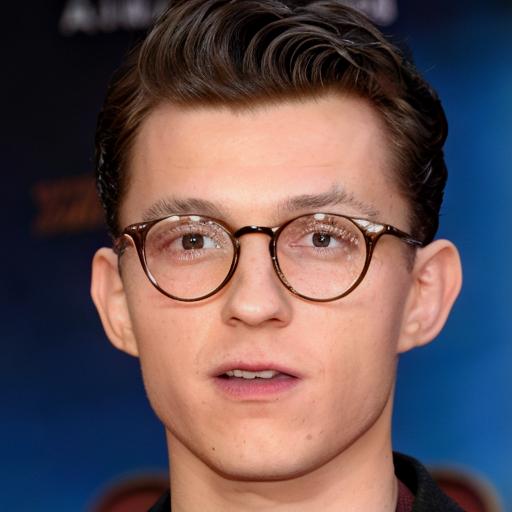} &
        		\includegraphics[width=0.16\textwidth]{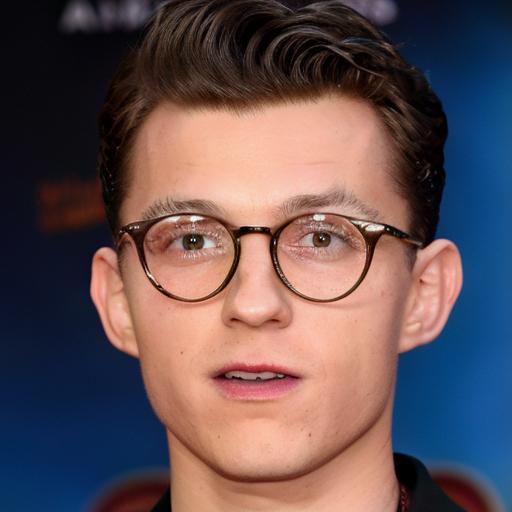} &
        		\includegraphics[width=0.16\textwidth]{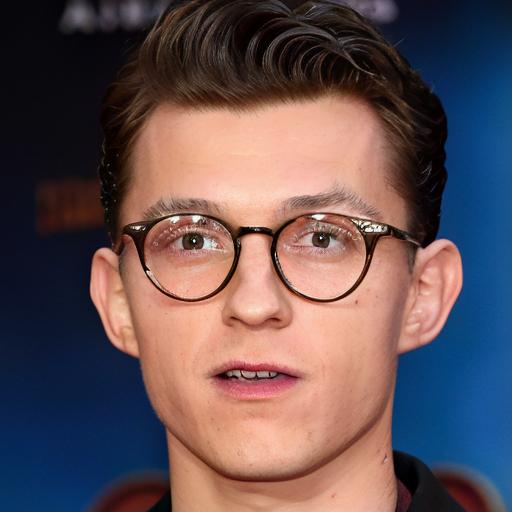} &
        		\includegraphics[width=0.16\textwidth]{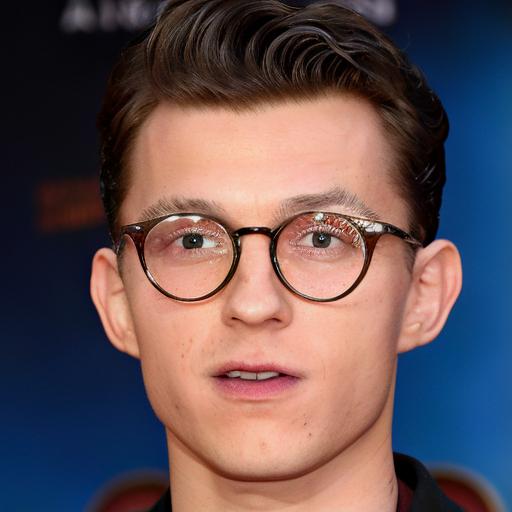} &
        		\includegraphics[width=0.16\textwidth]{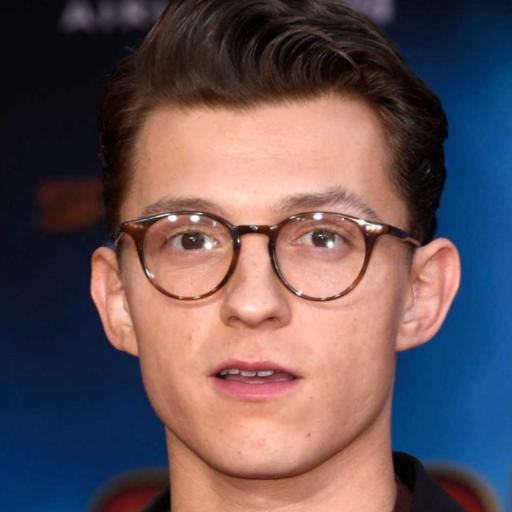}
        		\\
        		\footnotesize{LQ-Light} & \footnotesize{Seed = 0} & \footnotesize{Seed = 1} & \footnotesize{Seed = 2} & \footnotesize{Seed = 3} & \footnotesize{GT}
        		\\
        		\includegraphics[width=0.16\textwidth]{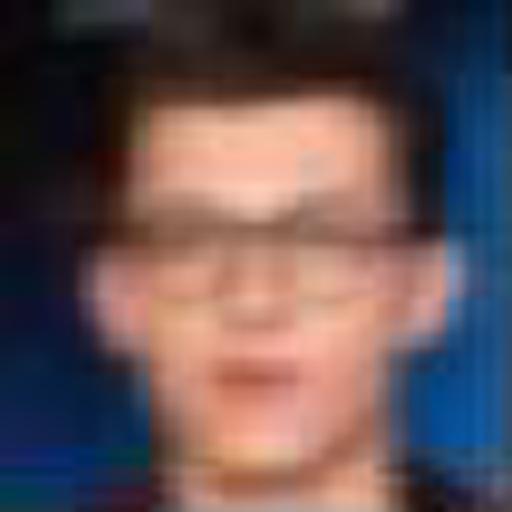} &
        		\includegraphics[width=0.16\textwidth]{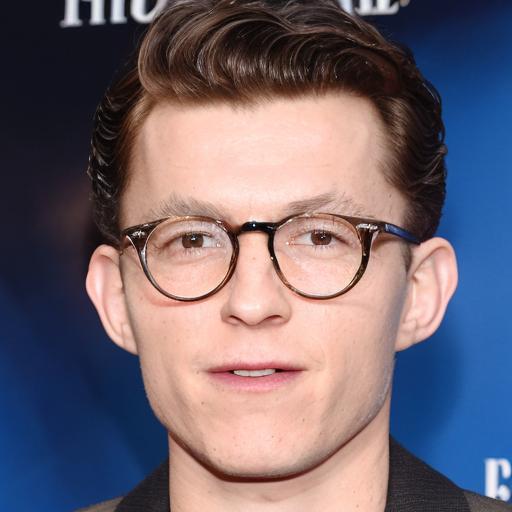} &
        		\includegraphics[width=0.16\textwidth]{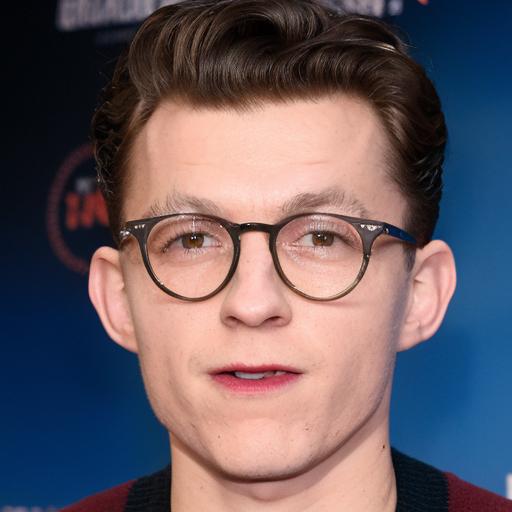} &
        		\includegraphics[width=0.16\textwidth]{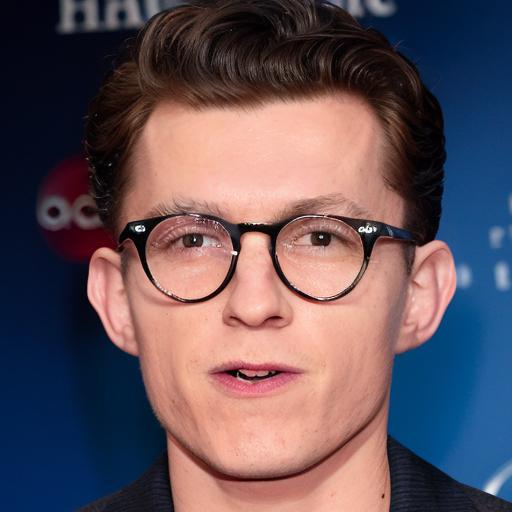} &
        		\includegraphics[width=0.16\textwidth]{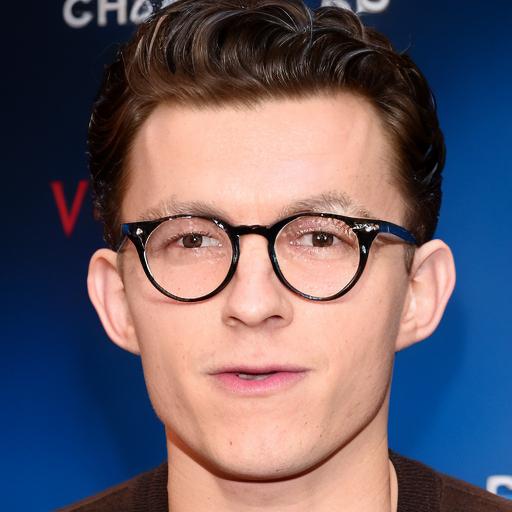} &
        		\includegraphics[width=0.16\textwidth]{figures/appendix/seed/320_8_gt.jpg}
        		\\
        		\footnotesize{LQ-Heavy} & \footnotesize{Seed = 0} & \footnotesize{Seed = 1} & \footnotesize{Seed = 2} & \footnotesize{Seed = 3} & \footnotesize{GT}
        		\\
        		\vspace{-1cm}
        	\end{tabular}
        \end{subfigure}%
	\end{center}
	\caption{
	Random seed effect.
	}
	\label{fig:seed}
\end{figure}

\paragraph{Randomness with Different Seeds}
Diffusion models are stochastic and notorious for unsatisfactory results with different random initializations. \Cref{fig:seed} contains results for an image with light and heavy degradations with four different seeds. For the light portion, the outputs tend to be mostly similar with small differences like skin texture. With the heavy portion, there are noticeable differences in the mouth, eyes and colors, although the identity is kept the same. Interestingly the background logo and text deviate largely, as they are not part of the learned personalization.

\begin{figure*}
	\begin{center}
    	\setlength{\tabcolsep}{1pt}
        \begin{subfigure}{0.98\textwidth}
        \hspace{-0.2cm}
        	\begin{tabular}{*6c}
        		\includegraphics[width=0.167\textwidth]{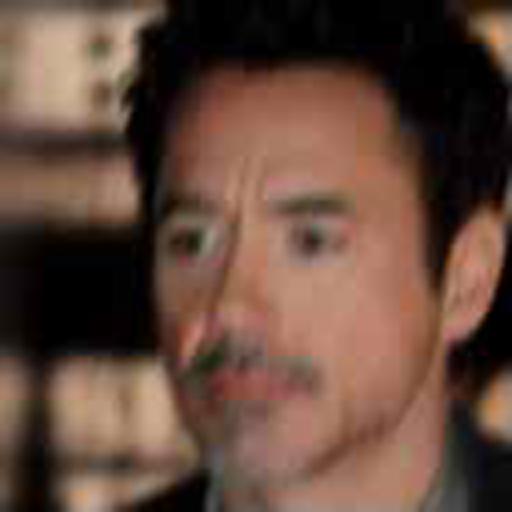} &
        		\includegraphics[width=0.167\textwidth]{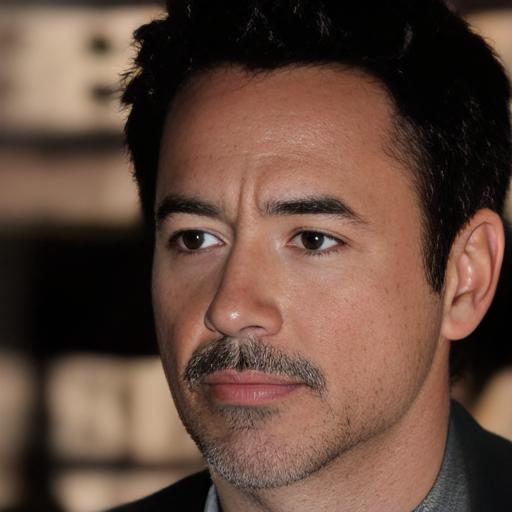} &
        		\includegraphics[width=0.167\textwidth]{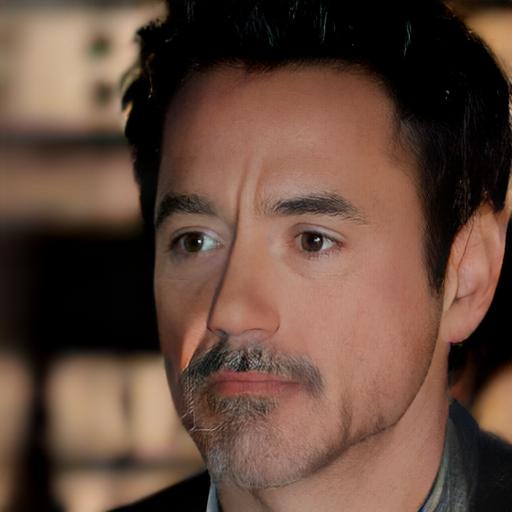} &
        		\includegraphics[width=0.167\textwidth]{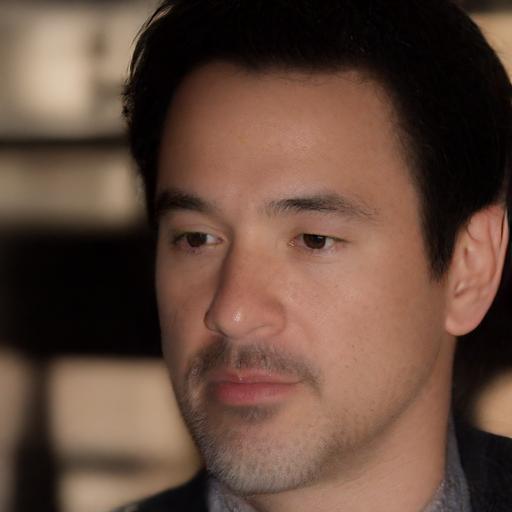} &
        		\includegraphics[width=0.167\textwidth]{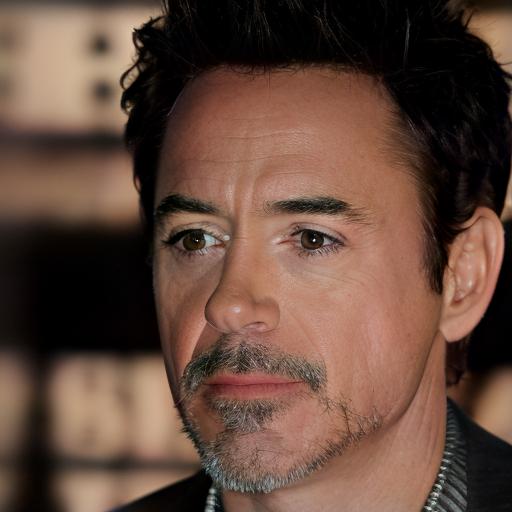} &
        		\includegraphics[width=0.167\textwidth]{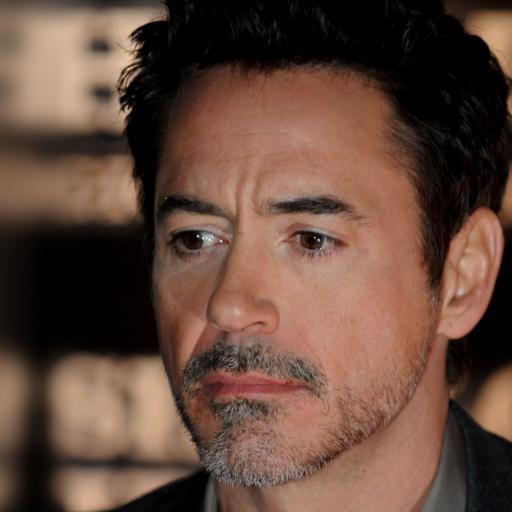}
        		\\
        		\includegraphics[width=0.167\textwidth]{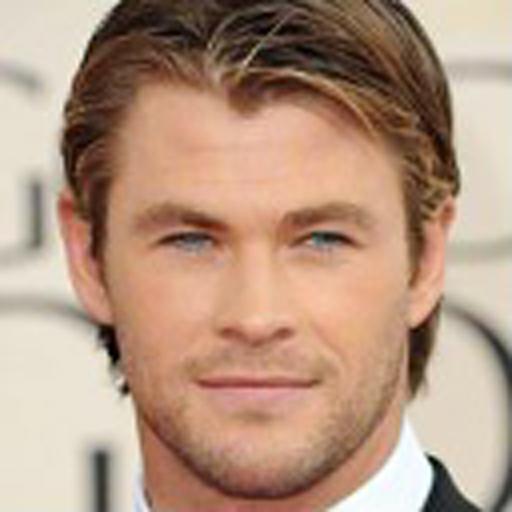} &
        		\includegraphics[width=0.167\textwidth]{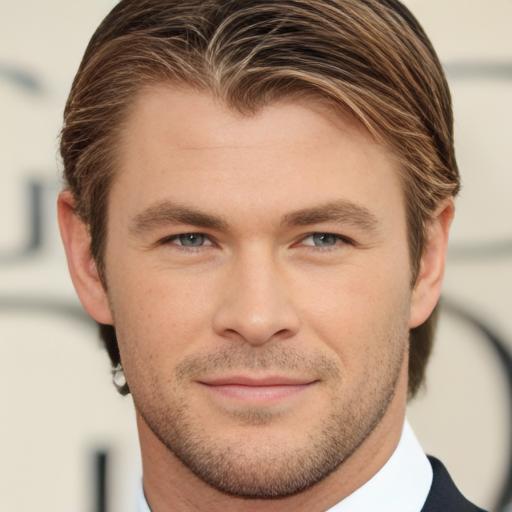} &
        		\includegraphics[width=0.167\textwidth]{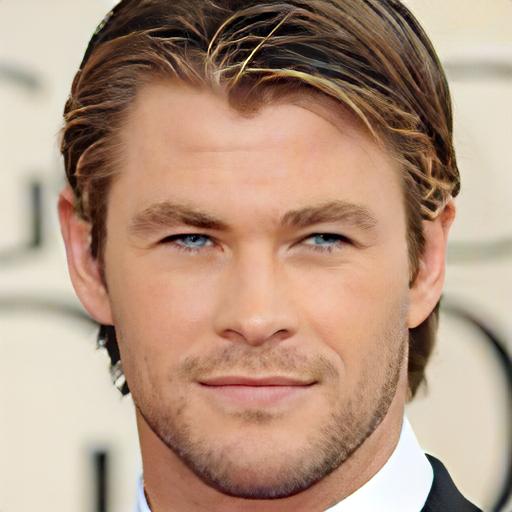} &
        		\includegraphics[width=0.167\textwidth]{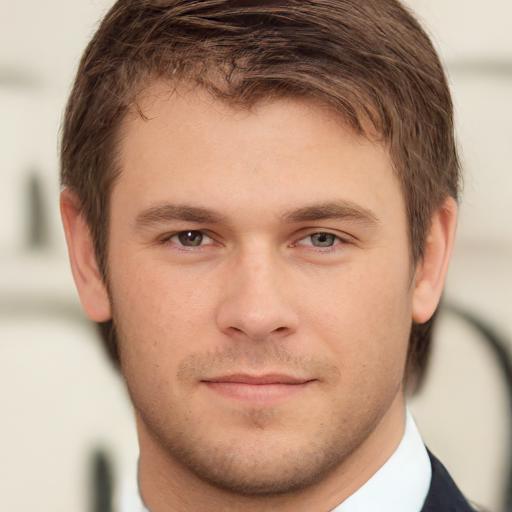} &
        		\includegraphics[width=0.167\textwidth]{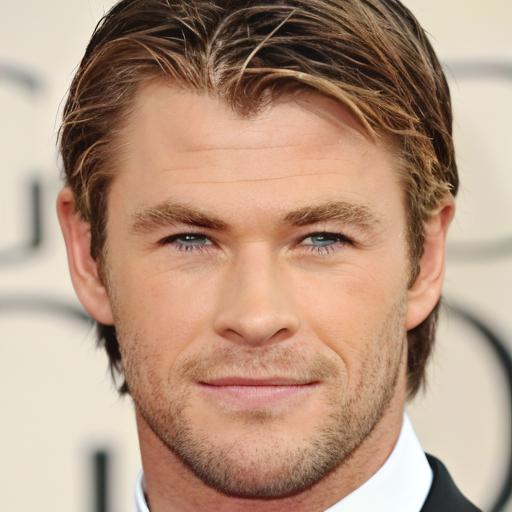} &
        		\includegraphics[width=0.167\textwidth]{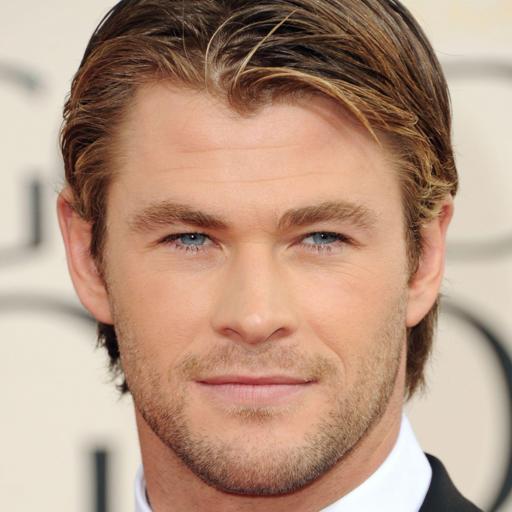}
        		\\
        		\includegraphics[width=0.167\textwidth]{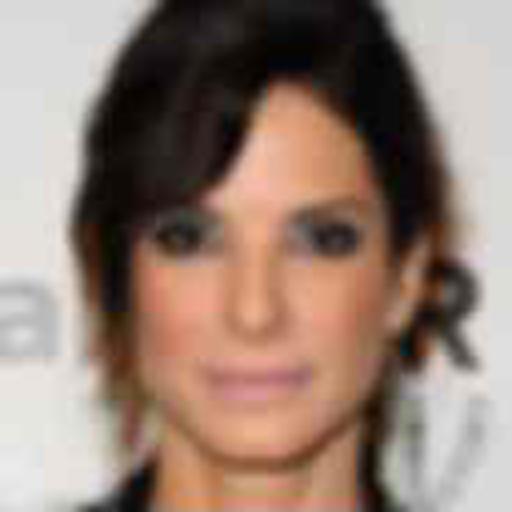} &
        		\includegraphics[width=0.167\textwidth]{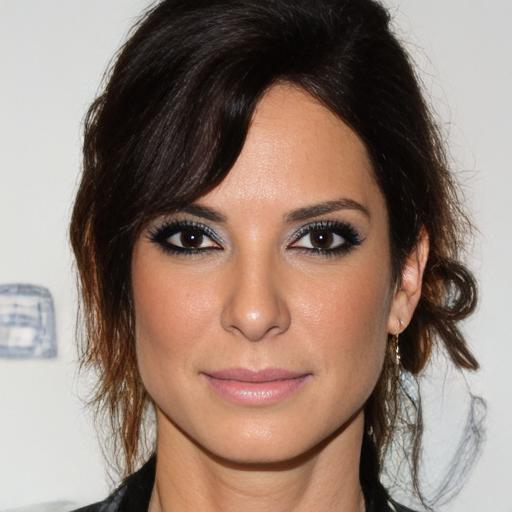} &
        		\includegraphics[width=0.167\textwidth]{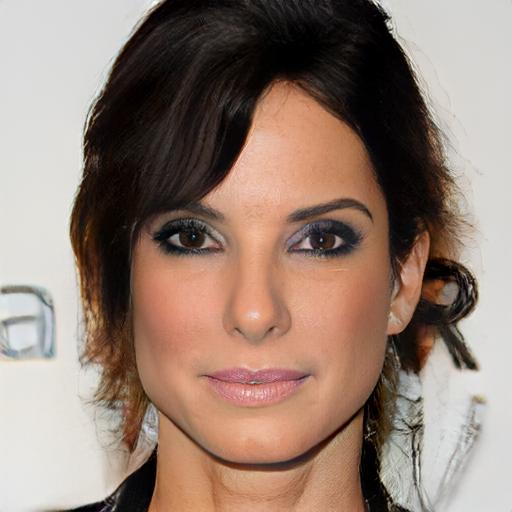} &
        		\includegraphics[width=0.167\textwidth]{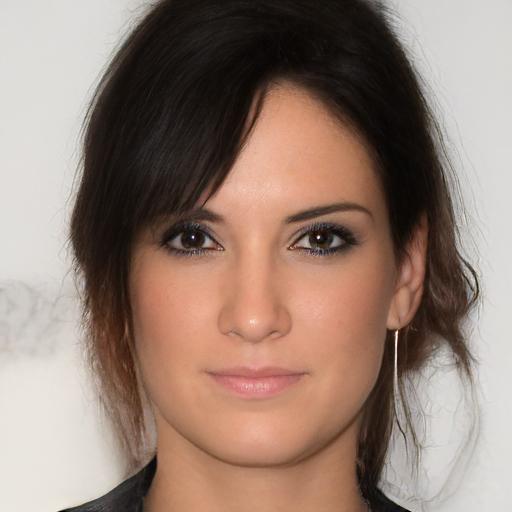} &
        		\includegraphics[width=0.167\textwidth]{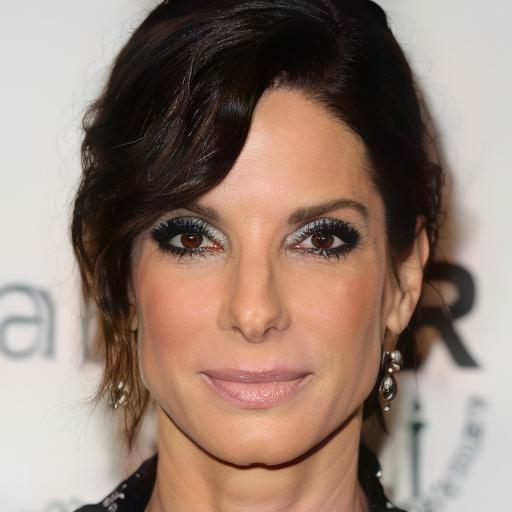} &
        		\includegraphics[width=0.167\textwidth]{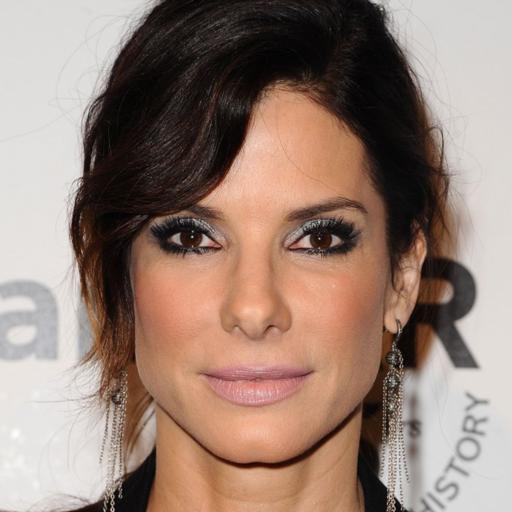}
        		\\
        		\includegraphics[width=0.167\textwidth]{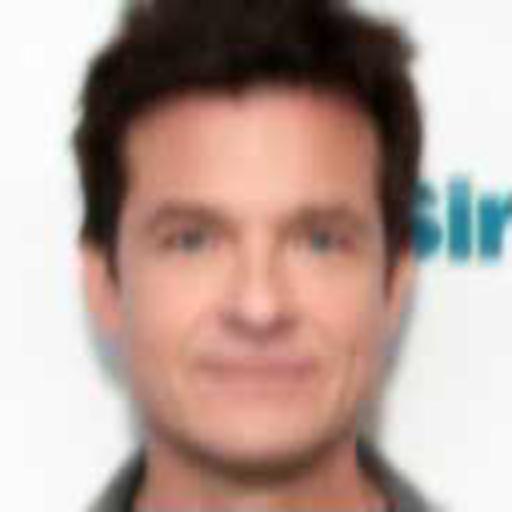} &
        		\includegraphics[width=0.167\textwidth]{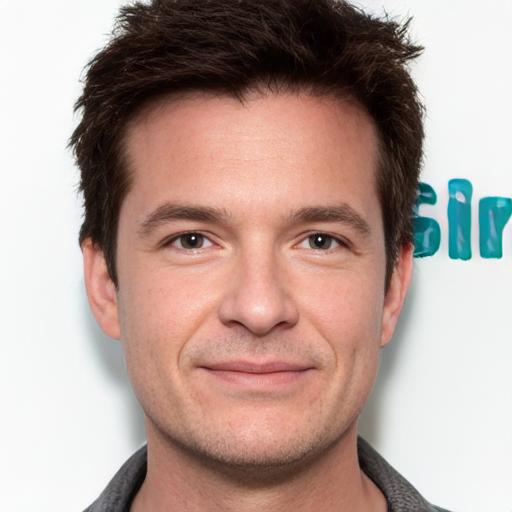} &
        		\includegraphics[width=0.167\textwidth]{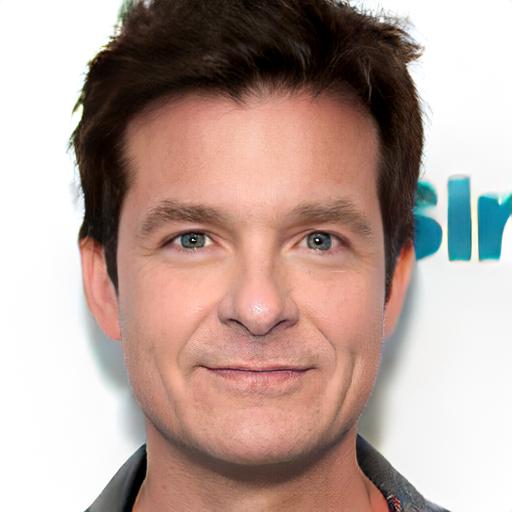} &
        		\includegraphics[width=0.167\textwidth]{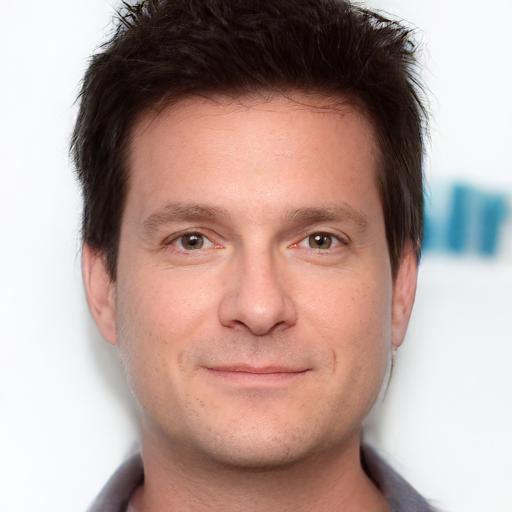} &
        		\includegraphics[width=0.167\textwidth]{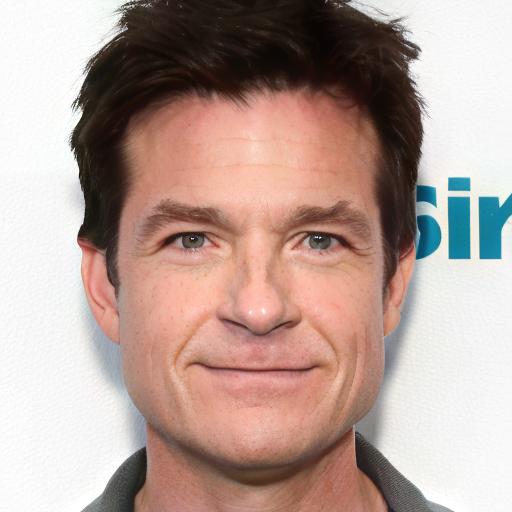} &
        		\includegraphics[width=0.167\textwidth]{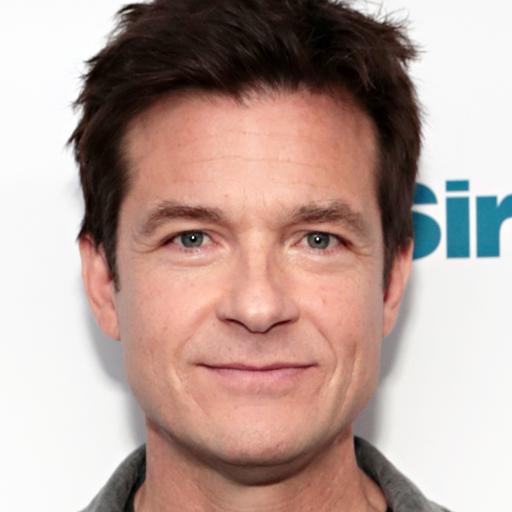}
        		\\
        		\includegraphics[width=0.167\textwidth]{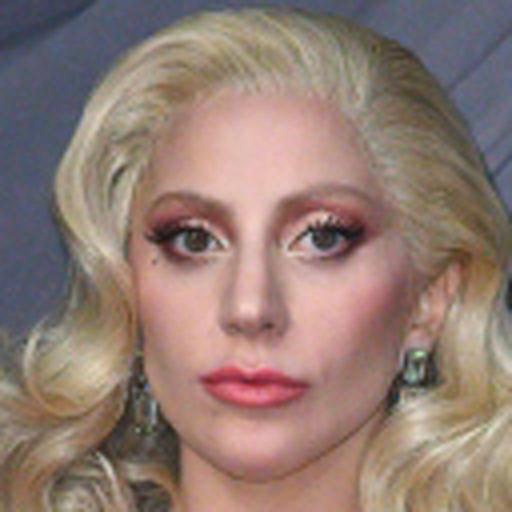} &
        		\includegraphics[width=0.167\textwidth]{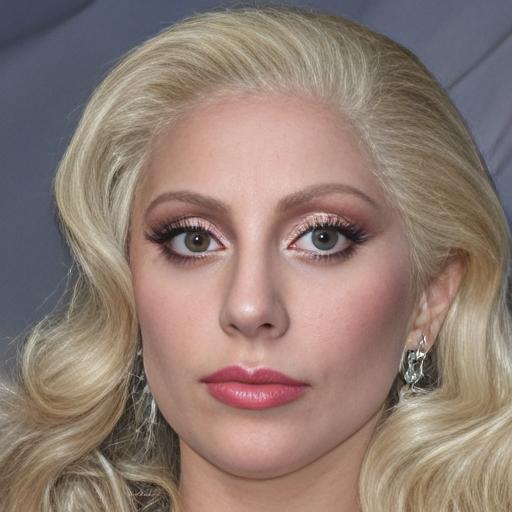} &
        		\includegraphics[width=0.167\textwidth]{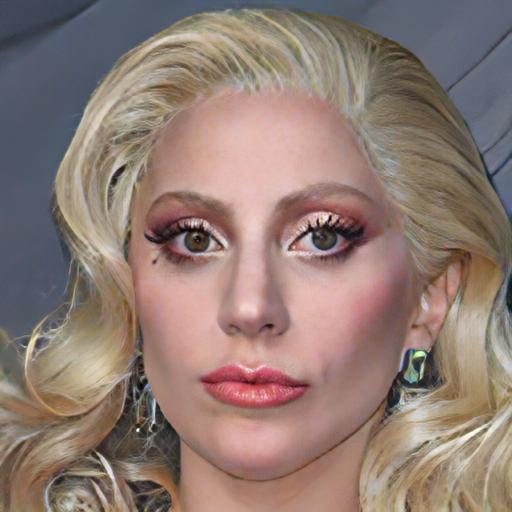} &
        		\includegraphics[width=0.167\textwidth]{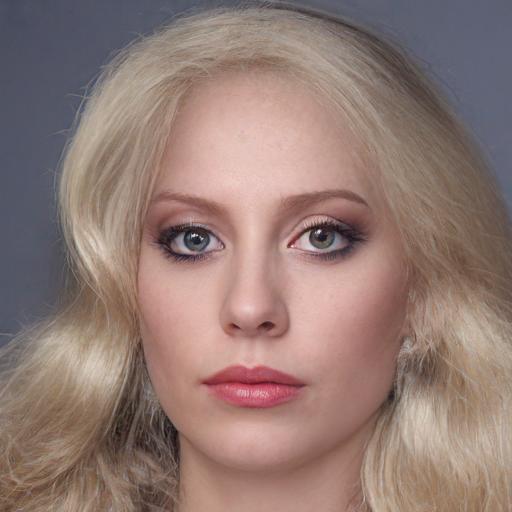} &
        		\includegraphics[width=0.167\textwidth]{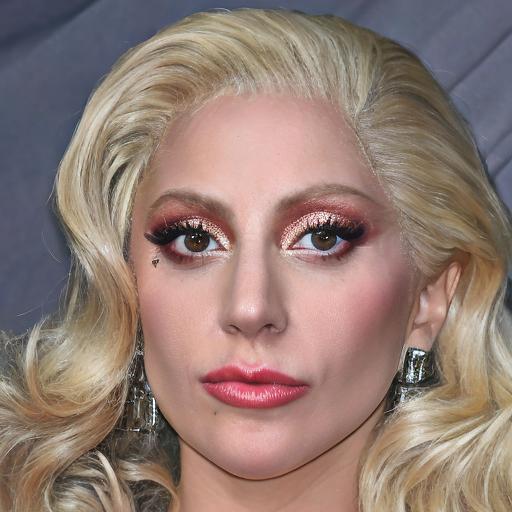} &
        		\includegraphics[width=0.167\textwidth]{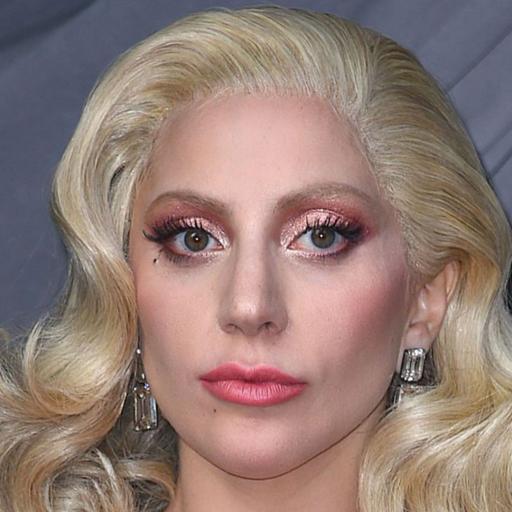}
        		\\
        		\includegraphics[width=0.167\textwidth]{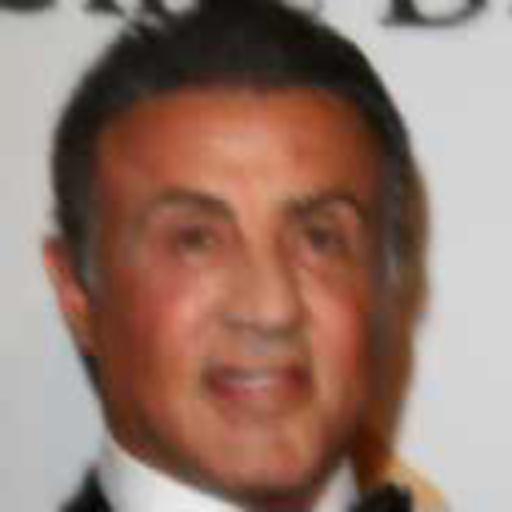} &
        		\includegraphics[width=0.167\textwidth]{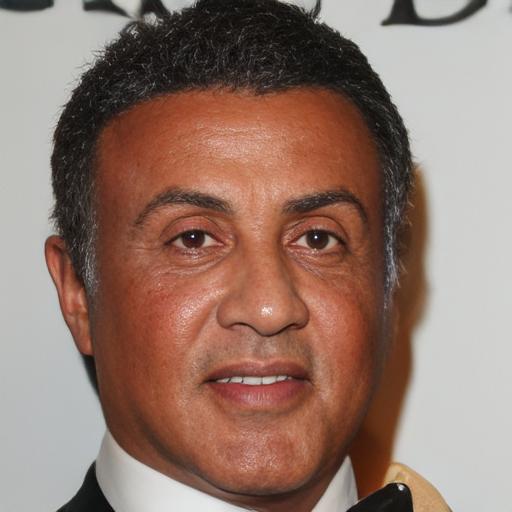} &
        		\includegraphics[width=0.167\textwidth]{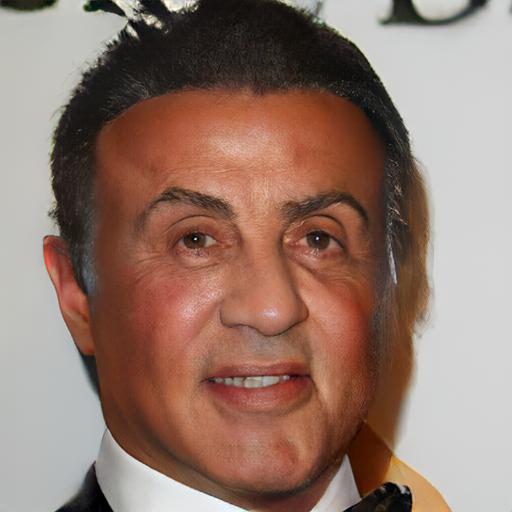} &
        		\includegraphics[width=0.167\textwidth]{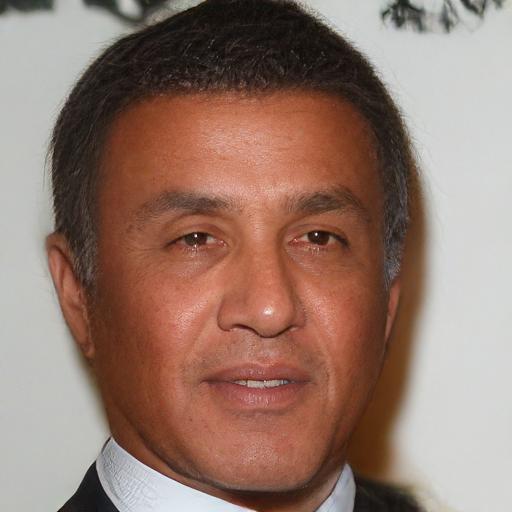} &
        		\includegraphics[width=0.167\textwidth]{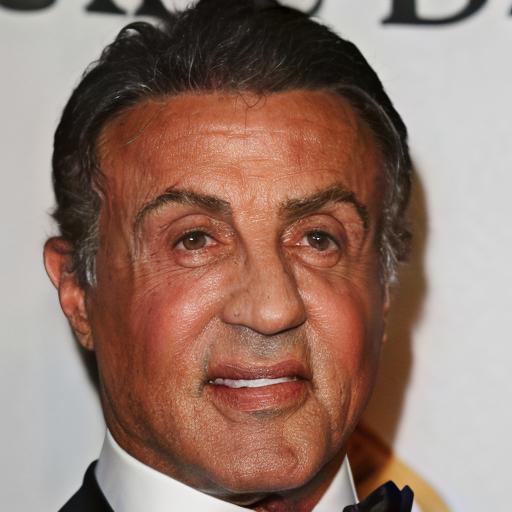} &
        		\includegraphics[width=0.167\textwidth]{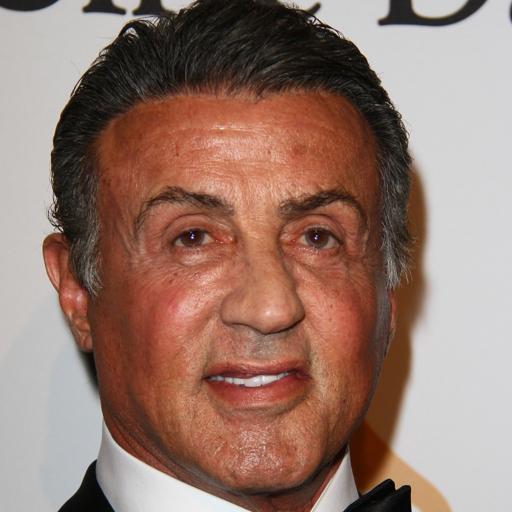}
        		\\
        		\footnotesize{Input} & \footnotesize{CodeFormer} & \footnotesize{DMDNet} & \footnotesize{DR2+SPAR} & \footnotesize{Ours} & \footnotesize{GT} \\
        		\vspace{-1cm}
        	\end{tabular}
        \end{subfigure}%
	\end{center}
	\caption{
	Qualitative comparison with state-of-the-art restoration models on Celeb-Ref dataset \cite{dmdnet} with synthetic light degradation.
	}
	\label{fig:celeb_ref_light2}
\end{figure*}

\begin{figure*}
	\begin{center}
    	\setlength{\tabcolsep}{1pt}
        \begin{subfigure}{0.98\textwidth}
        \hspace{-0.2cm}
        	\begin{tabular}{*6c}
        		\includegraphics[width=0.167\textwidth]{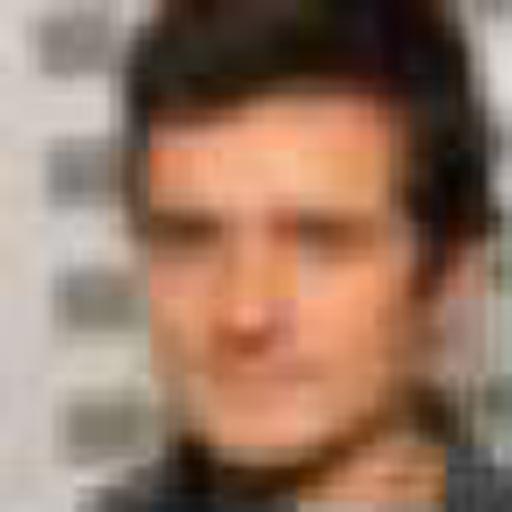} &
        		\includegraphics[width=0.167\textwidth]{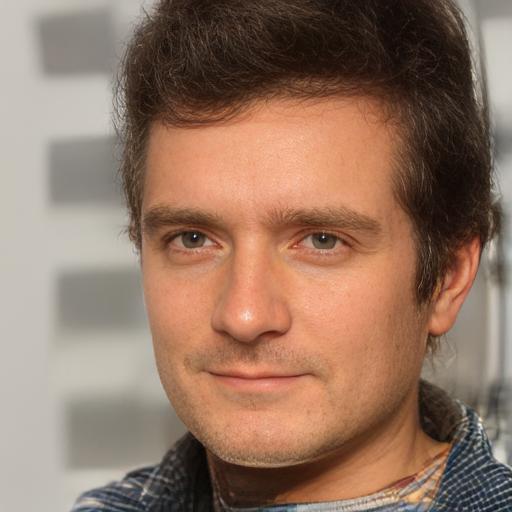} &
        		\includegraphics[width=0.167\textwidth]{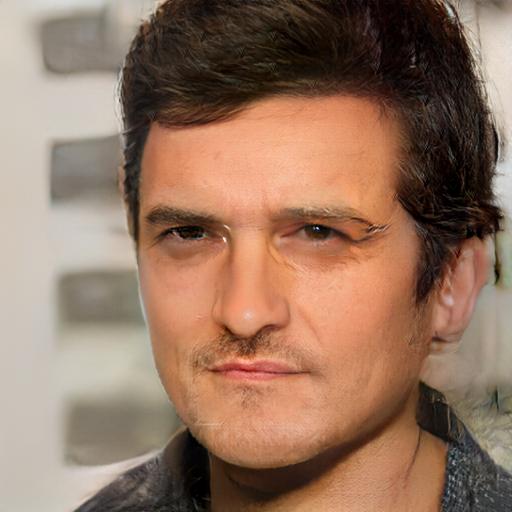} &
        		\includegraphics[width=0.167\textwidth]{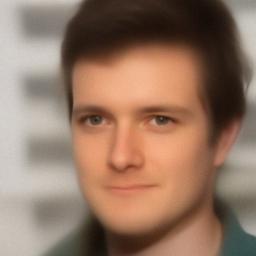} &
        		\includegraphics[width=0.167\textwidth]{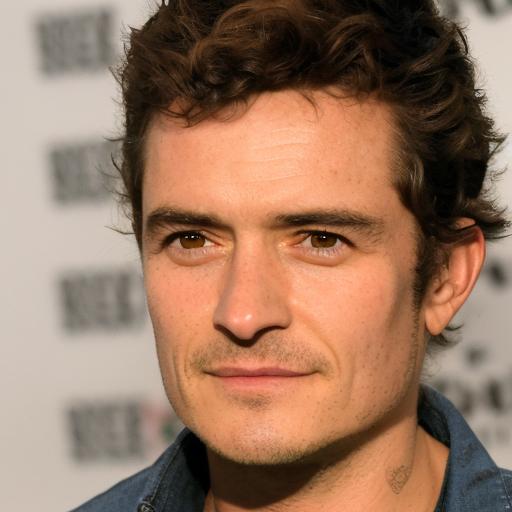} &
        		\includegraphics[width=0.167\textwidth]{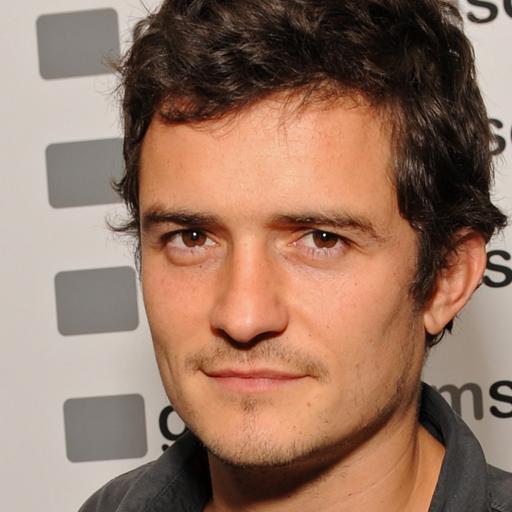}
        		\\
        		\includegraphics[width=0.167\textwidth]{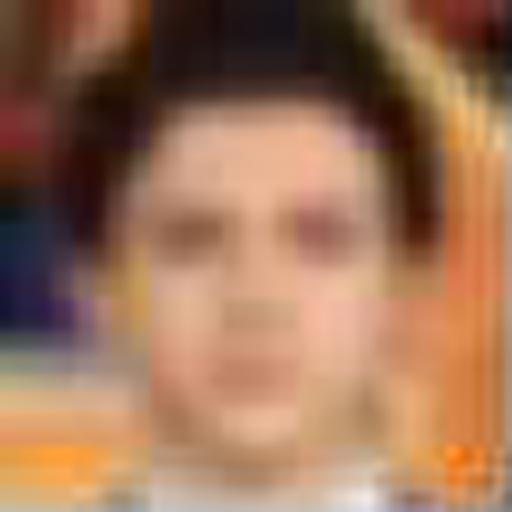} &
        		\includegraphics[width=0.167\textwidth]{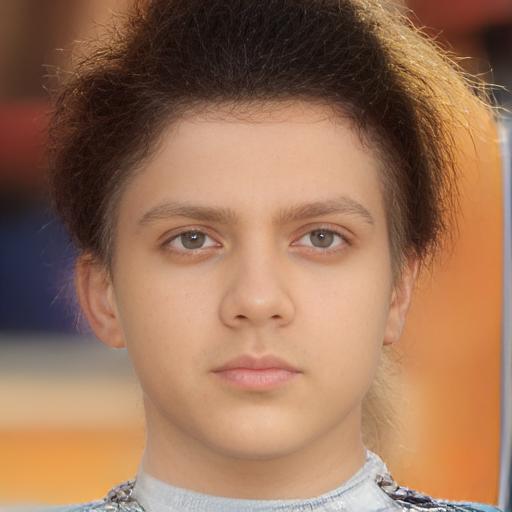} &
        		\includegraphics[width=0.167\textwidth]{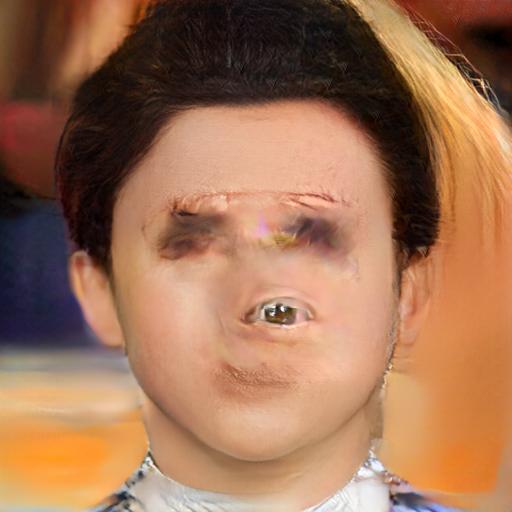} &
        		\includegraphics[width=0.167\textwidth]{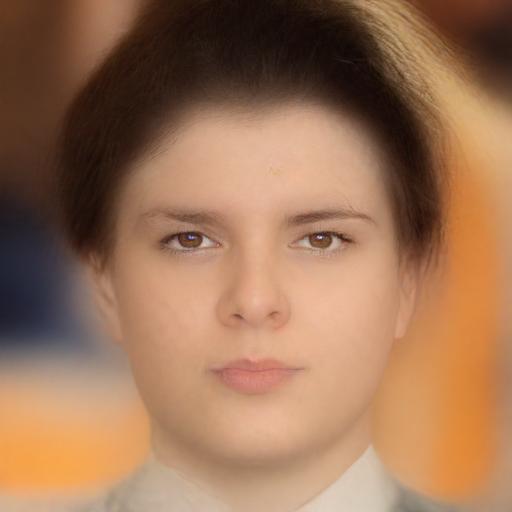} &
        		\includegraphics[width=0.167\textwidth]{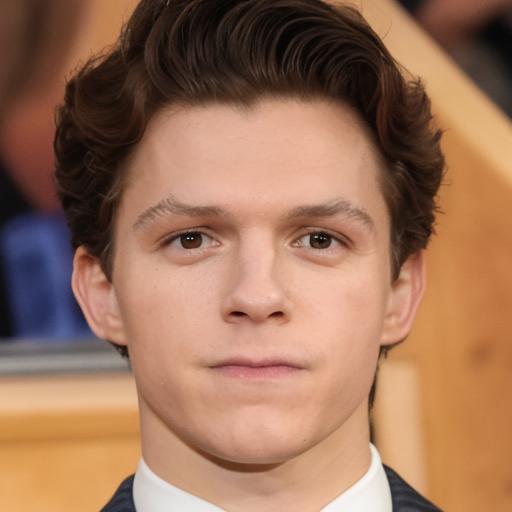} &
        		\includegraphics[width=0.167\textwidth]{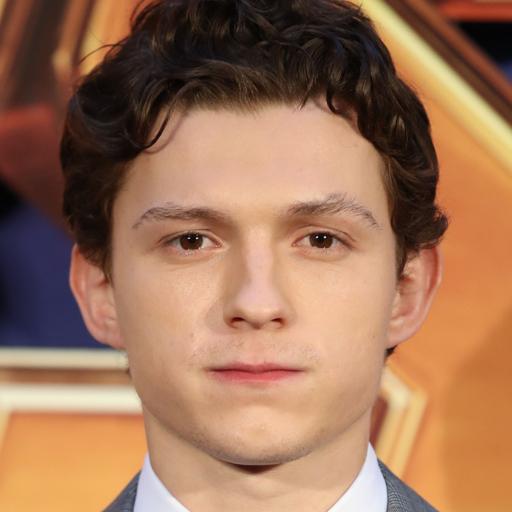}
        		\\
        		\includegraphics[width=0.167\textwidth]{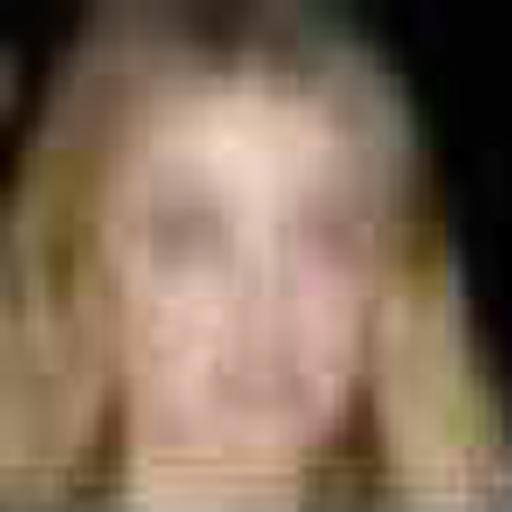} &
        		\includegraphics[width=0.167\textwidth]{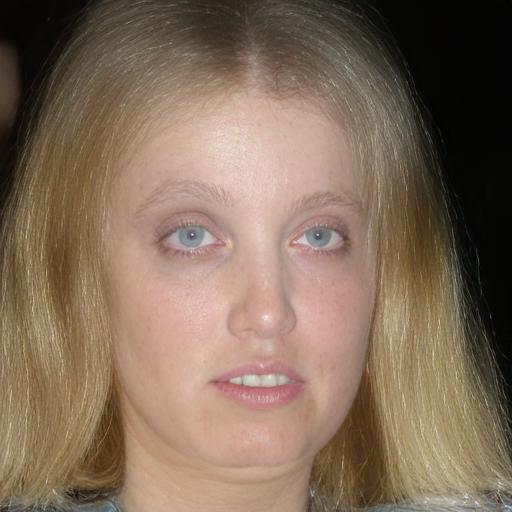} &
        		\includegraphics[width=0.167\textwidth]{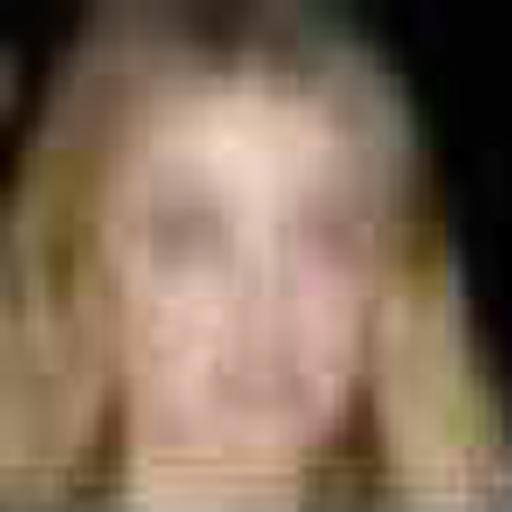} &
        		\includegraphics[width=0.167\textwidth]{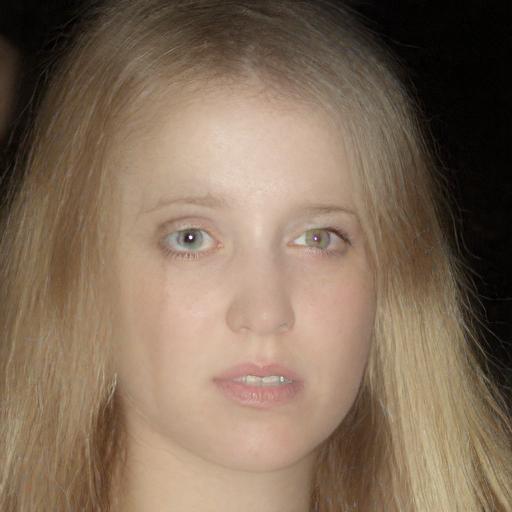} &
        		\includegraphics[width=0.167\textwidth]{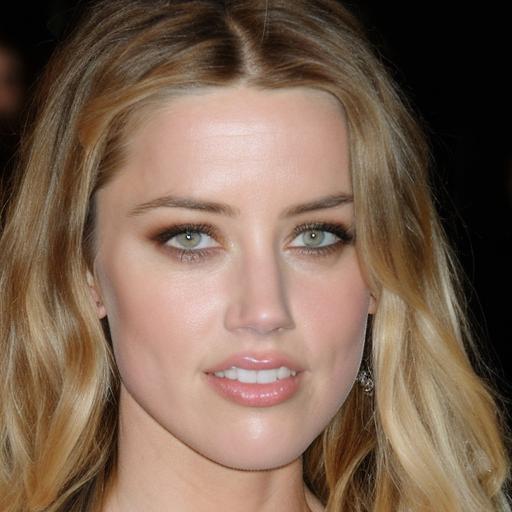} &
        		\includegraphics[width=0.167\textwidth]{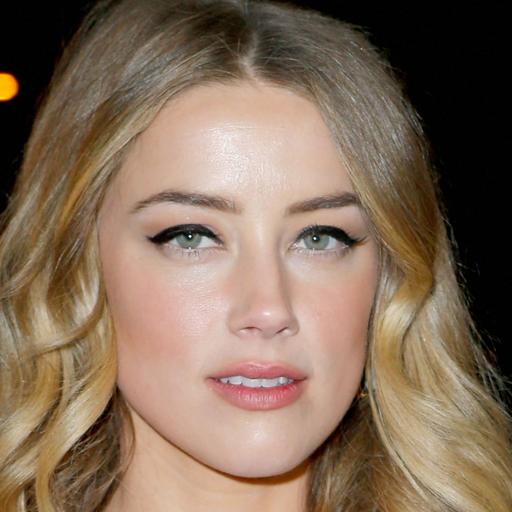}
        		\\
        		\includegraphics[width=0.167\textwidth]{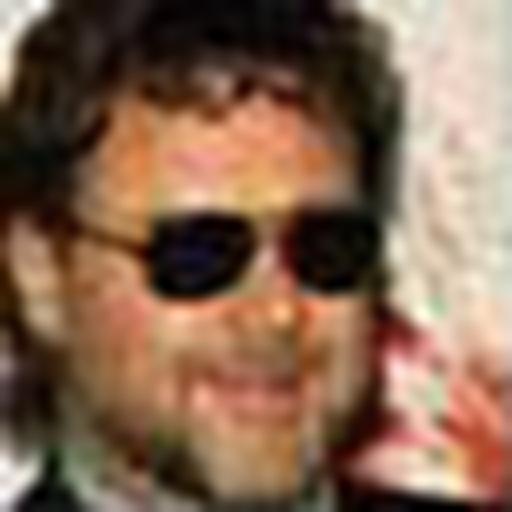} &
        		\includegraphics[width=0.167\textwidth]{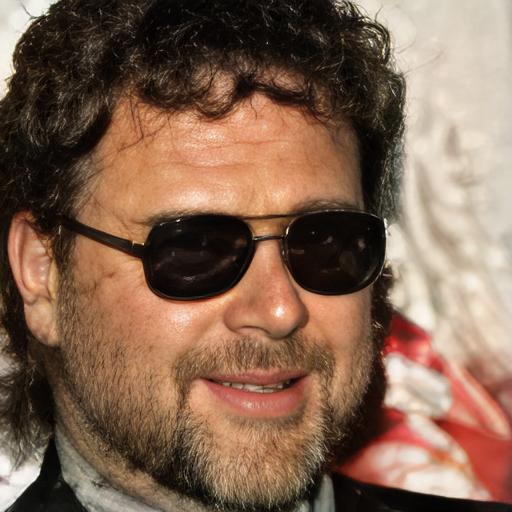} &
        		\includegraphics[width=0.167\textwidth]{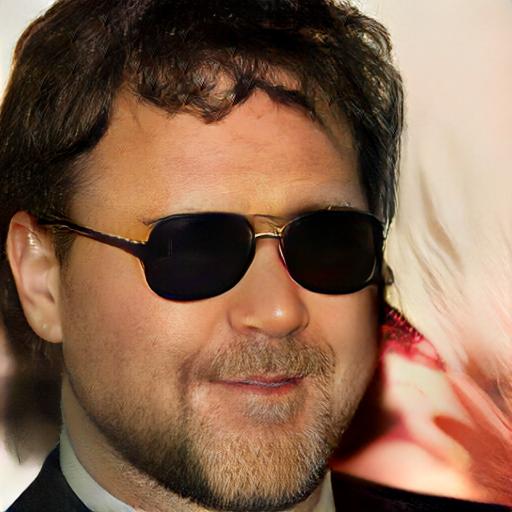} &
        		\includegraphics[width=0.167\textwidth]{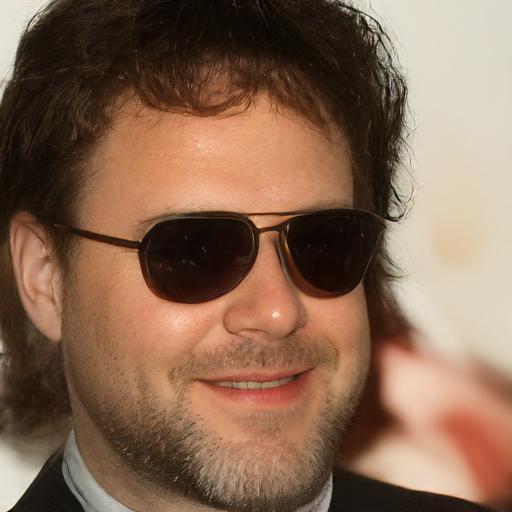} &
        		\includegraphics[width=0.167\textwidth]{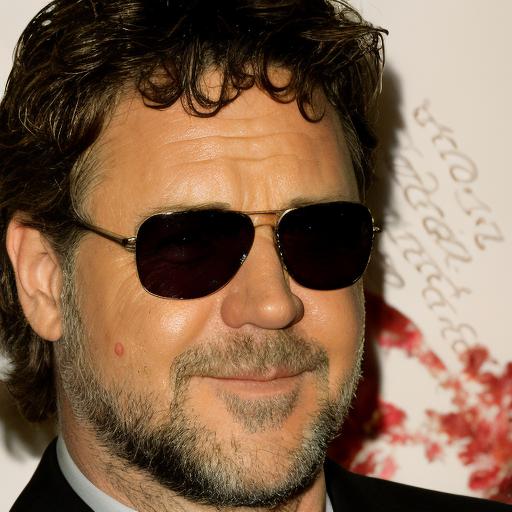} &
        		\includegraphics[width=0.167\textwidth]{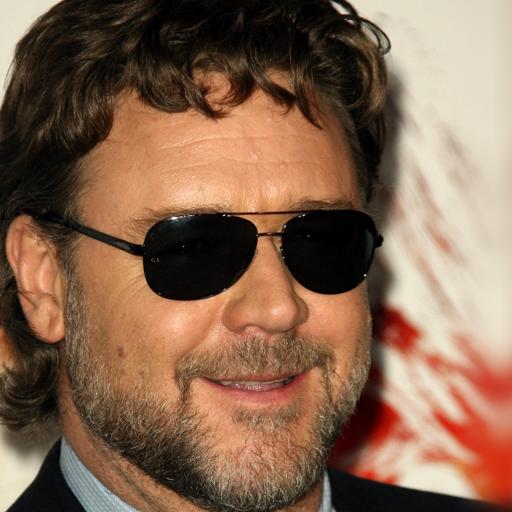}
        		\\
        		\includegraphics[width=0.167\textwidth]{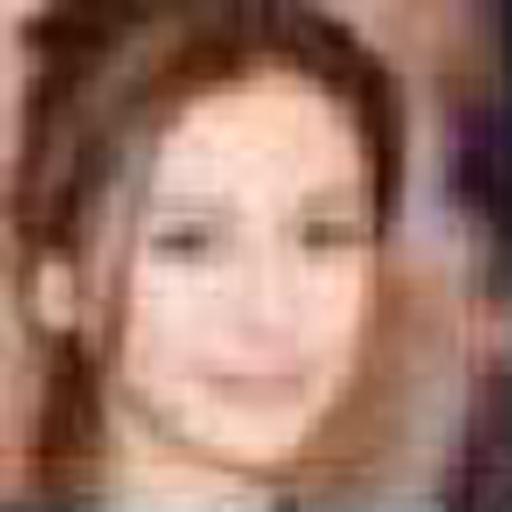} &
        		\includegraphics[width=0.167\textwidth]{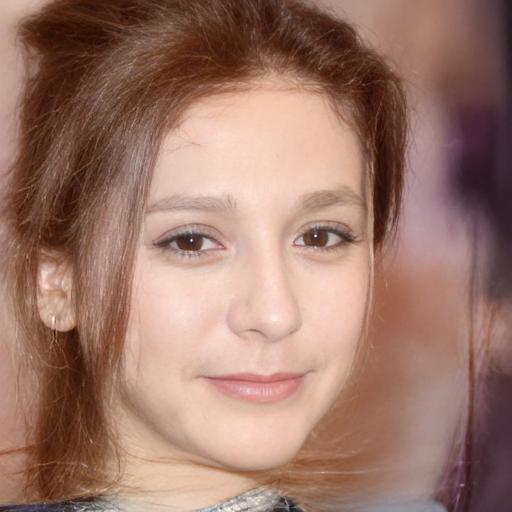} &
        		\includegraphics[width=0.167\textwidth]{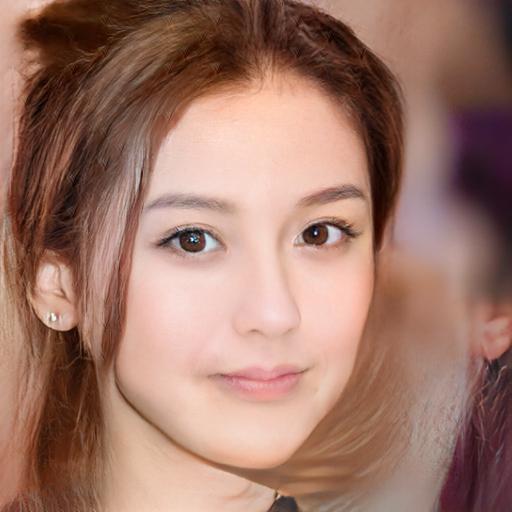} &
        		\includegraphics[width=0.167\textwidth]{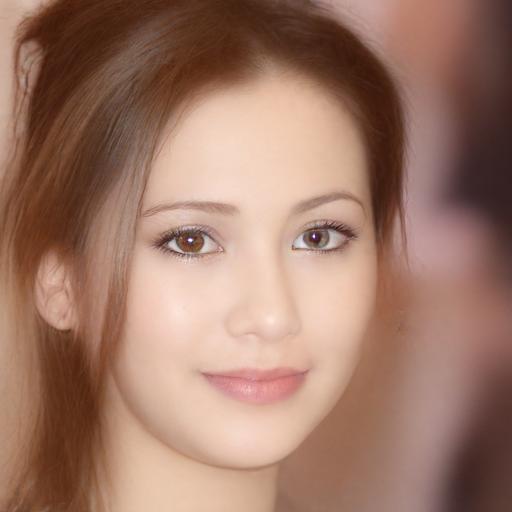} &
        		\includegraphics[width=0.167\textwidth]{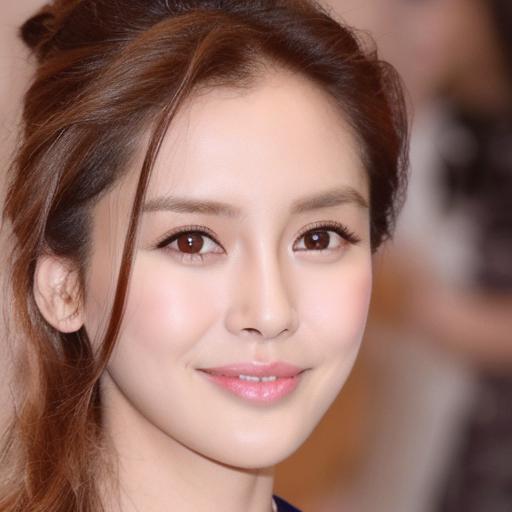} &
        		\includegraphics[width=0.167\textwidth]{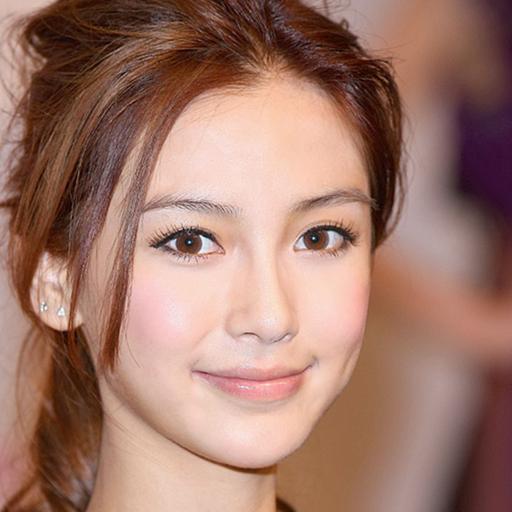}
        		\\
        		\includegraphics[width=0.167\textwidth]{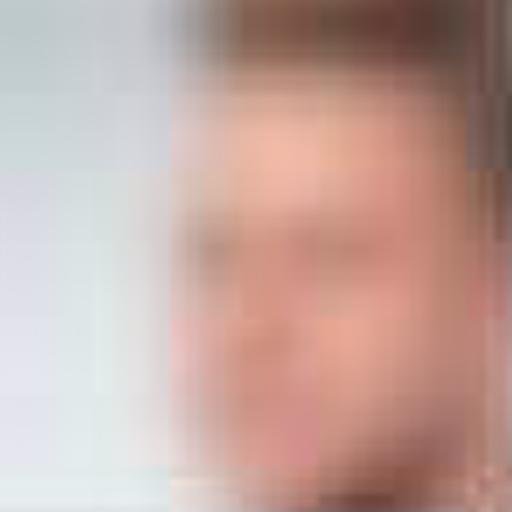} &
        		\includegraphics[width=0.167\textwidth]{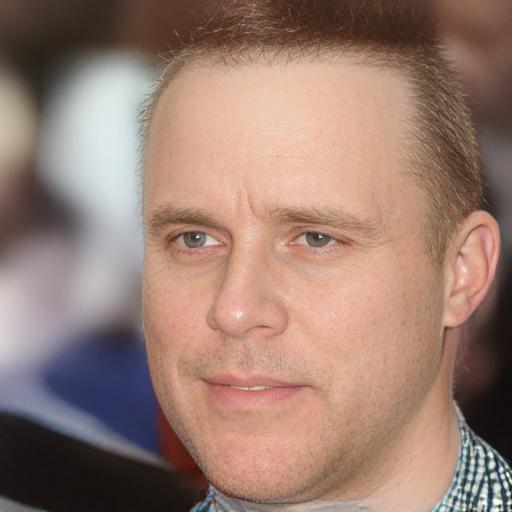} &
        		\includegraphics[width=0.167\textwidth]{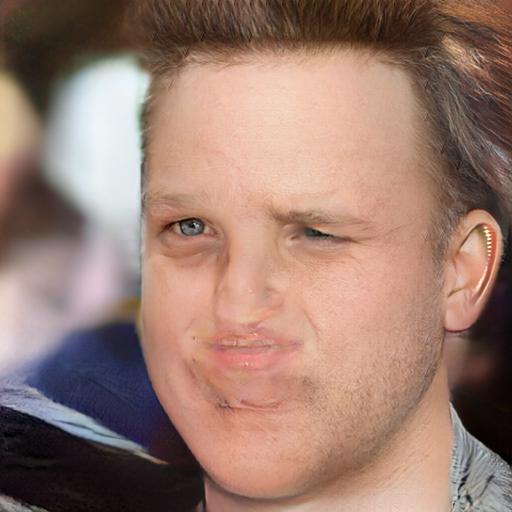} &
        		\includegraphics[width=0.167\textwidth]{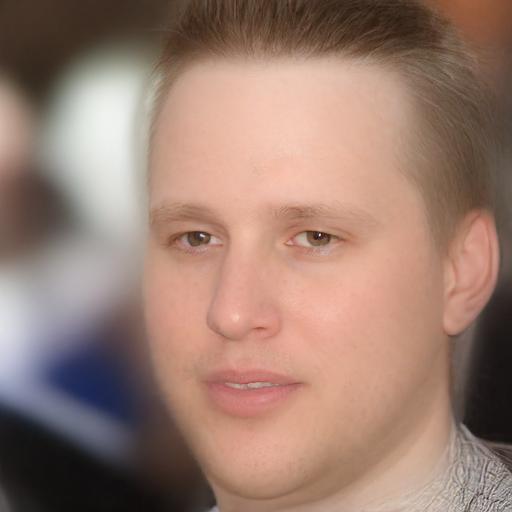} &
        		\includegraphics[width=0.167\textwidth]{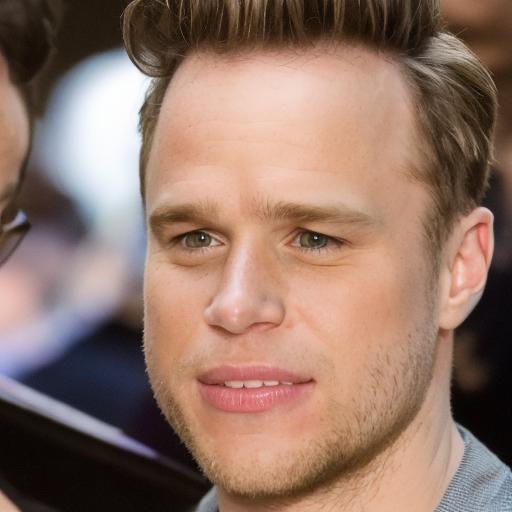} &
        		\includegraphics[width=0.167\textwidth]{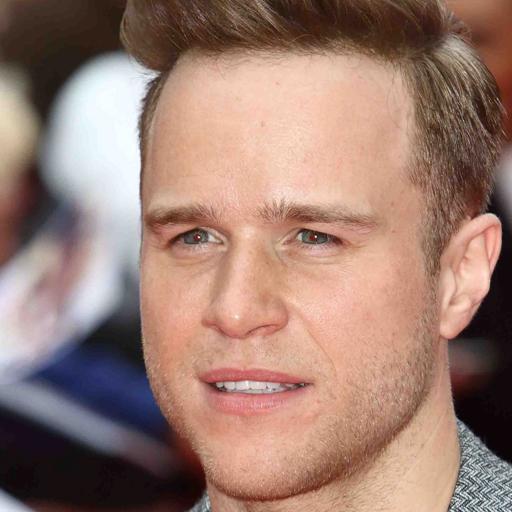}
        		\\
        		\footnotesize{Input} & \footnotesize{CodeFormer} & \footnotesize{DMDNet} & \footnotesize{DR2+SPAR} & \footnotesize{Ours} & \footnotesize{GT} \\
        		\vspace{-1cm}
        	\end{tabular}
        \end{subfigure}%
	\end{center}
	\caption{
	Qualitative comparison with state-of-the-art restoration models on Celeb-Ref dataset \cite{dmdnet} with synthetic heavy degradation.
	}
	\label{fig:celeb_ref_heavy2}
\end{figure*}

\begin{figure*}
	\begin{center}
    	\setlength{\tabcolsep}{1pt}
        \begin{subfigure}{0.98\textwidth}
        \hspace{-0.2cm}
        	\begin{tabular}{*6c}
        		\includegraphics[width=0.167\textwidth]{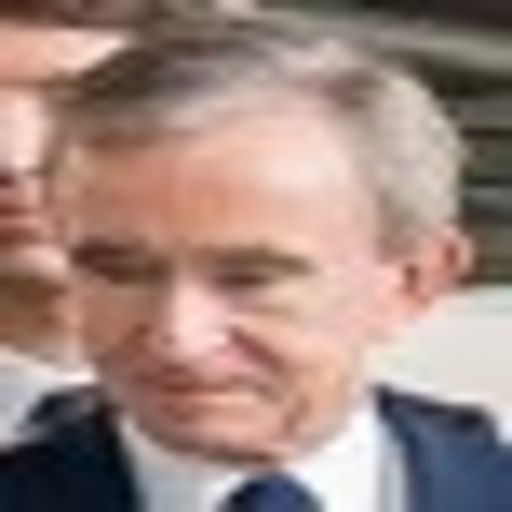} &
        		\includegraphics[width=0.167\textwidth]{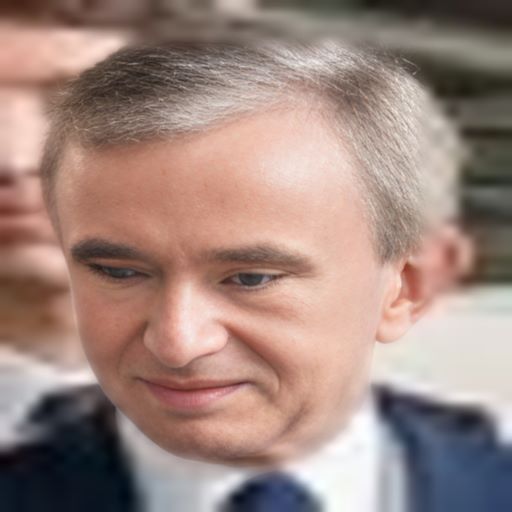} &
        		\includegraphics[width=0.167\textwidth]{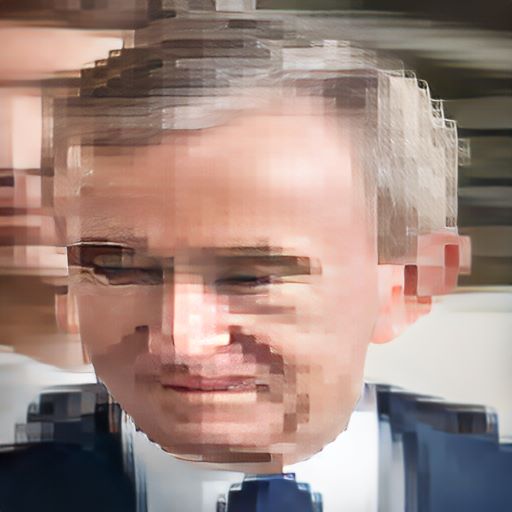} &
        		\includegraphics[width=0.167\textwidth]{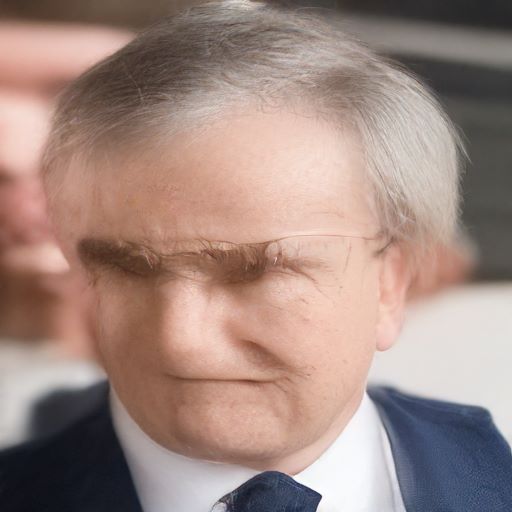} &
        		\includegraphics[width=0.167\textwidth]{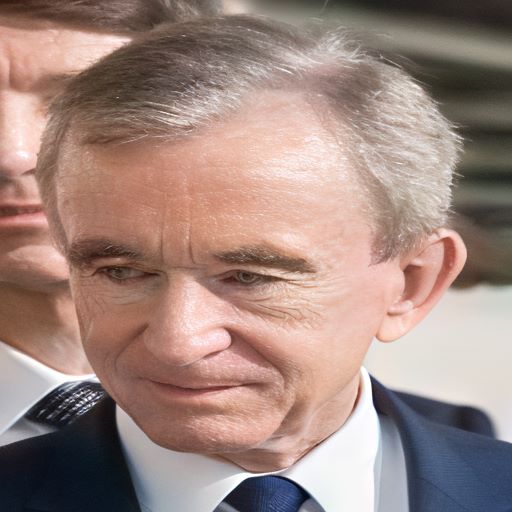} &
        		\includegraphics[width=0.167\textwidth]{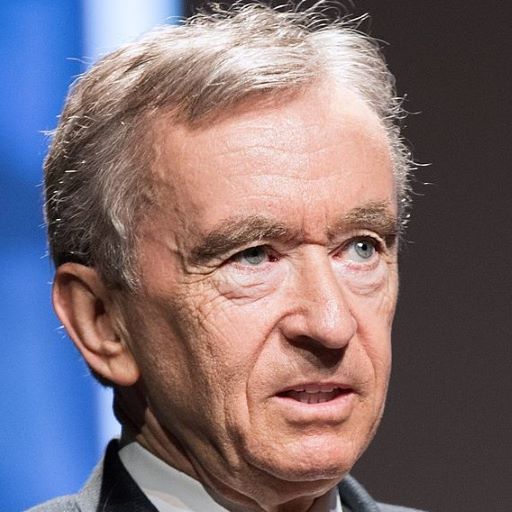}
        		\\
        		\includegraphics[width=0.167\textwidth]{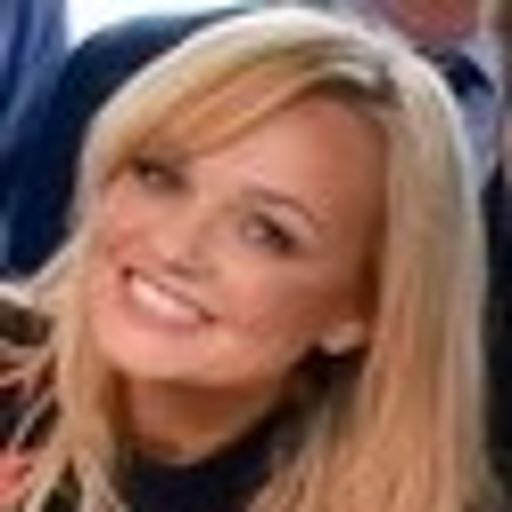} &
        		\includegraphics[width=0.167\textwidth]{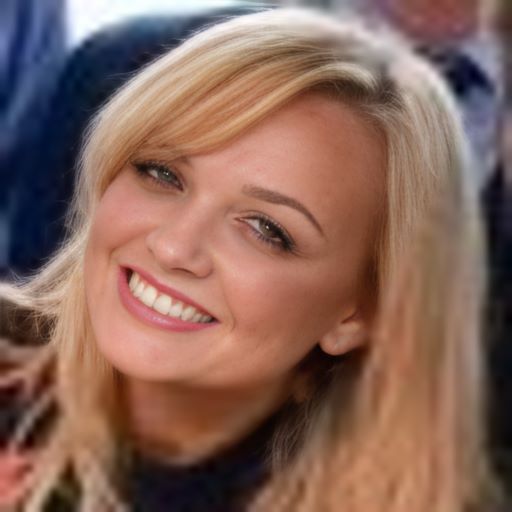} &
        		\includegraphics[width=0.167\textwidth]{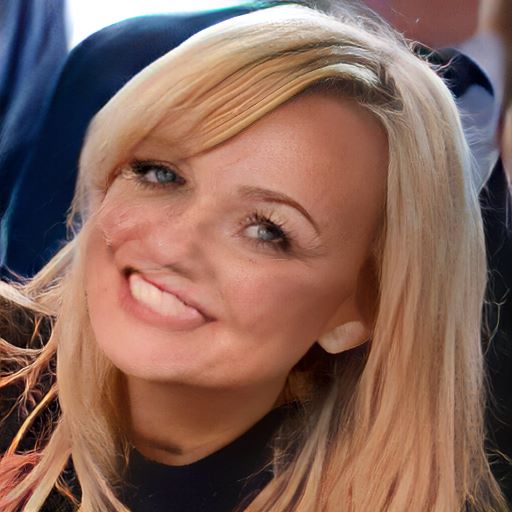} &
        		\includegraphics[width=0.167\textwidth]{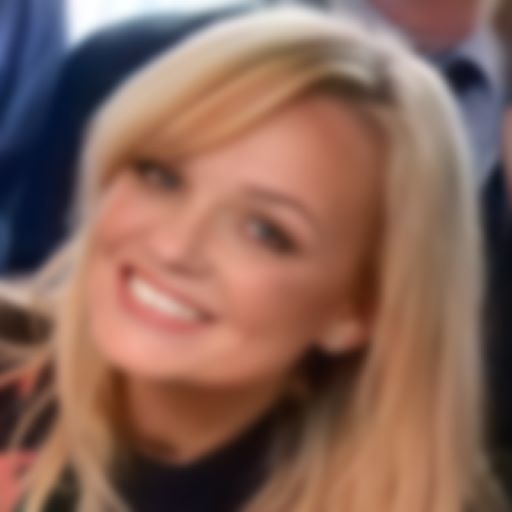} &
        		\includegraphics[width=0.167\textwidth]{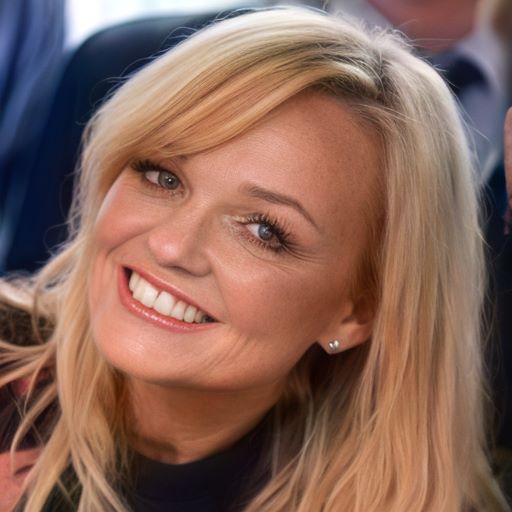} &
        		\includegraphics[width=0.167\textwidth]{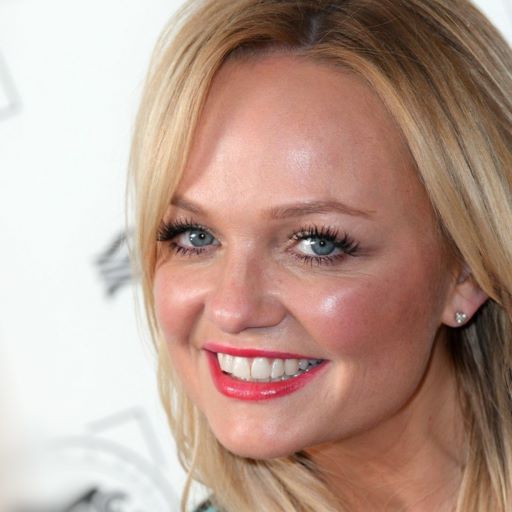}
        		\\
        		\includegraphics[width=0.167\textwidth]{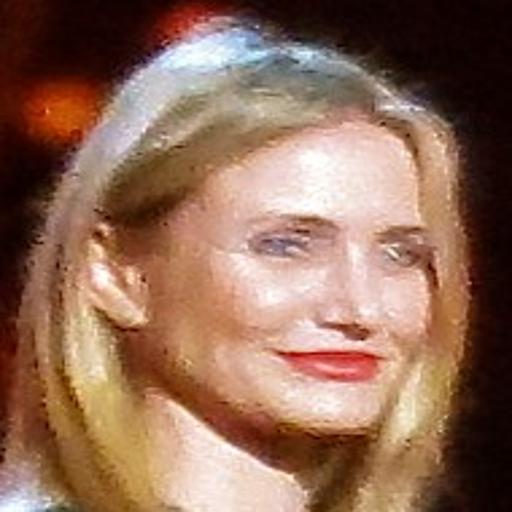} &
        		\includegraphics[width=0.167\textwidth]{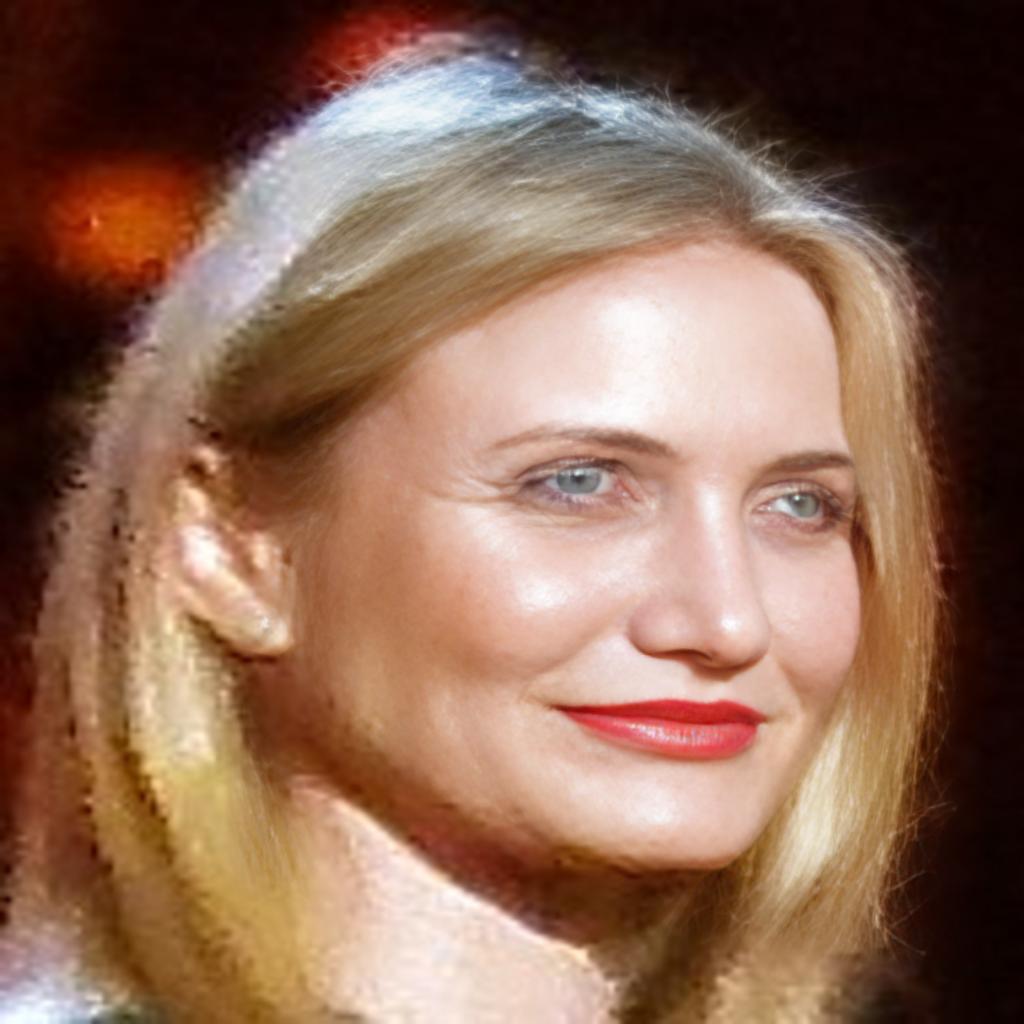} &
        		\includegraphics[width=0.167\textwidth]{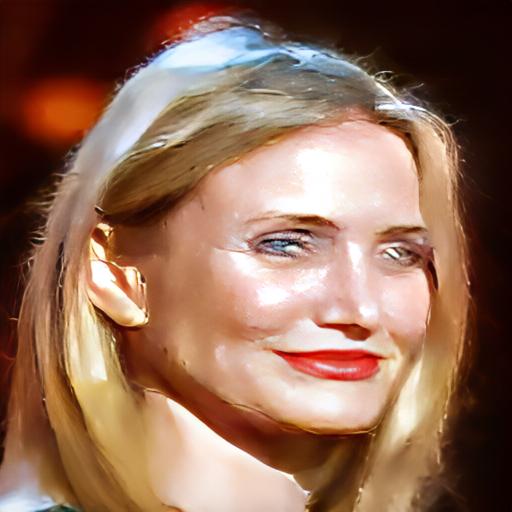} &
        		\includegraphics[width=0.167\textwidth]{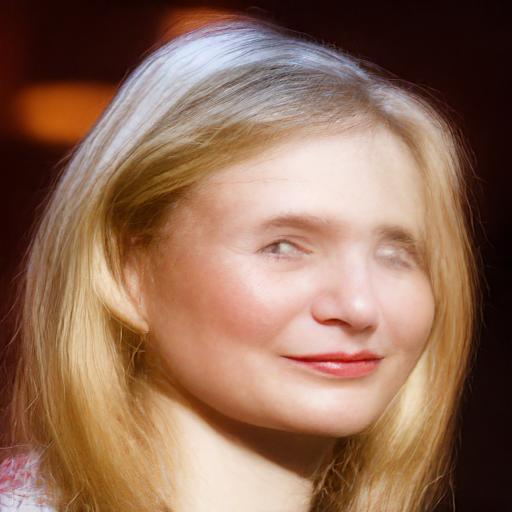} &
        		\includegraphics[width=0.167\textwidth]{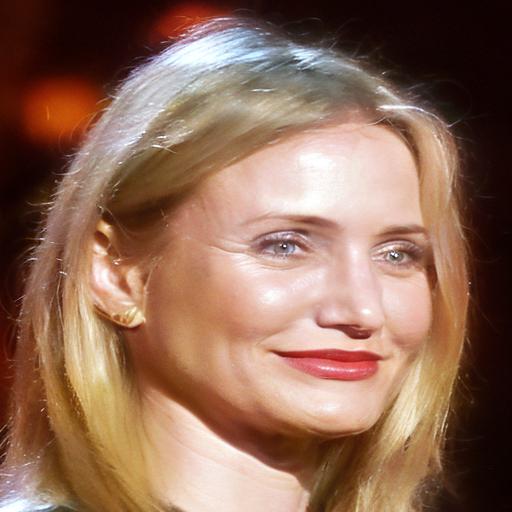} &
        		\includegraphics[width=0.167\textwidth]{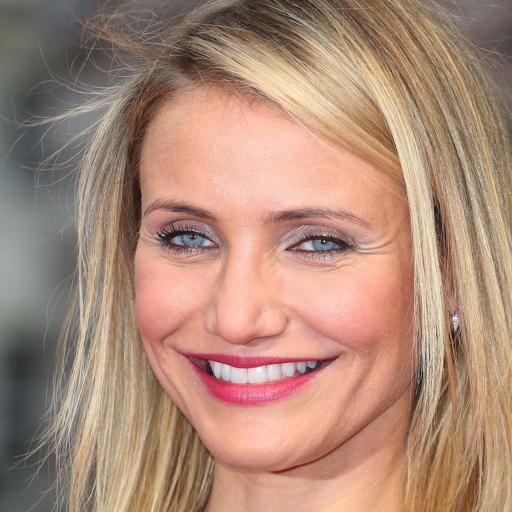}
        		\\
        		\includegraphics[width=0.167\textwidth]{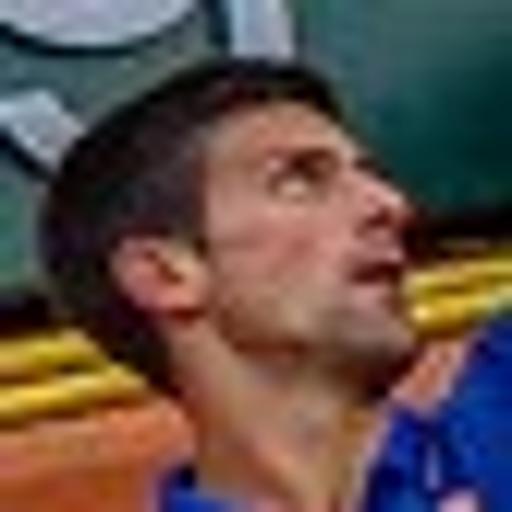} &
        		\includegraphics[width=0.167\textwidth]{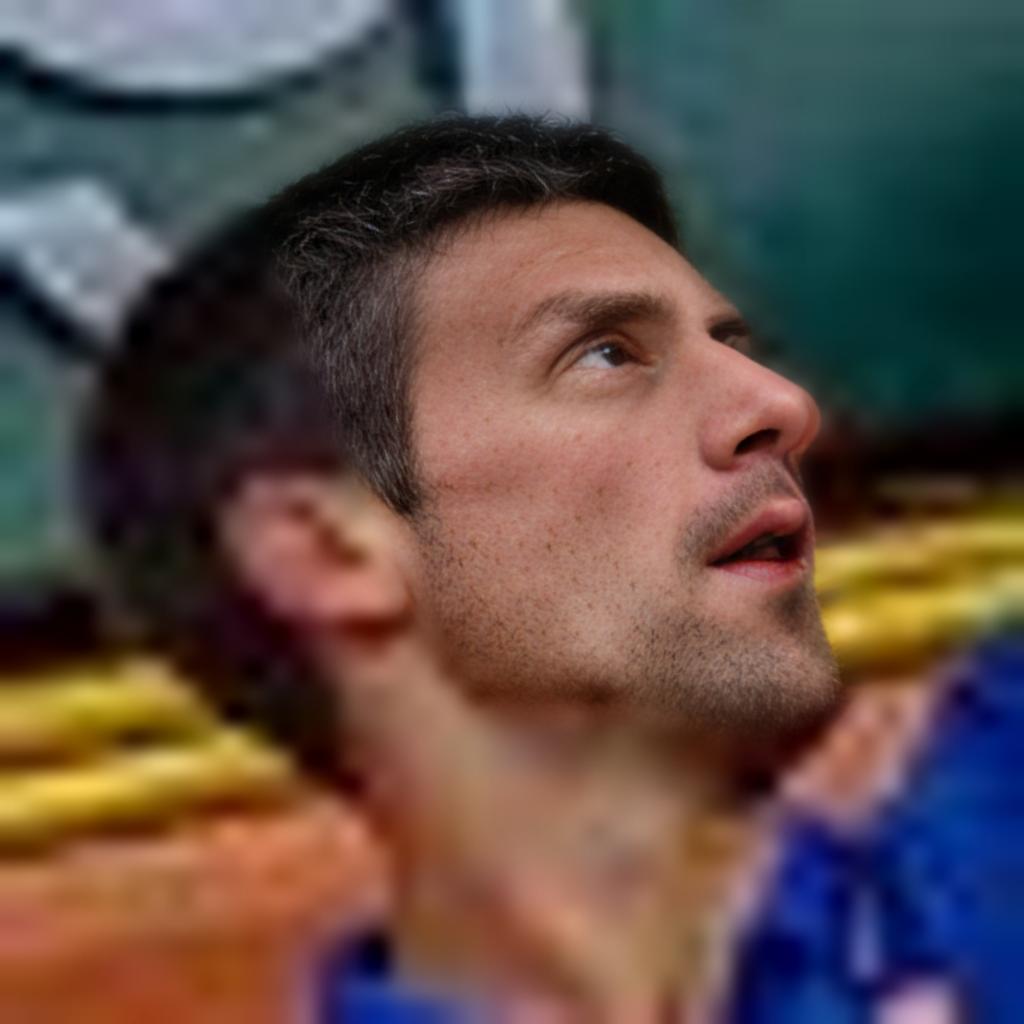} &
        		\includegraphics[width=0.167\textwidth]{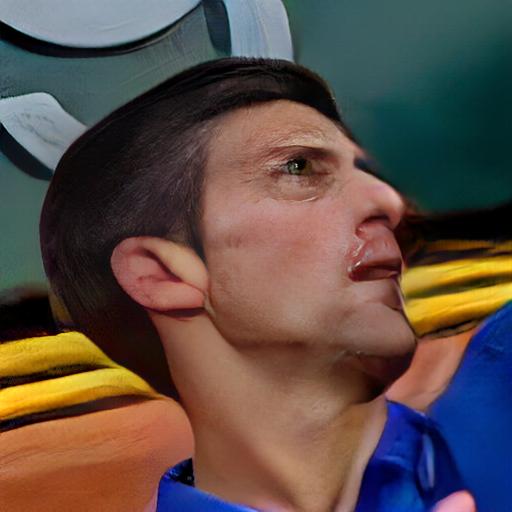} &
        		\includegraphics[width=0.167\textwidth]{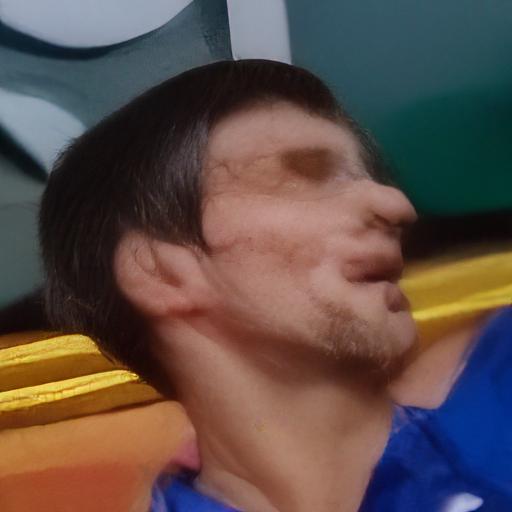} &
        		\includegraphics[width=0.167\textwidth]{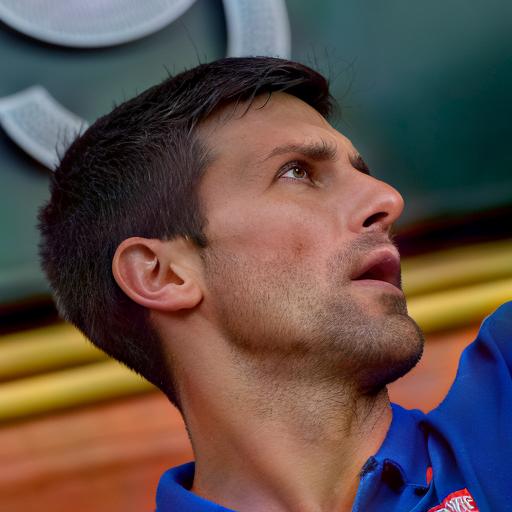} &
        		\includegraphics[width=0.167\textwidth]{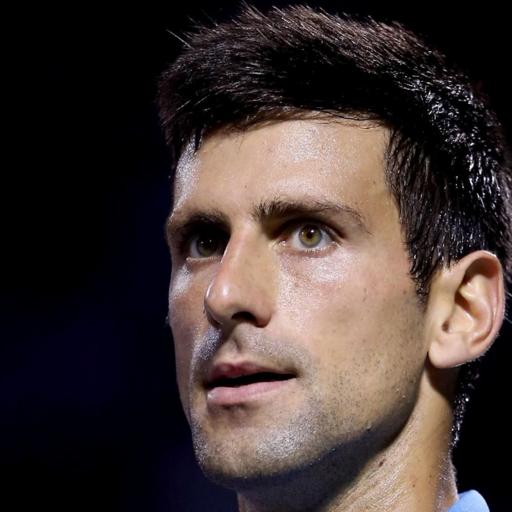}
        		\\
        		\includegraphics[width=0.167\textwidth]{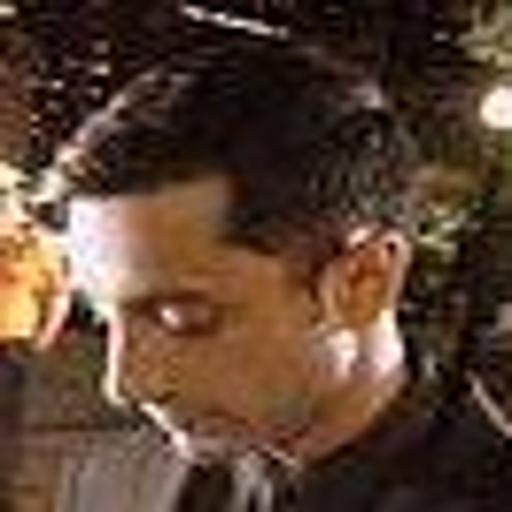} &
        		\includegraphics[width=0.167\textwidth]{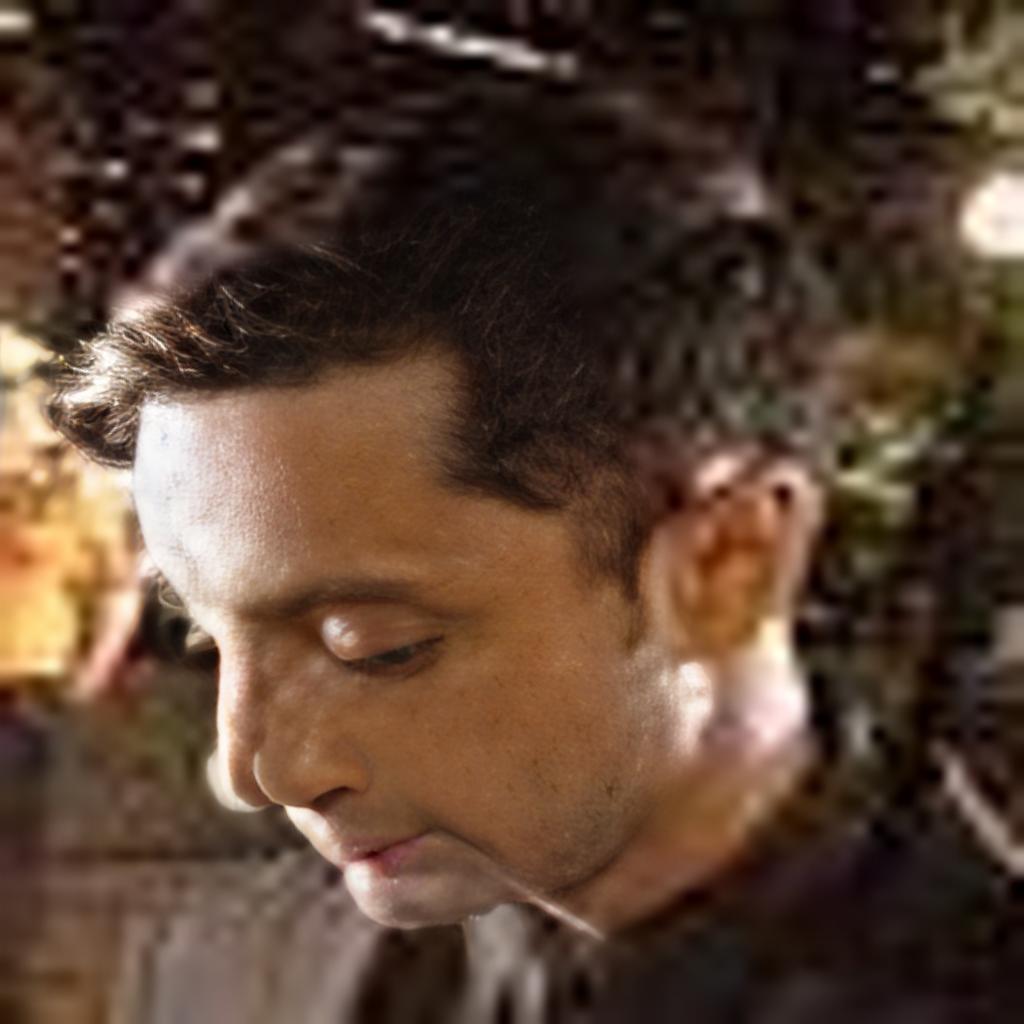} &
        		\includegraphics[width=0.167\textwidth]{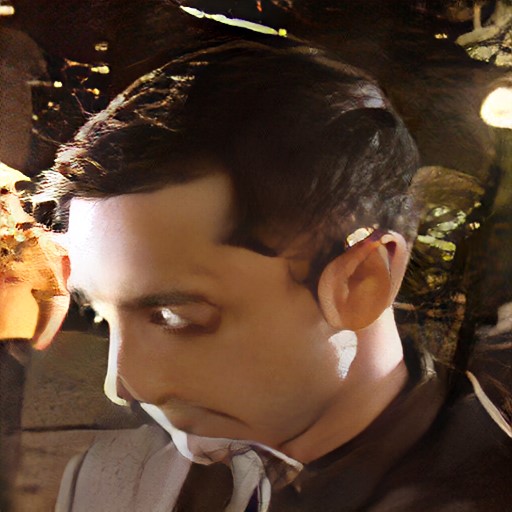} &
        		\includegraphics[width=0.167\textwidth]{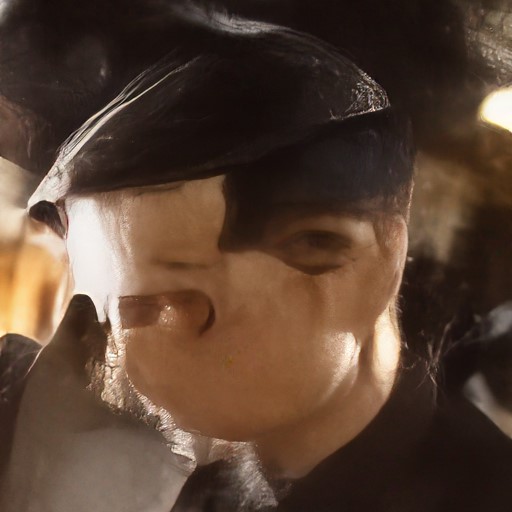} &
        		\includegraphics[width=0.167\textwidth]{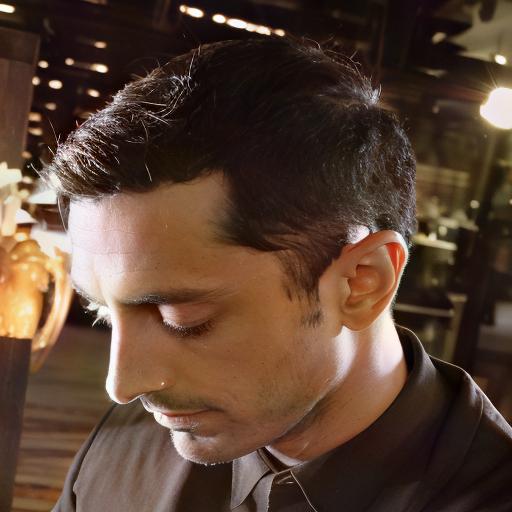} &
        		\includegraphics[width=0.167\textwidth]{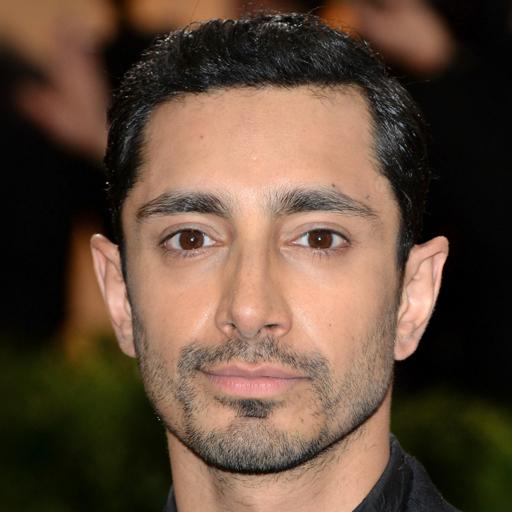}
        		\\
        		\includegraphics[width=0.167\textwidth]{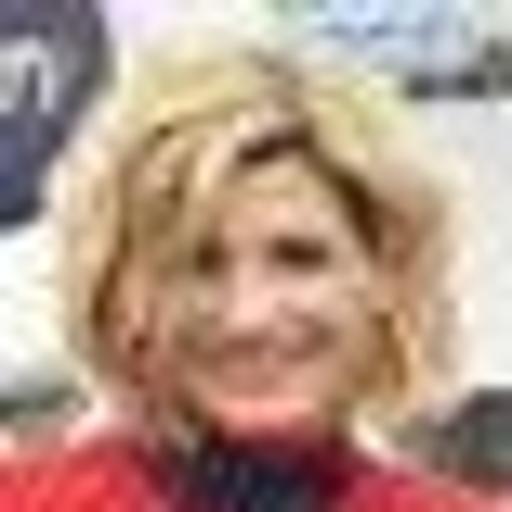} &
        		\includegraphics[width=0.167\textwidth]{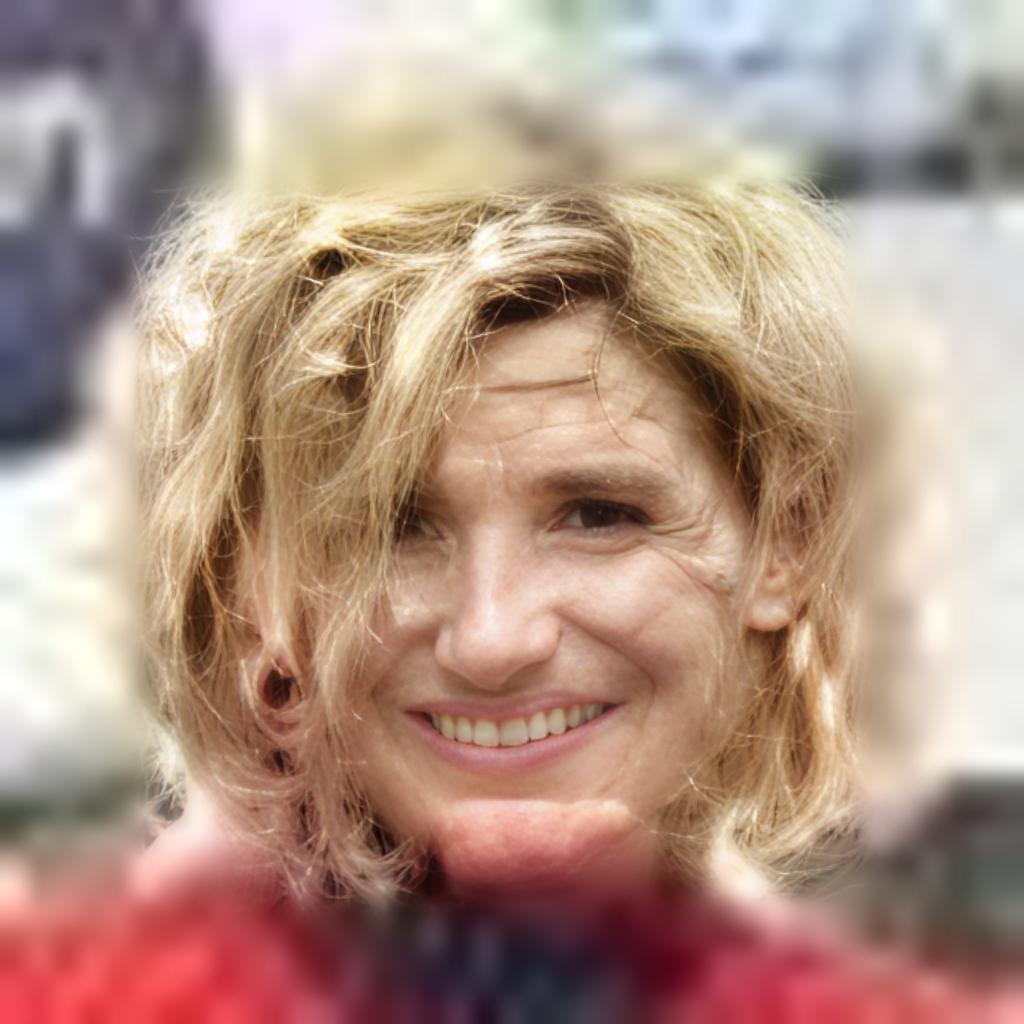} &
        		\includegraphics[width=0.167\textwidth]{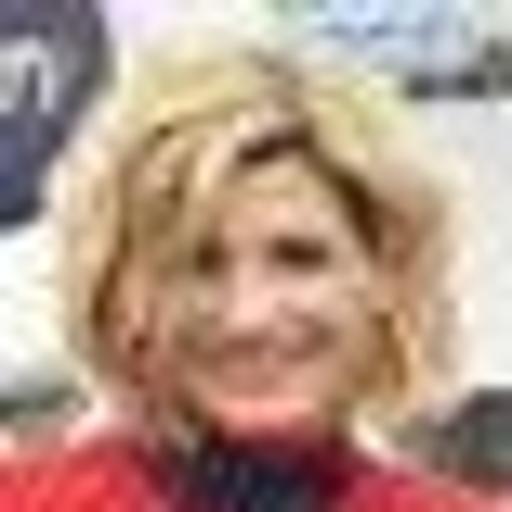} &
        		\includegraphics[width=0.167\textwidth]{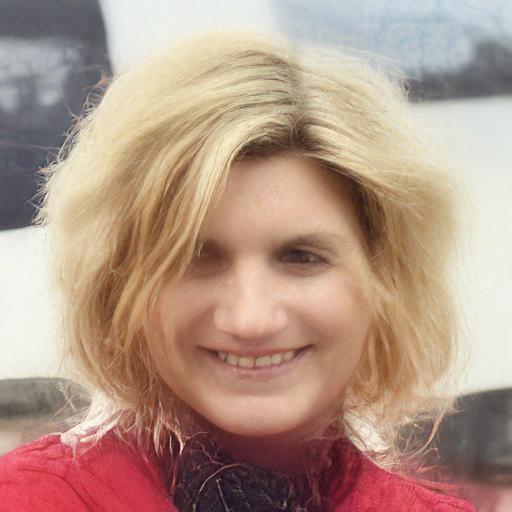} &
        		\includegraphics[width=0.167\textwidth]{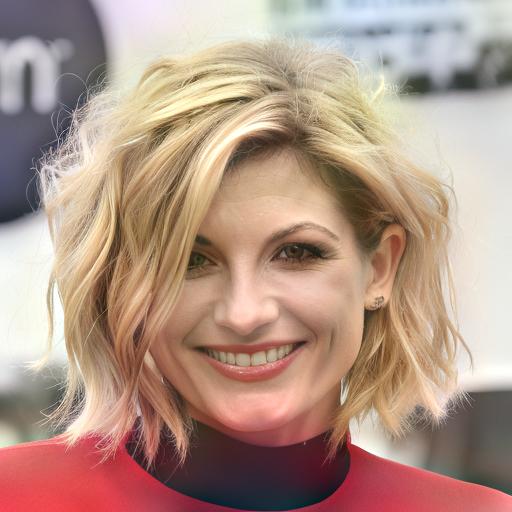} &
        		\includegraphics[width=0.167\textwidth]{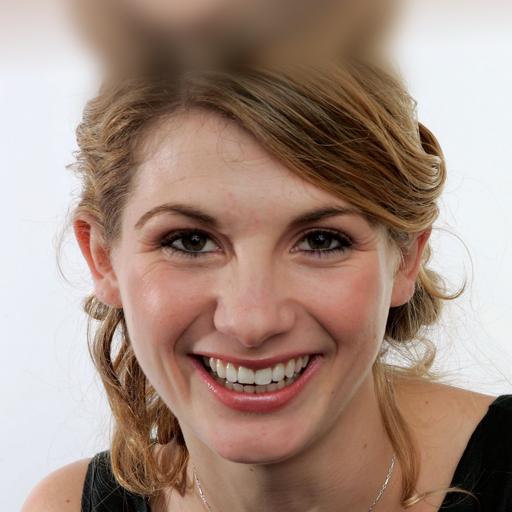}
        		\\
        		\footnotesize{Input} & \footnotesize{CodeFormer} & \footnotesize{DMDNet} & \footnotesize{DR2+SPAR} & \footnotesize{Ours} & \footnotesize{Pseudo-GT} \\
        		\vspace{-1cm}
        	\end{tabular}
        \end{subfigure}%
	\end{center}
	\caption[null]{
	Qualitative comparison with state-of-the-art restoration models on Celeb-Ref dataset \cite{dmdnet} with real degradation.
	}
	\label{fig:celeb_ref_real2}
\end{figure*}

\paragraph{Additional Qualitative Results}
Here we provide additional results on the light, heavy and real portions of the Celeb-Ref \cite{dmdnet} dataset. From \cref{fig:celeb_ref_light2} row three, we can see that the DMDNet is able to better preserver the identity compared to CodeFormer, but compared to ours it is still missing fine-grained details such as the notch in the chin. Despite achieving good results with light degradations, DMDNet struggles with the heavy and real degradations in \cref{fig:celeb_ref_heavy2,fig:celeb_ref_real2}. Although DR2 provides poor results in several cases, it works well on row 4 of \cref{fig:celeb_ref_heavy2}.

\begin{figure}[h]
	\begin{center}
    	\setlength{\tabcolsep}{1pt}
        \begin{subfigure}{0.46\textwidth}
        \hspace{-0.2cm}
        	\begin{tabular}{*5c}
        		\includegraphics[width=0.2\textwidth]{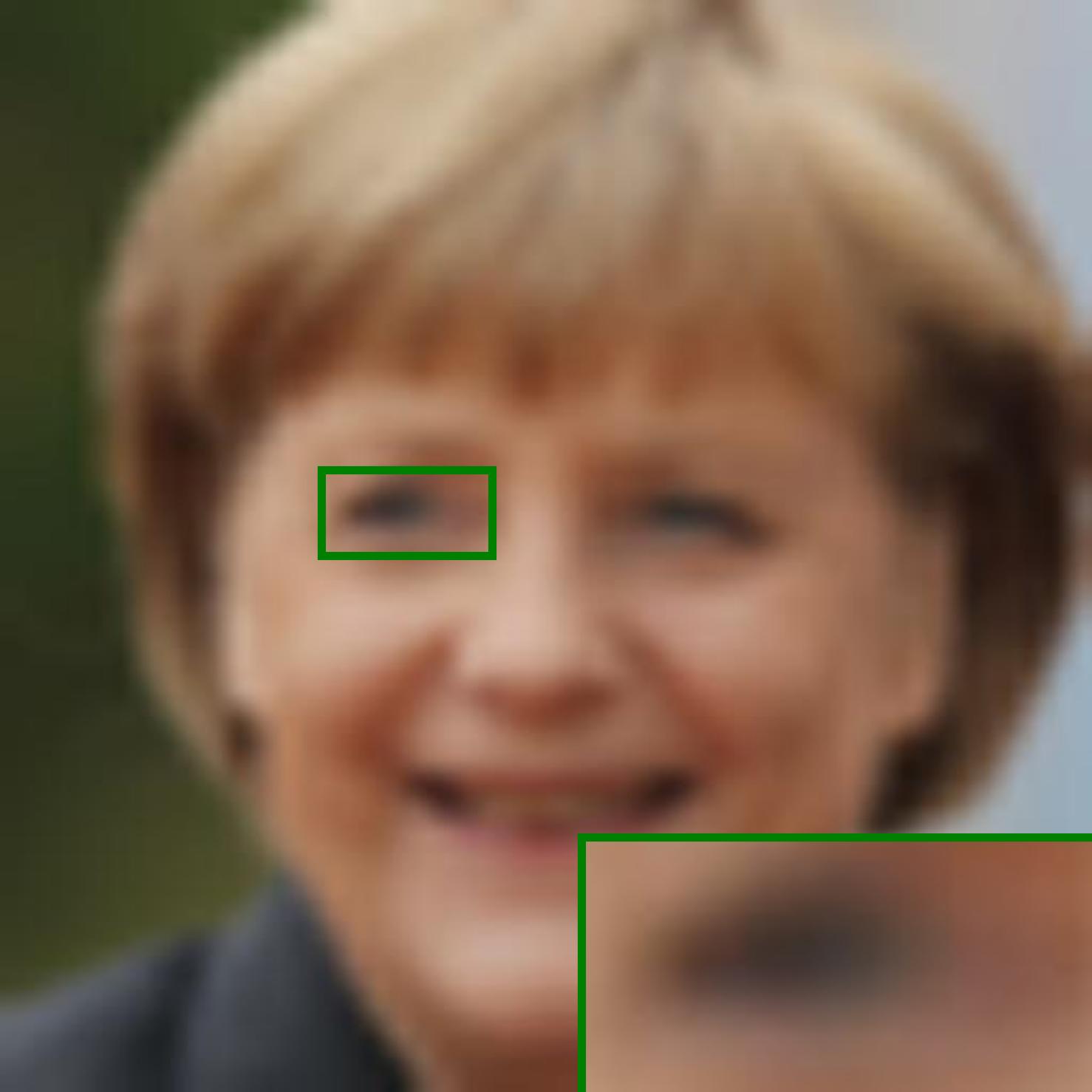} &
        		\includegraphics[width=0.2\textwidth]{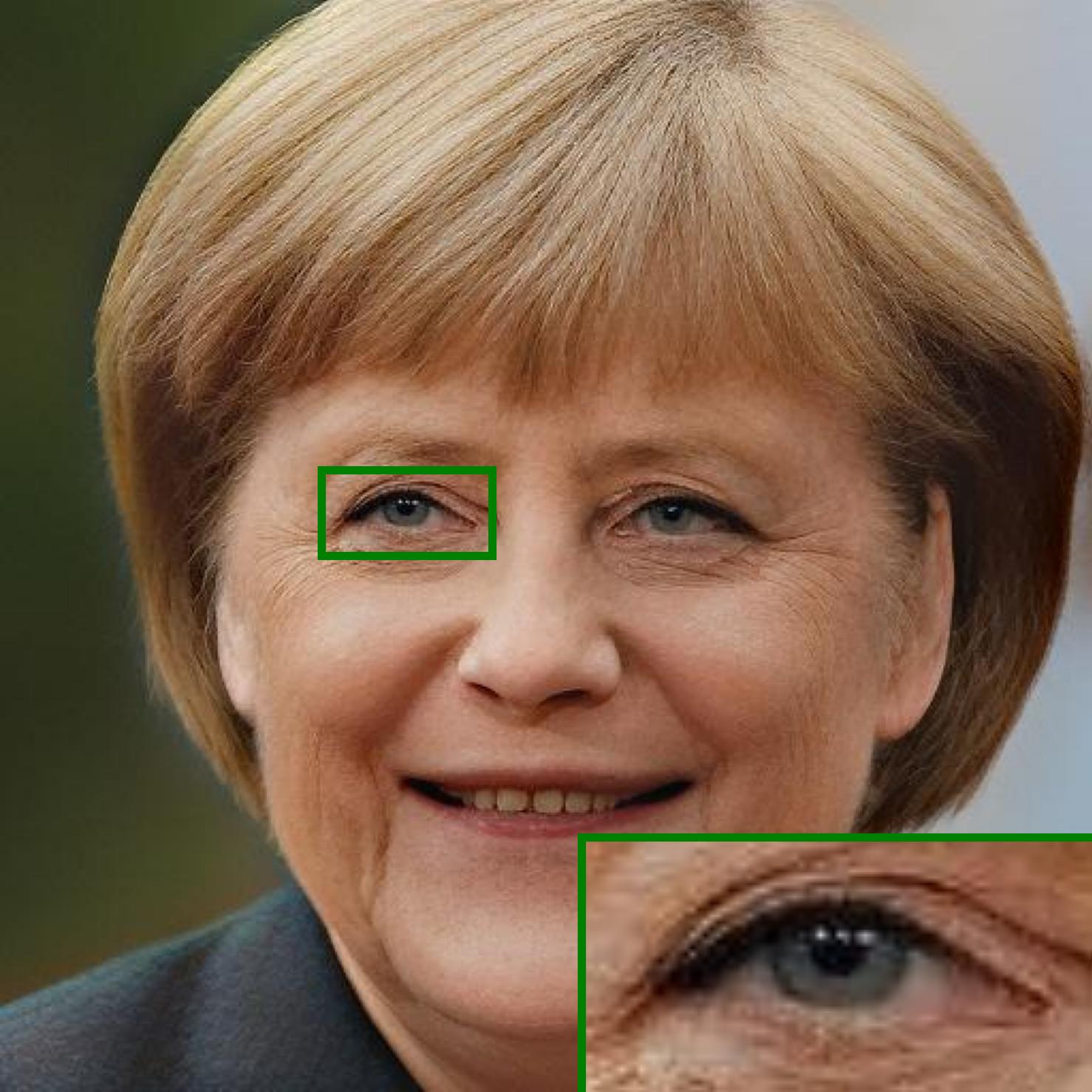} &
        		\includegraphics[width=0.2\textwidth]{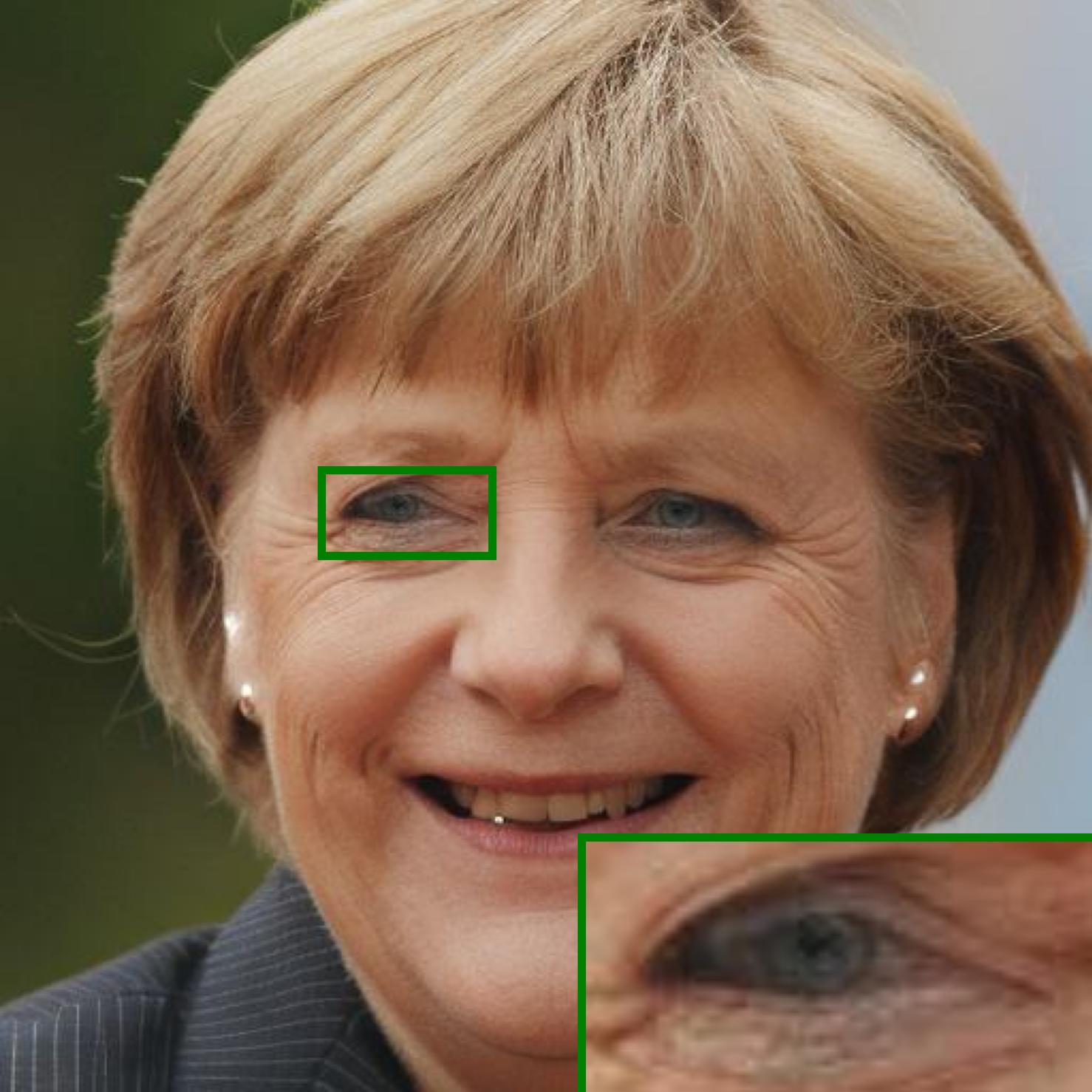} &
        		\includegraphics[width=0.2\textwidth]{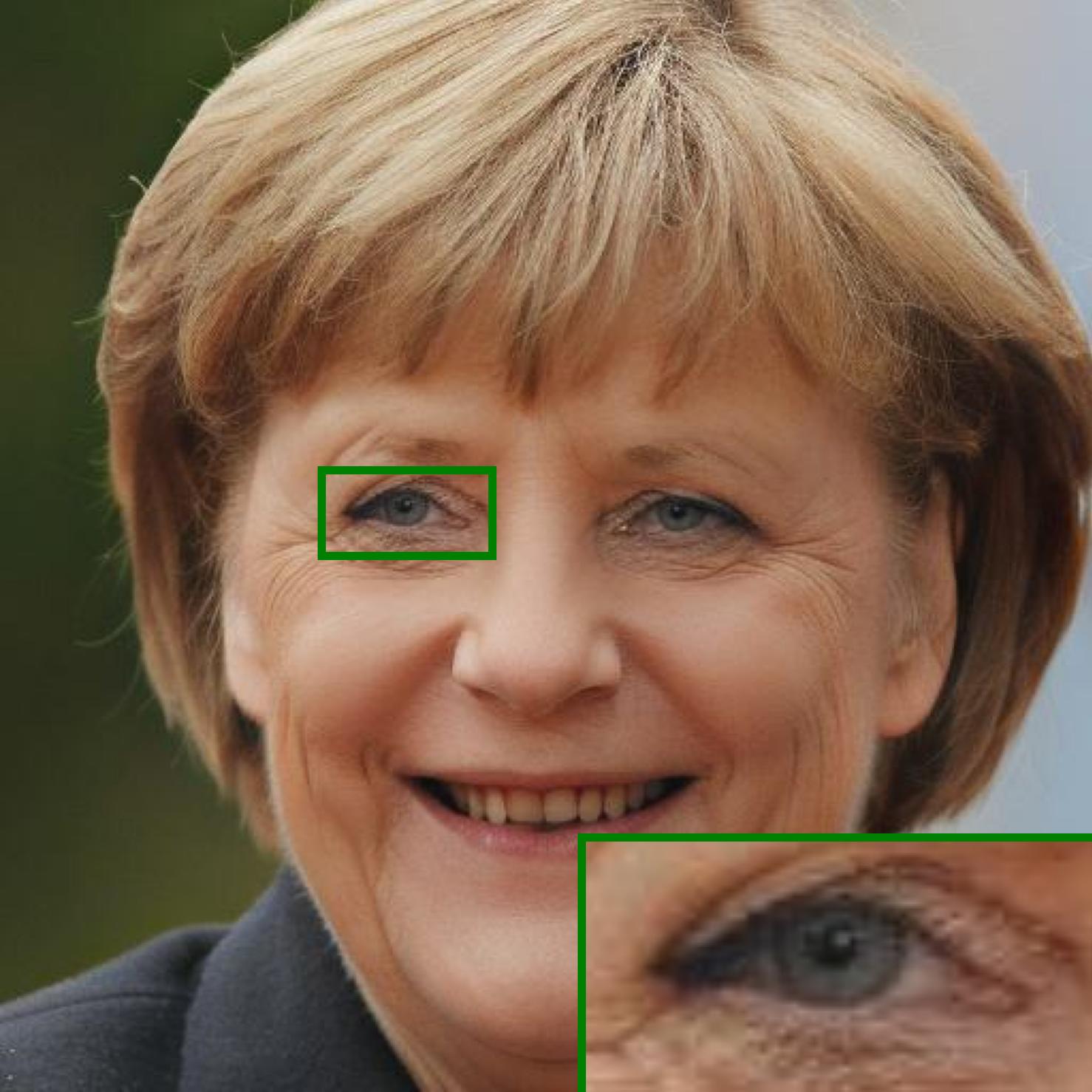} &
        		\includegraphics[width=0.2\textwidth]{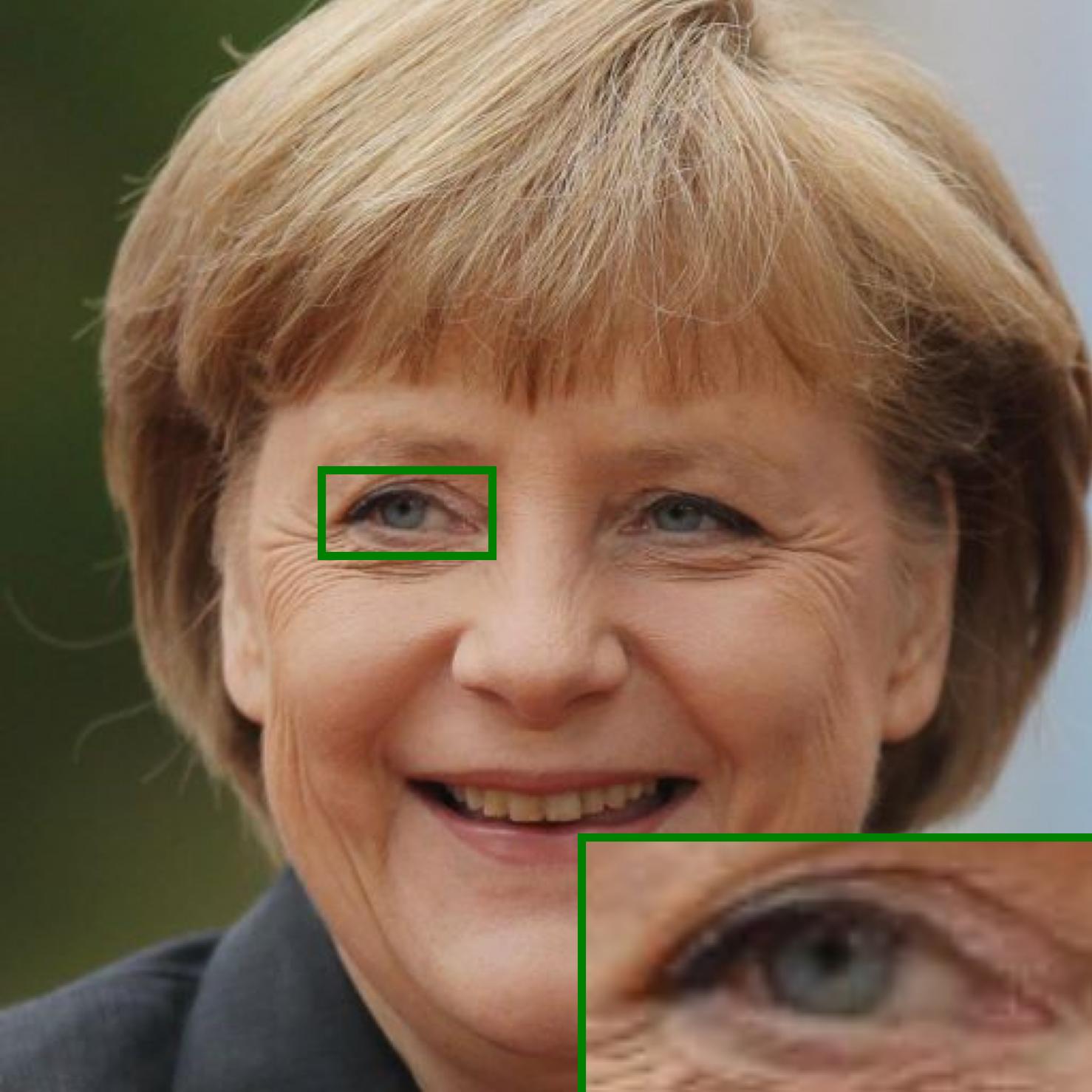}
        		\\
        		\includegraphics[width=0.2\textwidth]{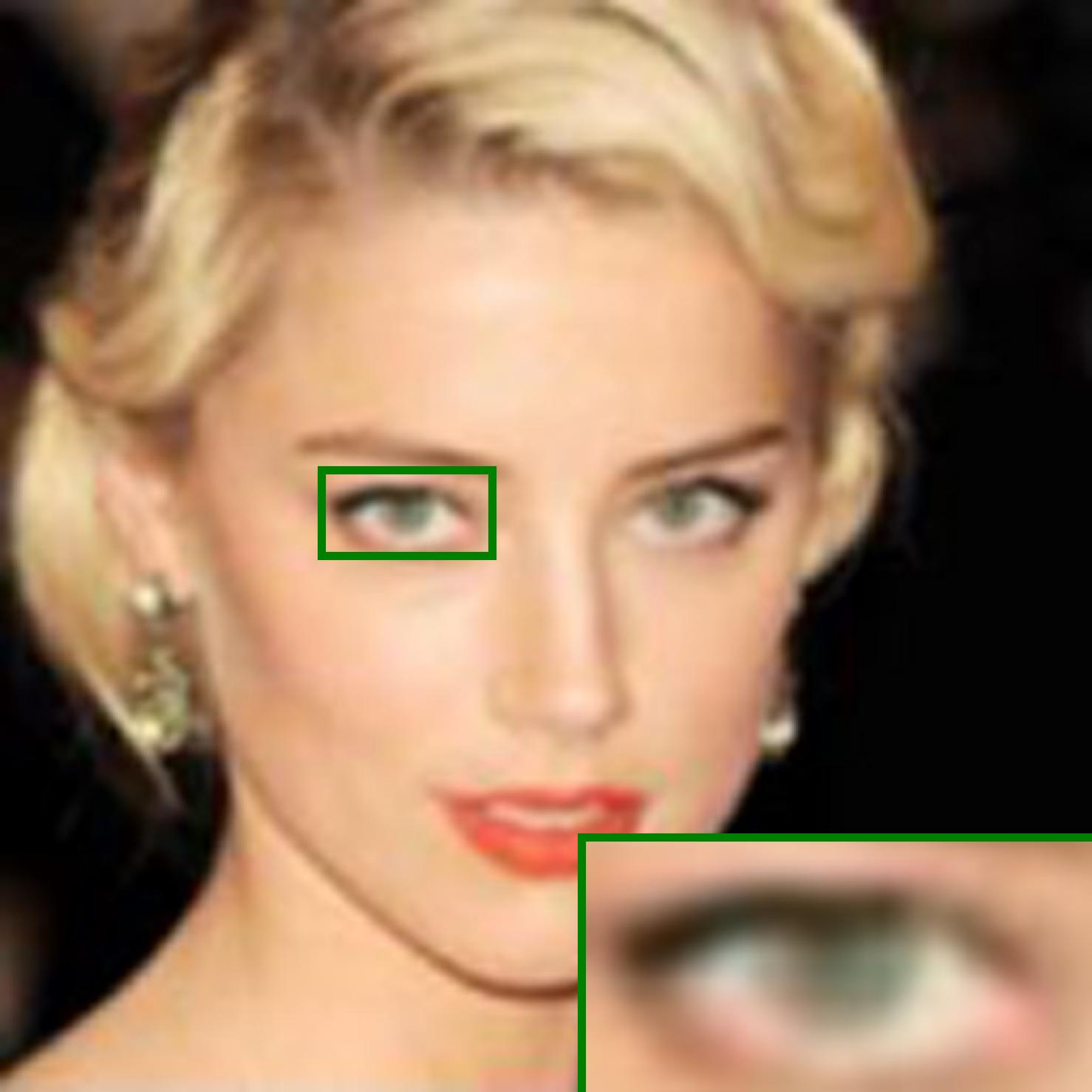} &
        		\includegraphics[width=0.2\textwidth]{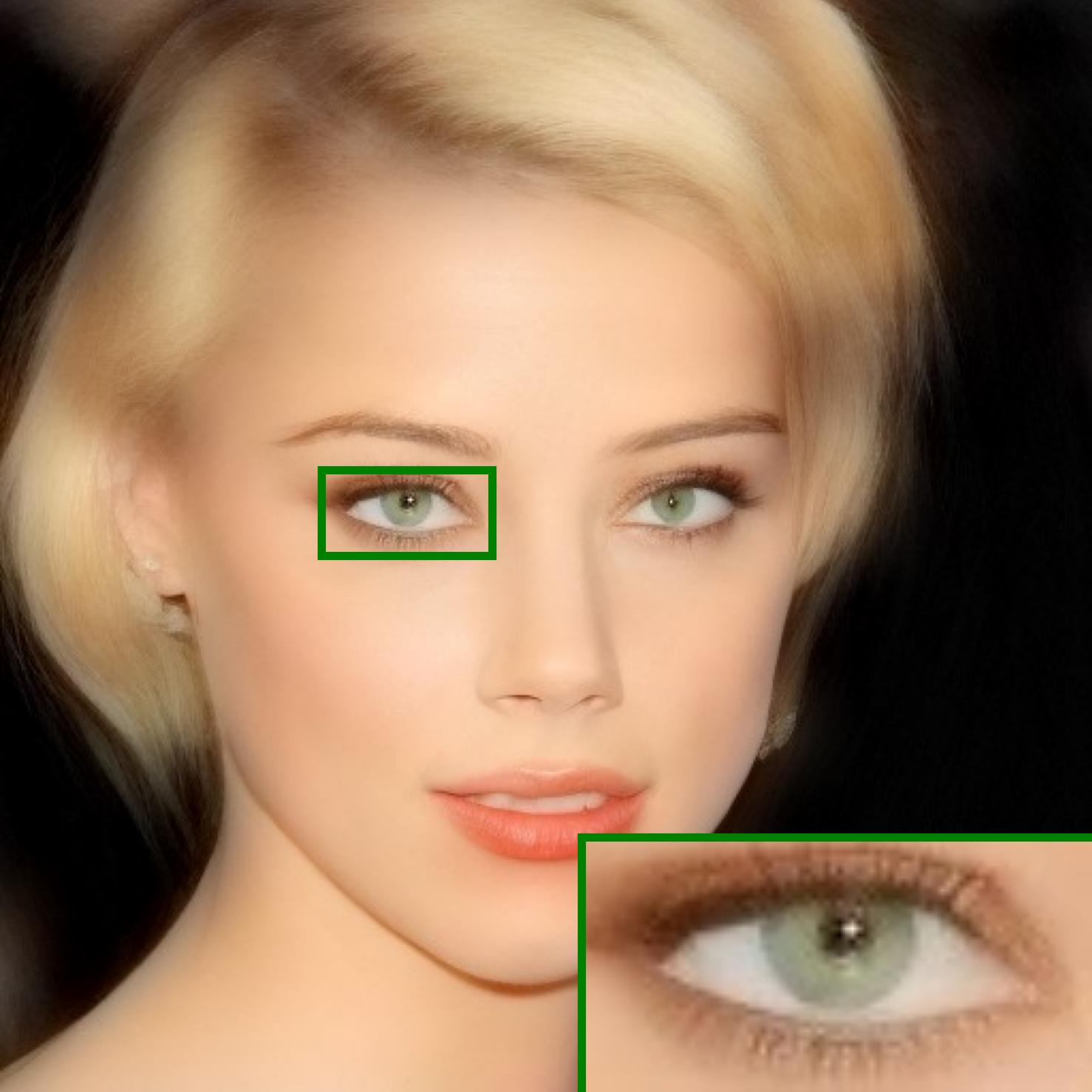} &
        		\includegraphics[width=0.2\textwidth]{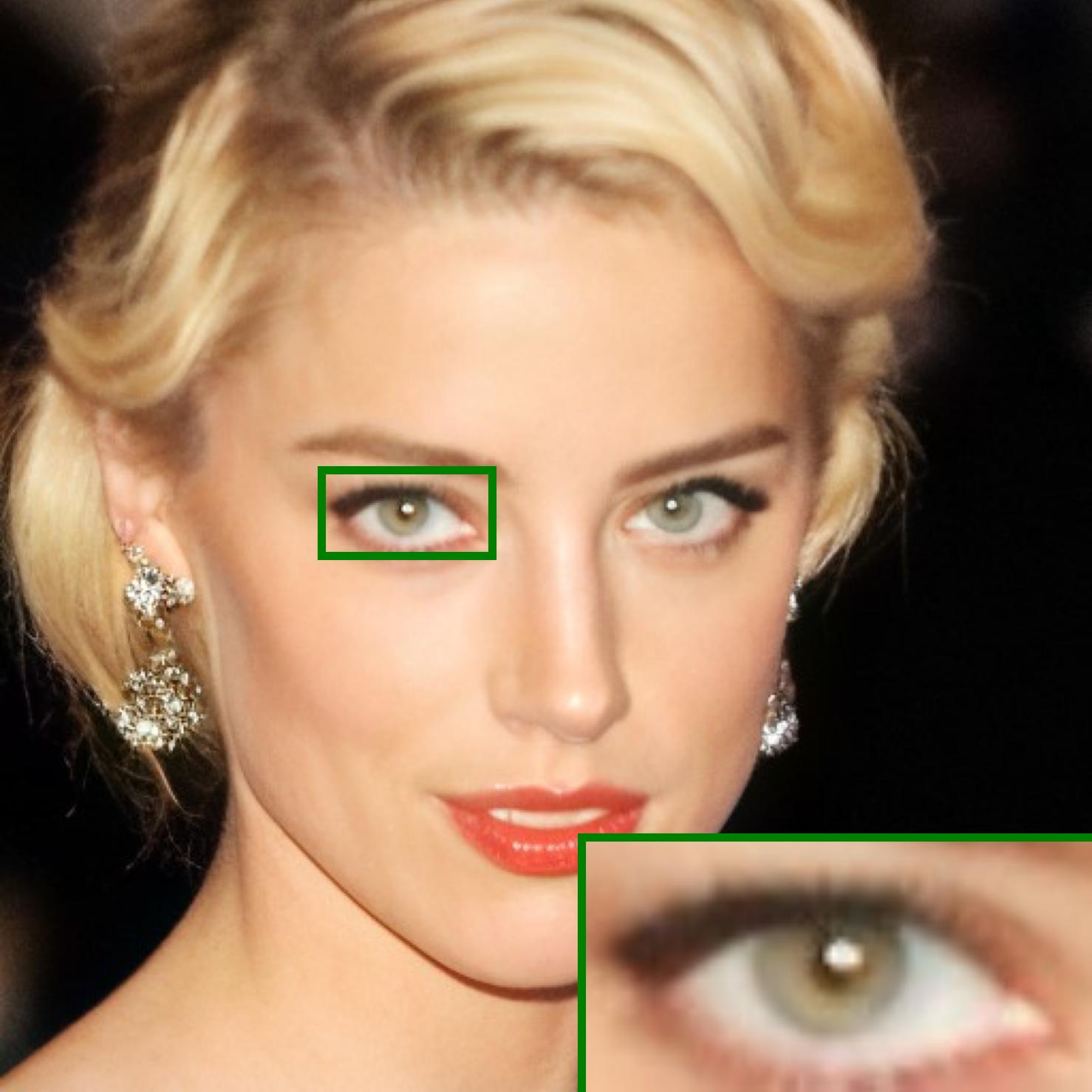} &
        		\includegraphics[width=0.2\textwidth]{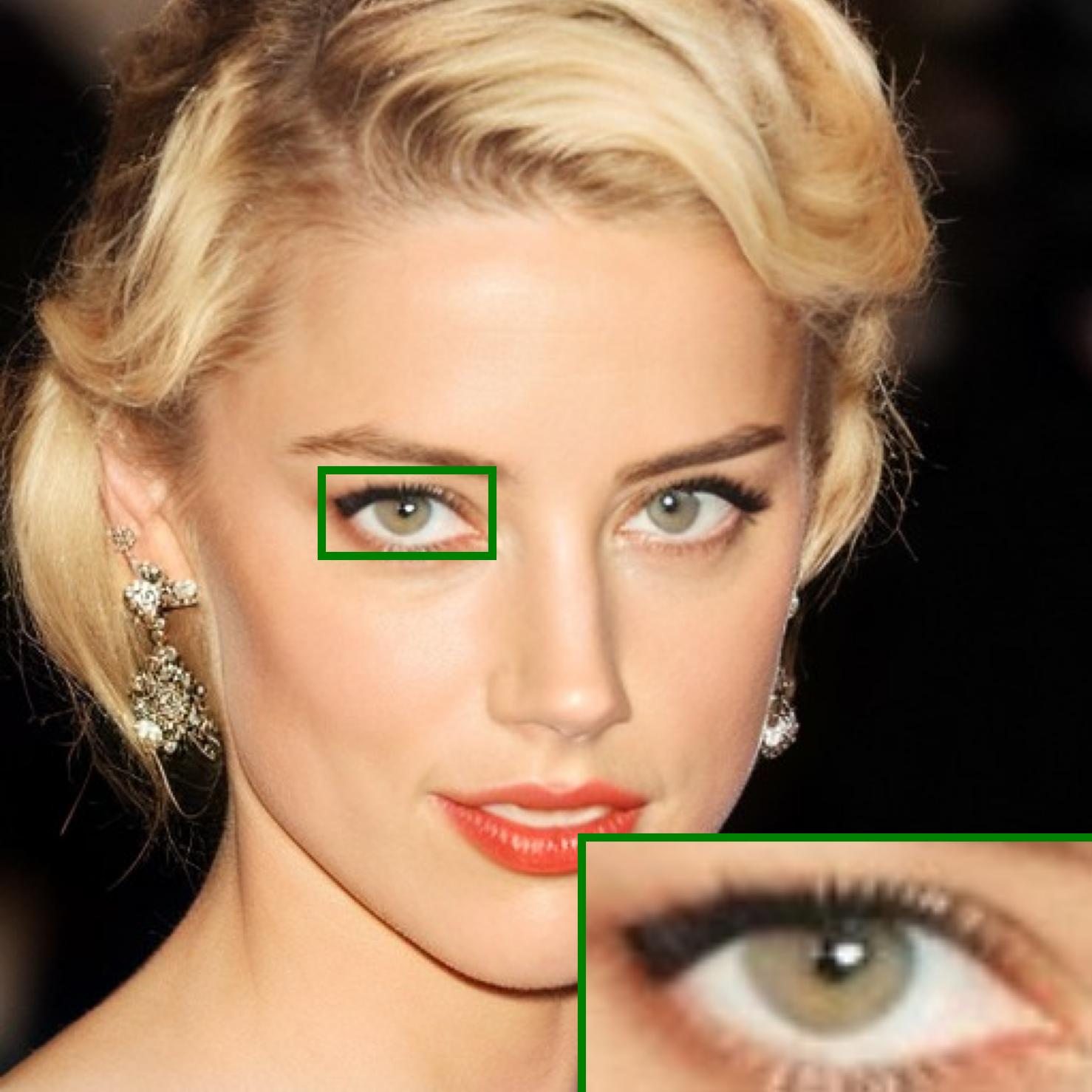} &
        		\includegraphics[width=0.2\textwidth]{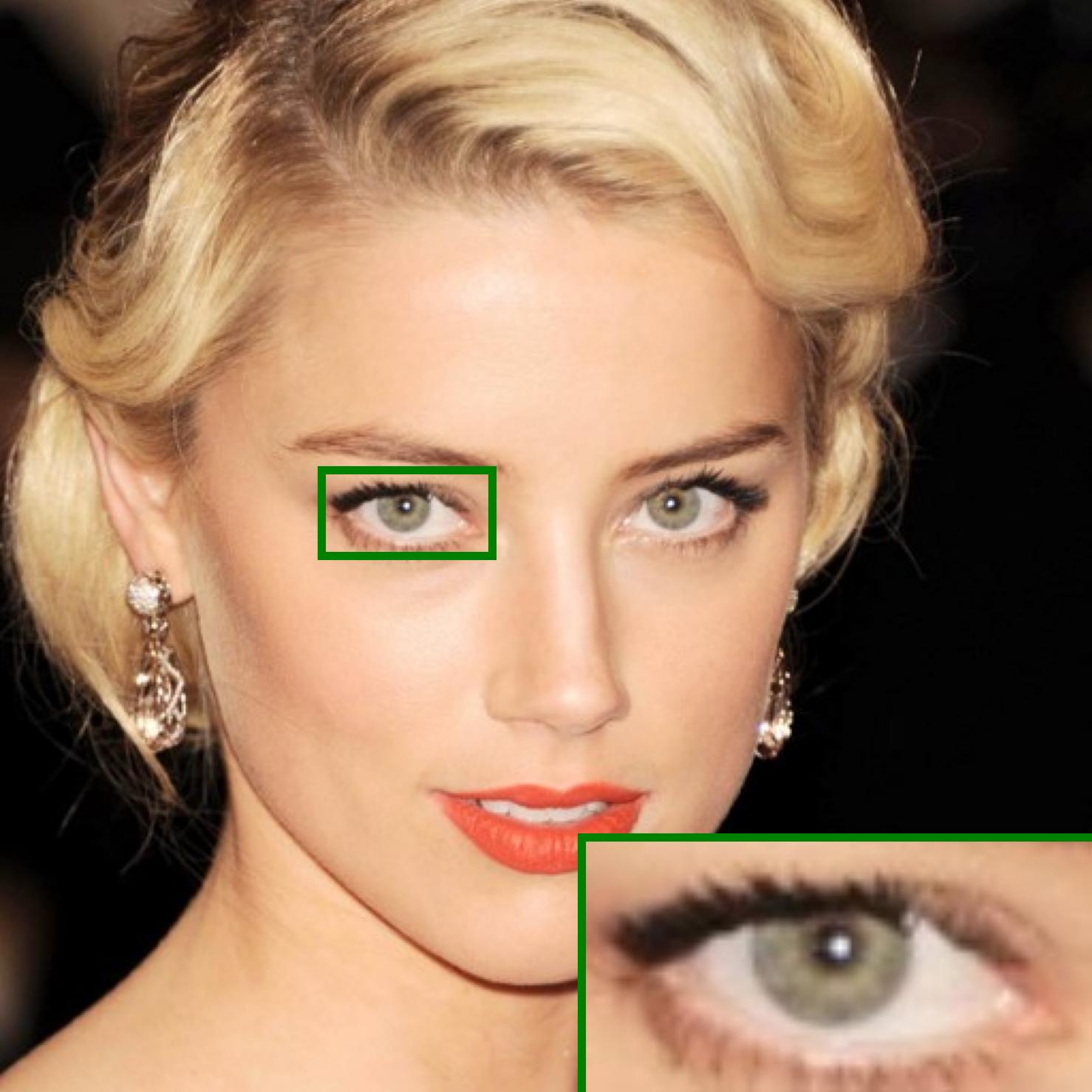}
        		\\
        		\footnotesize{Input} & \footnotesize{Base Model} & \footnotesize{Base Model} & \footnotesize{Ours} & \footnotesize{GT} \\
        		\vspace{-9mm}
        		 & \footnotesize{+ DreamBooth} & \footnotesize{+ ViCo} & &  \\
        		\vspace{-0cm}
        	\end{tabular}
        \end{subfigure}%
	\end{center}
	\caption{
	Results using different personalization techniques combined with a base restoration model. DreamBooth \cite{dreambooth} is not able to capture all the details and can result in poor-quality images. ViCo is better able to capture most details, but can still result in blurry images. Ours is able to capture fine-grained details without hurting the restoration performance of the Base Model. Zoom in for best view.
	}
	\label{fig:adapter_comparison}
\end{figure}

\begin{figure*}[h]
	\begin{center}
    	\setlength{\tabcolsep}{1pt}
        \begin{subfigure}{0.98\textwidth}
        \hspace{-0.2cm}
        	\begin{tabular}{*5c}
        		\includegraphics[width=0.2\textwidth]{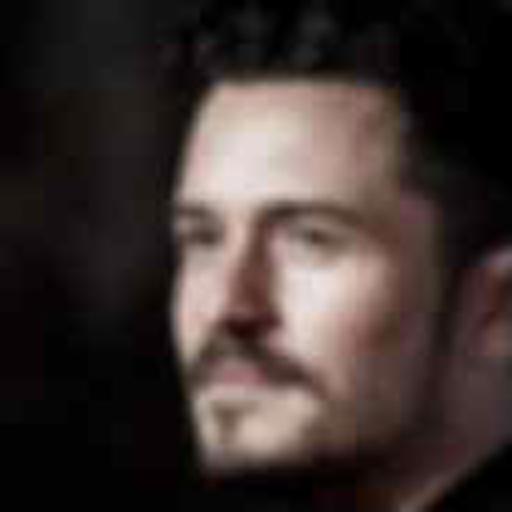} &
        		\includegraphics[width=0.2\textwidth]{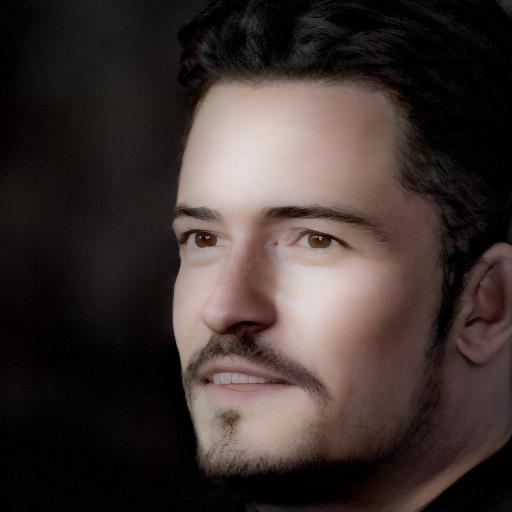} &
        		\includegraphics[width=0.2\textwidth]{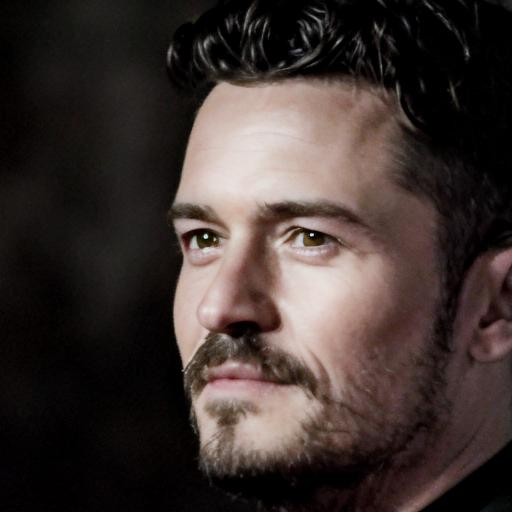} &
        		\includegraphics[width=0.2\textwidth]{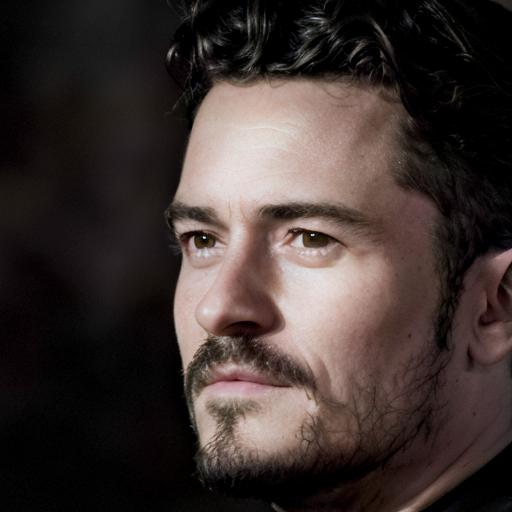} &
        		\includegraphics[width=0.2\textwidth]{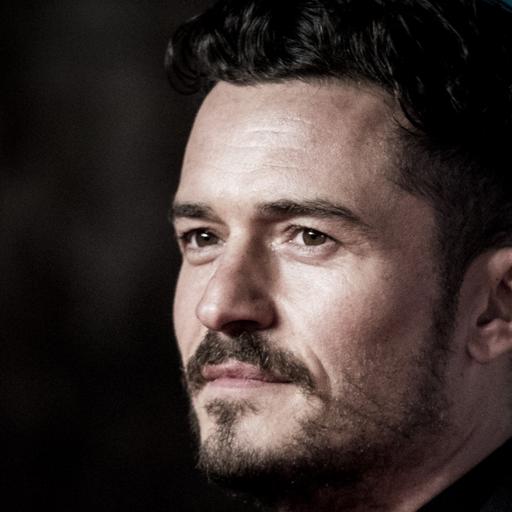}
        		\\
        		\includegraphics[width=0.2\textwidth]{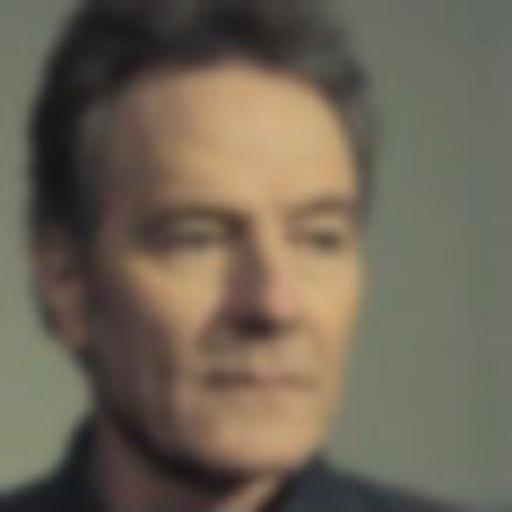} &
        		\includegraphics[width=0.2\textwidth]{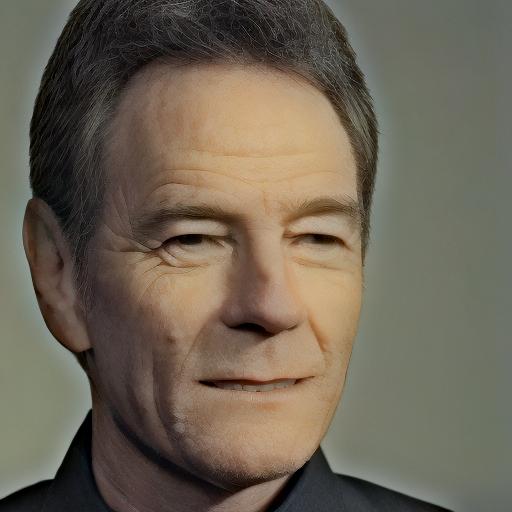} &
        		\includegraphics[width=0.2\textwidth]{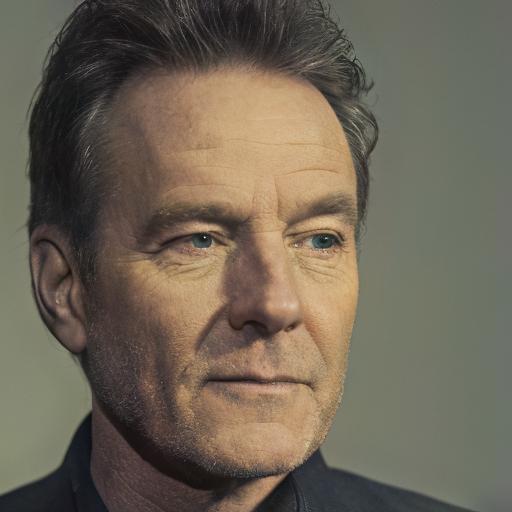} &
        		\includegraphics[width=0.2\textwidth]{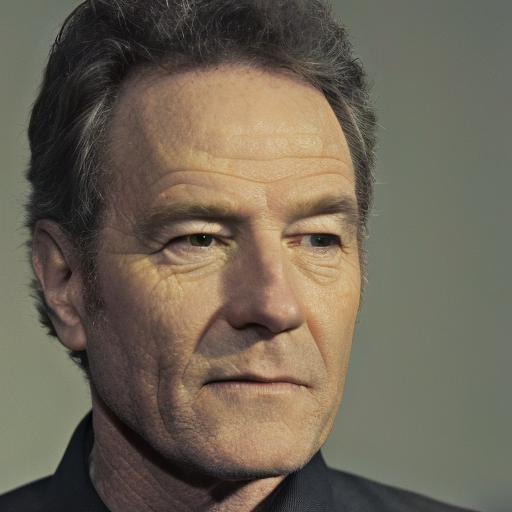} &
        		\includegraphics[width=0.2\textwidth]{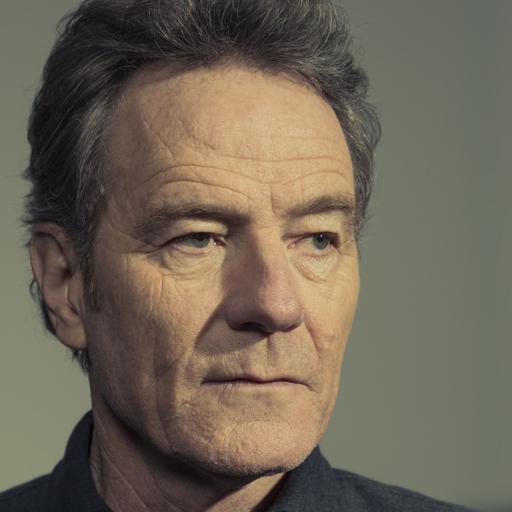}
        		\\
        		\includegraphics[width=0.2\textwidth]{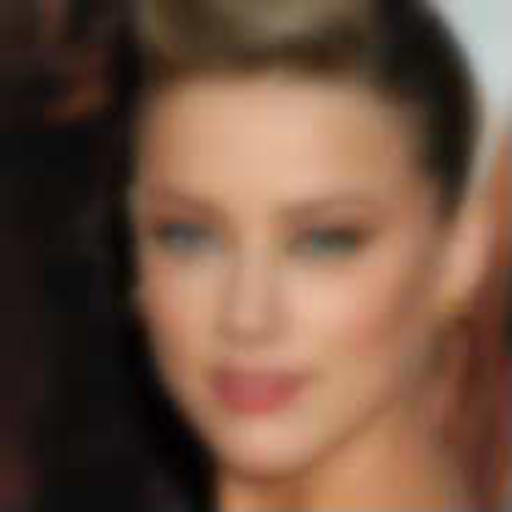} &
        		\includegraphics[width=0.2\textwidth]{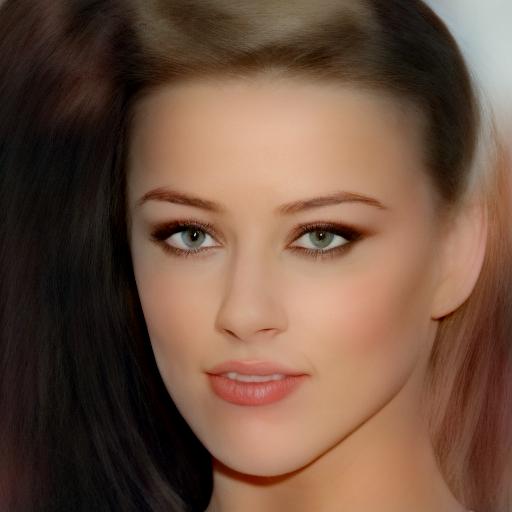} &
        		\includegraphics[width=0.2\textwidth]{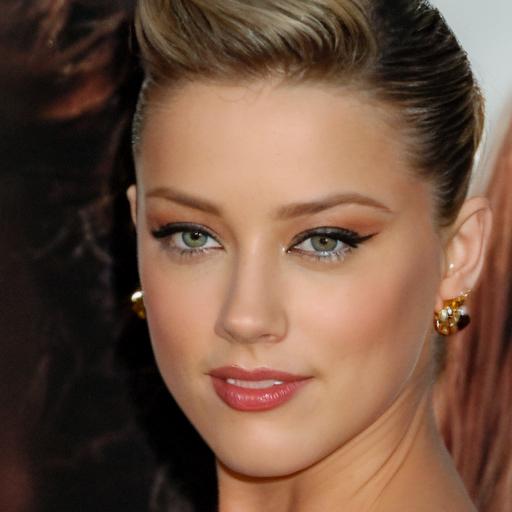} &
        		\includegraphics[width=0.2\textwidth]{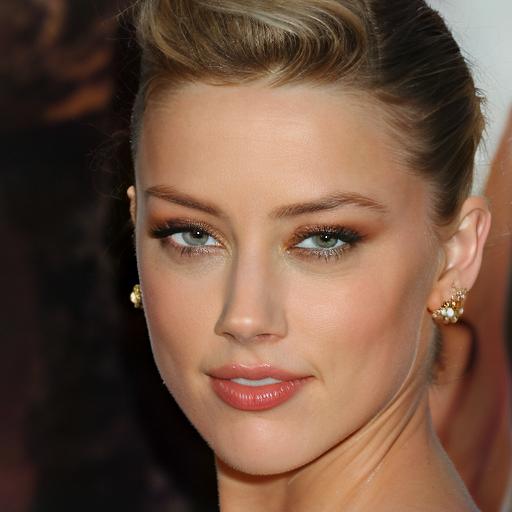} &
        		\includegraphics[width=0.2\textwidth]{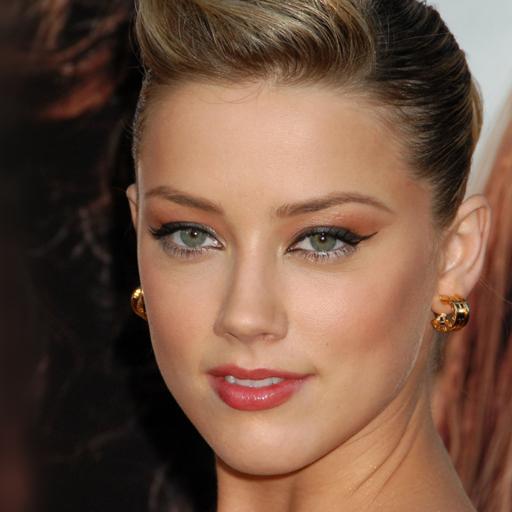}
        		\\
        		\includegraphics[width=0.2\textwidth]{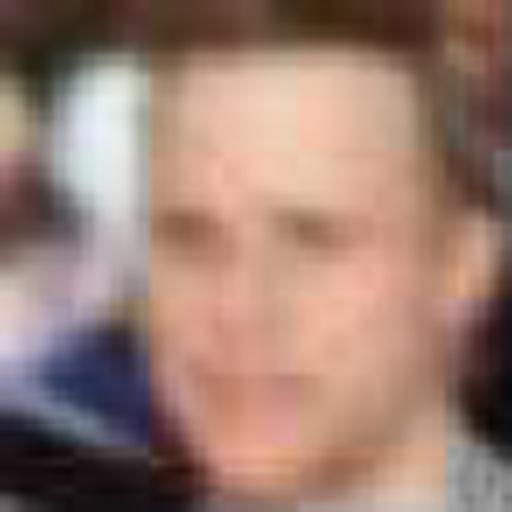} &
        		\includegraphics[width=0.2\textwidth]{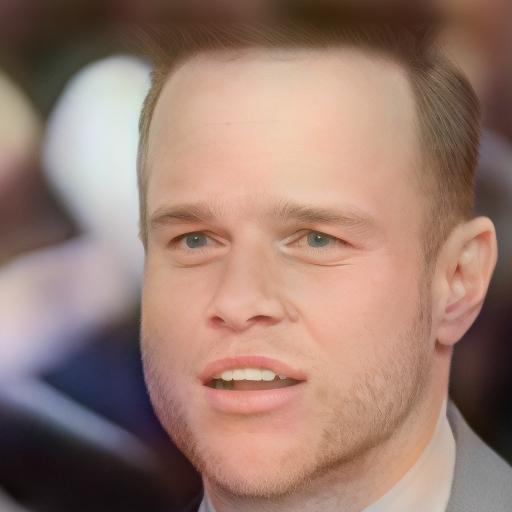} &
        		\includegraphics[width=0.2\textwidth]{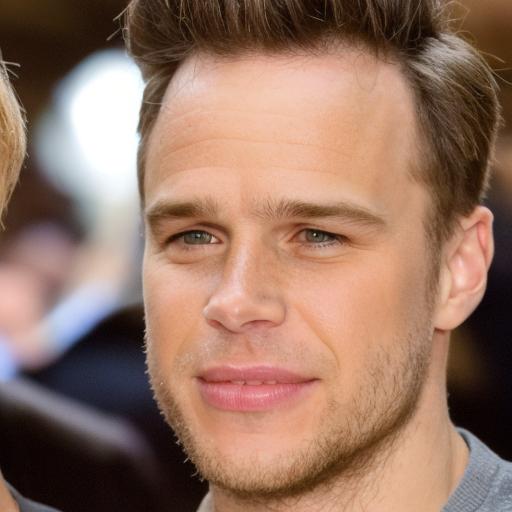} &
        		\includegraphics[width=0.2\textwidth]{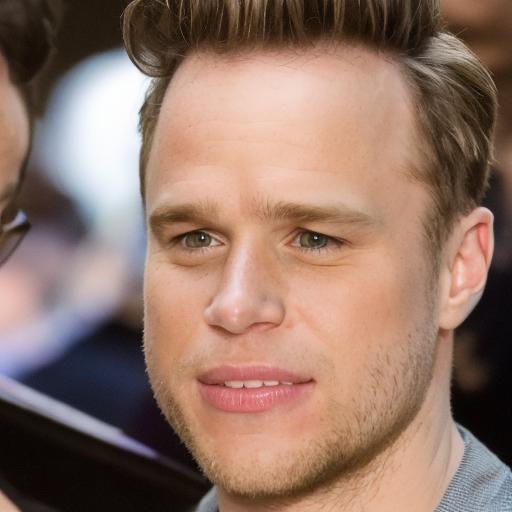} &
        		\includegraphics[width=0.2\textwidth]{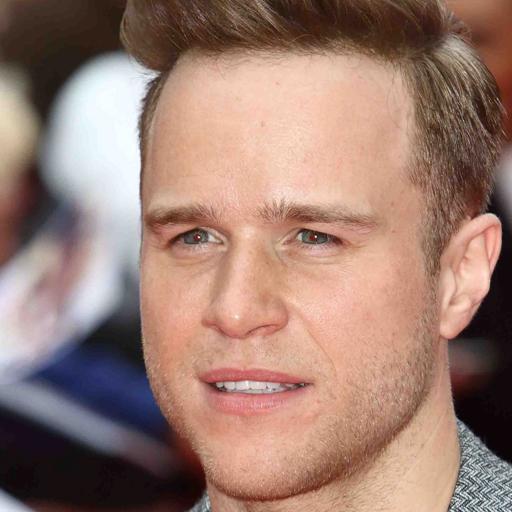}
        		\\
        		\includegraphics[width=0.2\textwidth]{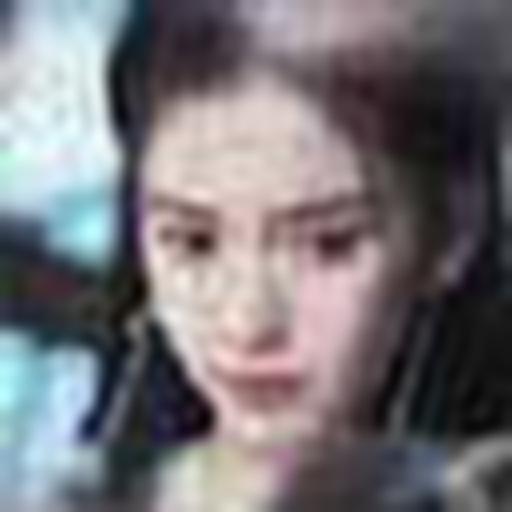} &
        		\includegraphics[width=0.2\textwidth]{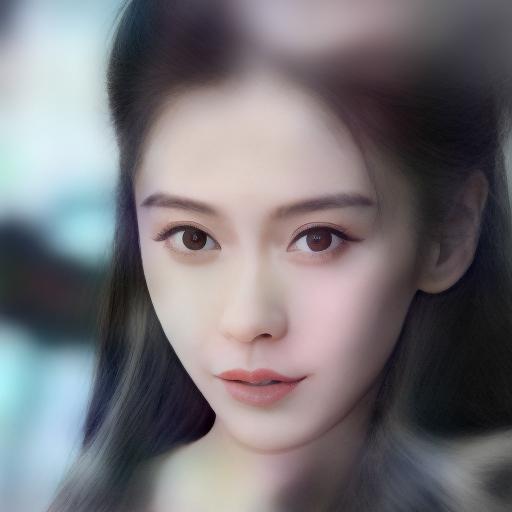} &
        		\includegraphics[width=0.2\textwidth]{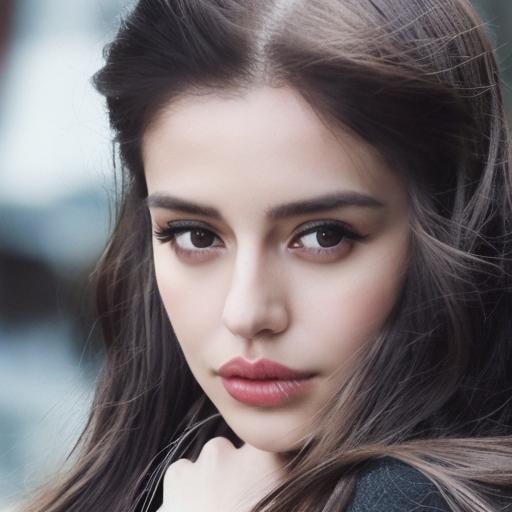} &
        		\includegraphics[width=0.2\textwidth]{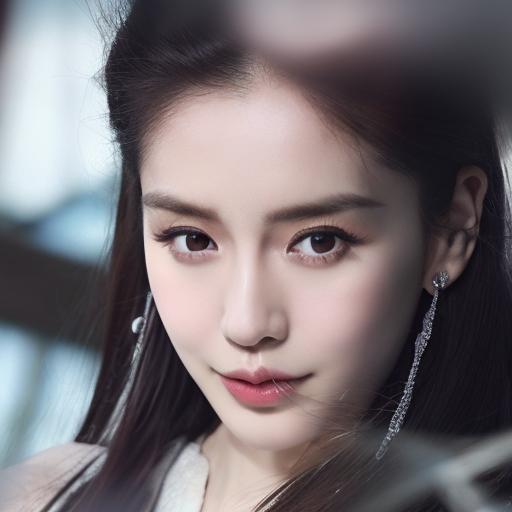} &
        		\includegraphics[width=0.2\textwidth]{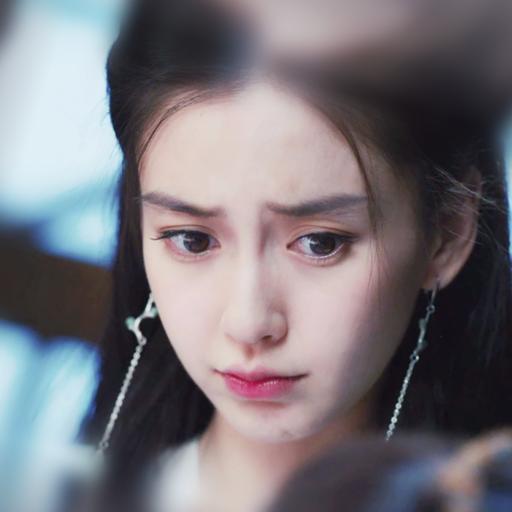}
        		\\
        		\includegraphics[width=0.2\textwidth]{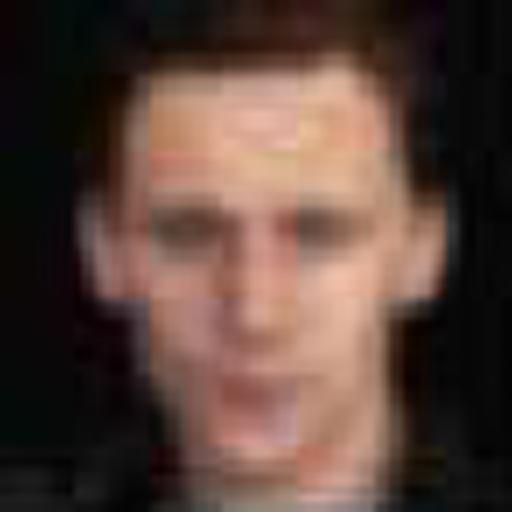} &
        		\includegraphics[width=0.2\textwidth]{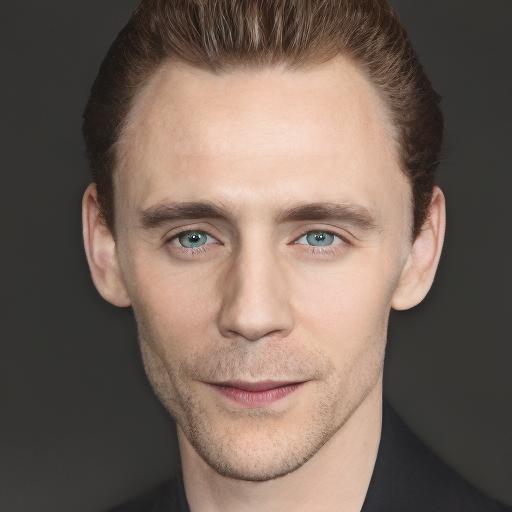} &
        		\includegraphics[width=0.2\textwidth]{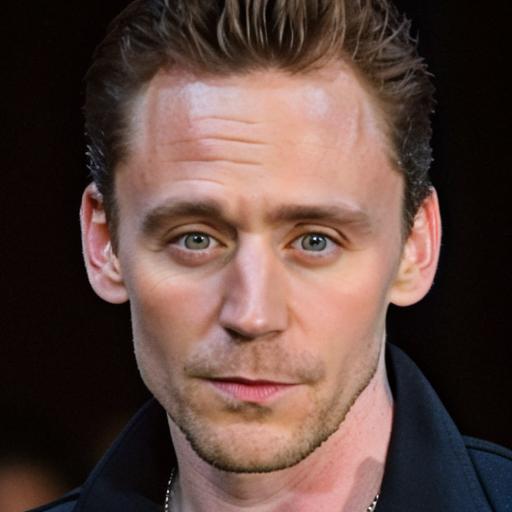} &
        		\includegraphics[width=0.2\textwidth]{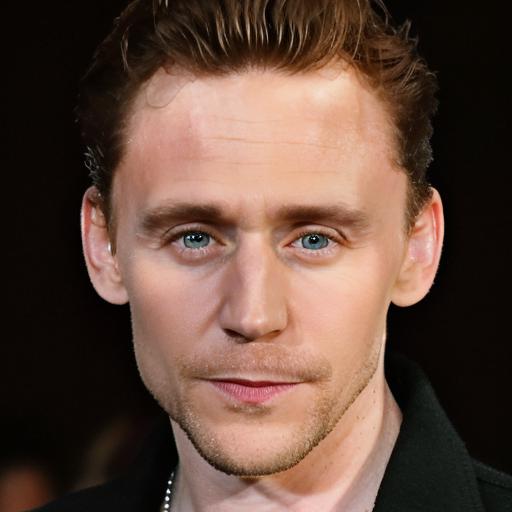} &
        		\includegraphics[width=0.2\textwidth]{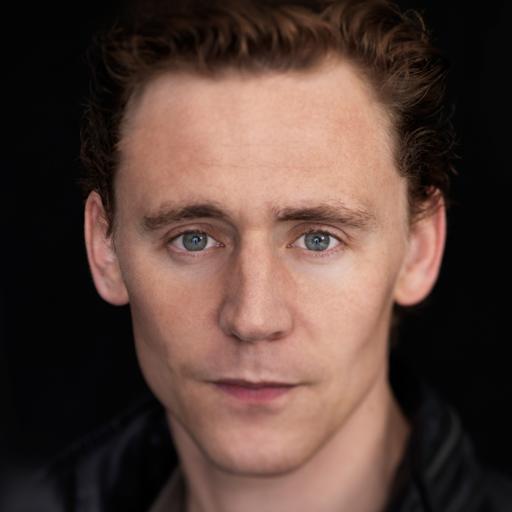}
        		\\
        		\footnotesize{Input} & \footnotesize{Base Model} & \footnotesize{Base Model} & \footnotesize{Ours} & \footnotesize{GT} \\
        		\vspace{-9mm}
        		 & \footnotesize{+ DreamBooth} & \footnotesize{+ ViCo} & &  \\
        		\vspace{-0cm}
        	\end{tabular}
        \end{subfigure}%
	\end{center}
	\caption{
	Results using different personalization techniques combined with a base restoration model with heavy degradation.
	}
	\label{fig:adapter_comparison_heavy}
\end{figure*}

\begin{figure*}[h]
	\begin{center}
    	\setlength{\tabcolsep}{1pt}
        \begin{subfigure}{0.98\textwidth}
        \hspace{-0.2cm}
        	\begin{tabular}{*5c}
        		\includegraphics[width=0.2\textwidth]{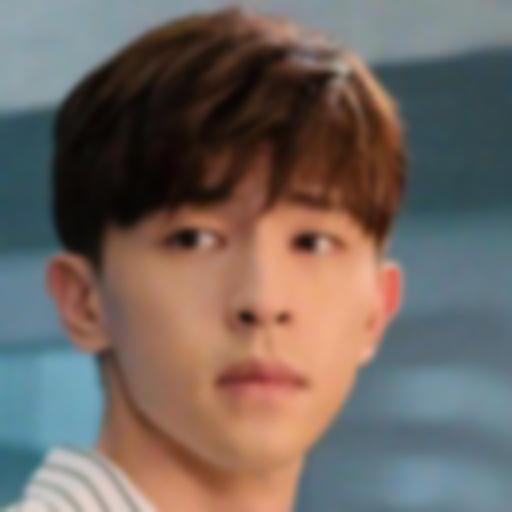} &
        		\includegraphics[width=0.2\textwidth]{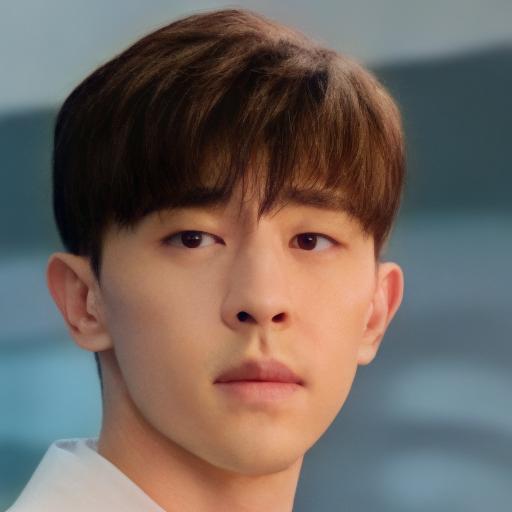} &
        		\includegraphics[width=0.2\textwidth]{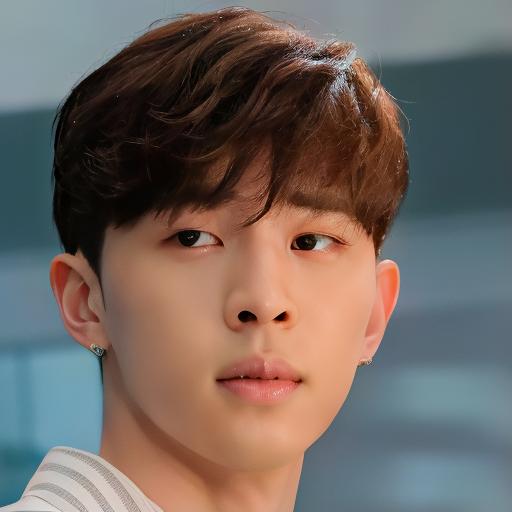} &
        		\includegraphics[width=0.2\textwidth]{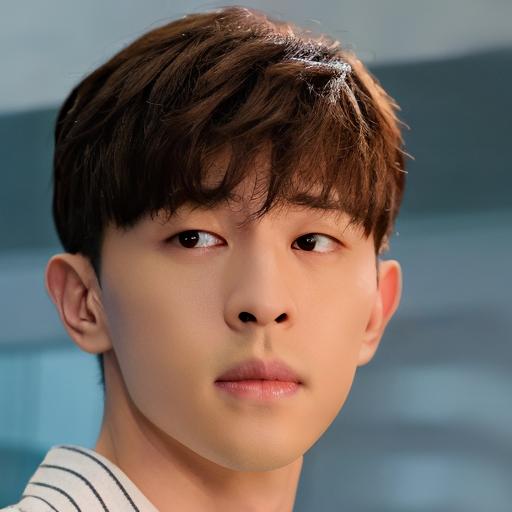} &
        		\includegraphics[width=0.2\textwidth]{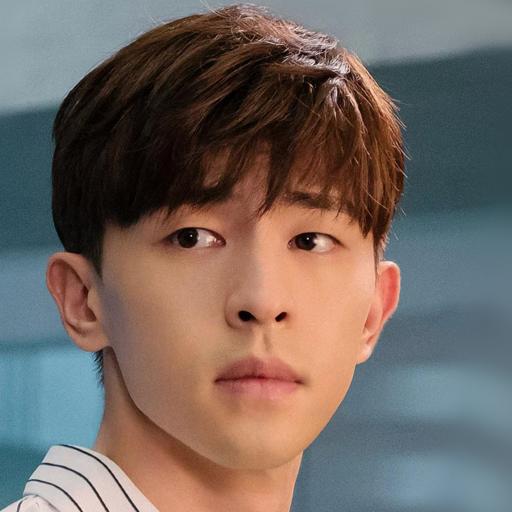}
        		\\
        		\includegraphics[width=0.2\textwidth]{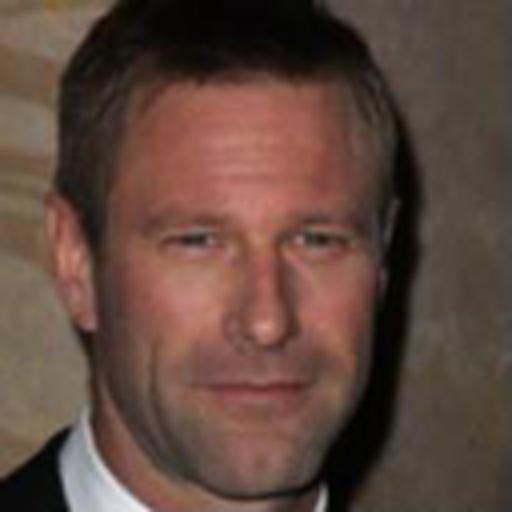} &
        		\includegraphics[width=0.2\textwidth]{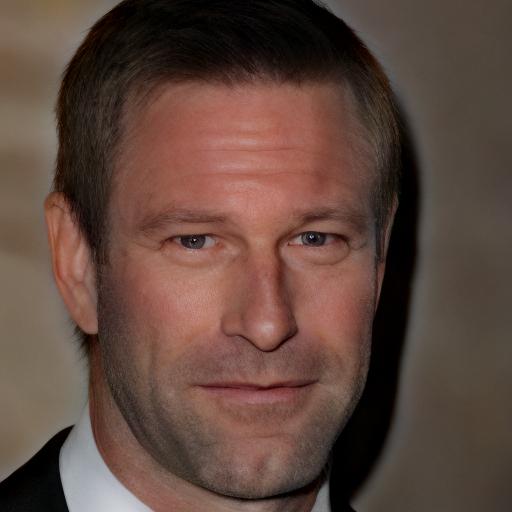} &
        		\includegraphics[width=0.2\textwidth]{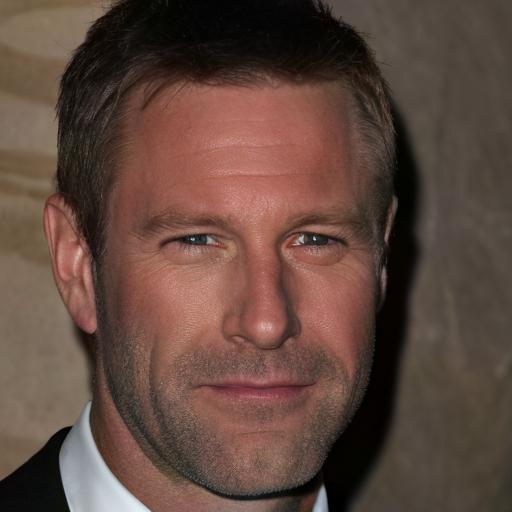} &
        		\includegraphics[width=0.2\textwidth]{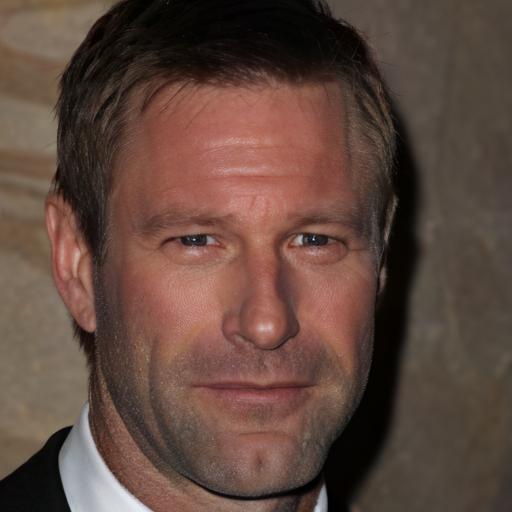} &
        		\includegraphics[width=0.2\textwidth]{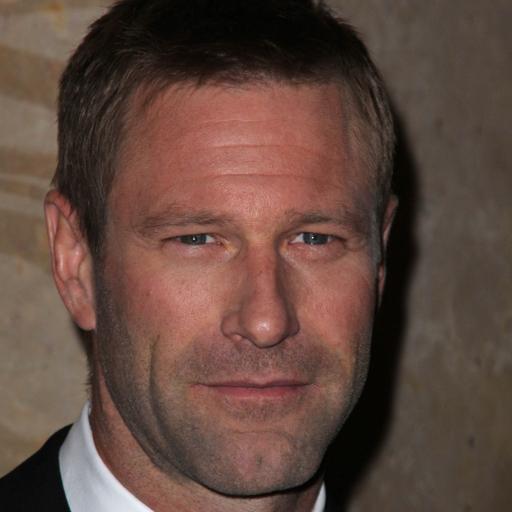}
        		\\
        		\includegraphics[width=0.2\textwidth]{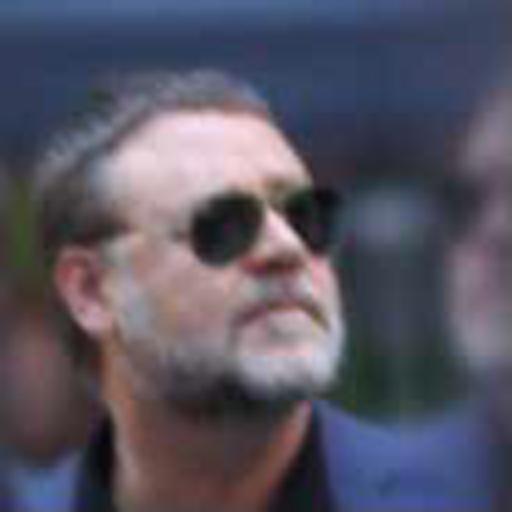} &
        		\includegraphics[width=0.2\textwidth]{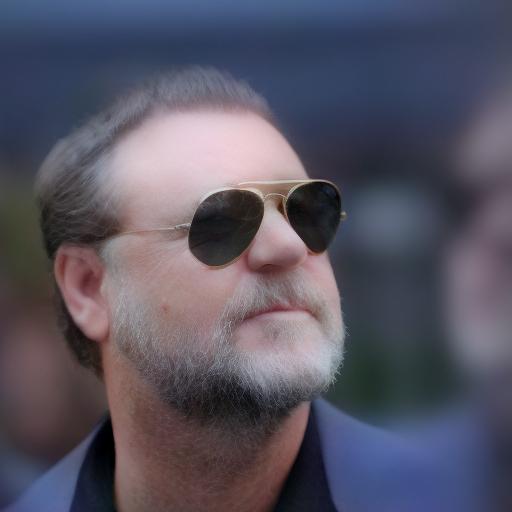} &
        		\includegraphics[width=0.2\textwidth]{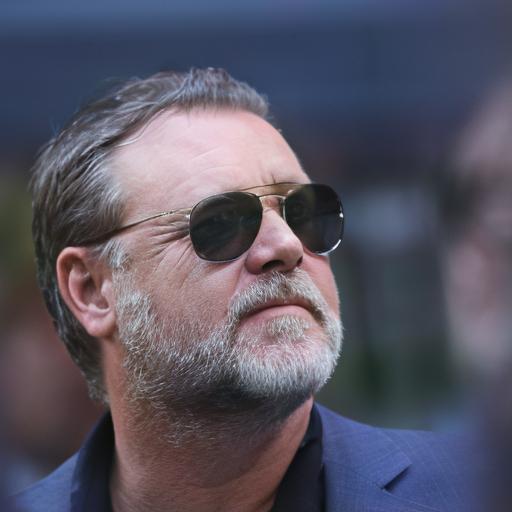} &
        		\includegraphics[width=0.2\textwidth]{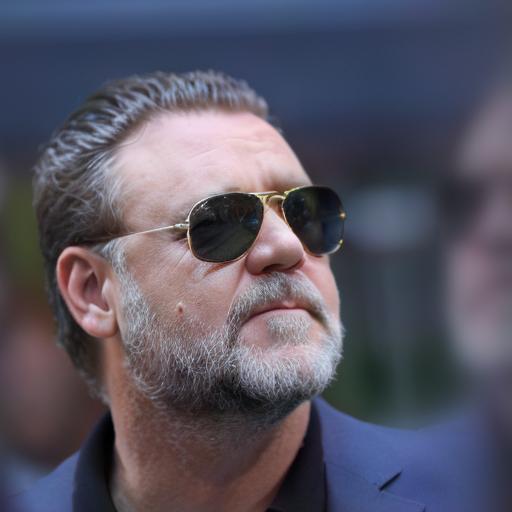} &
        		\includegraphics[width=0.2\textwidth]{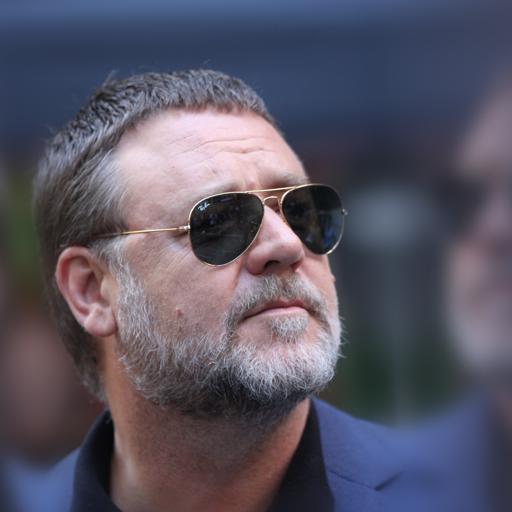}
        		\\
        		\includegraphics[width=0.2\textwidth]{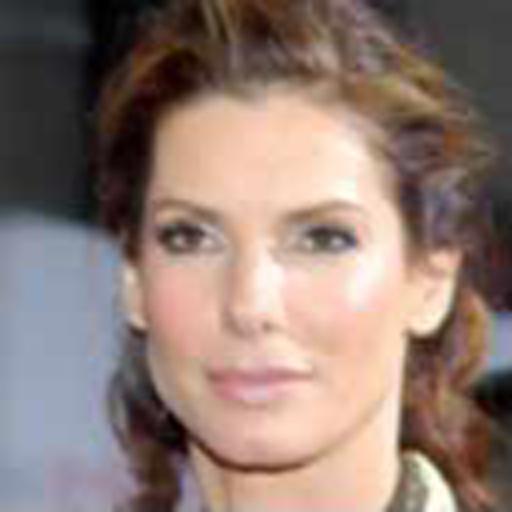} &
        		\includegraphics[width=0.2\textwidth]{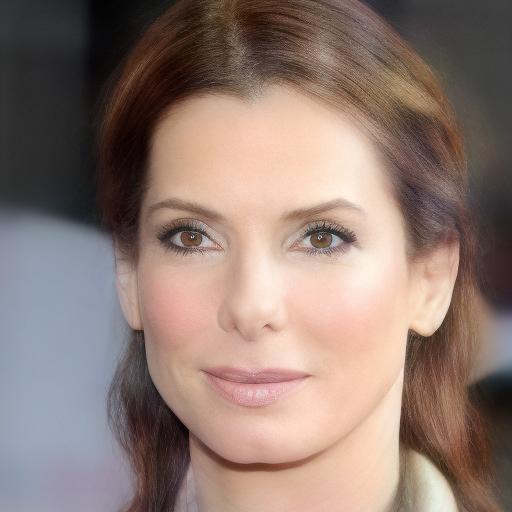} &
        		\includegraphics[width=0.2\textwidth]{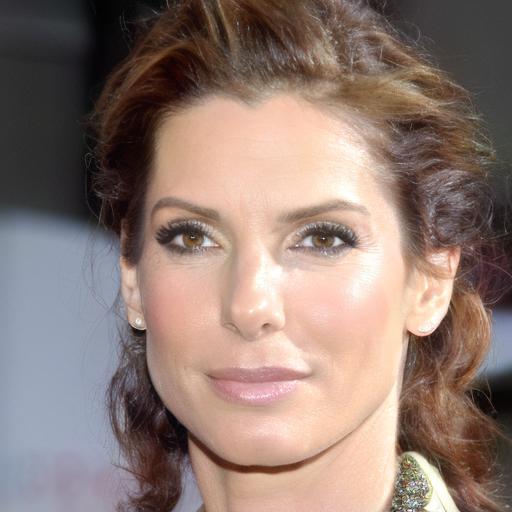} &
        		\includegraphics[width=0.2\textwidth]{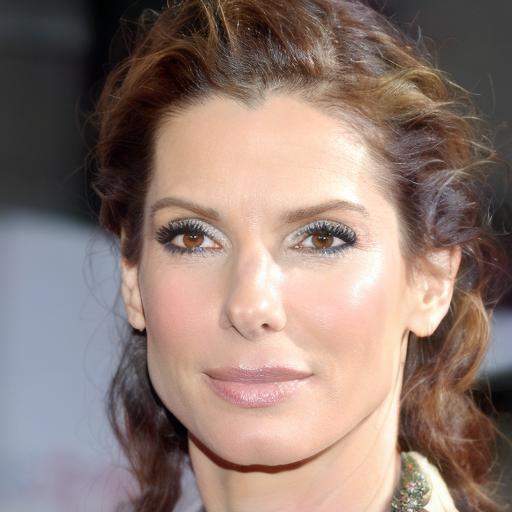} &
        		\includegraphics[width=0.2\textwidth]{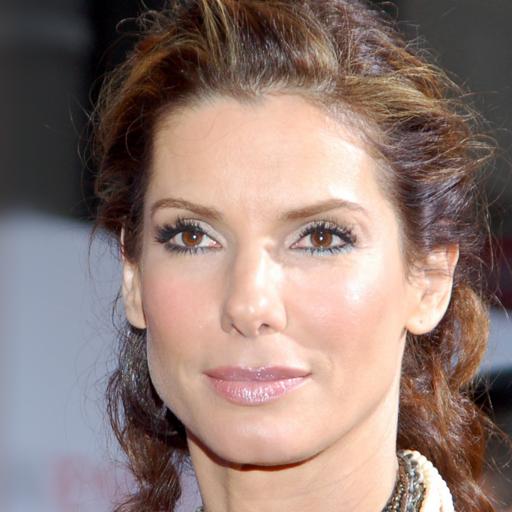}
        		\\
        		\includegraphics[width=0.2\textwidth]{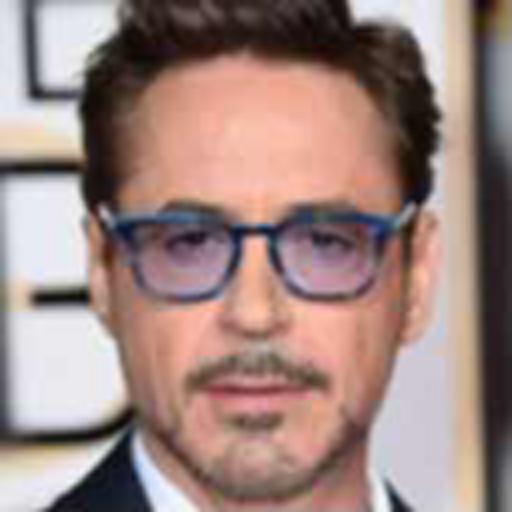} &
        		\includegraphics[width=0.2\textwidth]{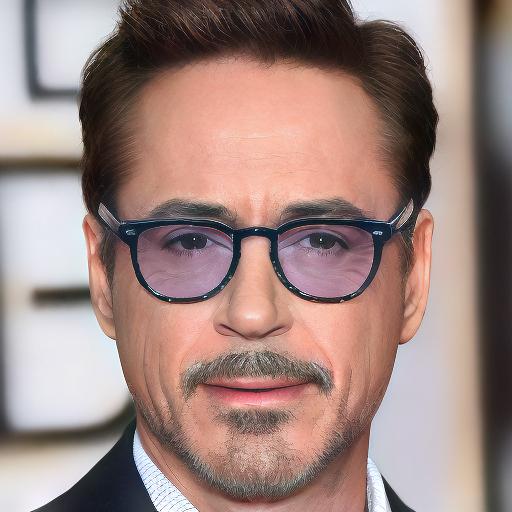} &
        		\includegraphics[width=0.2\textwidth]{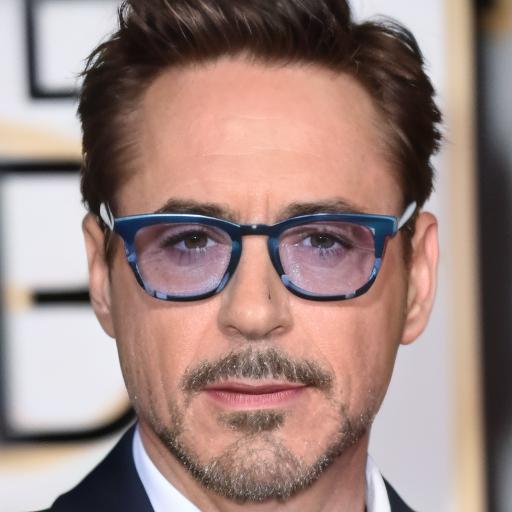} &
        		\includegraphics[width=0.2\textwidth]{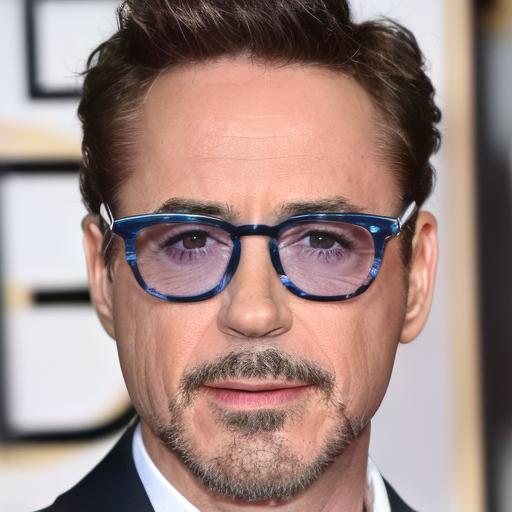} &
        		\includegraphics[width=0.2\textwidth]{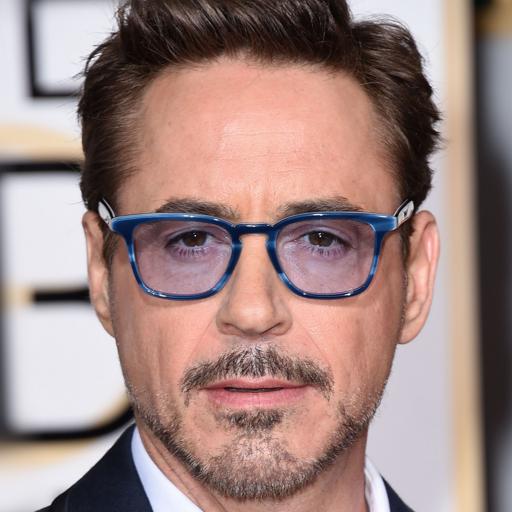}
        		\\
        		\includegraphics[width=0.2\textwidth]{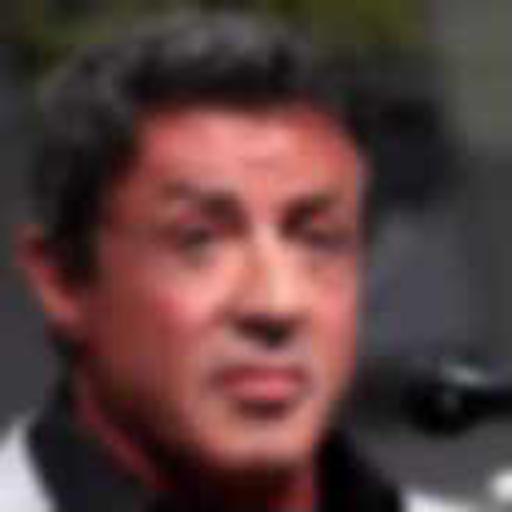} &
        		\includegraphics[width=0.2\textwidth]{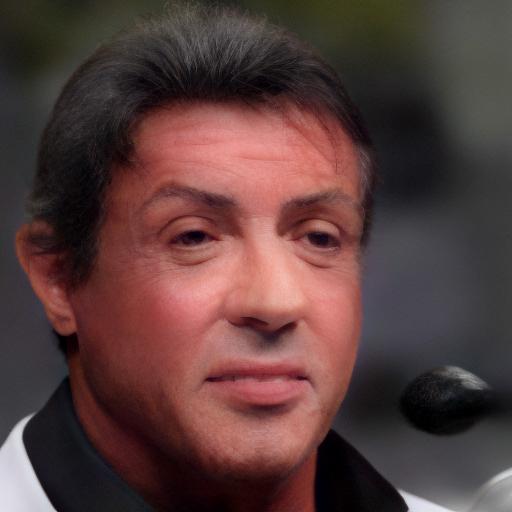} &
        		\includegraphics[width=0.2\textwidth]{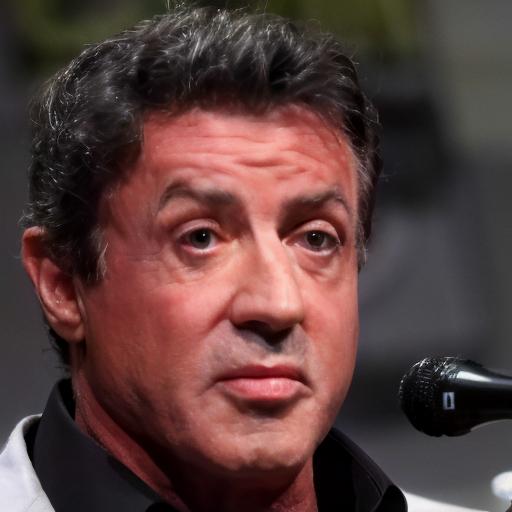} &
        		\includegraphics[width=0.2\textwidth]{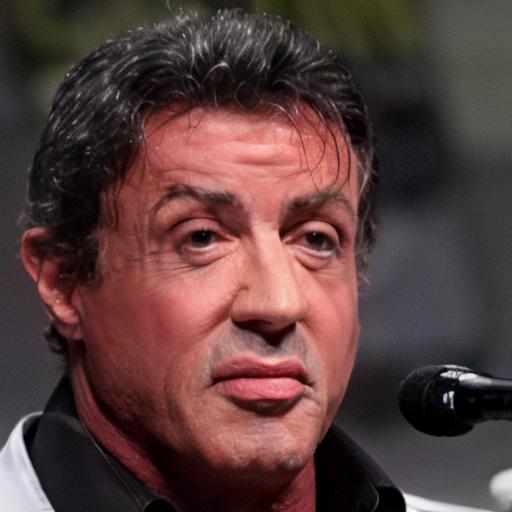} &
        		\includegraphics[width=0.2\textwidth]{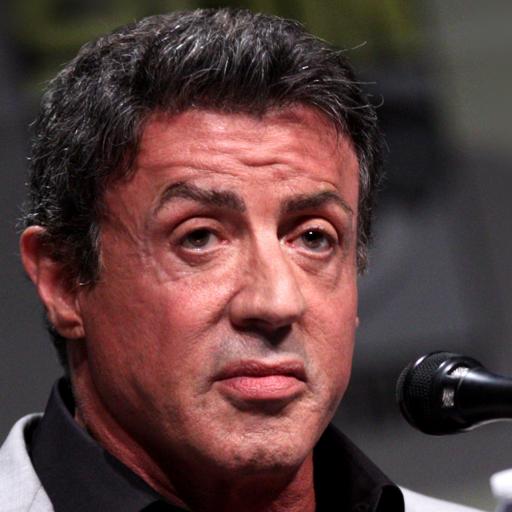}
        		\\
        		\footnotesize{Input} & \footnotesize{Base Model} & \footnotesize{Base Model} & \footnotesize{Ours} & \footnotesize{GT} \\
        		\vspace{-9mm}
        		 & \footnotesize{+ DreamBooth} & \footnotesize{+ ViCo} & &  \\
        		\vspace{-0cm}
        	\end{tabular}
        \end{subfigure}%
	\end{center}
	\caption{
	Results using different personalization techniques combined with a base restoration model with light degradation.
	}
	\label{fig:adapter_comparison_light}
\end{figure*}

Despite the input being very noisy and small in size, our result is faithful with the identity, while codeformer struggles due to requiring alignment. \Cref{fig:adapter_comparison} contains a qualitative comparison between different personalization techniques and more results are provided in \cref{fig:adapter_comparison_heavy,fig:adapter_comparison_light}. 

\paragraph{Additional Quantitative Results}
To complete the quantitative results of heavy portion from \cref{tab:quantitative_heavy}, the results of light and real portions are presented.

\begin{table}
\caption{Quantitative results for the real portion of the data. \color{red} Red \color{black} indicates the best and \color{blue} blue \color{black} indicates the second best}
\resizebox{0.48\textwidth}{!}{%
\begin{tabular}{c|c|c|c} 
 \hline
 Methods & Ref & MUSIQ $\uparrow$ & $^*$ID $\uparrow$ \\ 
 \hline
 Input & & 24.84 & 20.37 \\
 \hline
 StableSR \cite{stablesr} & & 51.59 & 23.39 \\
 Base Model & & 60.27 & 24.29 \\
 Base Model + DreamBooth & \checkmark & 55.31 & 29.78 \\
 Base Model + ViCo & \checkmark & 57.67 & 24.71 \\
 \hline
 \hline
 DMDNet \cite{dmdnet} & \checkmark & \color{blue} 58.36 & 22.29 \\
 DR2 \cite{dr2} & & 29.18 & 18.38 \\
 CodeFormer \cite{codeformer} & & 44.60 & \color{blue} 22.90 \\
 \hline
 PFStorer (Ours) & \checkmark & \color{red} 60.11 & \color{red} 25.01 \\
\end{tabular}}
\label{tab:quantitative_real}
\scriptsize{\\$^*$ Compare with a reference image.}
\end{table}

\Cref{tab:quantitative_real} tabulates the results for the real portion. As no GT is available, we only use MUSIQ \cite{musiq} and ID \cite{arcface} as metrics. For the ID we use a reference image of the same person. As can be seen from the results, the ID metric drops significantly compared to the heavy portion, where ID used GT image. Despite this the rankings of the results remain similar with ours as first and DMDNet and CodeFormer being close with similar results and DR2 achieving the lowest due to blurry results. Base Model + DreamBooth \cite{dreambooth} achieves the best result in ID which is likely due to overfitting to the identity, with poor restoration results. \Cref{tab:quantitative_light} presents results for the light partition. Our method is consistently among the second best performers, although the differences between the methods are minor.

\paragraph{Value of Learnable $\gamma$ after training}
Each layer $l$ has vector $\gamma^l$ with the size depending on the layer hidden dimension. The values of the mean of the vector for each layer is around 0.2 and 0.5. The higher importance of 0.5 values is from the middle of the UNet layers, where the resolution is lowest and the lower values of 0.2 at the higher resolution layers.

\section*{Non-Personalized Base Model Experiments}
In this section we cover the training details and results with the Base Model without personalization. The model is a pre-trained StableSR \cite{stablesr} without any modifications to the architecture. The personalized models all use the fine-tuned Base Model described in this section as their starting point.

\paragraph{Training}
The training is performed on a facial dataset FFHQ \cite{ffhq}, which contains 70,000 facial images in the resolution of $1024 \times 1024$. 50\% of the data is resized randomly to $512 \times 512$ and the other 50\% are taken as random crops of the same resolution. Fine-tuning is performed for 12 epochs. At this moment the personalizion adapter is not attached to the model. We synthesize training data in the same manner as the personalized model with the heavy degradation.

\begin{table}
\caption{Quantitative results for the light portion of the data. \color{red} Red \color{black} indicates the best and \color{blue} blue \color{black} indicates the second best}
\resizebox{0.48\textwidth}{!}{%
\begin{tabular}{c|c|cc|cc|c|c} 
 \hline
 Methods & Ref & PSNR $\uparrow$ & SSIM $\uparrow$ & LPIPS $\downarrow$ & MUSIQ $\uparrow$ & LMSE $\downarrow$ & ID $\uparrow$ \\ 
 \hline
 Input & & 22.56 & 0.719 & 0.615 & 58.83 & 9.74 & 21.85 \\
 \hline
 StableSR & & 27.18 & 0.767 & 0.337 & 62.96 & 5.73 & 70.62 \\
 Base Model & & 27.72 & 0.767 & 0.318 & 64.16 & 4.93 & 72.43 \\
 Base Model + DreamBooth & \checkmark & 24.77 & 0.721 & 0.419 & 62.01 & 14.29 & 62.57 \\
 Base Model + ViCo & \checkmark & 27.58 & 0.765 & 0.325 & 63.04 &  4.88 & 73.26 \\
 \hline
 \hline
 DMDNet \cite{dmdnet} & \checkmark & \color{red} 27.72 & \color{red} 0.780 & 0.312 & 63.07 & 6.43 & \color{blue} 72.66 \\
 DR2 \cite{dr2} & & 22.17 & 0.701 & 0.449 & 47.36 & 13.13 & 30.01 \\
 CodeFormer \cite{codeformer} & & 27.19 & 0.759 & \color{red} 0.293 & \color{red} 66.00 & \color{blue} 5.91 &  69.13 \\
 \hline
 PFStorer (Ours) & \checkmark & \color{blue} 27.71 & \color{blue} 0.767 & \color{blue} 0.309 & \color{blue} 63.31 & \color{red} 4.79 & \color{red} 75.39 \\
 \hline
 GT & & $\infty$ & 1 & 0 & 62.37 & 0 & 100 \\
\end{tabular}}
\label{tab:quantitative_light}
\end{table}

\begin{figure*}
	\begin{center}
    	\setlength{\tabcolsep}{1pt}
        \begin{subfigure}{0.98\textwidth}
        \hspace{-0.2cm}
        	\begin{tabular}{*4c}
        		\includegraphics[width=0.245\textwidth]{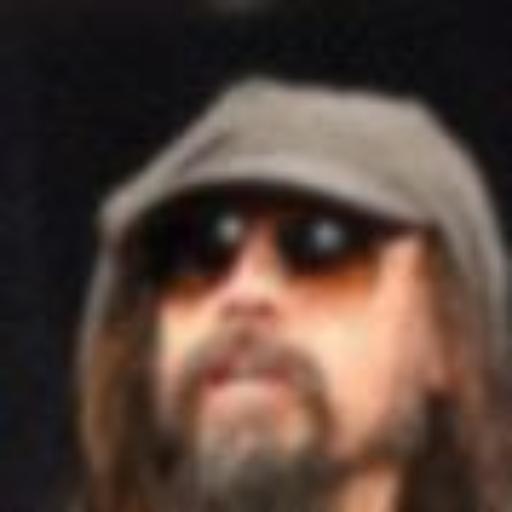} &
        		\includegraphics[width=0.245\textwidth]{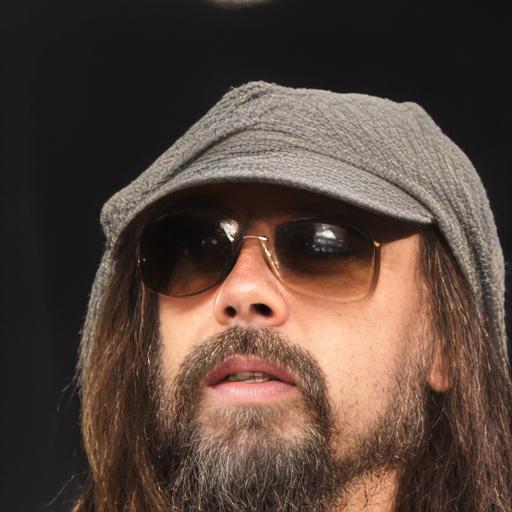} &
        		\includegraphics[width=0.245\textwidth]{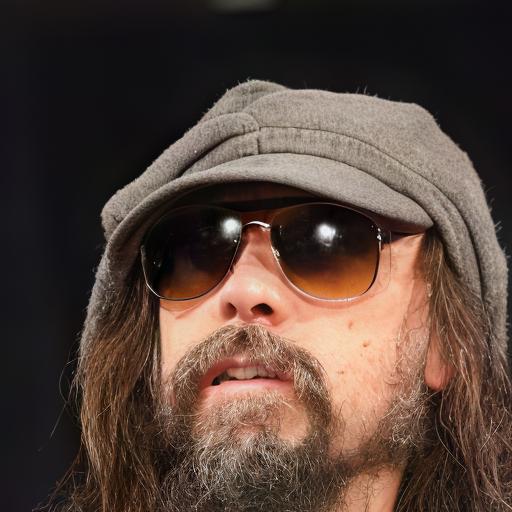} &
        		\includegraphics[width=0.245\textwidth]{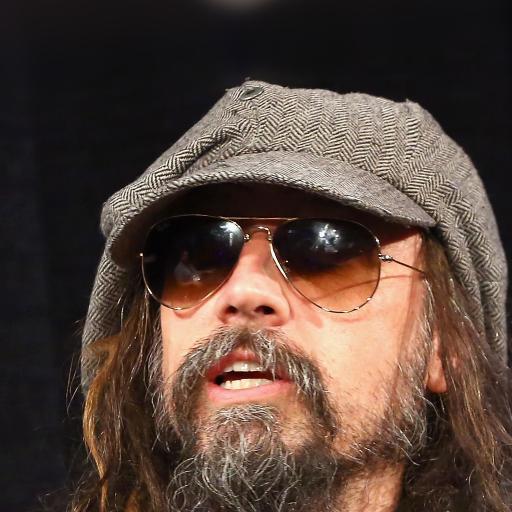}
        		\\
        		\includegraphics[width=0.245\textwidth]{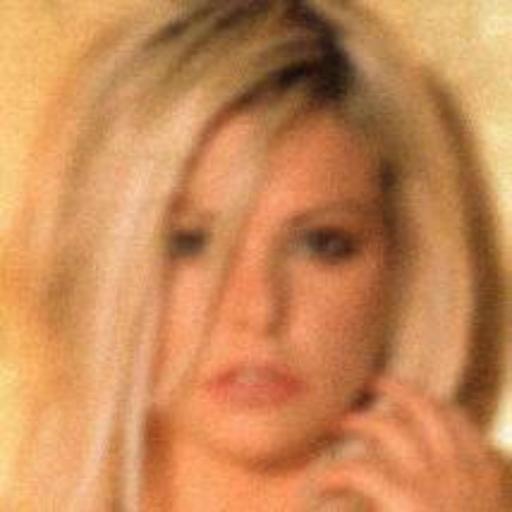} &
        		\includegraphics[width=0.245\textwidth]{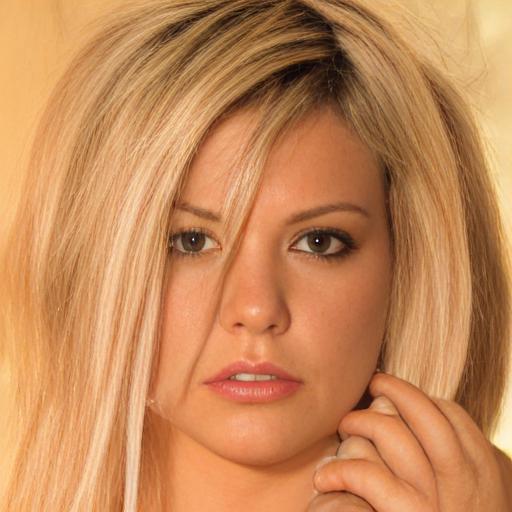} &
        		\includegraphics[width=0.245\textwidth]{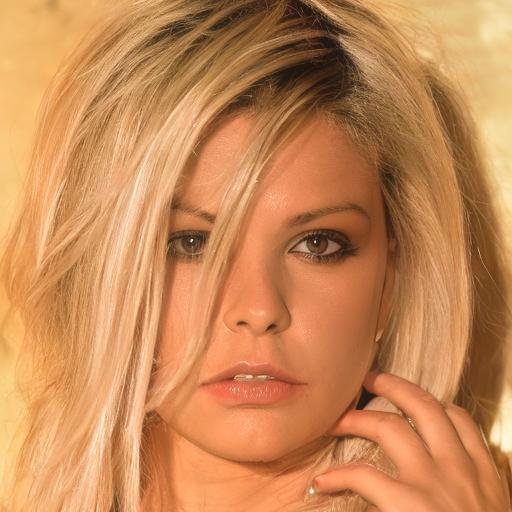} &
        		\includegraphics[width=0.245\textwidth]{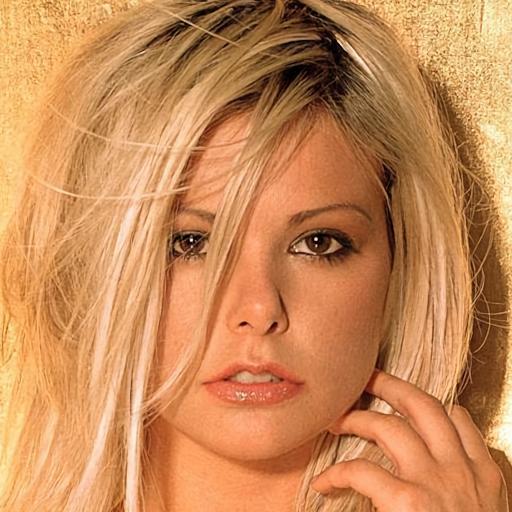}
        		\\
        		\footnotesize{Input} & \footnotesize{CodeFormer} & \footnotesize{Base Model} & \footnotesize{GT} \\
        		\vspace{-1cm}
        	\end{tabular}
        \end{subfigure}%
	\end{center}
	\caption{
	Qualitative comparison with state-of-the-art restoration models on CelebA-Test \cite{celeba_test} with synthetic degradation.
	}
	\label{fig:celeba_test}
\end{figure*}

\begin{figure*}
	\begin{center}
    	\setlength{\tabcolsep}{1pt}
        \begin{subfigure}{0.98\textwidth}
        \hspace{-0.2cm}
        	\begin{tabular}{*6c}
        		\includegraphics[width=0.167\textwidth]{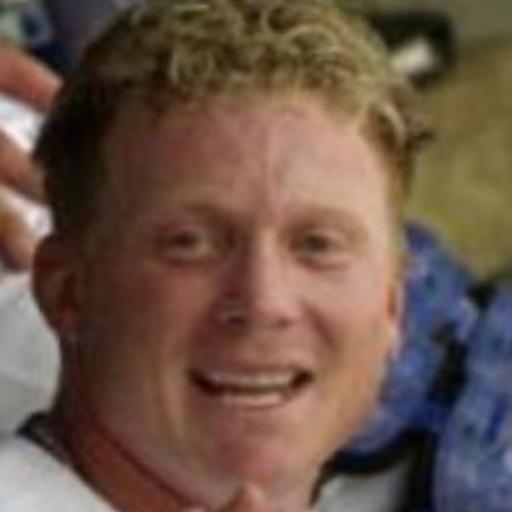} &
        		\includegraphics[width=0.167\textwidth]{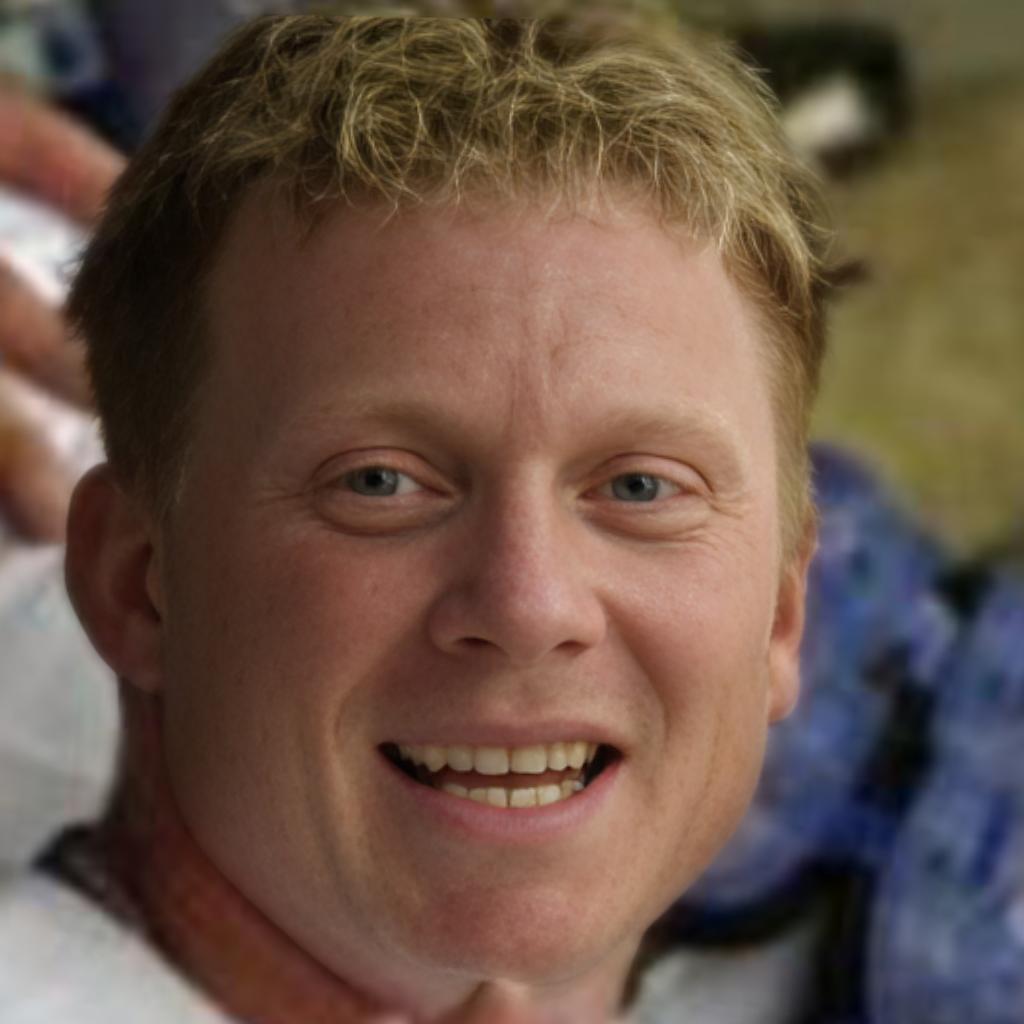} &
        		\includegraphics[width=0.167\textwidth]{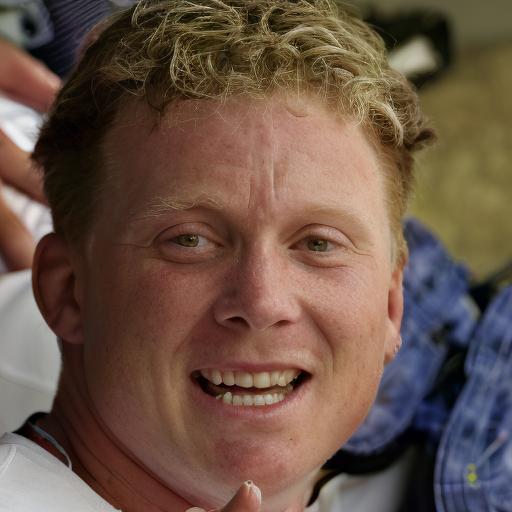} &
        		\includegraphics[width=0.167\textwidth]{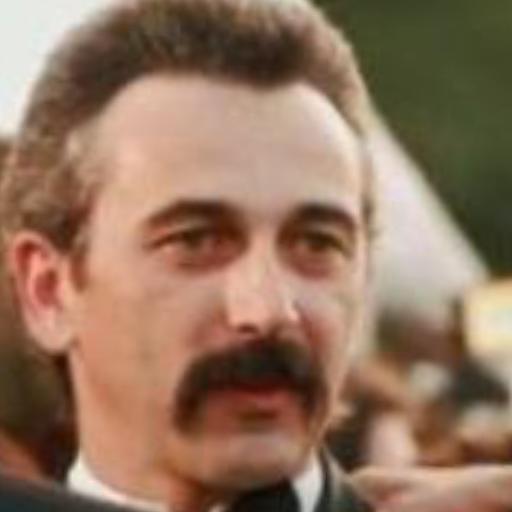} &
        		\includegraphics[width=0.167\textwidth]{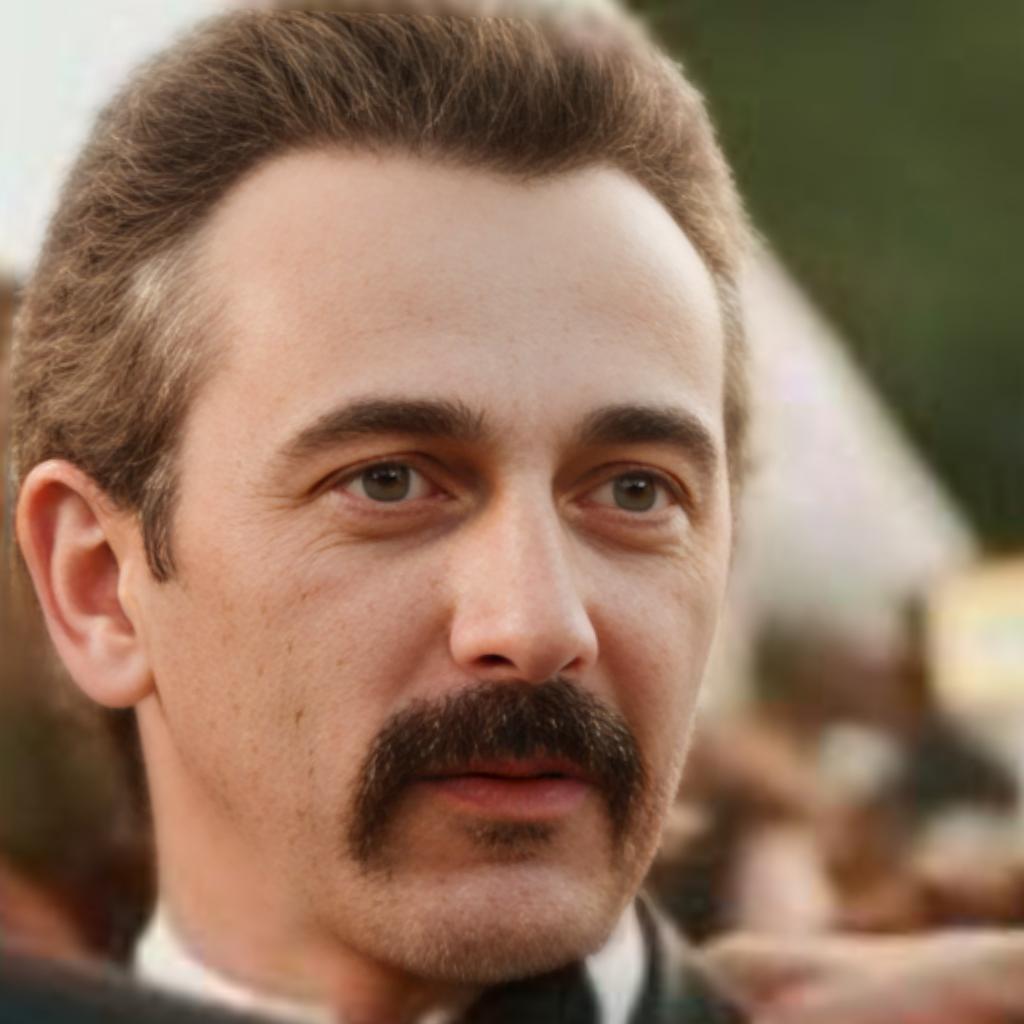} &
        		\includegraphics[width=0.167\textwidth]{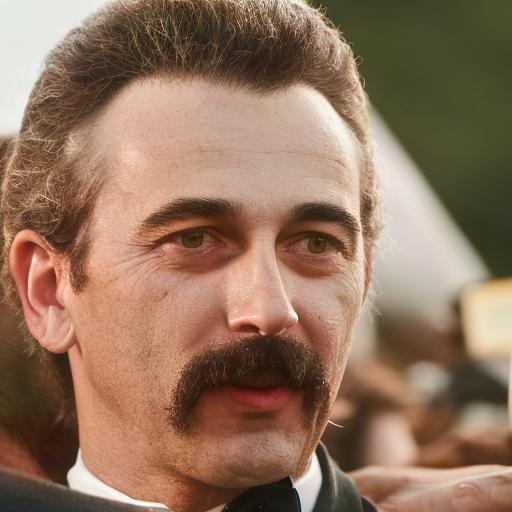}
        		\\
        		\footnotesize{Input} & \footnotesize{CodeFormer} & \footnotesize{Base Model)} & \footnotesize{Input} & \footnotesize{CodeFormer} & \footnotesize{Base Model} \\
        		\multicolumn{3}{c}{LFW-Test} & \multicolumn{3}{c}{LFW-Test} \\
        		
        		\includegraphics[width=0.167\textwidth]{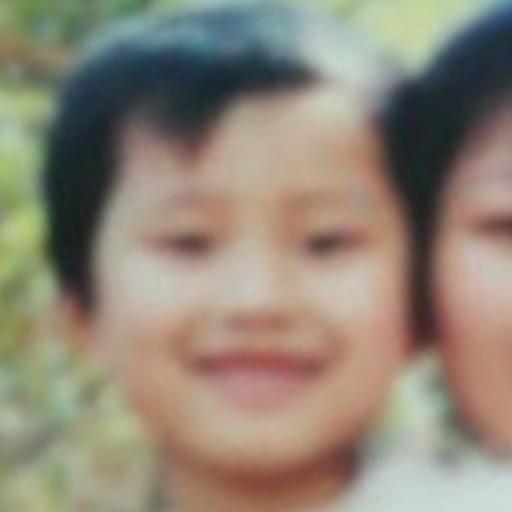} &
        		\includegraphics[width=0.167\textwidth]{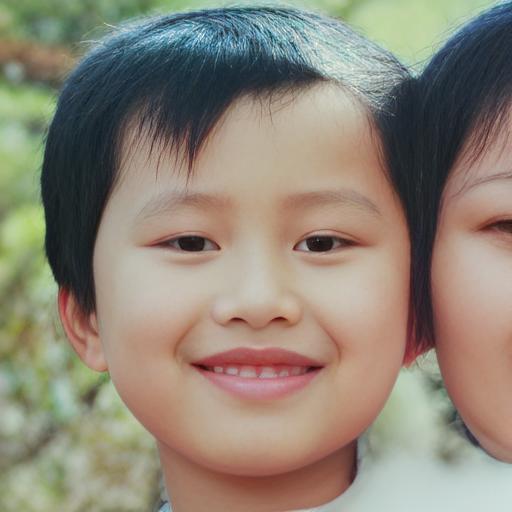} &
        		\includegraphics[width=0.167\textwidth]{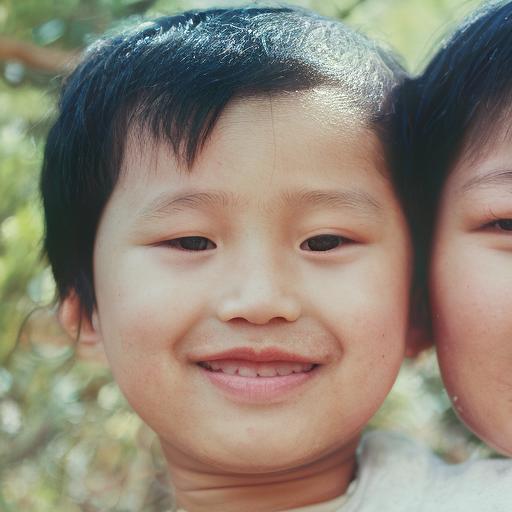} &
        		\includegraphics[width=0.167\textwidth]{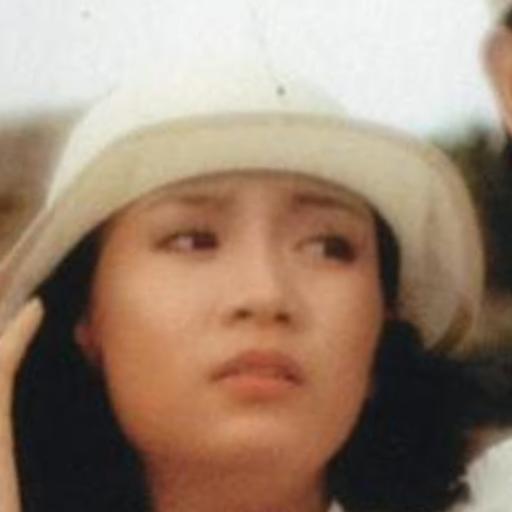} &
        		\includegraphics[width=0.167\textwidth]{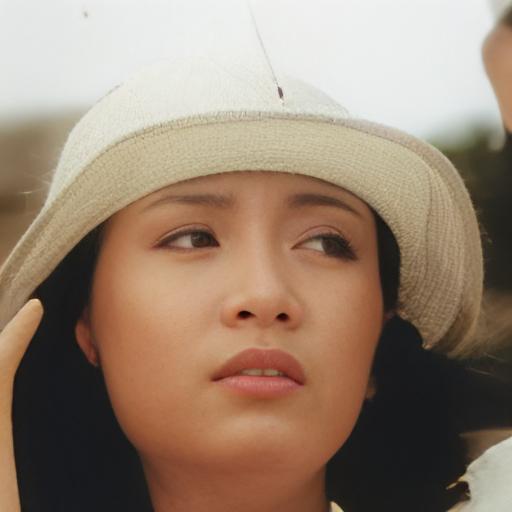} &
        		\includegraphics[width=0.167\textwidth]{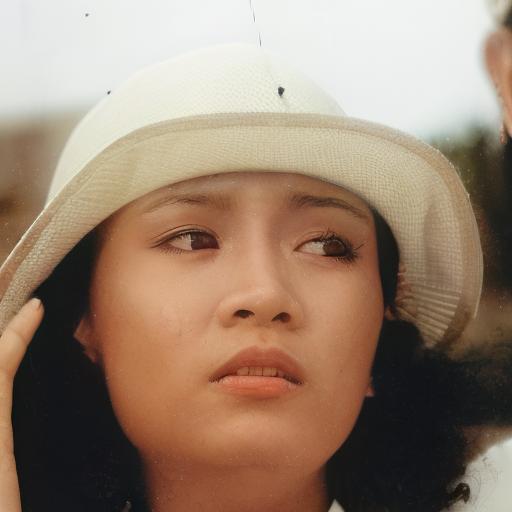}
        		\\
        		\footnotesize{Input} & \footnotesize{CodeFormer} & \footnotesize{Base Model} & \footnotesize{Input} & \footnotesize{CodeFormer} & \footnotesize{Base Model} \\
        		\multicolumn{3}{c}{WebPhoto} & \multicolumn{3}{c}{WebPhoto} \\
        		
        		\includegraphics[width=0.167\textwidth]{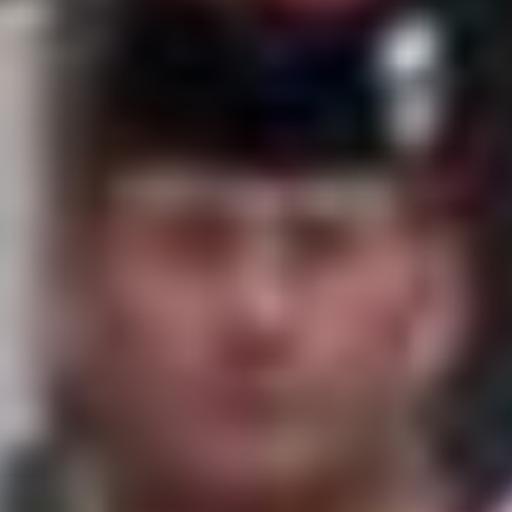} &
        		\includegraphics[width=0.167\textwidth]{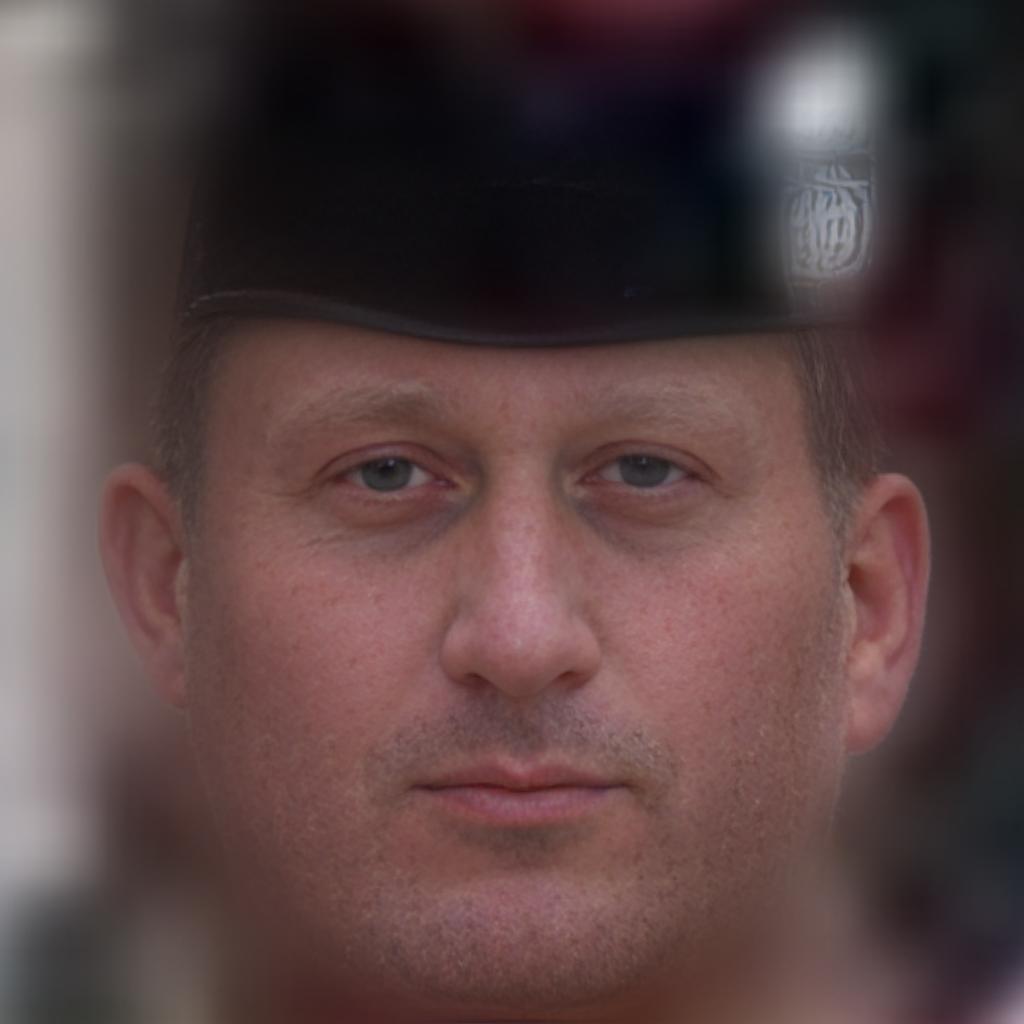} &
        		\includegraphics[width=0.167\textwidth]{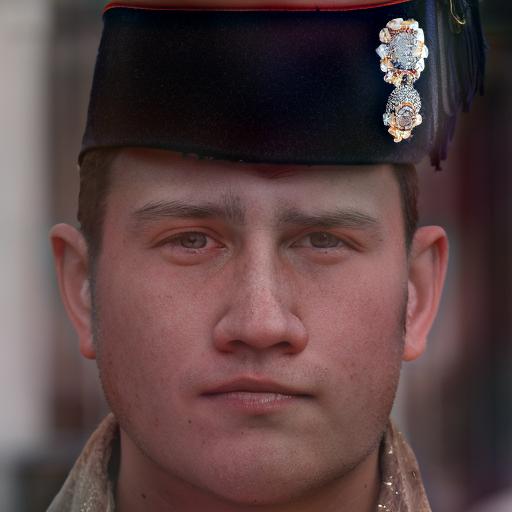} &
        		\includegraphics[width=0.167\textwidth]{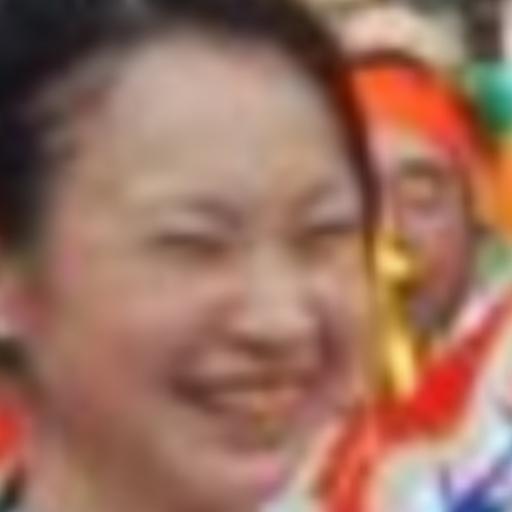} &
        		\includegraphics[width=0.167\textwidth]{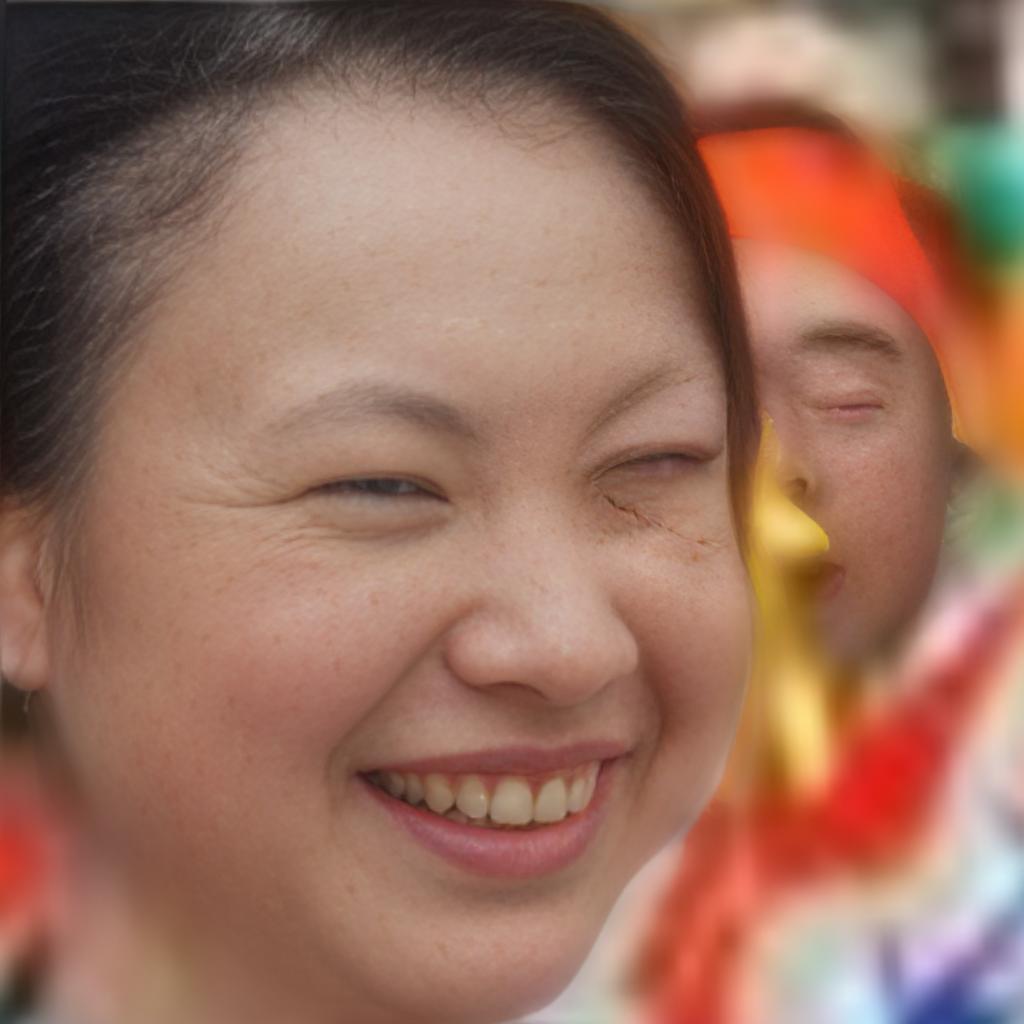} &
        		\includegraphics[width=0.167\textwidth]{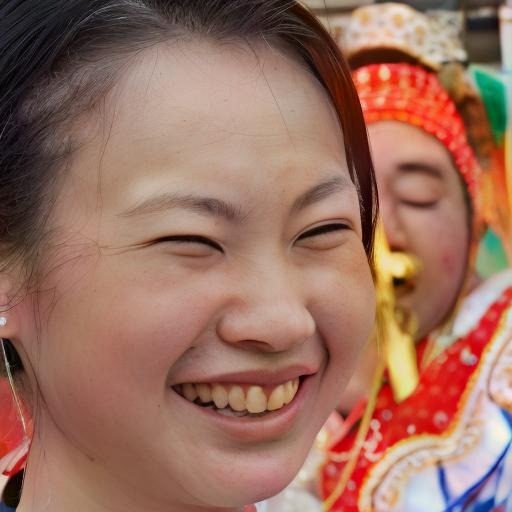}
        		\\
        		\footnotesize{Input} & \footnotesize{CodeFormer} & \footnotesize{Base Model} & \footnotesize{Input} & \footnotesize{CodeFormer} & \footnotesize{Base Model)} \\
        		\multicolumn{3}{c}{Wider-Test} & \multicolumn{3}{c}{Wider-Test} \\
        		\vspace{-1cm}
        	\end{tabular}
        \end{subfigure}%
	\end{center}
	\caption{
	Qualitative comparison with state-of-the-art restoration models on LFW \cite{lfw}, WebPhoto \cite{gfp-gan} and Wider-Test \cite{codeformer}.
	}
	\label{fig:real_datasets}
\end{figure*}

\paragraph{Qualitative Results}
For qualitative results on synthetic degradation on CelebA-Test split \cite{celeba_test}, see \cref{fig:celeba_test}. The synthetic degradation for Celeba-Test is obtained from \cite{gfp-gan}. Compared to CodeFormer \cite{codeformer} our method is able to generate more fine-grained details, while being more faithful to the low-quality image, \eg, the color of facial hair on top. Results from real-world datasets, LFW \cite{lfw}, WebPhoto \cite{gfp-gan} and Wider-Test \cite{codeformer} are shown in \cref{fig:real_datasets}. In LFW, which contains less severe degradations, compared to CodeFormer our method is able to generate more details with sharper textures. Our method struggles with WebPhoto, as it contains old images with scratches, color degradation and other untypical degradations. With severe degradation on Wider-Test, our method is able to generate realistic images, while CodeFormer struggles with artifacts.

\paragraph{Quantitative Results}
We provide quantitative results with standard metrics. \Cref{tab:celeba_test} tabulates results from CelebA-Test, where the results are taken from \cite{gfp-gan}, except for CodeFormer and ours. In most of the metrics the results are similar between GFP-GAN, CodeFormer and ours. In the real-world datasets, \cref{tab:real_datasets}, our method obtains the best FID for LFW and WIDER.

\begin{table}[h]
\caption{Quantitative results for CelebA-Test with non-personalized model. \color{red} Red \color{black} indicates the best and \color{blue} blue \color{black} indicates the second best}
\resizebox{0.48\textwidth}{!}{%
\begin{tabular}{c|cc|ccc|c}
	Methods & PSNR $\uparrow$ & SSIM $\uparrow$ & LPIPS $\downarrow$ & MUSIQ $\uparrow$ & FID $\downarrow$ & ID $\uparrow$ \\
	\hline
	Input & 25.35 & \color{blue} 0.684 & 0.486 & 58.83 & 143.98 & 52.06 \\
	DFDNet & 23.68 &  0.662 & 0.4341 & N/A & 59.08 & 59.69 \\
	PULSE & 21.61 & 0.620 & 0.4851 & N/A & 67.56 & 30.45 \\
	GFP-GAN & 25.08 & 0.677 & \color{blue} 0.3646 & N/A & \color{blue} 42.62 & \color{red} 65.40 \\
	CodeFormer & \color{red} 26.77 & \color{red} 0.719 & \color{red} 0.343 & \color{blue} 66.54 & 52.44 & 62.73 \\
	Base Model & \color{blue} 26.03 & 0.680 & 0.392 & \color{red} 66.57 & \color{red} 40.36 & \color{blue} 63.89 \\
	\hline
	GT & $\infty$ & 1 & 0 & 63.43 & 43.43 & 1 \\
\end{tabular}}
\label{tab:celeba_test}
\end{table}

\begin{table}[h]
\caption{Quantitative results for real-world datasets with non-personalized model. \color{red} Red \color{black} indicates the best and \color{blue} blue \color{black} indicates the second best}
\resizebox{0.48\textwidth}{!}{%
\begin{tabular}{c|cc|cc|cc}
	Dataset & \multicolumn{2}{c|}{\textbf{LFW-Test}} & \multicolumn{2}{c|}{\textbf{WebPhoto-Test}}  & \multicolumn{2}{c}{\textbf{WIDER-Test}} \\ Degradation & \multicolumn{2}{c|}{mild} & \multicolumn{2}{c|}{medium}  & \multicolumn{2}{c}{heavy} \\
	Methods       & FID$\downarrow$  & MUSIQ$\uparrow$  & FID$\downarrow$  & MUSIQ$\uparrow$   &FID$\downarrow$  & MUSIQ$\uparrow$   \\ 
	\hline
	Input     & 137.56 & 25.05 &  170.11  & 19.24 & 202.06 & 15.57 \\
	PULSE~&   64.86 & 66.98  &     {86.45}     &    66.57 & 73.59 & {65.36}\\
	DFDNet  &  62.57  &  {67.95} &      100.68     &63.81 & 57.84& 59.34 \\
	GFP-GAN \cite{gfp-gan} & \color{blue} 49.96 & \color{blue} {68.95}   &  87.35 & \color{blue} {68.04}   &40.59 & \color{blue} {68.26}\\ 
	CodeFormer \cite{codeformer} & {52.02} &  \color{red} 71.43    &   \color{red}  78.87     & \color{red} 70.51 & \color{blue} 39.06  & \color{red} 69.31\\ 
	Base Model & \color{red} 44.11 & 66.57 & \color{blue} 80.90 & 62.69 & \color{red} 34.72 & 63.91 \\
	\hline
	Light degradation & 44.02 & 62.69 & 84.81 & 57.64 & 82.93 & 51.66 \\
\end{tabular}}
\label{tab:real_datasets}
\end{table}

\paragraph{Ablation: Heavy Degradation}
We show that with more complex degradations the method is able to perform better in cases with severe degradation. The results are tabulated in bottom of \cref{tab:real_datasets}. \textit{Base Model} uses the heavy degradation, where as the \textit{Light degradation} does not. For LFW, which has relatively mild degradations, the performance between Base Model and simple degradation does not change drastically as expected. However, for WIDER-Test we can see a large difference as the FID more than doubles from 34.72 to 82.93, meaning a significant decrease in quality. Using heavy degradation results in higher quality outputs under severe degradation, while having minimal effect on mild cases.

\section*{Societal Impact}
Machine learning models can learn biases from their datasets. We show that our model is capable of working with different ethnicities and skin tones, while acknowledging that the testing is limited. We also note that since our model is built upon previous models, it inherits any biases these models may contain. To avoid misunderstanding of the capabilites of our models, \textit{e.g.}, using it for enhancing security footage for criminal investigations, we have shown the limitations in experiments and emphasize that the identity of the restored individual needs to be known beforehand. Malicious users may want to mislead viewers with generated images, which is a common common issue with existing similar methods. However, recent approaches in detecting fake imagery are improving rapidly.

\paragraph{Privacy and Image Copyrights}
In this paper we showcase several pictures of individuals. Several images are from the publicly available Celeb-Ref dataset \cite{dmdnet}. Images shown from the collected 20 image dataset are of well-known celebrities and are under a Creative Commons license. Real world images not part of the collected dataset are under public domain or a Creative Commons license.

